%% file: main.tex
\documentclass[10pt,twocolumn,letterpaper]{article}

\usepackage{cvpr}              

\usepackage{graphicx,array, multirow,xcolor}
\usepackage{amsmath}
\usepackage{amssymb}
\usepackage{booktabs}

\usepackage{algorithm}
\usepackage{algorithmicx}
\usepackage[noend]{algpseudocode}
\algrenewcommand\algorithmicprocedure{}

\usepackage{custom_formats}

\usepackage[pagebackref,breaklinks,colorlinks]{hyperref}

\usepackage[capitalize]{cleveref}
\crefname{section}{Sec.}{Secs.}
\Crefname{section}{Section}{Sections}
\Crefname{table}{Table}{Tables}
\crefname{table}{Tab.}{Tabs.}

\def\cvprPaperID{11220} 
\def\confName{CVPR}
\def\confYear{2022}

\input{defs}
\input{macros}

\begin{document}

\title{How Much More Data Do I Need?\\Estimating Requirements for Downstream Tasks} 

\author{
Rafid Mahmood$^1$~~~~~
James Lucas$^{1,2,3}$~~~~~
David Acuna$^{1,2,3}$~~~~~
Daiqing Li$^1$~~~~~ 
Jonah Philion$^{1,2,3}$ \\
Jose M. Alvarez$^1$~~~~~
Zhiding Yu$^1$~~~~~
Sanja Fidler$^{1,2,3}$~~~~~
Marc T. Law$^1$ \\
$^1$NVIDIA~~~$^2$University of Toronto~~~$^3$Vector Institute \\
{\tt\footnotesize \{rmahmood, jalucas, dacunamarrer, daiqingl, jphilion, josea, zhidingy, sfidler, marcl\}@nvidia.com} \\
\url{https://nv-tlabs.github.io/estimatingrequirements/}\vspace{-1em}
}
\maketitle

\input{sections/abstract}
\input{sections/intro}

\input{sections/related}
\input{sections/problem}
\input{sections/experiments/experiments}

\input{sections/analysis}

\input{sections/discussion}

{\small
\bibliographystyle{ieee_fullname}
\bibliography{refs}
}

%
\input{sections/supplement}

\end{document}

%% file: defs.tex
\newcommand{\dataset}{\mathcal{D}}
\newcommand{\fir}[1]{\mathbf{#1}}
\newcommand{\und}[1]{\underline{#1}}

\newcommand{\regressionset}[2]{\{(#1, #2)\}}

%% file: macros.tex
\newif\ifcomments
\commentstrue

\newcommand{\authcomment}[3]{\ifcomments {\color{#2} [#1: #3]} \else \fi}

\newcommand{\rrm}[1]{\authcomment{RRM}{blue}{#1}}
\newcommand{\ML}[1]{\authcomment{ML}{cyan}{#1}}
\newcommand{\JL}[1]{\authcomment{JL}{orange}{#1}}

\newcommand{\SF}[1]{\authcomment{SF}{green}{#1}}

%% file: sections/abstract.tex
\begin{abstract}
   Given a small training data set and a learning algorithm, how much more data is necessary to reach a target validation or test performance? This question is of critical importance in applications such as autonomous driving or medical imaging where collecting data is expensive and time-consuming. Overestimating or underestimating data requirements incurs substantial costs that could be avoided with an adequate budget. Prior work on neural scaling laws suggest that the power-law function can fit the validation performance curve and extrapolate it to larger data set sizes. We find that this does not immediately translate to the more difficult downstream task of estimating the required data set size to meet a target performance. In this work, we consider a broad class of computer vision tasks and systematically investigate a family of functions that generalize the power-law function to allow for better estimation of data requirements. Finally, we show that incorporating a tuned correction factor and collecting over multiple rounds significantly improves the performance of the data estimators. Using our guidelines, practitioners can accurately estimate data requirements of machine learning systems to gain savings in both development time and data acquisition costs.
\end{abstract}

%% file: sections/intro.tex
\vspace{-7mm}
\section{Introduction}
\label{sec:intro}

Before deploying a deep learning model, designers may mandate that the model meet a baseline performance, such as a target metric over a held out validation or test set. 
For example, an object detector may require a minimum mean average precision before being deployed in a safety-critical application. 
One of the most effective ways of meeting the target performance is by collecting more training data for a given model. 
However, how much more data is needed?

Overestimating data requirements can incur costs from unnecessary collection, cleaning, and annotation. 
For example, annotating segmentation data sets may require $15$ to $40$ seconds per object~\cite{acuna2018efficient}, meaning annotating a driving data set of $100,000$ images with on average $10$ cars per image can take between $170$ and $460$ days-equivalent of time.
On the other hand, underestimating means having to collect more data at a later stage, incurring future costs and workflow delays. For instance in autonomous vehicle applications, each period of data collection requires managing a fleet of drivers to record driving videos. 
Thus, accurately estimating how much data is needed for a given task can reduce both costs and delays in the deep learning workflow. 

\begin{figure}[!t]
\begin{center}
\begin{minipage}{0.99\linewidth}
\includegraphics[width=1\textwidth]{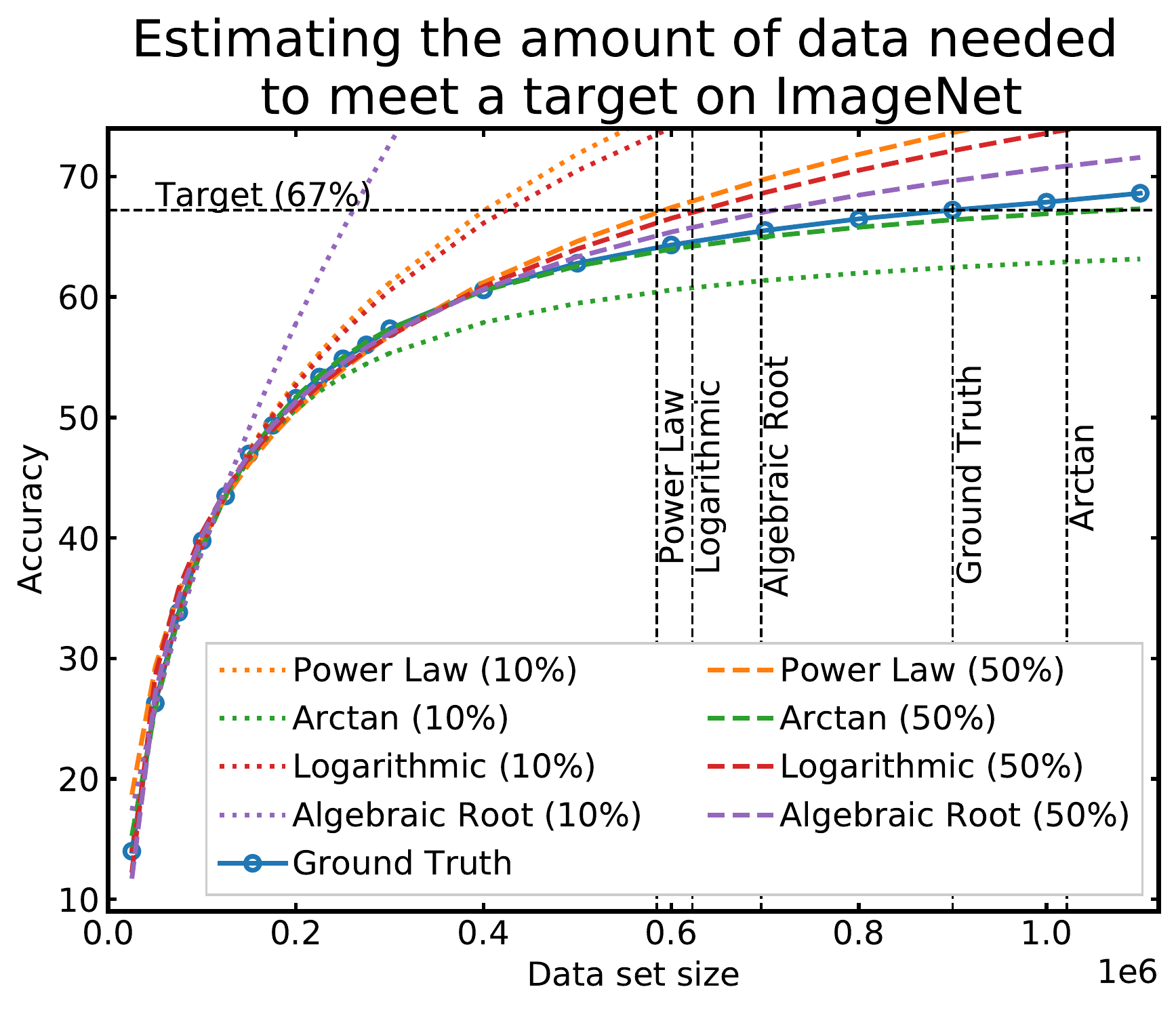}
\end{minipage}
    \vspace{-8mm}
\end{center}
    \caption{\label{fig:regression_summary} 
    Extrapolating accuracy on ImageNet \cite{deng2009imagenet} as a function of data set size from $10\%$ of the data set ($125,000$ images; dotted) and $50\%$ ($600,000$ images; dashed) using four regression functions. 
    The vertical dashed lines show how much data is needed to meet a target 67\% validation accuracy according to each dashed curve.
    All the dashed curves can accurately extrapolate performance as they are given a sufficient amount of images. Although the functions have an error of 1-6\% from the ground truth (67\% at $900,000$ images), they mis-estimate the data requirement by $120,000$ to $310,000$ images. 
    }
    \vspace{-6mm}
\end{figure}

There is a growing body of literature on estimating the sample complexity of machine learning models~\cite{frey1999modeling, gu2001modelling, bisla2021theoretical}.
Recently proposed \emph{neural scaling laws} suggest that generalization scales with the data set size according to a power law~\cite{hestness2017deep, rosenfeld2019constructive, kaplan2020scaling, hoiem2021learning, bahri2021explaining}. Rosenfield \etal \cite{rosenfeld2019constructive} propose fitting a power law function using the performance statistics from a small data set to extrapolate the performance for larger data sets; while not a focus of their paper, they suggest this can be used to estimate the data requirements. 
However, the power law function is not the only possible choice. 
We propose in this paper to use it with similar functions that can be more accurate in practice. 
Figure~\ref{fig:regression_summary} illustrates the data collection process in image classification with the ImageNet data set~\cite{deng2009imagenet} for the power law function and several effective alternatives. 
When using small data sets to extrapolate, the fitted functions may diverge in different ways from the ground truth performance curve. 
More importantly, even a small error in extrapolating accuracy can lead to large errors in over or under-estimating the data requirements, which may present huge operational costs.

\begin{figure*}[!t]
\begin{center}
\begin{minipage}{0.63\linewidth}
\includegraphics[width=1\textwidth, trim={2.5cm 6cm 4.5cm 2.5cm}, clip]{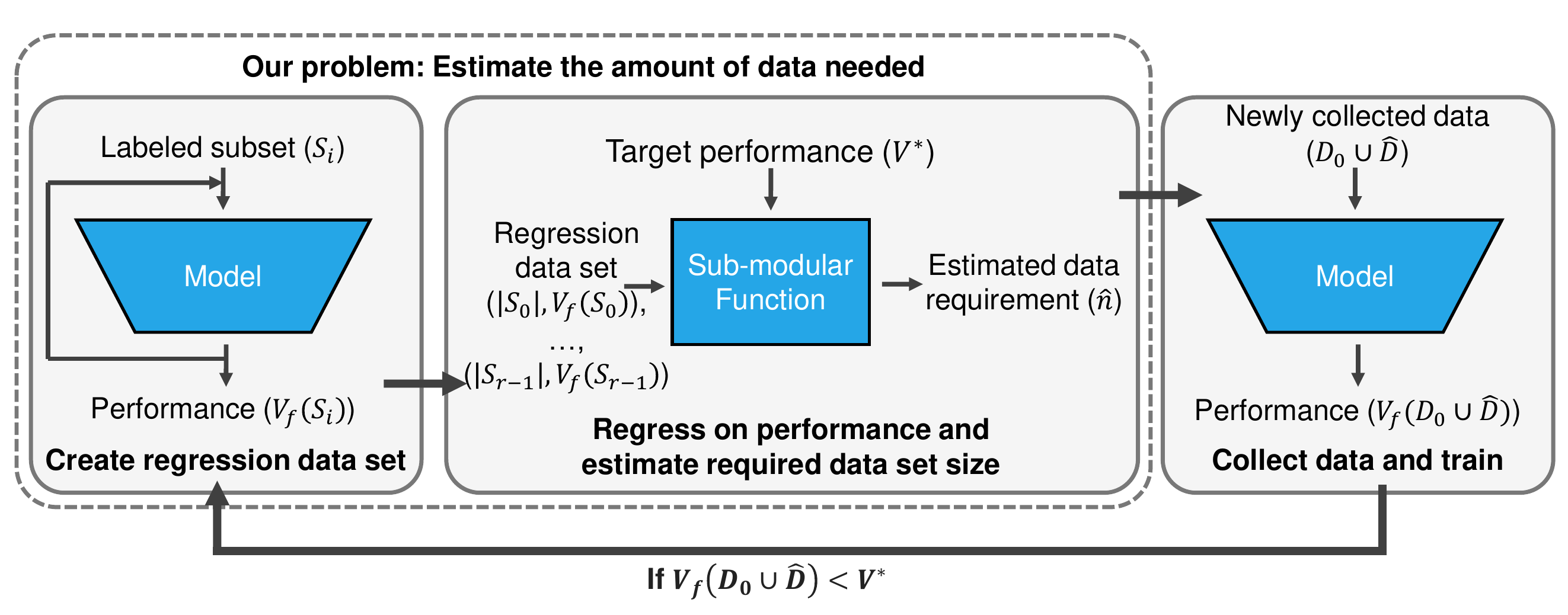} 
    \vspace{-8mm}
    \caption{\label{fig:flowchart} The iterative data collection loop.
    } 
\end{minipage}
\begin{minipage}{0.35\linewidth} 
    \small
    \resizebox{0.99\textwidth}{!}{
    \begin{tabular}{ll} \toprule
        Regression Function    & $\hv(n; \btheta)$ \\ \midrule
        Power Law   & $\theta_1 n^{\theta_2} + \theta_3$  \\
        Arctan      & $\displaystyle \frac{200}{\pi} \arctan\left(\theta_1 \frac{\pi}{2} n + \theta_2 \right) + \theta_3$ \\
        Logarithmic & $\theta_1 \log ( n + \theta_2 ) + \theta_3$ \\
        Algebraic Root & $\displaystyle \frac{100 n}{(1 + | \theta_1 n |^{\theta_2})^{1/\theta_2}} +\theta_3$
        \\ \bottomrule
    \end{tabular}
    }
    \captionof{table}{\label{tab:submodular_funcs} Four concave monotonic increasing regression functions explored in this paper. The set of learnable parameters is  $\btheta := \{ \theta_1, \theta_2, \theta_3 \}$.}
\end{minipage}
\end{center}
    \vspace{-9mm}
\end{figure*}


In this paper, we ask: given a small training data set and a model not yet meeting target performance in some metric, what is the least amount of data we should collect to meet the target? Generalizing the estimation of data requirements from power laws, we investigate several alternate regression functions and show that all of them are well-suited towards estimating model performance. Moreover, each function is almost always either overly optimistic (\ie under-estimating the data requirement) or pessimistic (\ie over-estimating), meaning that there is no unique best regression function for all situations, but using all of the different functions, we can approximately bound the true data requirement.
Through a simulation of the data collection workflow, we show that incrementally collecting data over multiple rounds is critical to meeting the requirement without significantly exceeding it.
Finally, we introduce a simple correction factor to help these functions meet data requirement more often; this factor can be learned by simulating on prior tasks. 
We explore classification, detection, and segmentation tasks with different data sets, models, and metrics to show that our results hold in every setting considered.

Altogether, our empirical findings and proposed enhancements yield easy-to-implement guidelines for data collection in real-world applications: practitioners should allocate for up to five rounds of data collection and use the correction factor introduced in this paper to augment an optimistic regression function (\eg Power Law, Logarithmic, Algebraic Root) in order to accurately estimate data requirements and ultimately collect only a relatively small amount more than the minimum data required to meet the desired performance. We believe that this approach can improve workflows and yield large cost savings in the future.


%% file: sections/related.tex
\section{Related work}
\label{sec:setup_related_work}

\noindent \textbf{Neural scaling laws.}
Prior work has estimated model performance as a function of data set size~\cite{frey1999modeling, sun2017revisiting, rosenfeld2019constructive, bisla2021theoretical, zhai2021scaling, abnar2021exploring}. 
The recent literature on neural scaling laws argues that model performance (usually defined as decreasing loss over a validation set) scales with data set size according to a power law function, \ie $V \propto \theta_1 n^{\theta_2}$ where $n$ is the data set size.
Hestness \etal~\cite{hestness2017deep} empirically validate that power laws accurately model negative validation set loss and top-1 error over different image classification, language, and audio tasks. 
Bahri \etal~\cite{bahri2021explaining} prove that for over-parametrized networks, under Lipschitz continuity of the loss function, model, and data distribution, the out-of-sample loss scales in $\set{O}(n^{-1/\theta})$.
Rosenfeld \etal~\cite{rosenfeld2019constructive} fit power law functions using small data subsets. 
Finally, Hoiem \etal~\cite{hoiem2021learning} use power laws to construct learning curves and investigate modeling questions. One key difference between these studies and our own is that we focus on estimating target data requirements given an approximate relationship between data size and model performance; such as a power law function.
More broadly, this area also relates to the study of learning curves in classical machine learning~\cite{jones2003introduction, figueroa2012predicting, viering2021shape}. Our work differentiates from this literature through a detailed simulation that investigates the operational costs of poorly estimating a learning curve.




\noindent \textbf{Active learning.} In this work, we consider collecting data over multiple rounds. This is related to active learning \cite{cohn1996active}, where a model selects which data to use during multiple rounds of training. The focus of active learning is to intelligently select this data given a fixed collection budget \cite{settles2009active, sener2017active, yoo2019learning, sinha2019variational, mahmood2021low}, sometimes with a focus on the performance on rare categories \cite{poms2021low}. However, the goal of this work is to predict the optimal collection budget itself. 
This paper focuses on random sampling, but includes experiments with active learning in the Appendix to demonstrate that our insights on estimating the data requirement hold independent of the sampling strategy.

\noindent \textbf{Statistical learning theory.} Loosely speaking, statistical learning theory seeks to relate model performance and data set size. Accurate theoretical characterizations of this relationship could be used to infer the target data requirements, but these results are typically only tight asymptotically; if at all. More recent work has explored empirically estimating this theoretical relationship~\cite{jiang2020neurips, jiang2021methods}. Bisla \etal~\cite{bisla2021theoretical} build models of generalization for deep neural networks under assumptions on the training and test behaviour that are validated empirically. Bisla \etal highlight the utility in being able to estimate data requirements from such a model, but do not explore this empirically as we do in this work.


%% file: sections/problem.tex

\section{Main problem}
\label{sec:setup}

In this section, we mathematically define the data collection problem and the general solution method.
The goal of this problem is to estimate the data set size that returns a desired performance in a limited number of rounds. 
We first model performance as a function of data set size and then solve for the data given an input performance.


\subsection{The data collection problem}
\label{sec:setup_datacollection}

Let $z \sim p(z)$ be data drawn from a distribution $p$. 
For instance, $z := (x, y)$ may correspond to images $x$ and labels $y$.
Consider a prediction problem for which we currently have an initial training data set $\dataset_0 := \{z_i\}_{i=1}^{n_0}$ of $n_0$ points and a model $f$. 
Let $V_f(\dataset)$ be a score function of the model after it is trained on a set $\dataset$. 
Our goal is to obtain a pre-determined target score $V^* > V_f(\dataset_0)$.

To achieve our goal, we sample $\hn$ additional points to create $\hat\dataset := \{ \hat{z}_i \}_{i=1}^{\hn} \sim p(z)$ and then evaluate $V_f(\dataset_0 \cup \hat\dataset)$. If we do not meet the target, we must determine a larger $\hn$ and augment $\hat\dataset$ with more data. 
Because each data point incurs a cost from collecting, cleaning, and labeling, we ideally want the fewest number of points $\hn$ that achieve the target.  
Furthermore, because initiating a round of data collection is itself expensive and time-consuming, we are only permitted a maximum of $T$ rounds; failing to meet the requirement within $T$ rounds means failing to solve the problem. 
This problem is summarized in the following iterative sequence. Initialize $\hat\dataset = \emptyset$. Then in each round, repeat: 

(1) Estimate the amount of additional data $\hn$ needed. 
 
(2) Sample points until $|\hat\dataset| = \hn$ and then evaluate the score. If $V_f(\dataset_0 \cup \hat\dataset) \geq V^*$, then terminate. Otherwise, repeat for another round up to $T$ rounds.

The objective of the data collection problem is to select the minimum $\hn$ such that $V_f(\dataset_0 \cup \hat\dataset) \geq V^*$ within $T$ rounds. 
This paper focuses on the first step of the loop: accurately estimating the $\hn$ required to meet $V^*$.

\subsection{Regressing performance using data set size}
\label{sec:setup_regression}

\begin{figure}[t!]
    \vspace{-2mm}
    \begin{algorithm}[H]
    \footnotesize
    \caption{The data collection problem}\label{alg:data_collection}
    \begin{algorithmic}[1]
    \State \textbf{Input:}  Initial data set $\dataset_0$, Score function $V_f(\dataset)$, Target score $V^*$, Maximum rounds $T$, Regression model $\hv(n;\btheta)$, Initial regression set size $r$
    \State Set $n_0 \gets | \dataset_0 |$, $\hat\dataset = \emptyset$
    \Procedure{Create regression data set}{}
    \State Sample subsets $\set{S}_0 \subset \set{S}_1 \subset \cdots \subset \set{S}_{r-1} = \dataset_0$ 
    \State Evaluate $V_f(\set{S}_i)$ and create $\set{R} \gets \regressionset{|\set{S}_i|}{V_f(\set{S}_i)}_{i=0}^{r-1}$
    \EndProcedure
    \Procedure{Perform data collection}{}
    \Repeat
    \State Fit $\btheta^* \gets \argmin_{\btheta} \sum_{(n, v) \in \set{R}} (v - \hv(n; \btheta) )^2$
    \State Minimize $\hn$ subject to $\hv(n_0 + \hn; \btheta^*) \geq V^*$
    \State Sample points from $p(z)$ until $|\hat\dataset| = \hn$
    \State Train model and evaluate score $V_f(\dataset_0 \cup \hat\dataset)$ 
    \State Update $\set{R} \gets \set{R} \cup \regressionset{n_0 + \hn}{V_f(\dataset_0 \cup \hat\dataset)}$
    \Until{$V_f(\dataset_0 \cup \hat\dataset) \geq V^*$ or $T$ rounds have passed}
    \EndProcedure
    \State \textbf{Output:} Final collected data set $\dataset_0 \cup \hat\dataset$
    \end{algorithmic}
    \end{algorithm}
    \vspace{-8mm}
\end{figure}


Figure~\ref{fig:flowchart} illustrates our data collection pipeline to estimate $\hn$, motivated by the following empirical observation.

\noindent\textbf{Observation from~\cite{frey1999modeling, rosenfeld2019constructive}.} \emph{
Let $\dataset_0 \subset \dataset_1 \subset \cdots$ be a growing sequence of data sets and let $n_i = |\dataset_i|$ for each $i$ in the sequence. Then, the piecewise linear function
\begin{align*}\label{eq:piecewiselinear}
    v(n) := \begin{cases} 
        \frac{V_f(\dataset_0)}{n_0} n , & n \leq n_0 \\ 
        \frac{V_f(\dataset_i) - V_f(\dataset_{i-1})}{n_i - n_{i-1}} \left( n - n_i \right) + V_f(\dataset_i), & n_{i-1} \leq n \leq n_i
    \end{cases}
\end{align*}
is concave and monotonically increasing.
}

Recall that  $V_f(\dataset_i)$ is the model score after it is trained on $\dataset_i$. 
We refer to $v(n)$ as the model score function over the training data set size. 
The observation implies that intuitively, as we collect more data, the marginal value of each additional data point should decrease (\eg Figure~\ref{fig:regression_summary}). Furthermore, we can model $v(n)$ by regression using concave, monotonically increasing functions. 
Within the data collection loop, we first estimate $\hn$ by using the available data, $\dataset_0$ and $\hat\dataset$, and the corresponding scores by fitting a regression model $\hv(n; \btheta)$ of $v(n)$, where $\btheta$ is the set of regression parameters. 
We consider four functions that satisfy the Observation (see Table~\ref{tab:submodular_funcs}) from the learning curve literature~\cite{viering2021shape}. 
While we could use more complicated models, we find these simpler structured functions with a small number of parameters are easier to fit to smaller data sets of learning statistics.
Using the fitted regression function, we solve for the smallest $\hn$ such that $\hv(n_0 + \hn; \btheta) \geq V^*$.

Algorithm~\ref{alg:data_collection} summarizes the main steps. We first create a regression data set by selecting $r$ subsets $\set{S}_0 \subset \set{S}_1 \subset \dots \subset \set{S}_{r-1} = \dataset_0$ and computing their scores; this yields a set of $r$ pairs $\set{R} := \{ (|\set{S}_i|, V_f(\set{S}_i))\}_{i=0}^{r-1}$. 
Then, in the data collection loop, we select a function $\hv(n; \btheta)$ from Table~\ref{tab:submodular_funcs} and fit the set of parameters $\btheta$ via least squares minimization. 
Finally, we minimize $\hn$ subject to $\hv(\hn; \btheta^*)~\geq~V^*$, and then collect $\hn$ new points.
In subsequent rounds of data collection as we obtain $\hat\dataset$, we augment $\set{R}$ with pairs $(|\dataset_0| + |\hat\dataset|, V_f(\dataset_0 \cup \hat\dataset))$, and then re-fit $\hv(n; \btheta)$.

The existing literature shows that power laws can estimate model accuracy using data set size, but the practical application of estimating the required data set size to meet a target score presents three major challenges. We highlight them below using the ImageNet data set in Figure~\ref{fig:regression_summary}.

\noindent\textbf{All of the functions in Table \ref{tab:submodular_funcs} fit the model score.} 
With enough data, all of the regression functions in Table~\ref{tab:submodular_funcs} can accurately fit $v(n)$. 
When fit using $|\dataset_0| = 600,000$ images ($\approx50\%$ of the data set), 
Figure~\ref{fig:regression_summary} shows that each link function (dashed curves) achieves at most $6\%$ error from the ground truth accuracy when extrapolating. 
Although power laws are theoretically motivated~\cite{bahri2021explaining,hutter2021learning}, is there empirical justification for using them over other functions?

\noindent\textbf{Extrapolating accuracy with small data sets is hard.} 
With limited data, all of the regression functions extrapolate $v(n)$ poorly. 
Figure~\ref{fig:regression_summary} shows how each curve (dotted curves) diverges significantly from the ground truth when fitting with $|\dataset_0| = 125,000$ images ($\approx10\%$ of the data set). Further, some curves provide better fit than power laws. 
This small data regime was observed in~\cite{hestness2017deep,rosenfeld2019constructive} who proposed jointly regressing on data set and model size;
while this improves extrapolating performance, it also requires a $2\times$ larger $\set{R}$ obtained by sampling subsets and modifying different models. 
This can grow computationally expensive and time-consuming; as a result, we focus on simple estimators using a small number of training statistics, i.e., $r \leq 10$.

\noindent\textbf{Small accuracy errors yield large data errors.} 
Suppose we must build a model meeting $67\%$ test set accuracy on ImageNet, which requires $900,000$ data points. 
Even though the functions fit using $600,000$ images achieve error $|67\% - \hv(900,000;\btheta)|$ between $1$ to $6\%$,
they mis-estimate the data requirement between $120,000$ to $310,000$ images---collecting up to $34\%$ less data than actually required. Since the tolerance for extrapolation errors is low, we must determine best practices for estimating data needs.


%% file: sections/experiments/experiments.tex
\section{Empirical findings}\label{sec:empirical}

We investigate the three challenges using regression and simulation over different data sets and tasks. We first summarize our experimental setup before analyzing the results.

\subsection{Data and methods}

\begin{table}[t!]
    \centering
    \footnotesize
\resizebox{\linewidth}{!}{
\begin{tabular}{lllc}
\toprule
             Data set & Task & Score & Full data set size \\
\midrule
    CIFAR10~\cite{krizhevsky2009learning}  & Classification & Accuracy & $50,000$     \\
    CIFAR100~\cite{krizhevsky2009learning} & Classification & Accuracy & $50,000$      \\
    ImageNet~\cite{deng2009imagenet} & Classification & Accuracy & $1,281,167$   \\ \midrule
    VOC~\cite{pascal-voc-2007, pascal-voc-2012} & 2-D Object Detection & Mean AP & $16,551$       \\ 
    nuScenes~\cite{nuscenes2019} & 3-D Object Detection & Mean AP & $28,130$ \\ \midrule
    BDD100K~\cite{yu2020bdd100k} & Semantic Segmentation & Mean IoU & $7,000$          \\
    nuScenes~\cite{nuscenes2019} & BEV Segmentation & Mean IoU & $28,130$    \\
\bottomrule
\end{tabular}
}
    \vspace{-1em}
    \caption{Data sets, tasks, and score functions considered.}
    \label{tab:tasks}
    \vspace{-2em}
\end{table}

\begin{table*}[t!]
\begin{minipage}{0.73\linewidth}
    \centering
    \footnotesize
\begin{tabular}{clllcccc}
\toprule
        & Data set &  $n_0$  & $r$ & Power Law &          Arctan &       Logarithmic &    Algebraic Root \\
\midrule
\multirow{9}{*}{\rotatebox[origin=c]{90}{Classification}} &
  CIFAR10                   & 10\% &  5 &     $39.02 \pm 20.3$ & $\fir{7.98 \pm 7.1}$ &     $32.28 \pm 13.1$ &  $33.63 \pm 22.1$  \\
& CIFAR10                   & 20\% & 10 &      $15.26 \pm 1.3$ &  $\fir{1.0 \pm 0.6}$ &      $11.53 \pm 1.5$ &    $4.97 \pm 1.6$  \\
& CIFAR10                   & 50\% & 17 &        $6.0 \pm 0.5$ & $\fir{0.38 \pm 0.3}$ &        $4.4 \pm 0.5$ &    $0.76 \pm 0.4$  \\ \cmidrule{2-8}
& CIFAR100                  & 10\% &  5 &     $34.38 \pm 35.1$ & $\fir{13.3 \pm 5.3}$ &     $17.25 \pm 21.8$ &  $26.29 \pm 16.8$  \\
& CIFAR100                  & 20\% & 10 &      $29.52 \pm 3.9$ & $\fir{4.71 \pm 2.0}$ &      $19.87 \pm 2.5$ &   $40.33 \pm 1.5$  \\
& CIFAR100                  & 50\% & 17 &       $5.49 \pm 0.2$ & $\fir{0.69 \pm 0.2}$ &       $5.42 \pm 0.2$ &    $3.65 \pm 0.3$  \\ \cmidrule{2-8}
& ImageNet                  & 10\% &  4 &      $23.89 \pm 7.4$ & $\fir{3.19 \pm 2.1}$ &       $17.2 \pm 3.2$ &    $60.1 \pm 1.1$  \\
& ImageNet                  & 20\% &  8 &      $10.12 \pm 0.4$ & $\fir{2.38 \pm 0.5}$ &       $9.46 \pm 0.6$ &    $7.61 \pm 1.0$  \\
& ImageNet                  & 50\% & 15 &       $5.06 \pm 0.1$ & $\fir{0.74 \pm 0.2}$ &       $3.81 \pm 0.2$ &    $1.64 \pm 0.2$  \\ \midrule
\multirow{6}{*}{\rotatebox[origin=c]{90}{Detection}} &
  VOC                       & 20\% &  4 &       $4.66 \pm 3.1$ & $\fir{2.98 \pm 1.6}$ &       $3.23 \pm 2.1$ &    $3.28 \pm 1.8$  \\
& VOC                       & 30\% &  6 &       $3.16 \pm 1.6$ & $\fir{2.31 \pm 1.2}$ &       $2.55 \pm 1.3$ &    $2.83 \pm 1.3$  \\
& VOC                       & 50\% & 10 &       $1.15 \pm 0.5$ & $\fir{0.79 \pm 0.5}$ &       $1.08 \pm 0.4$ &    $1.13 \pm 0.5$  \\ \cmidrule{2-8}
& nuScenes                  & 10\% &  2 &       $6.57 \pm 0.5$ &      $13.43 \pm 0.3$ & $\fir{0.79 \pm 0.2}$ &    $4.53 \pm 0.4$  \\
& nuScenes                  & 20\% &  4 &       $2.10 \pm 2.1$ & $\fir{1.65 \pm 1.0}$ &       $1.73 \pm 1.3$ &    $2.32 \pm 1.6$  \\
& nuScenes                  & 50\% &  6 &       $0.69 \pm 0.2$ &       $0.71 \pm 0.1$ &       $0.51 \pm 0.2$ & $\fir{0.36 \pm 0.2}$ \\ \midrule
\multirow{6}{*}{\rotatebox[origin=c]{90}{Segmentation}} &
  BDD100K                   & 10\% &  5 &       $9.85 \pm 7.9$ &       $8.12 \pm 7.6$ &       $9.18 \pm 8.9$ & $\fir{5.82 \pm 2.3}$  \\
& BDD100K                   & 20\% & 10 &       $2.98 \pm 1.2$ & $\fir{0.76 \pm 0.3}$ &       $1.60 \pm 0.9$ &    $2.76 \pm 1.2$  \\
& BDD100K                   & 50\% & 17 &       $1.30 \pm 0.5$ &       $0.95 \pm 0.3$ & $\fir{0.82 \pm 0.2}$ &    $1.10 \pm 0.5$  \\ \cmidrule{2-8}
& nuScenes                  & 10\% &  5 &       $2.78 \pm 0.0$ &       $2.30 \pm 0.7$ &       $2.03 \pm 0.9$ & $\fir{1.47 \pm 0.6}$  \\
& nuScenes                  & 20\% & 10 & $\fir{0.61 \pm 0.2}$ &       $3.34 \pm 0.6$ &       $0.91 \pm 0.7$ &    $2.31 \pm 1.0$  \\
& nuScenes                  & 50\% & 17 &       $0.38 \pm 0.3$ &       $2.40 \pm 0.1$ & $\fir{0.28 \pm 0.2}$ &    $1.77 \pm 1.7$  \\ 
\bottomrule
\end{tabular}
\end{minipage} 
      \begin{minipage}{0.27\linewidth}
    \caption{\label{tab:regression_RMSE_tasks}Mean$\pm$standard deviation of multiple runs evaluating the RMSE on extrapolating performance in each task when trained on small subsets of the data. We report $n_0$ in terms of the percentage of the true data set. The lowest error for each setting is bolded. 
    We provide regression plots and alternate error metrics in the supplementary content.
    Given $50\%$ of the data, every function achieves low regression error, whereas, with $10\%$ of the data all of the functions have significant error in their estimation. Furthermore, the alternative functions typically outperform the Power Law across different values of $n_0$ and over different tasks.
    }
      \end{minipage}
\vspace{-4mm}
\end{table*}

We assess the data collection problem on image classification, object detection, and semantic segmentation tasks summarized in Table~\ref{tab:tasks}. In classification, we train ResNets~\cite{he2016deep} on the CIFAR10~\cite{krizhevsky2009learning}, CIFAR100~\cite{krizhevsky2009learning}, and ImageNet~\cite{deng2009imagenet} data sets, where we determine the amount of data needed to meet a target validation set accuracy. We train SSD300~\cite{liu2016ssd} for  2-D object detection using the PASCAL VOC data sets~\cite{pascal-voc-2007, pascal-voc-2012}, where we determine the amount of data needed to meet a target mean average precision (AP).
For 3-D object detection, we train the FCOS3D network architecture \cite{wang2021fcos3d} 
on different subsets of the nuScenes training set. 
We report mean average precision (mAP) following the nuScenes 3D detection evaluation protocol\cite{nuscenes2019}. Samples are obtained randomly across different scenes. 
We explore semantic segmentation using BDD100K \cite{yu2020bdd100k}, which is a large-scale driving dataset collected over 50K drives with various geographic, environmental, and weather conditions.
For multi-view Bird's-Eye-View (BEV) segmentation, we train the ``Lift Splat'' architecture~\cite{liftsplat} on the nuScenes data set~\cite{nuscenes2019}. 
Here, we report mean intersection-over-union (IoU).
For each task, we fix the architecture of the model and learning algorithm including data sampling. Details are in the supplementary content.

For each data set and task, we have an initial dataset $\dataset_0$ (\eg $n_0 = 10\%$ of the training data set). 
In our analyses, we report $n_0$ in terms of the relative size of $\dataset_0$ \wrt the full training data set.
We first create a regression data set $\set{R}$ according to Algorithm~\ref{alg:data_collection} by sampling $r$ subsets that grow linearly in size (\ie each $|\set{S}_i| = |\dataset_0| (i + 1)/r$ for all $i \in \{0, \dots, r-1\}$).
To ensure that this regression procedure is inexpensive, we use a small $r \leq 10$.
Then to evaluate our regression functions on extrapolating performance and estimating data requirements, we sample larger subsets $\dataset_1 \subset \dataset_2 \subset \cdots$ of growing size (\eg $10\%$, $20\%$, $30\%$, \dots, $100\%$ of the full training data set). For each subset, we train our model and evaluate the score $V_f(\dataset_i)$. Using these sets, we construct the piecewise linear score function $v(n)$, which we use as a ground truth.



We perform two types of experiments. In the first, preliminary, analysis we fit each regression function from Table~\ref{tab:submodular_funcs} using $\set{R}$ and then evaluate their error with respect to predicting $V_f(\dataset_i)$ for all $|\dataset_i| > |\dataset_0|$. This analysis reveals how well each of the regression functions can extrapolate the model's score for larger data sets.
Our second, main, analysis is a simulation of the data collection problem in Algorithm~\ref{alg:data_collection} where we initialize with $n_0 = 10\%$ of the full training data set ($n_0=20\%$ for VOC) and estimate how much data is needed to obtain different target values $V^*$ within $T=1, 3, 5$ rounds. 
Here, we repeat the same steps described in the Data Collection stage of Algorithm~\ref{alg:data_collection}, except with one difference. 
In our simulations, rather than sampling more data and evaluating $V_f(\dataset_0 \cup \hat\dataset)$ in each round (\eg lines 11--13), we evaluate $v(n_0 + \hn)$ to obtain the model score.
This simulation approximates the true data collection problem, while simplifying experimentation since we do not have to repeatedly re-train our model.

\subsection{Analysis}
\label{sec:empirical_analysis}

\noindent\textbf{Regression.}
Table~\ref{tab:regression_RMSE_tasks} summarizes the Root Mean Squared Error (RMSE) of each regression function when extrapolating the score for larger data sets. 
In each data set and task, we perform three runs with different random seeds, showing how well we can extrapolate with small, medium, and large subsets of the data. 
In the supplement, we provide regression plots for $v(n)$ versus $\hv(n;\btheta^*)$ and a table summarizing regression error in terms of relative error ratio.

We validate that the first two challenges mentioned in Section~\ref{sec:setup_regression} hold for every task that we consider. 
Given a sufficient amount of initial data $\dataset_0$ to fit a regression model (\ie when $n_0$ is equal to $50\%$ of the full data set size), every link function achieves a low RMSE (whose range is the interval $[0,100]$). 
Moreover, there is always at least one regression function that achieves an RMSE less than $1$. 
When $n_0$ is equal to $10\%$ of the full data set size, most of the link functions yield high RMSE, suggesting that the functions are susceptible to diverging from the true $v(n)$ when fitted on a small data set. 
Finally, for most data sets, our alternative regression functions consistently yield low RMSE. In particular, the Arctan function is the best for all of the classification data sets, and often cuts the RMSE from the Power Law by half.
These results show that extrapolating model performance from small data sets is difficult, but furthermore, other regression functions instead of the Power Law may obtain more accurate regressions of the score.

\begin{figure*}[!t]
\vspace{-1mm}
\begin{center}
\begin{minipage}{0.16\linewidth}\includegraphics[width=1\textwidth]{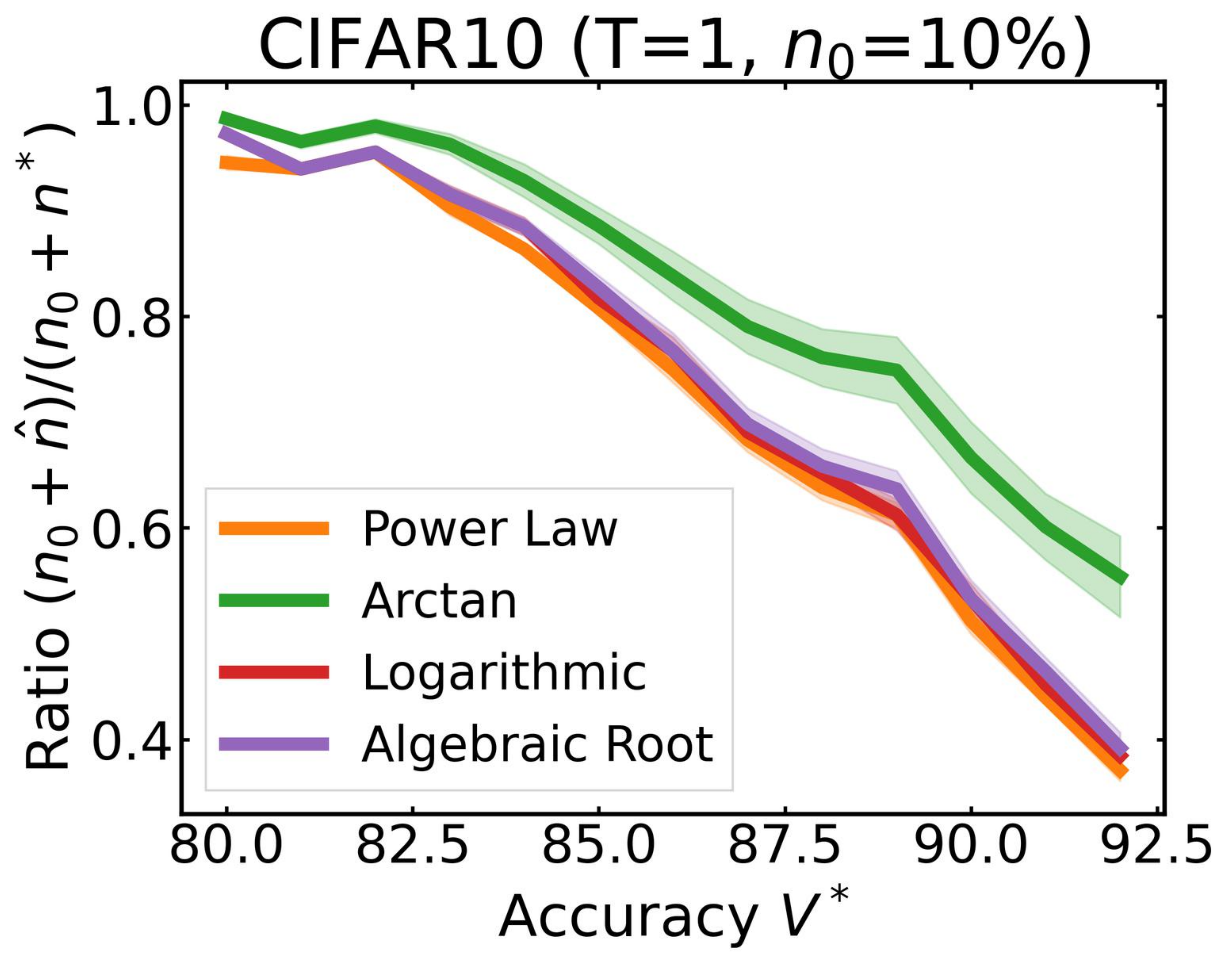} \end{minipage}
\begin{minipage}{0.16\linewidth}\includegraphics[width=1\textwidth]{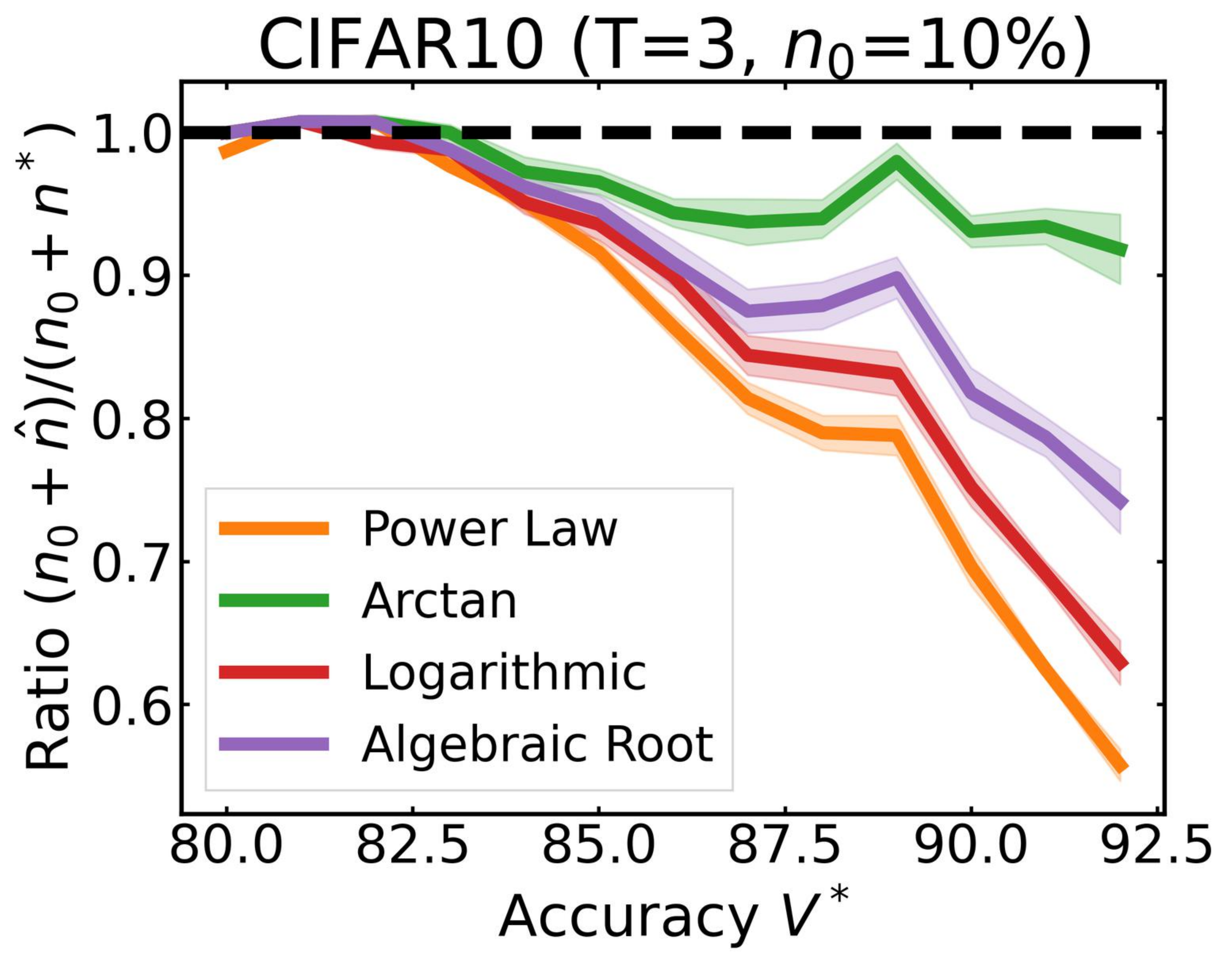} \end{minipage}
\begin{minipage}{0.16\linewidth}\includegraphics[width=1\textwidth]{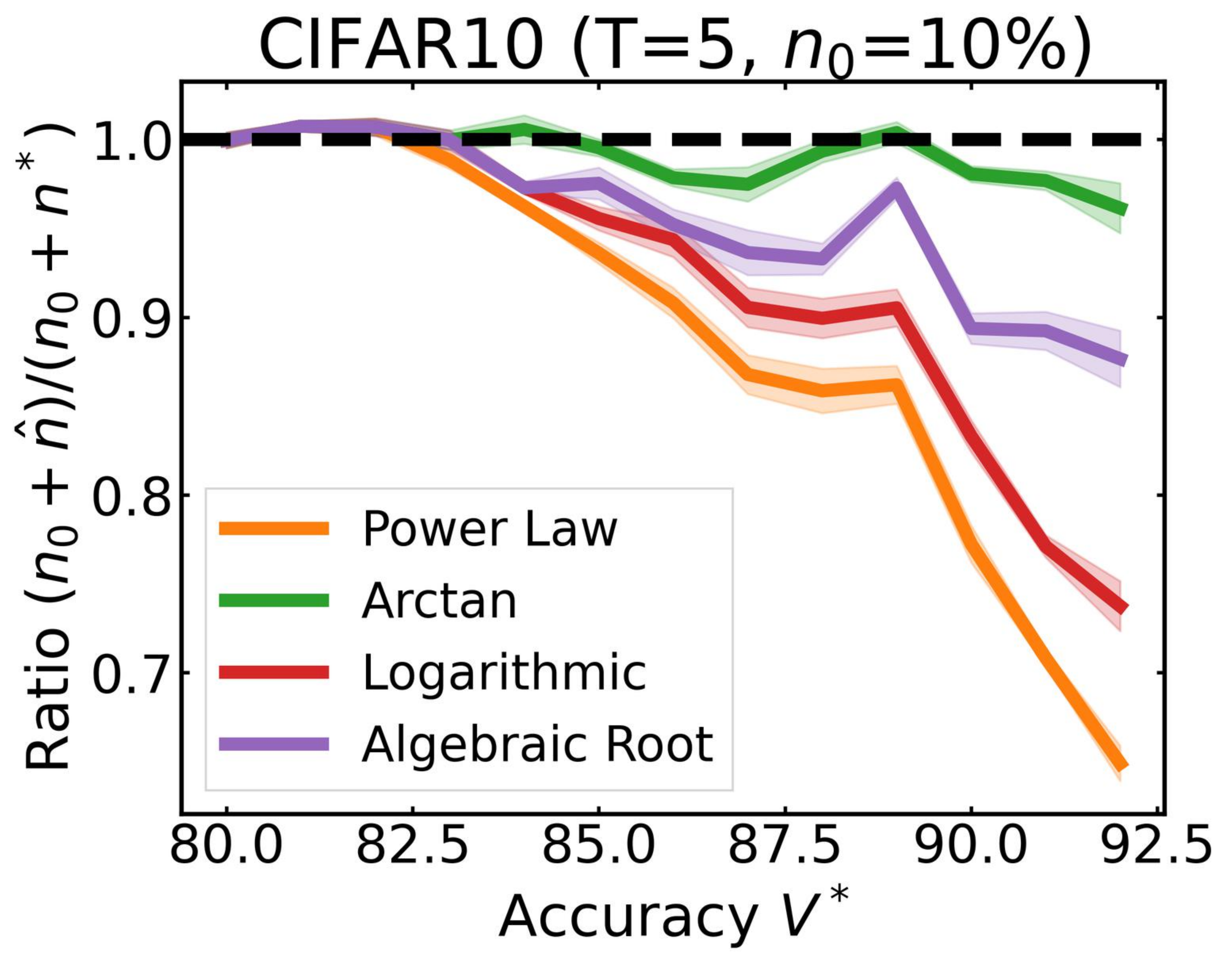} \end{minipage}
\begin{minipage}{0.16\linewidth}\includegraphics[width=1\textwidth]{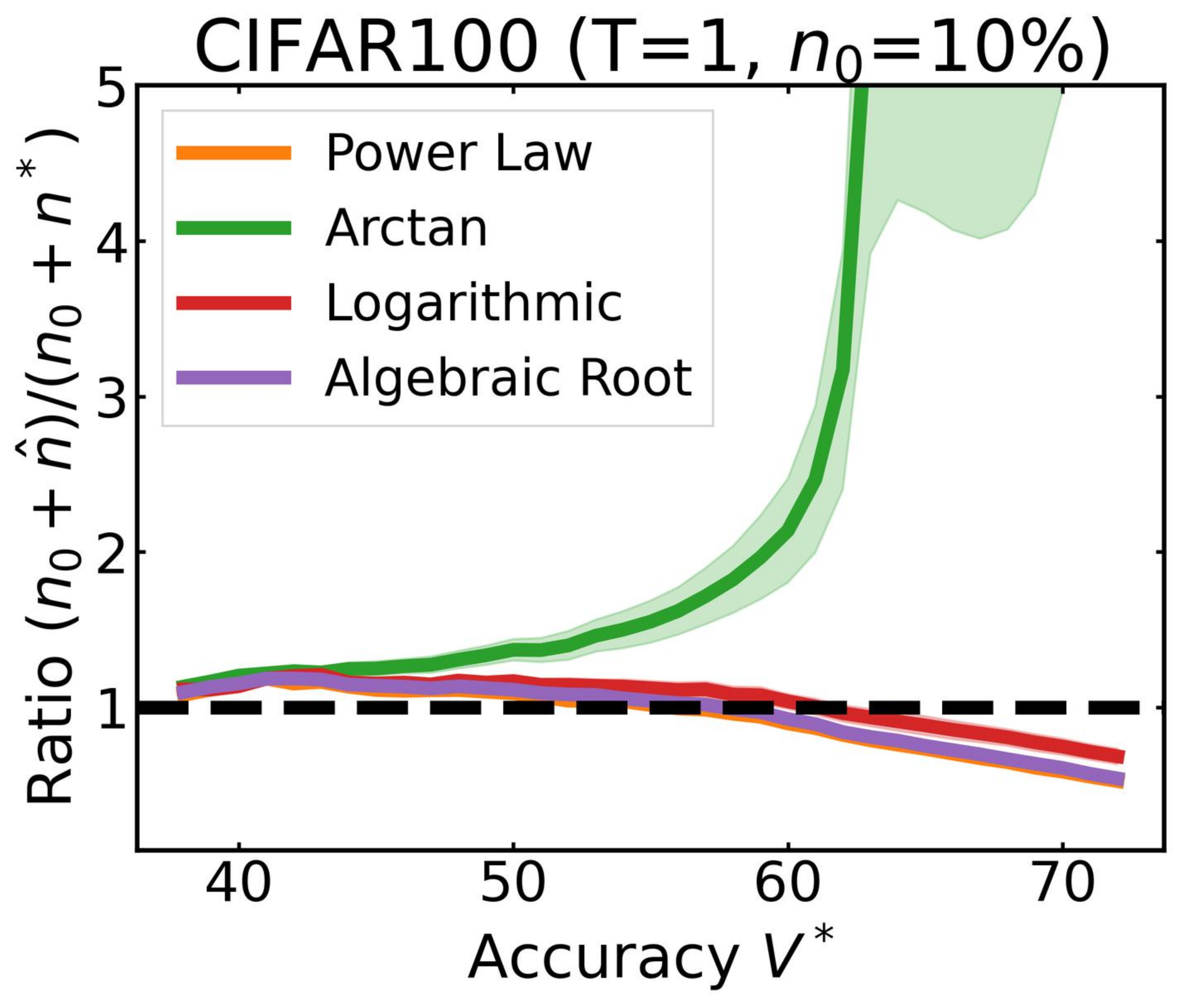} \end{minipage}
\begin{minipage}{0.16\linewidth}\includegraphics[width=1\textwidth]{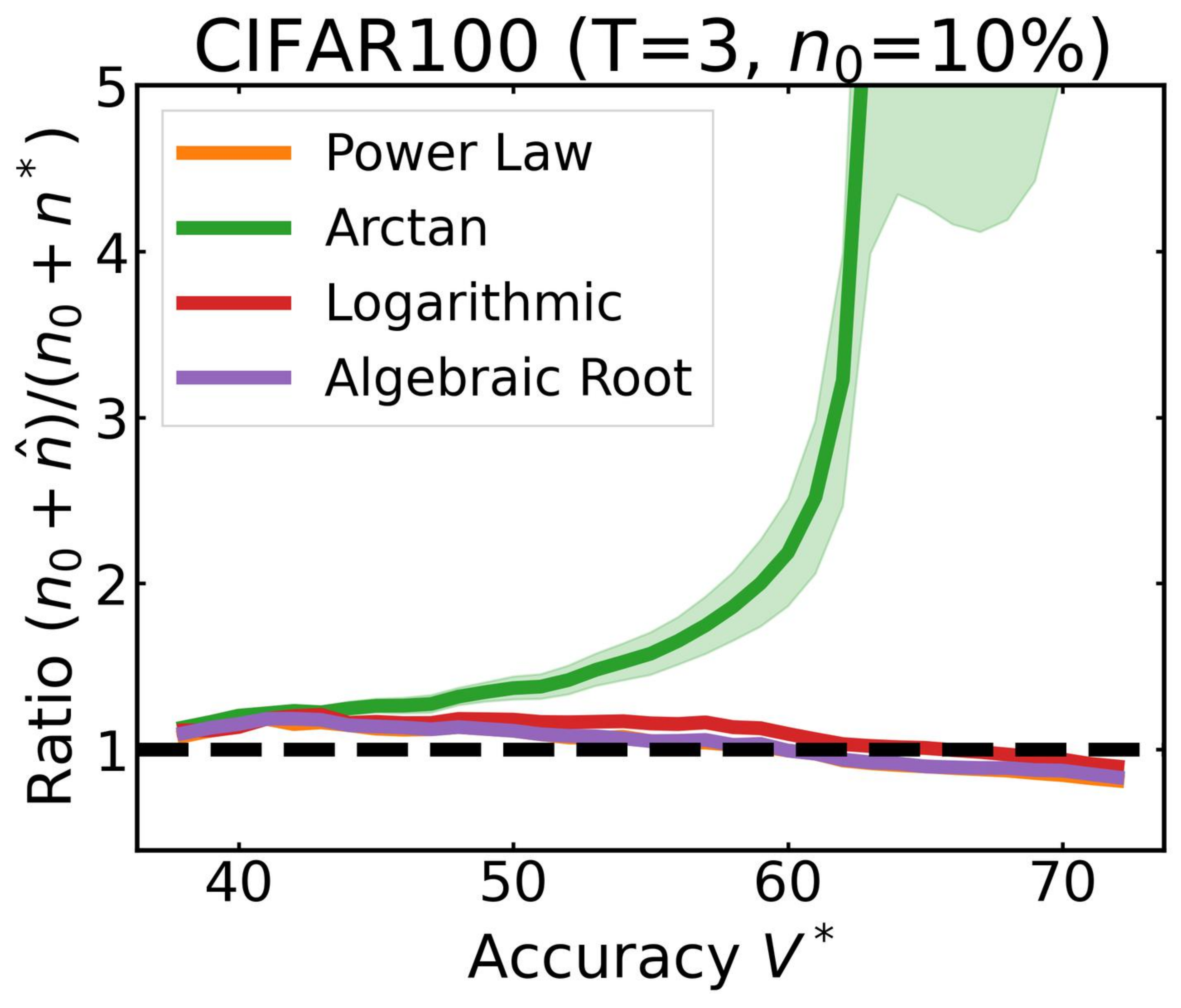} \end{minipage}
\begin{minipage}{0.16\linewidth}\includegraphics[width=1\textwidth]{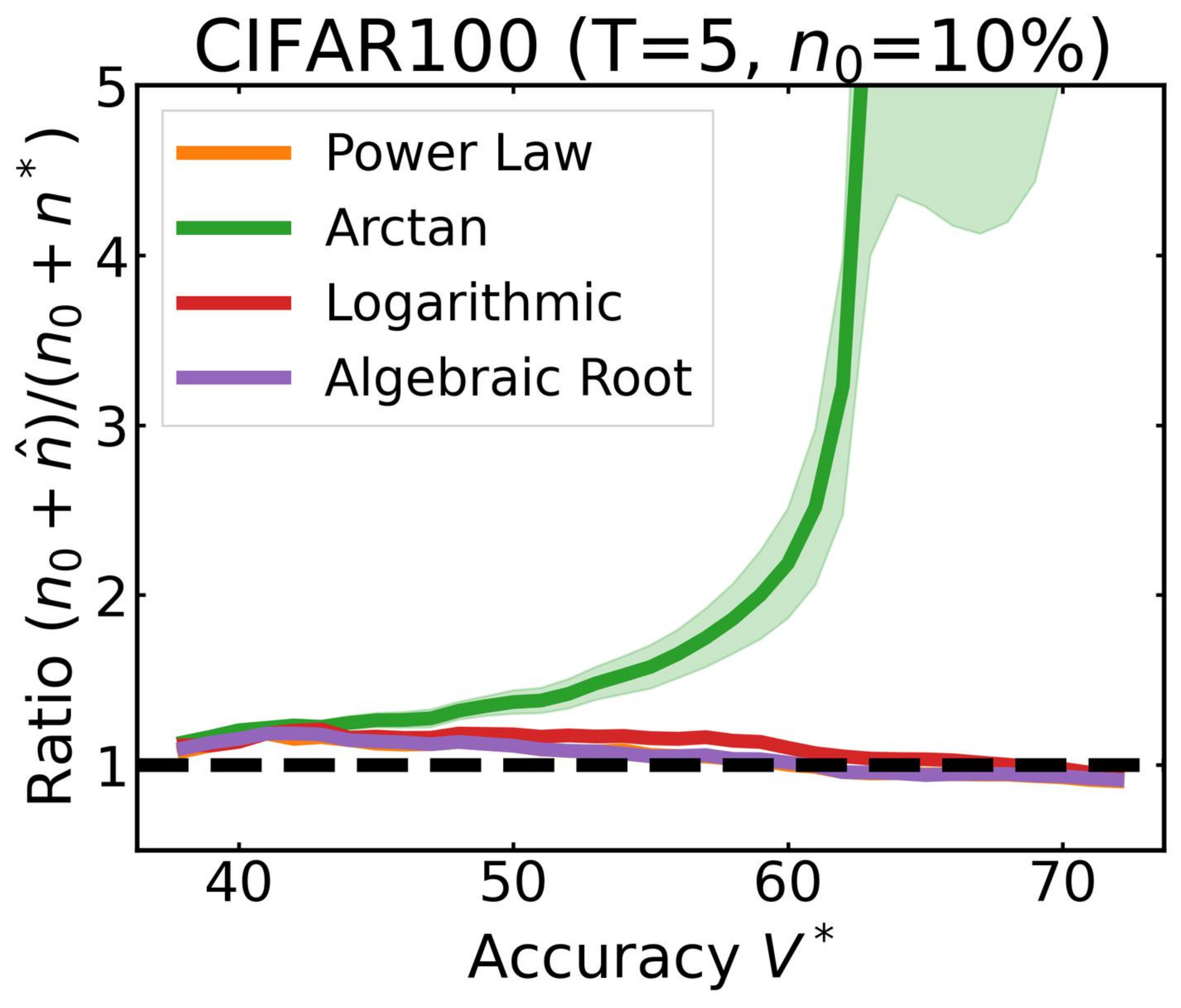} \end{minipage}
\begin{minipage}{0.16\linewidth}\includegraphics[width=1\textwidth]{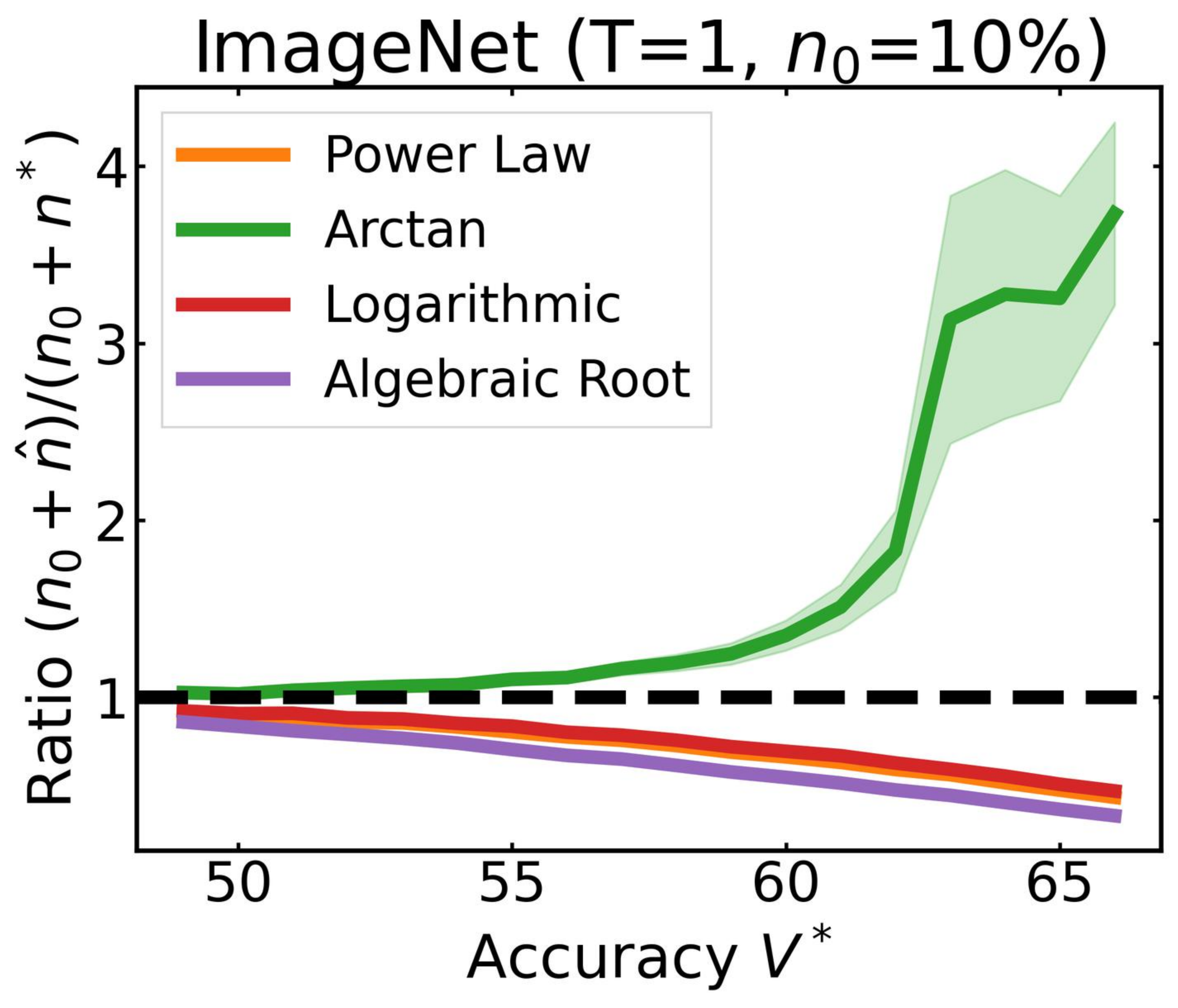} \end{minipage}
\begin{minipage}{0.16\linewidth}\includegraphics[width=1\textwidth]{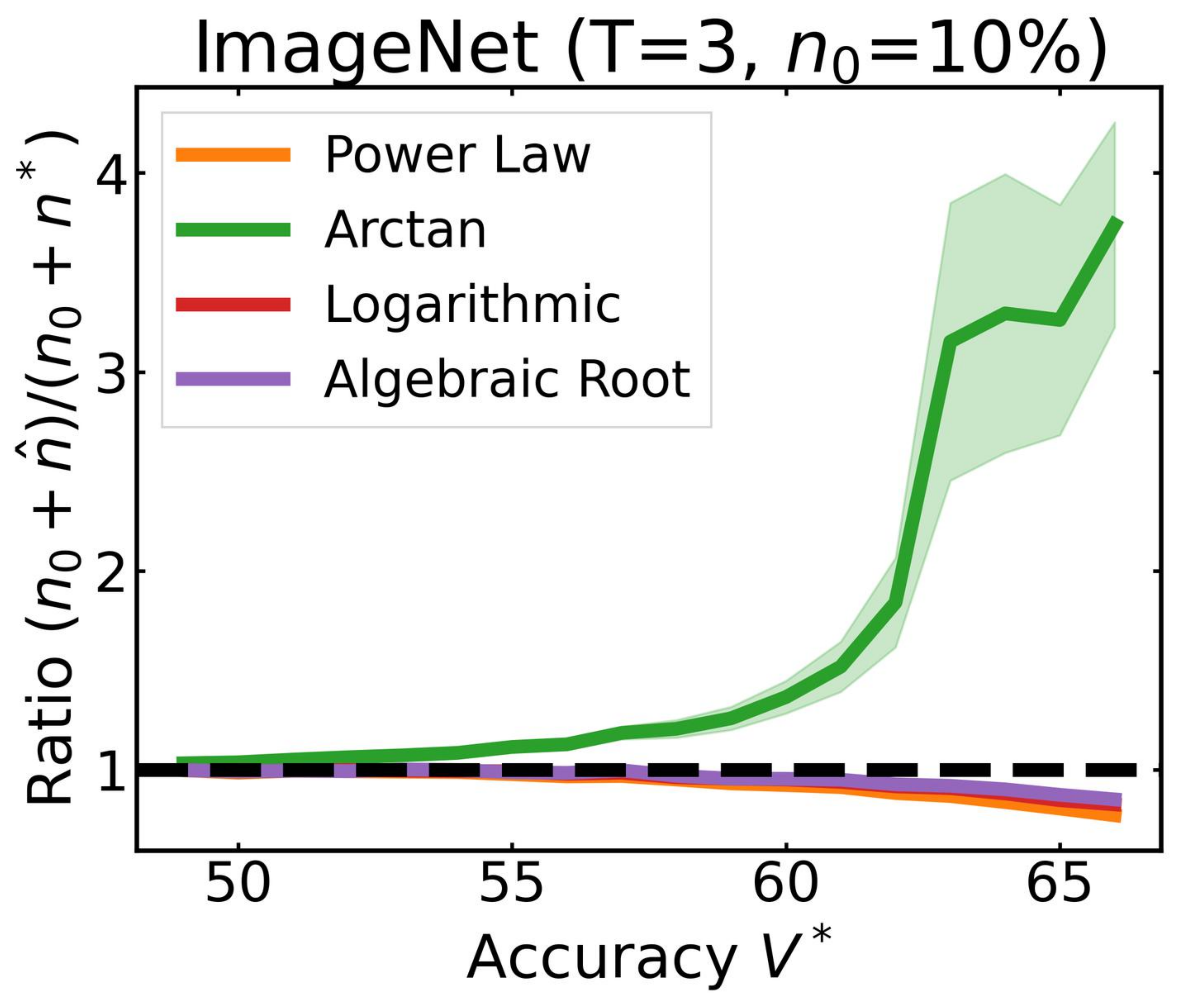} \end{minipage}
\begin{minipage}{0.16\linewidth}\includegraphics[width=1\textwidth]{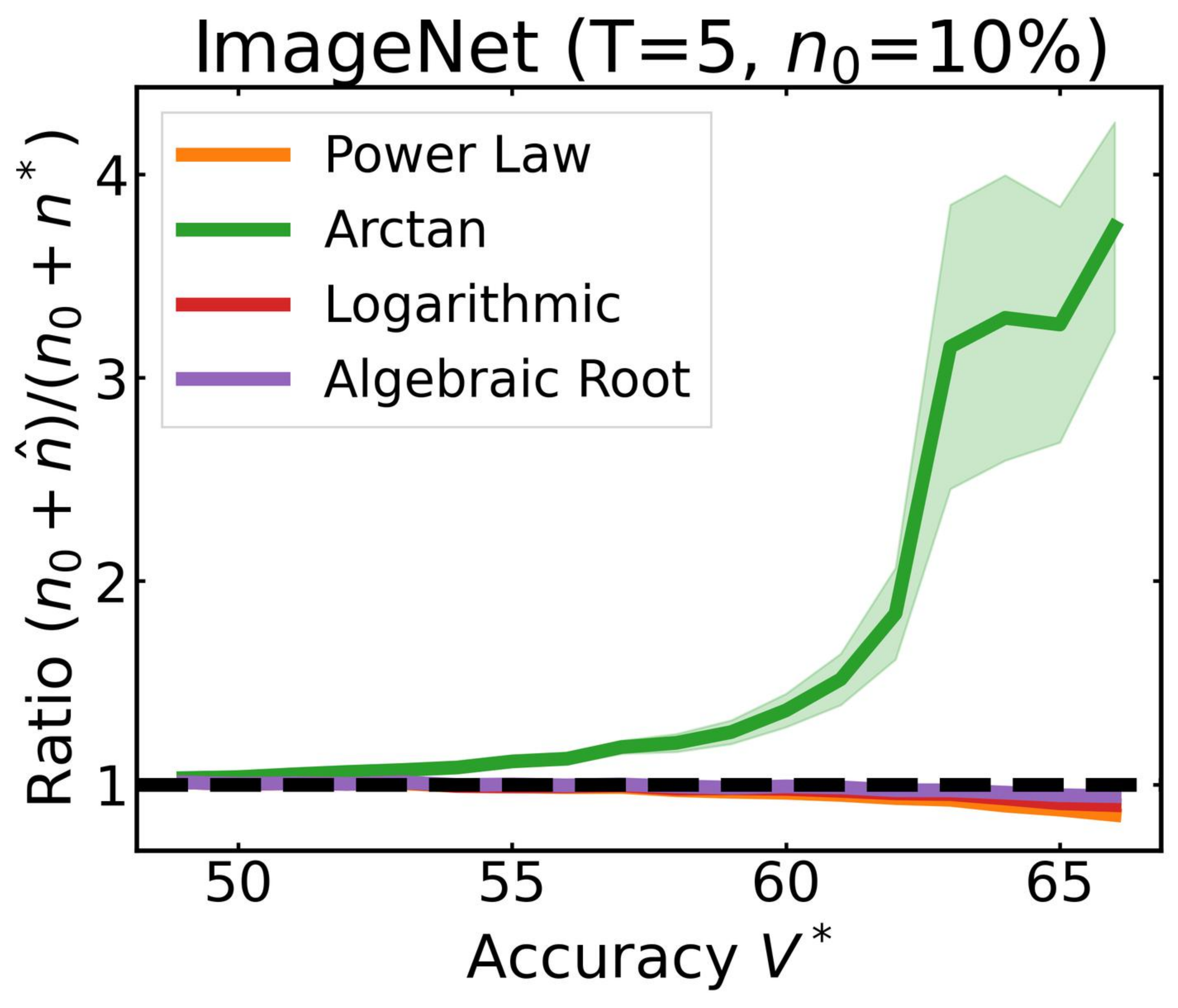} \end{minipage}
\begin{minipage}{0.16\linewidth}\includegraphics[width=1\textwidth]{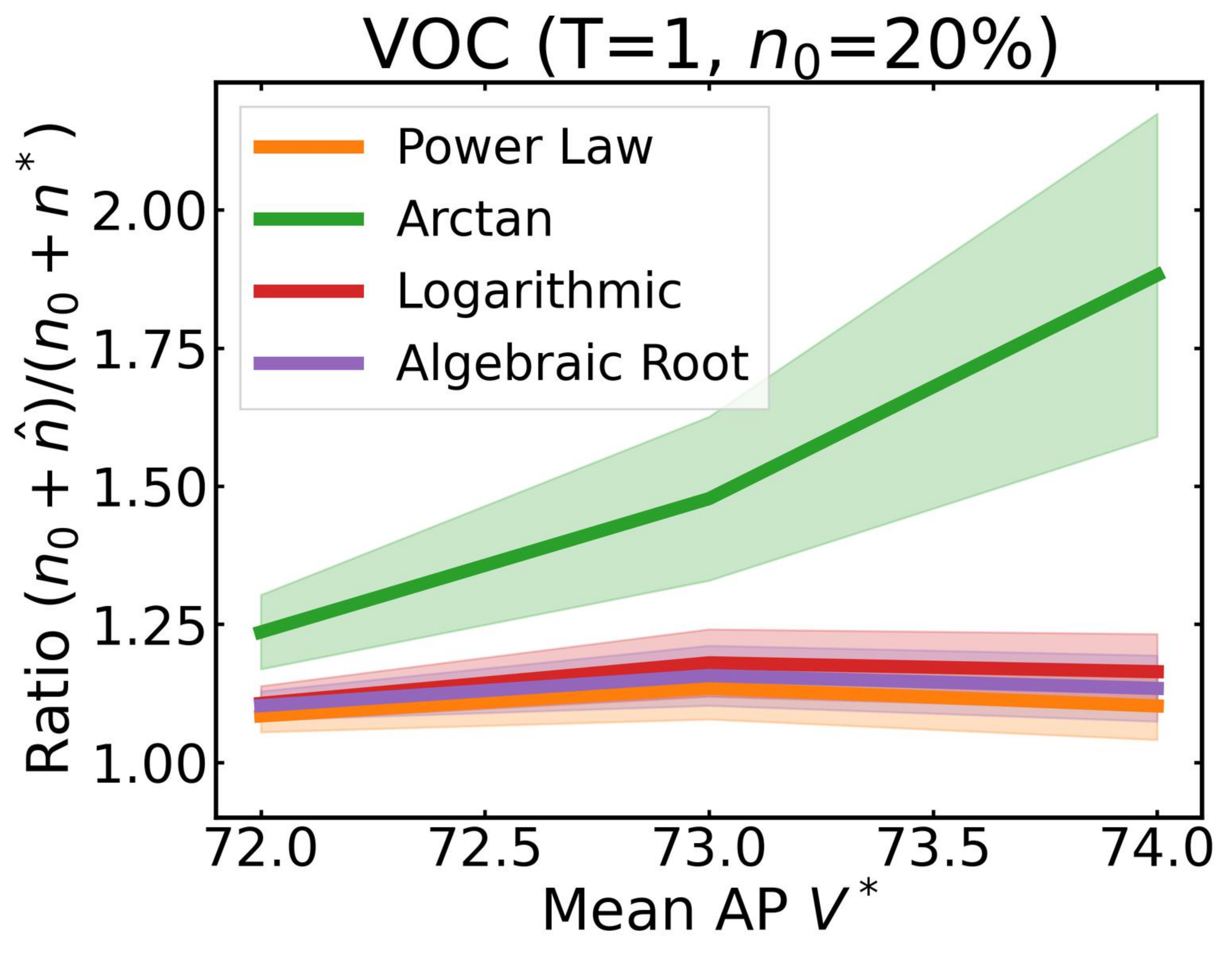} \end{minipage}
\begin{minipage}{0.16\linewidth}\includegraphics[width=1\textwidth]{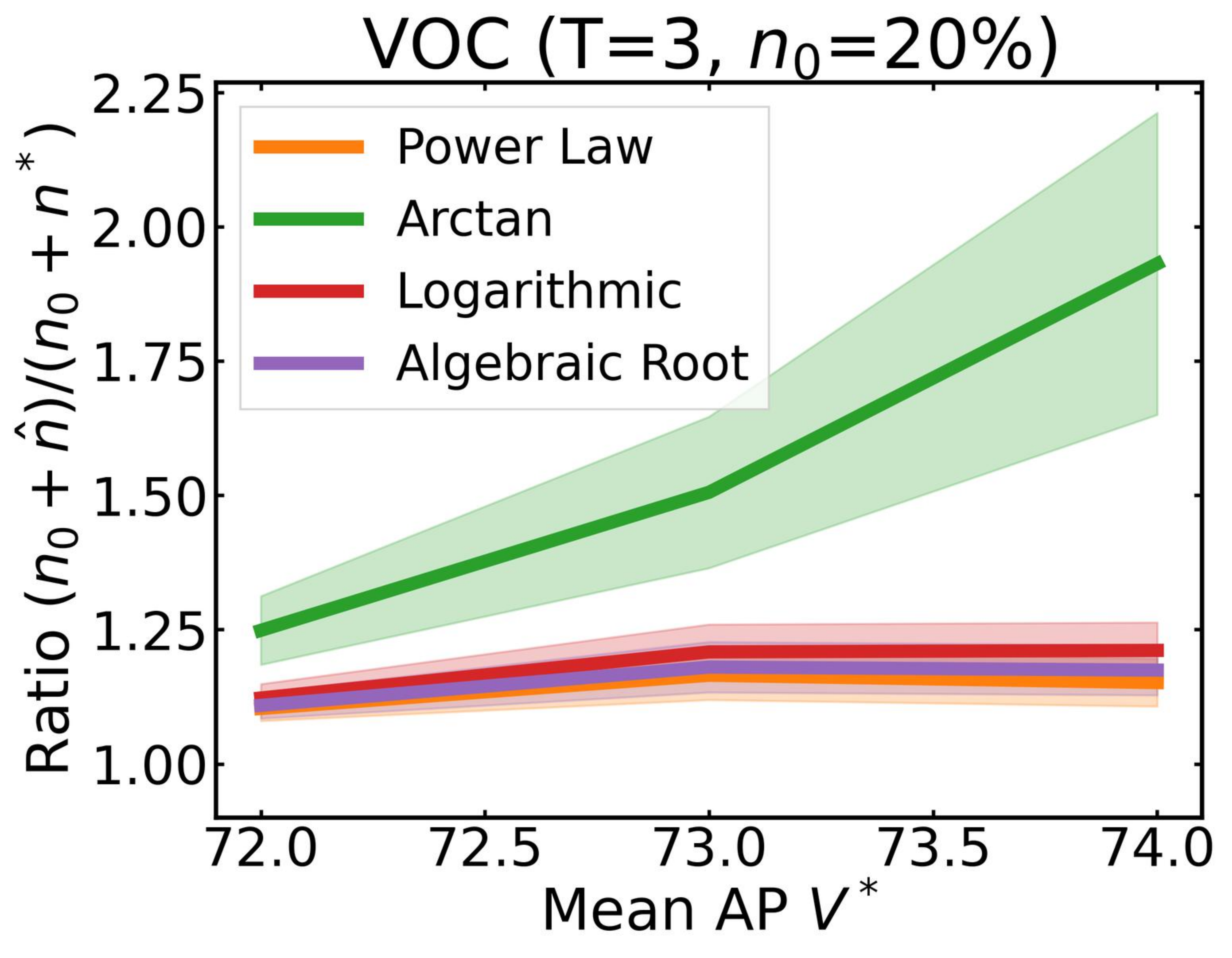} \end{minipage}
\begin{minipage}{0.16\linewidth}\includegraphics[width=1\textwidth]{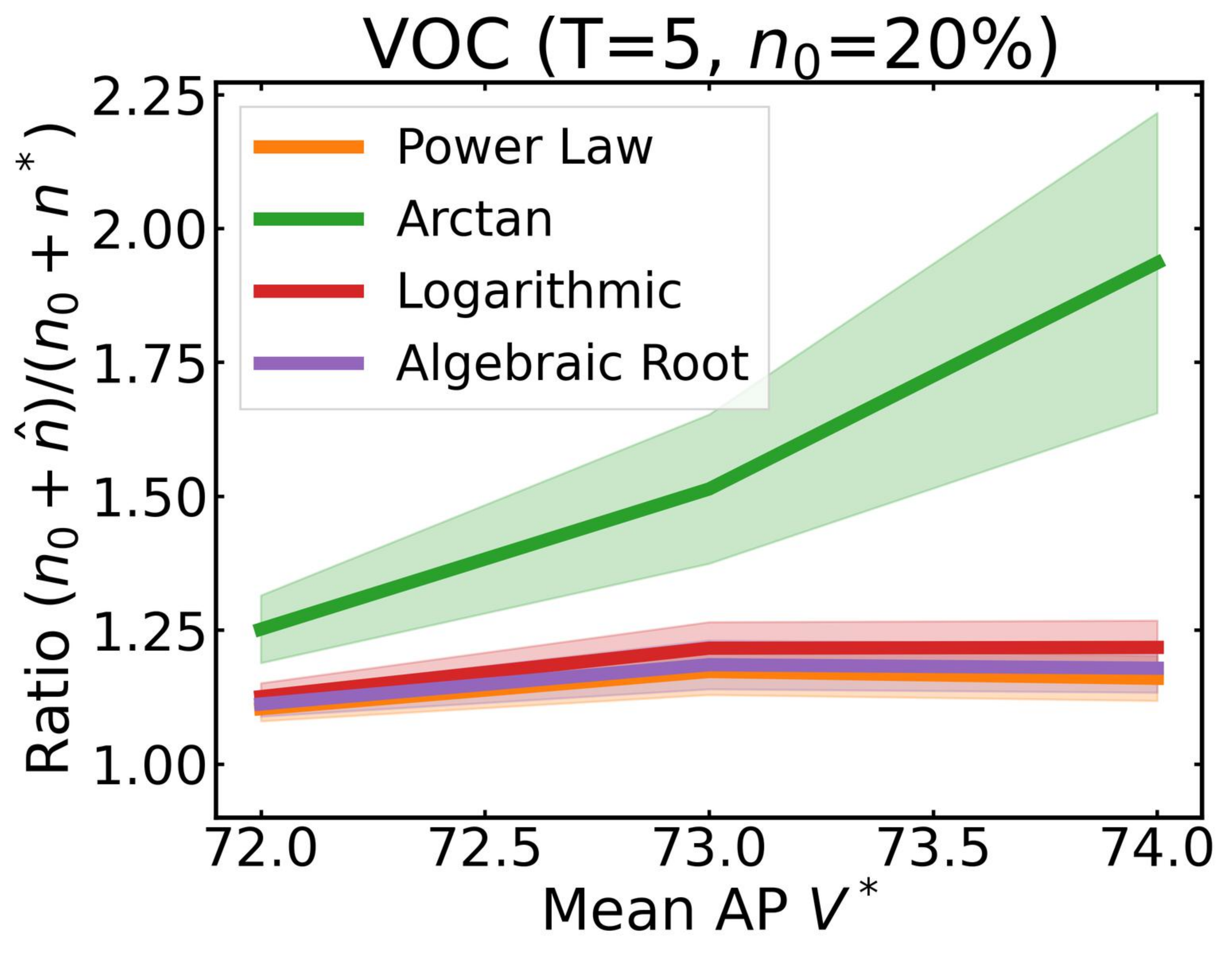} \end{minipage}
\begin{minipage}{0.16\linewidth}\includegraphics[width=1\textwidth]{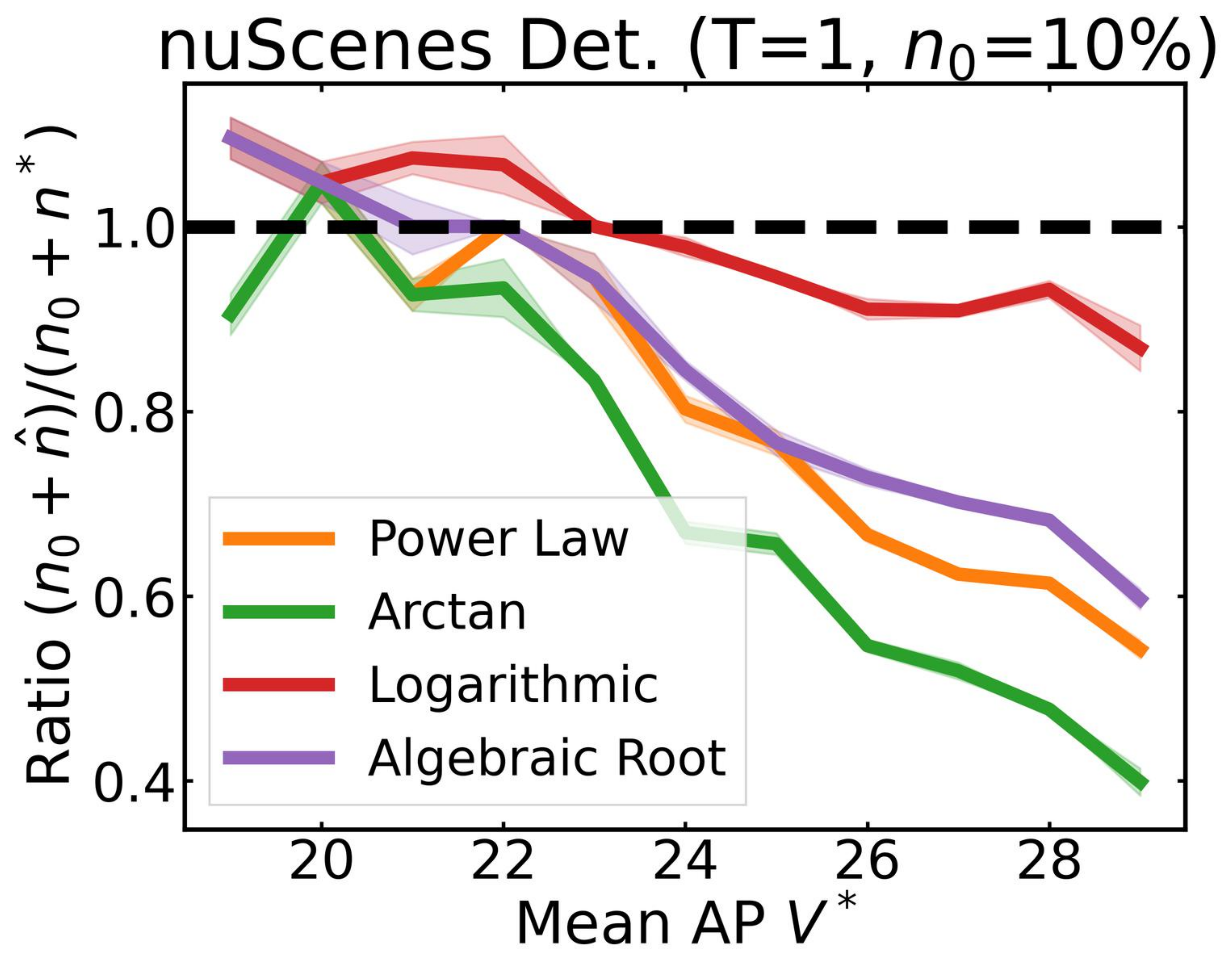} \end{minipage}
\begin{minipage}{0.16\linewidth}\includegraphics[width=1\textwidth]{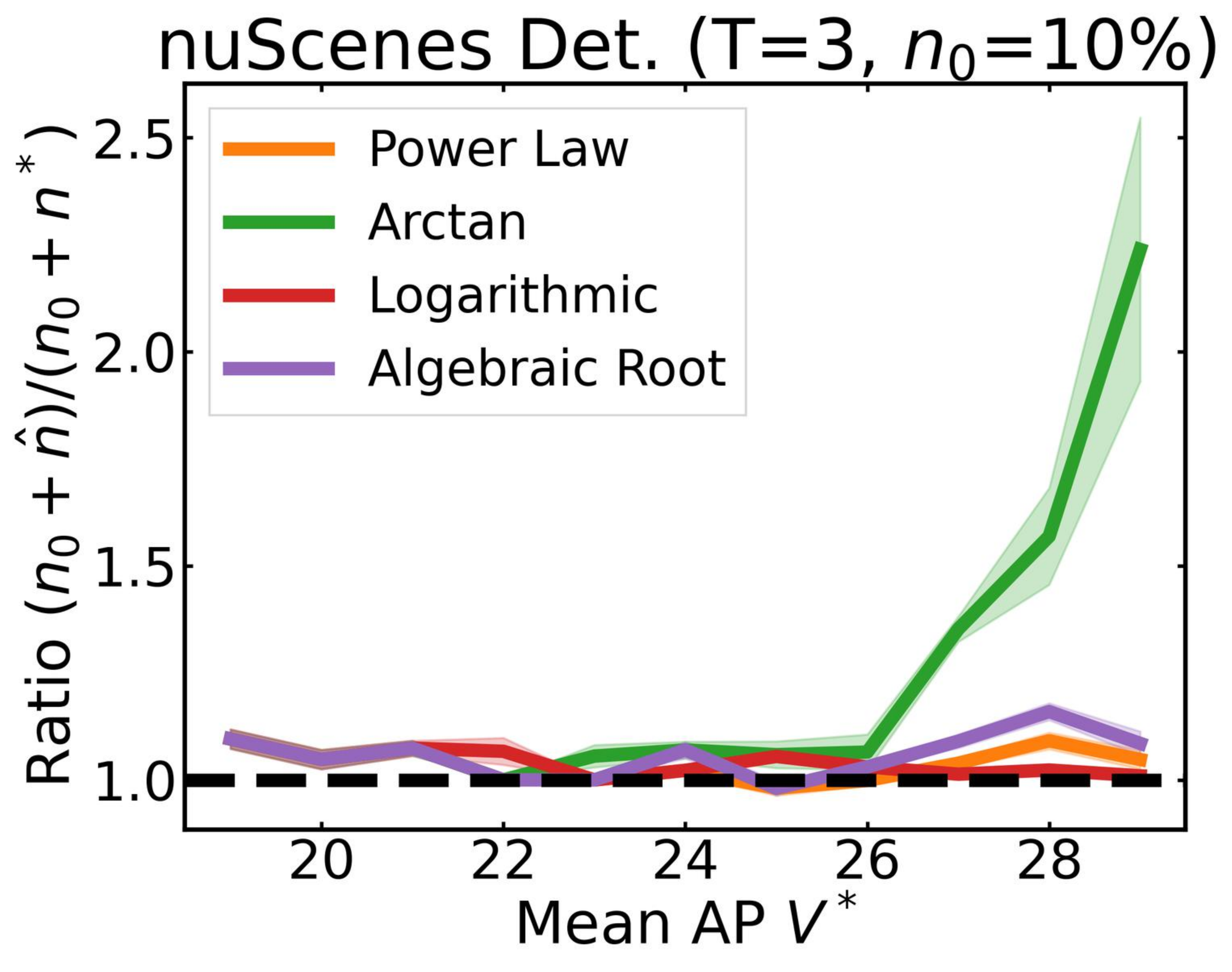} \end{minipage}
\begin{minipage}{0.16\linewidth}\includegraphics[width=1\textwidth]{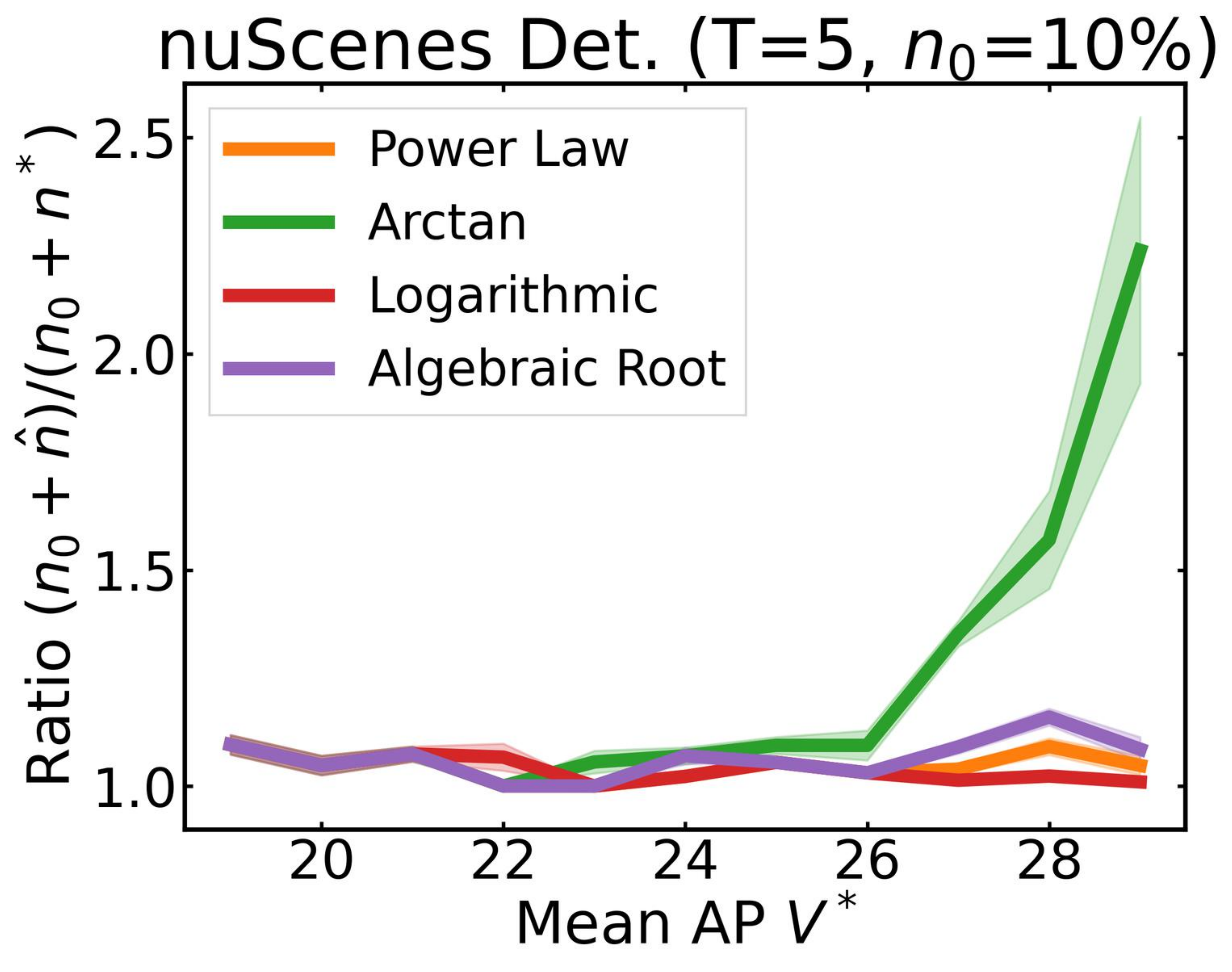} \end{minipage}
\begin{minipage}{0.16\linewidth}\includegraphics[width=1\textwidth]{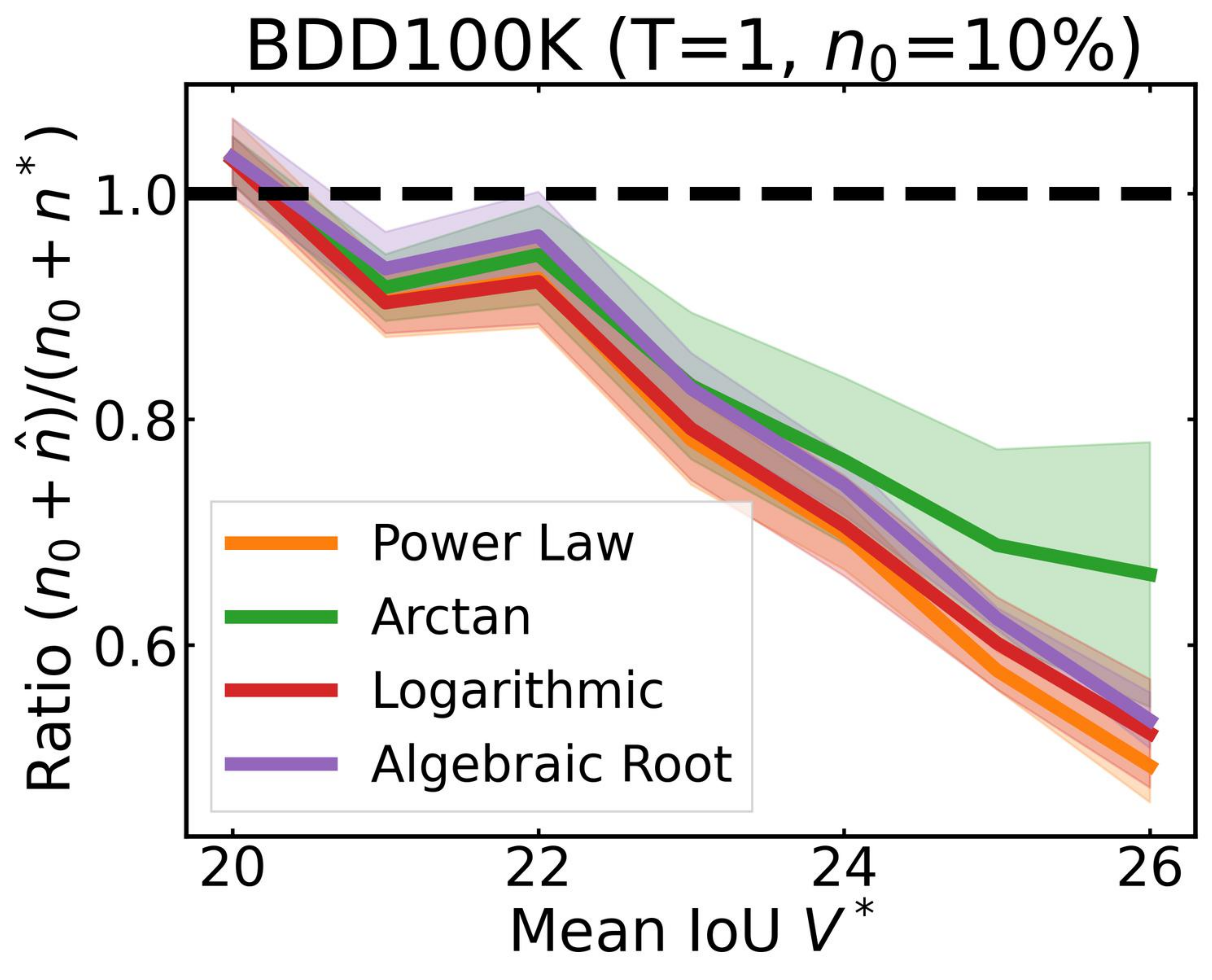} \end{minipage}
\begin{minipage}{0.16\linewidth}\includegraphics[width=1\textwidth]{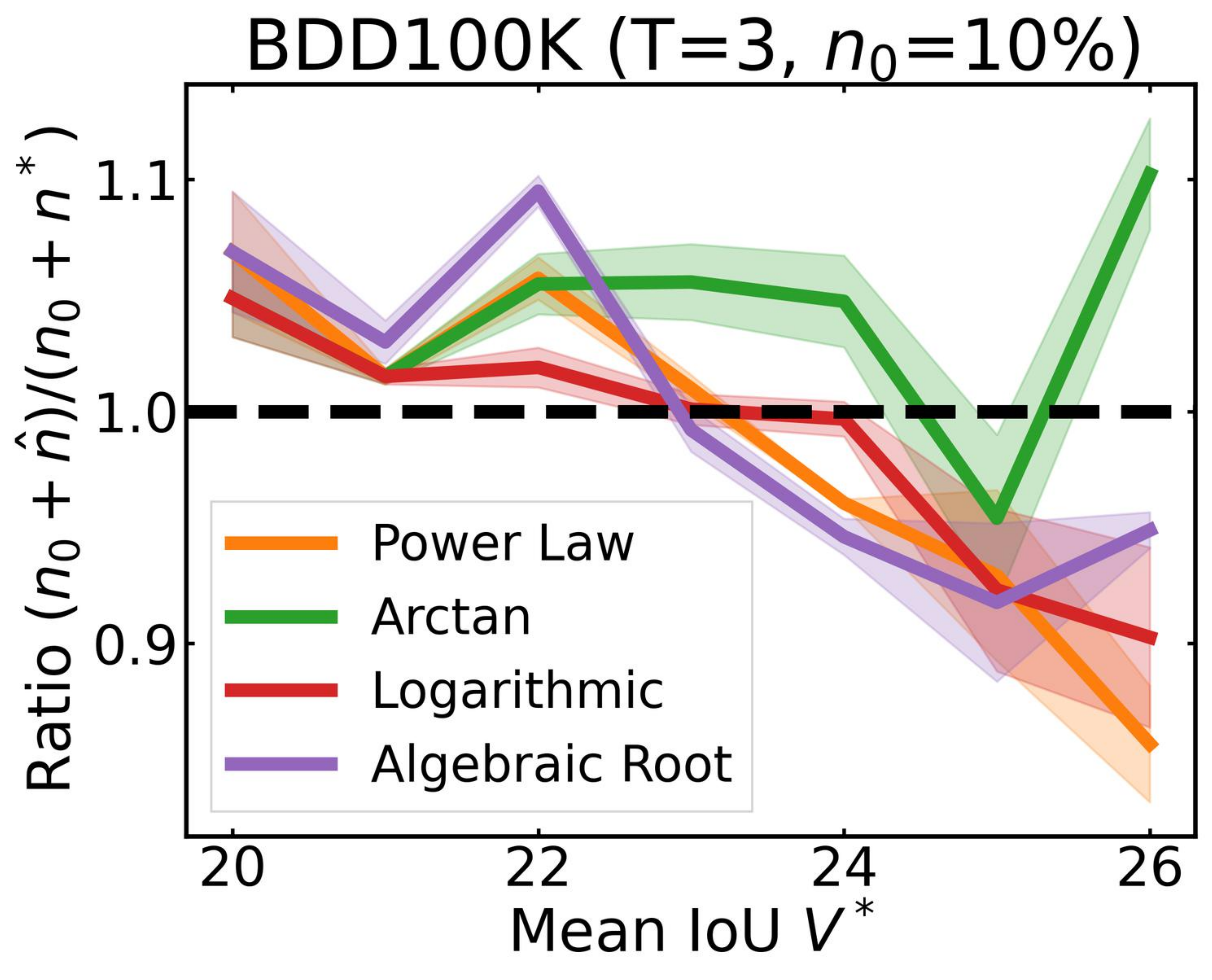} \end{minipage}
\begin{minipage}{0.16\linewidth}\includegraphics[width=1\textwidth]{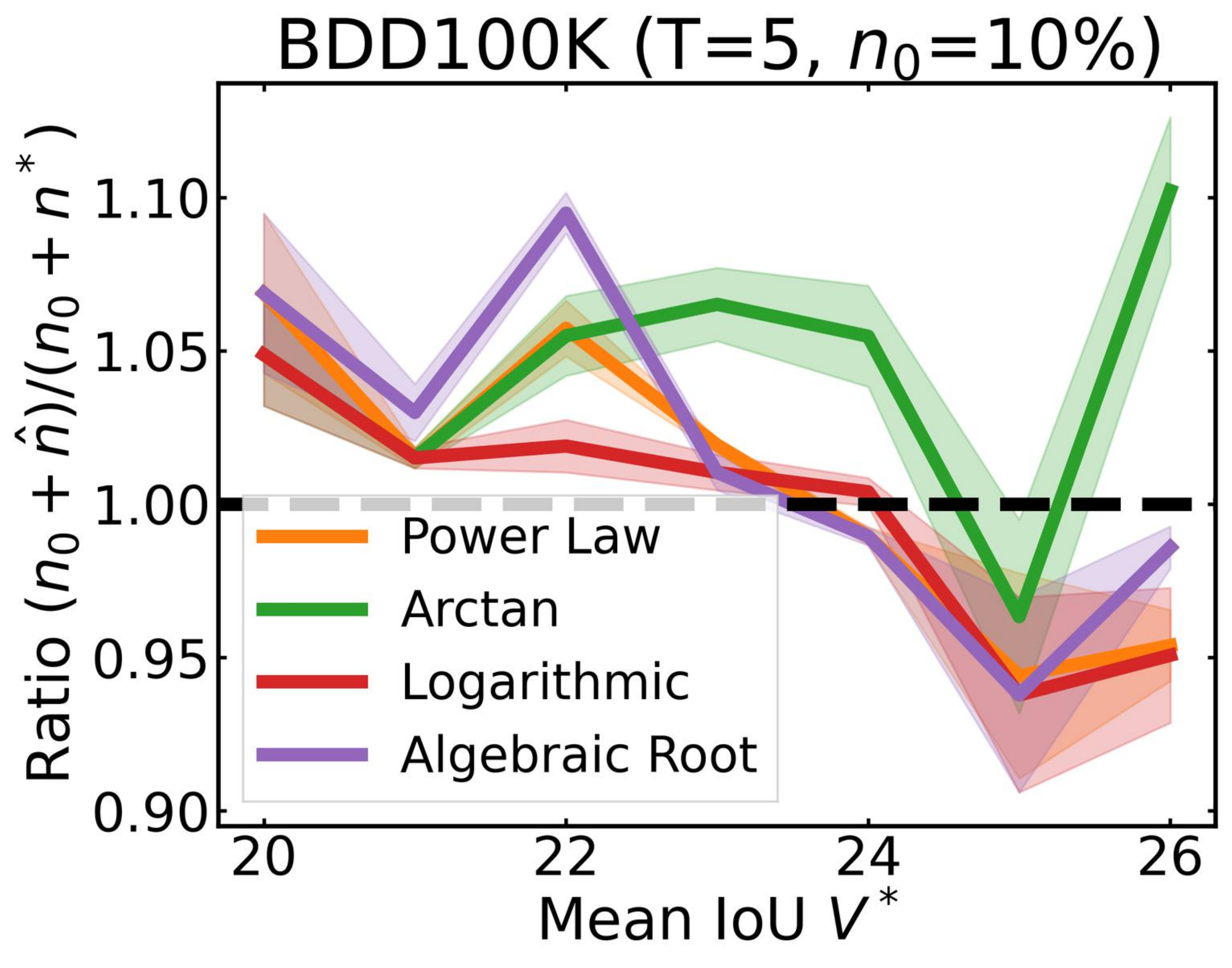} \end{minipage}
\begin{minipage}{0.16\linewidth}\includegraphics[width=1\textwidth]{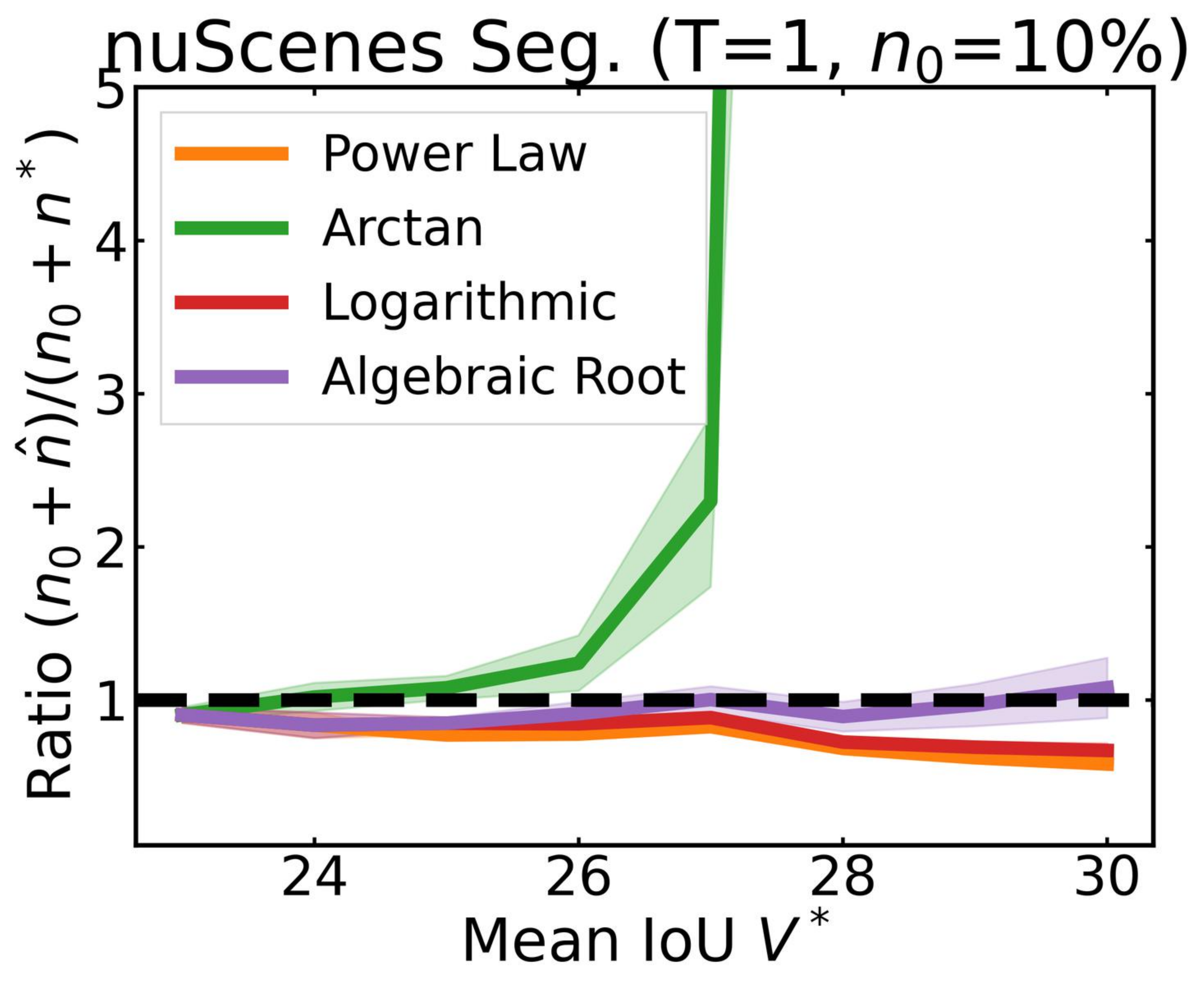} \end{minipage}
\begin{minipage}{0.16\linewidth}\includegraphics[width=1\textwidth]{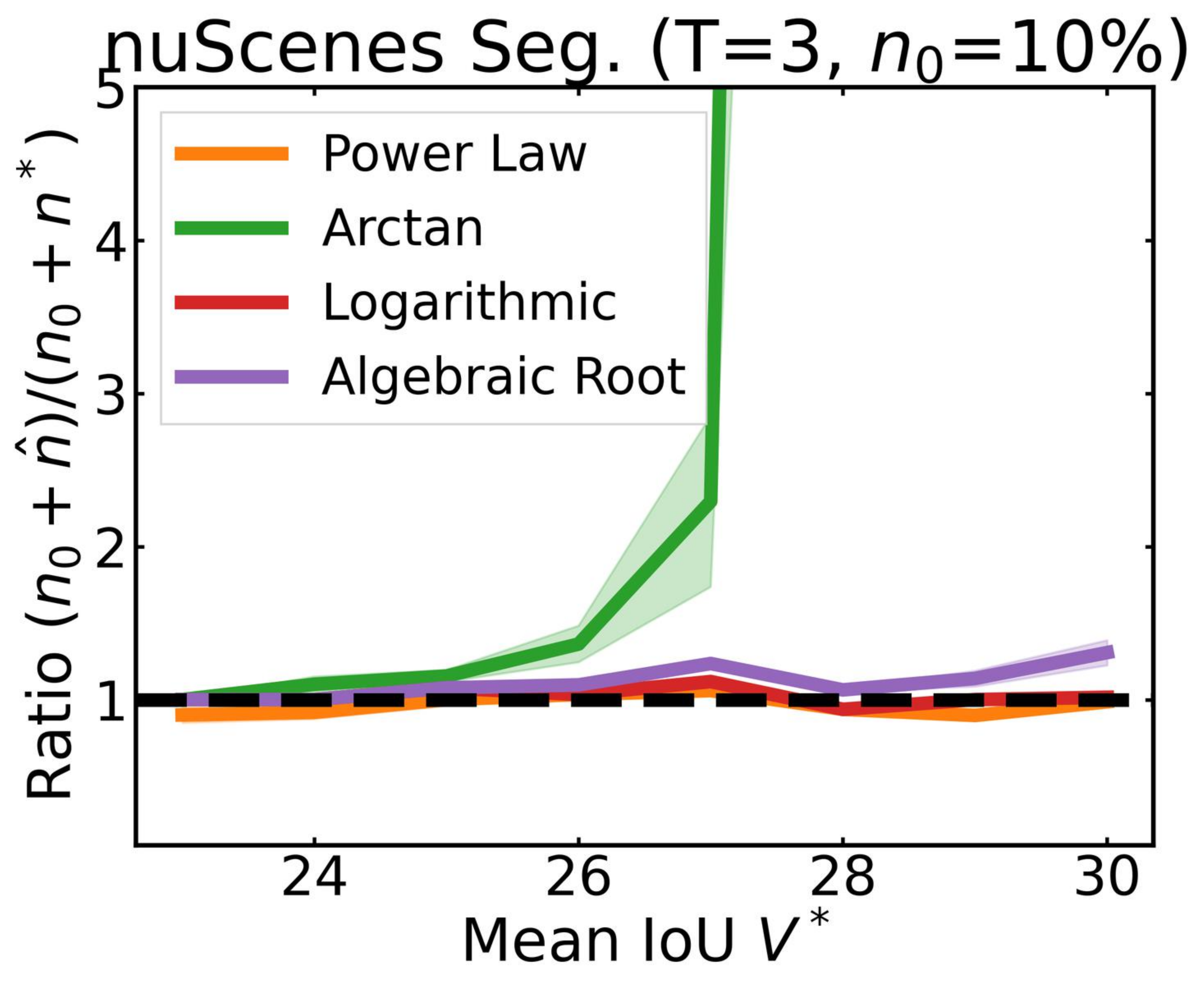} \end{minipage}
\begin{minipage}{0.16\linewidth}\includegraphics[width=1\textwidth]{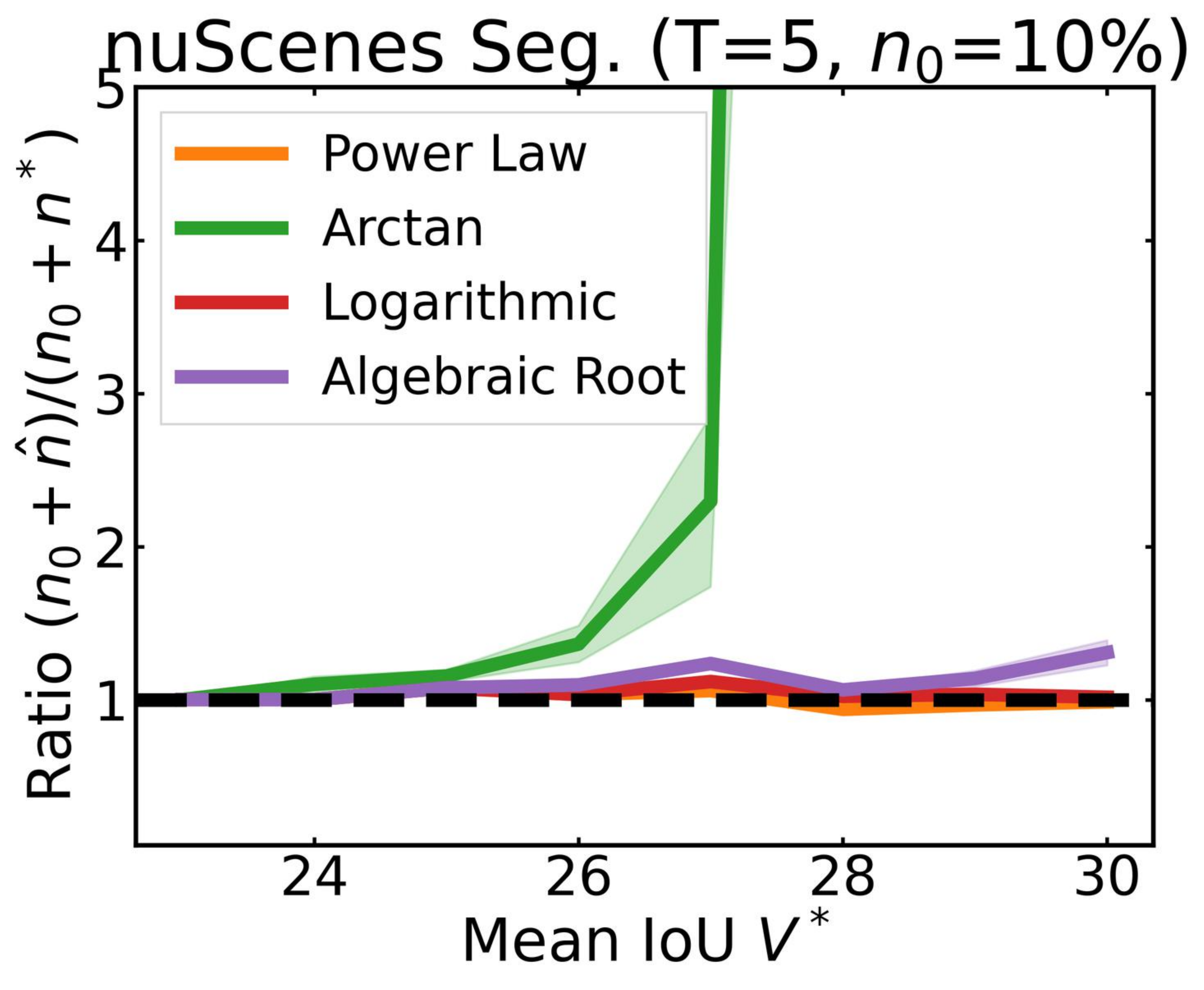} \end{minipage}
\begin{minipage}{0.48\linewidth}
\caption{\label{fig:simulation_all} The ratio of the amount of data collected versus the minimum data needed (y-axis) for different target $V^*$ (x-axis) in simulations initializing with $n_0 = 10\%$ of the data set ($n_0 = 20\%$ for VOC). For each data set, we show simulations for $T=1, 3, 5$ maximum rounds. The dashed black line corresponds to collecting the least amount of data needed to reach $V^*$.
}
\end{minipage}
\end{center}
\vspace{-11mm}
\end{figure*}

\noindent\textbf{Simulation.} 
We simulate data collection for each of the different regression functions by sweeping a range of targets $V^*$ when $n_0$ and $T$ are given.
Figure~\ref{fig:simulation_all} reports the ratio of the final data collected by each function versus the minimum data required according to the ground truth score, \ie $(n_0 + \hn)/(n_0 + n^*)$ where $n^*$ is the smallest value satisfying $v(n_0 + n^*) = V^*$. The value of $n^*$ is easy to find since $v(n)$ is a piecewise linear monotonically increasing function.

In evaluating how each regression function collects data, there are two scenarios to consider. 
If the ratio is less than one, the function is described as an optimistic predictor of the score that under-estimates how much data will be needed. 
A ratio less than one means that using this regression function, we will not collect enough data to meet $V^*$ within $T$ rounds, thereby failing to solve the problem. 
On the other hand if the ratio is greater than one, the function is a pessimistic predictor that over-estimates how much data will be needed.
An ideal data collection policy will achieve the smallest ratio greater than one.
Our experiments show that in general, the Arctan function is the most pessimistic and often achieves the largest ratios by a large margin.

We first validate the third challenge from Section~\ref{sec:setup_regression} and remark that low regression error does not necessarily translate to better data collection.
On CIFAR100, ImageNet, and VOC, using Arctan may lead to collecting up to five times more data than is actually needed; with BEV segmentation on nuScenes, it may lead to over 10 times more. Recall from Figure~\ref{fig:regression_summary} that on ImageNet, we require approximately $900,000$ images to reach a target $V^*=67\%$.
Using Arctan when initialized with $n_0=10\%$ of the data would result in collecting approximately $4.5$ million images in the first round alone, whereas all of the other regression functions achieve a ratio approximately equal to $1$. 
Although Table~\ref{tab:regression_RMSE_tasks} showed that Arctan achieved the lowest RMSE ($3.19$) of all functions in regression, using it to estimate data requirements would lead to an unnecessarily expensive data collection procedure.
This reveals that simply analyzing regression error is insufficient when determining good data collection policies, necessitating our simulation approach.


For most regression functions, collecting enough data requires multiple rounds. When $T=1$, the Power Law, Logarithmic, and Algebraic Root functions under-estimate the data requirement for all data sets and tasks except for VOC. 
However when $T=5$, for every data set except for CIFAR10, all of the functions yield ratios greater than $0.9$ over the entire range of $V^*$. That is, we can consistently reach at least $90\%$ of the data needed with any of the functions.

Ultimately, even with $T=5$, these estimators can still under-estimate the requirement when $V^*$ is large (\eg on ImageNet, the Power Law, Logarithmic, and Algebraic Root functions achieve ratios less than $1$ for $V^* \geq 62\%$). From an operational perspective, although these methods do not incur large costs, they also fail to solve the problem. In the next section, we show simple techniques to correct these estimators and better guide data collection.

\noindent\textbf{Ablations.}
In the supplement, we perform ablations that evaluate regressions and simulations on different model depths and widths for CIFAR100.
We also consider alternate score functions such as collecting enough data to meet a target performance on a specific class using nuScenes. 
Finally, we explore estimating requirements when using active learning rather than random sampling for CIFAR100.
Our results indicate the same trends, further supporting the challenges towards estimating the data requirement. 

%% file: sections/analysis.tex
\section{Towards better estimates of data}


We previously showed that some optimistic estimators fail to collect enough data to meet $V^*$ whereas other pessimistic estimators lead to collecting far more data than required. Here, we first introduce a correction factor, which is a bias term that addresses the problem of under-estimating data requirements. 
We then show how analyzing both the optimistic and pessimistic regression functions considered in this paper can lead to a collection of estimates that often bound the true data requirement.

\begin{table*}[t]
\begin{minipage}{0.76\linewidth}
    \centering
    \scriptsize
\begin{tabular}{clllcccccccc}
\toprule
       & Data set &   $n_0$ &  $T$ & \multicolumn{2}{c}{Power Law}  & \multicolumn{2}{c}{Arctan}  & \multicolumn{2}{c}{Logarithmic}  & \multicolumn{2}{c}{Algebraic Root}  \\
              &   &   &   & Without & With & Without & With & Without & With & Without & With \\
\midrule
            \multirow{6}{*}{\rotatebox[origin=c]{90}{Classification}}        
               & CIFAR100 &  10\% &  $1$ &  $0.53$ &        $0.91$ &   $\fir{1.13}$ &    $ 1.36$ &   $0.68$ &         $\und{1.54}$ &   $0.54$ &    $ 0.82$ \\
               & CIFAR100 &  10\% &  $3$ &  $0.81$ &  $\und{1.09}$ &         $1.13$ &    $ 1.19$ &   $ 0.9$ &   $\fir{\und{1.08}}$ &   $0.83$ &    $ 0.94$ \\
               & CIFAR100 &  10\% &  $5$ &  $ 0.9$ &  $\und{1.03}$ &         $1.13$ &    $ 1.19$ &   $0.94$ &         $\und{1.11}$ &   $0.91$ &    $\fir{\und{1.01}}$ \\ 
               \cmidrule{2-12}
               & ImageNet &  10\% &  $1$ &  $0.43$ &  $\und{1.16}$ &   $\fir{1.02}$ &    $ 1.35$ &   $0.47$ &    $\und{1.28}$ &   $0.33$ &    $  0.5$ \\
              &  ImageNet &  10\% &  $3$ &  $0.77$ &  $\und{1.10}$ &   $\fir{1.03}$ &    $ 1.08$ &   $0.83$ &    $\und{1.06}$ &   $0.85$ &    $\fir{\und{1.03}}$ \\
              &  ImageNet &  10\% &  $5$ &  $0.85$ &  $\und{1.07}$ &   $\fir{1.03}$ &    $ 1.08$ &   $ 0.9$ &    $\und{1.06}$ &   $0.94$ &    $\fir{\und{1.03}}$ \\ 
               \midrule
              \multirow{6}{*}{\rotatebox[origin=c]{90}{Detection}}
                &    VOC &  20\% &  $1$ &  $\fir{1.08}$ &  $ 6.42$ &   $1.24$ &    $ 5.05$ &   $1.11$ &    $  7.4$ &   $ 1.1$ &    $ 6.03$ \\
                &   VOC &  20\% &  $3$ &   $\fir{1.1}$ &  $ 2.75$ &   $1.25$ &    $ 1.64$ &   $1.12$ &    $ 2.23$ &   $1.11$ &    $ 1.54$ \\
                &  VOC &  20\% &  $5$ &   $\fir{1.1}$ &  $ 2.03$ &   $1.25$ &    $ 1.64$ &   $1.13$ &    $ 2.23$ &   $1.11$ &    $ 1.54$ \\ 
                \cmidrule{2-12}
  & nuScenes &  10\% &  $1$ & $0.56$ &   $\und{2.9}$ &   $0.39$ &    $0.51$ &  $0.83$ & $\und{32.45}$ &   $0.61$ &    $\und{2.9}$ \\
  & nuScenes &  10\% &  $3$ & $0.94$ &  $\und{1.05}$ &   $ 1.0$ &     $1.0$ &   $1.0$ &       $ 1.68$ &   $0.94$ &    $\und{1.07}$ \\
  & nuScenes &  10\% &  $5$ &  $1.0$ &        $1.09$ &   $ 1.0$ &     $1.0$ &   $1.0$ &       $ 1.68$ &   $ 1.0$ &    $1.07$ \\ 
                \midrule
         \multirow{6}{*}{\rotatebox[origin=c]{90}{Segmentation}}  
         &   BDD100K &  10\% &  $1$ &  $0.49$ &  $\und{2.45}$ &   $0.66$ &    $\fir{\und{1.79}}$ &   $0.52$ &    $\und{ 5.2}$ &   $0.53$ &    $\und{2.17}$ \\
         &   BDD100K &  10\% &  $3$ &  $0.86$ &  $\und{1.76}$ &   $0.95$ &    $\fir{\und{ 1.2}}$ &   $ 0.9$ &    $\und{1.58}$ &   $0.92$ &    $\und{1.19}$ \\
         &  BDD100K &  10\% &  $5$ &  $0.94$ &  $\und{1.48}$ &   $0.96$ &    $\fir{\und{ 1.2}}$ &   $0.94$ &    $\und{1.58}$ &   $0.94$ &    $\und{1.19}$ \\ 
                \cmidrule{2-12}
  & nuScenes &  10\% &  $1$ &  $0.58$ &  $\und{24.58}$ &         $0.9$ &    $  4.63$ &       $0.67$ &    $\und{27.46}$ &        $0.83$ &    $\und{51.12}$ \\
 &  nuScenes &  10\% &  $3$ &  $ 0.9$ &  $\und{ 1.42}$ &   $\fir{1.0}$ &    $  1.25$ &       $0.94$ &    $\und{ 1.31}$ &   $\fir{1.0}$ &    $ 1.25$ \\
  & nuScenes &  10\% &  $5$ &  $0.94$ &  $\und{ 1.07}$ &   $\fir{1.0}$ &    $  1.25$ &  $\fir{1.0}$ &    $\und{ 1.31}$ &   $\fir{1.0}$ &    $ 1.25$ \\ 
\bottomrule
\end{tabular}
\end{minipage} 
      \begin{minipage}{0.23\linewidth}
    \caption{\label{tab:tolerance_values}The minimum ratio $\frac{n_0 + \hn}{n_0 + n^*}$ for each regression function without (baseline) and with using $\tau$ when estimating data requirements. The best ratio (\ie smallest value greater than $1$) for each data set is bolded. Instances where using $\tau$ for a given regression function increased the ratio from below to above $1$ are underlined.
    Power Law, Logarithmic, and Algebraic Root improve for nearly every setting to obtain ratios above $1$. Furthermore, these functions achieve their best performance when $T=5$. 
    }
      \end{minipage}
\vspace{-3mm}
\end{table*}

\begin{figure*}[!t]
\begin{center}
\begin{minipage}{0.16\linewidth}\includegraphics[width=1\textwidth]{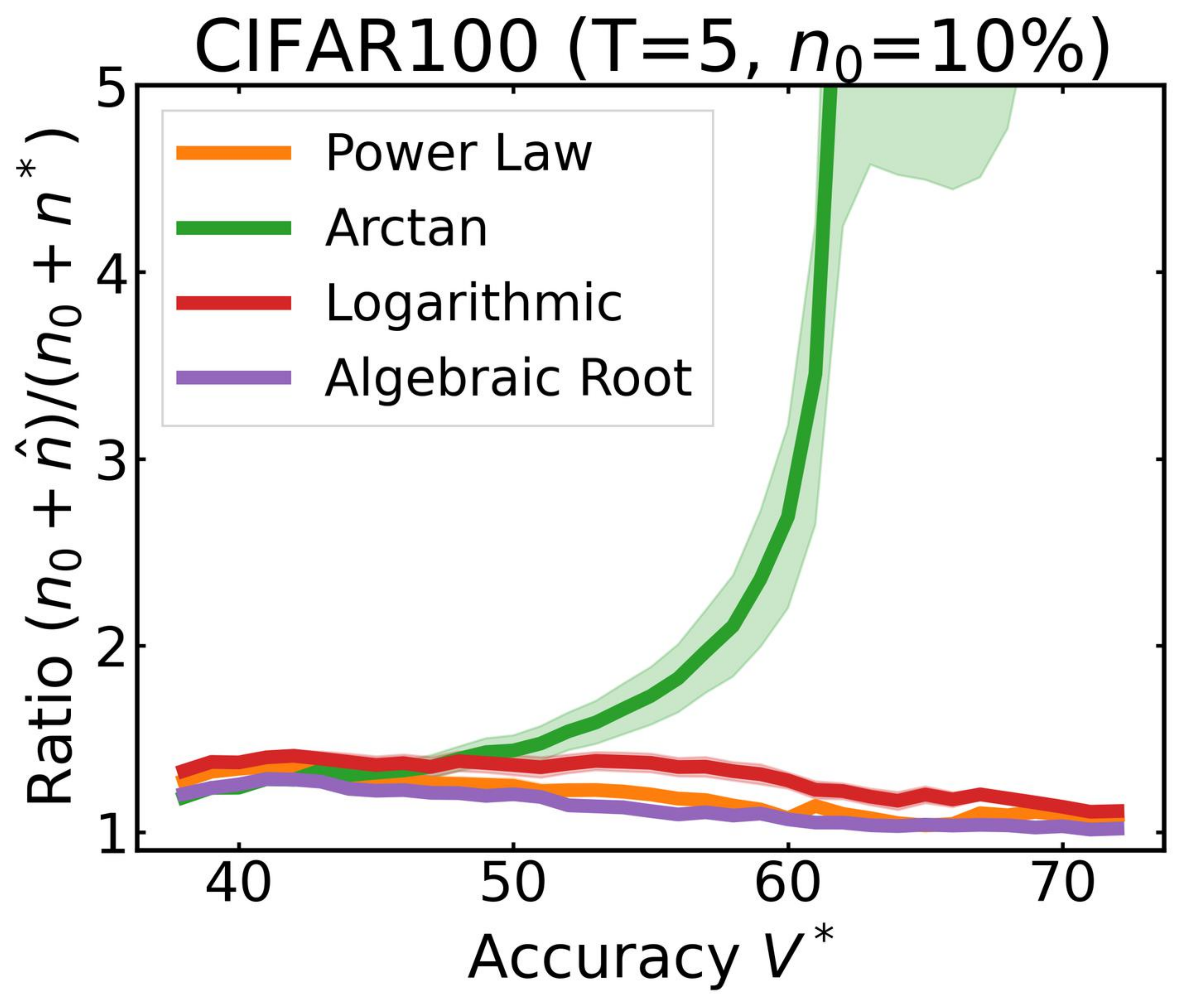} \end{minipage}
\begin{minipage}{0.16\linewidth}\includegraphics[width=1\textwidth]{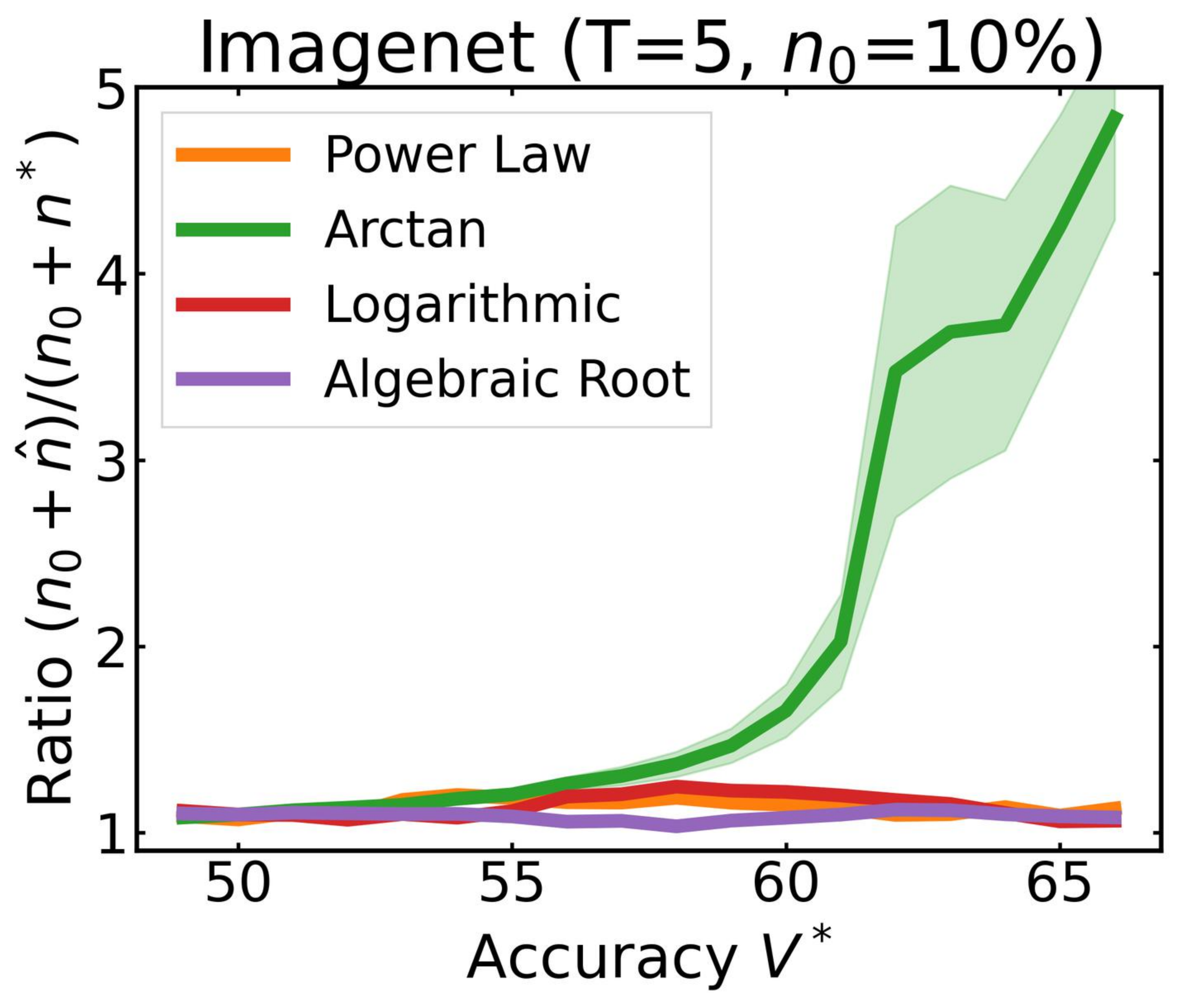} \end{minipage}
\begin{minipage}{0.16\linewidth}\includegraphics[width=1\textwidth]{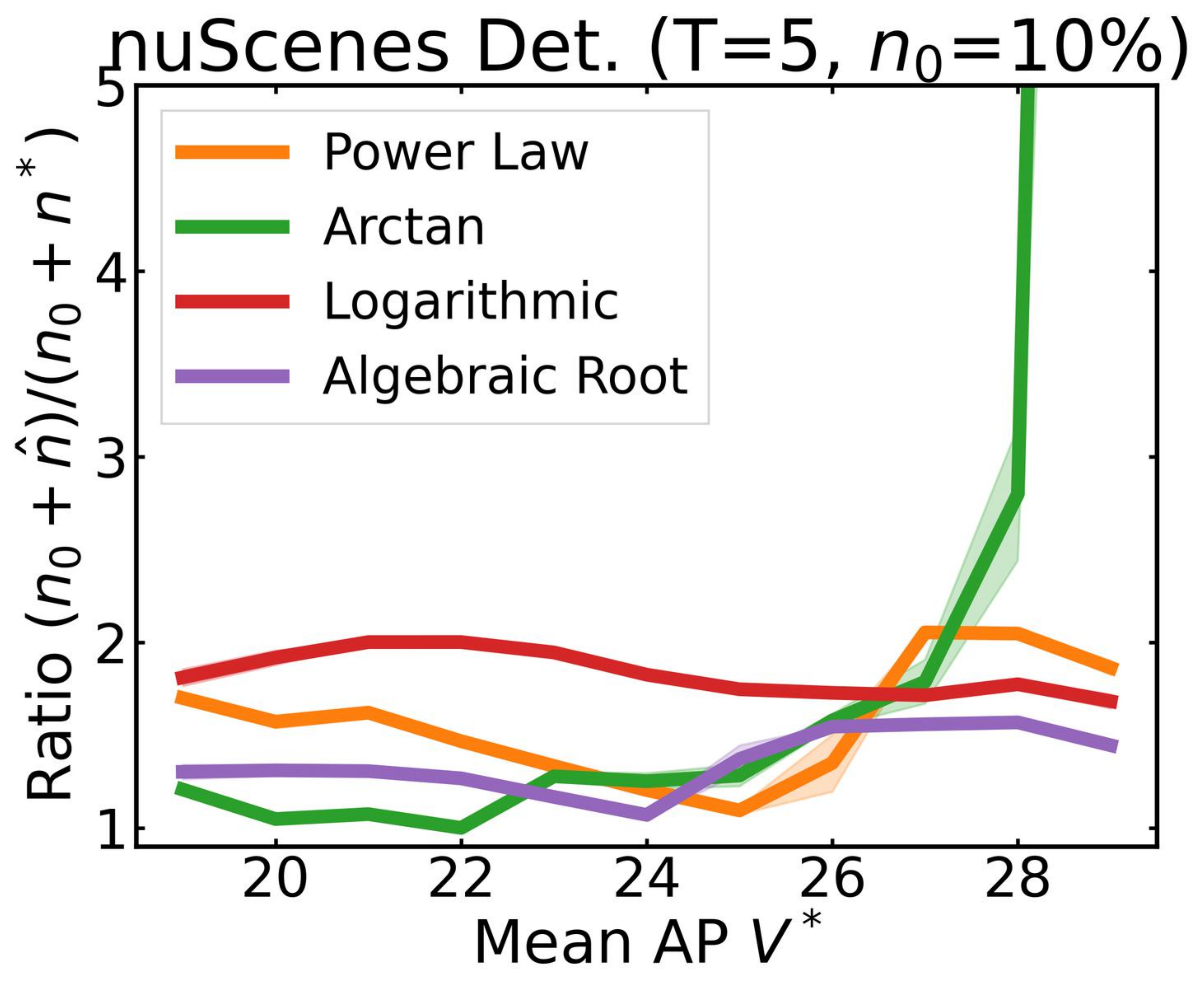} \end{minipage}
\begin{minipage}{0.16\linewidth}\includegraphics[width=1\textwidth]{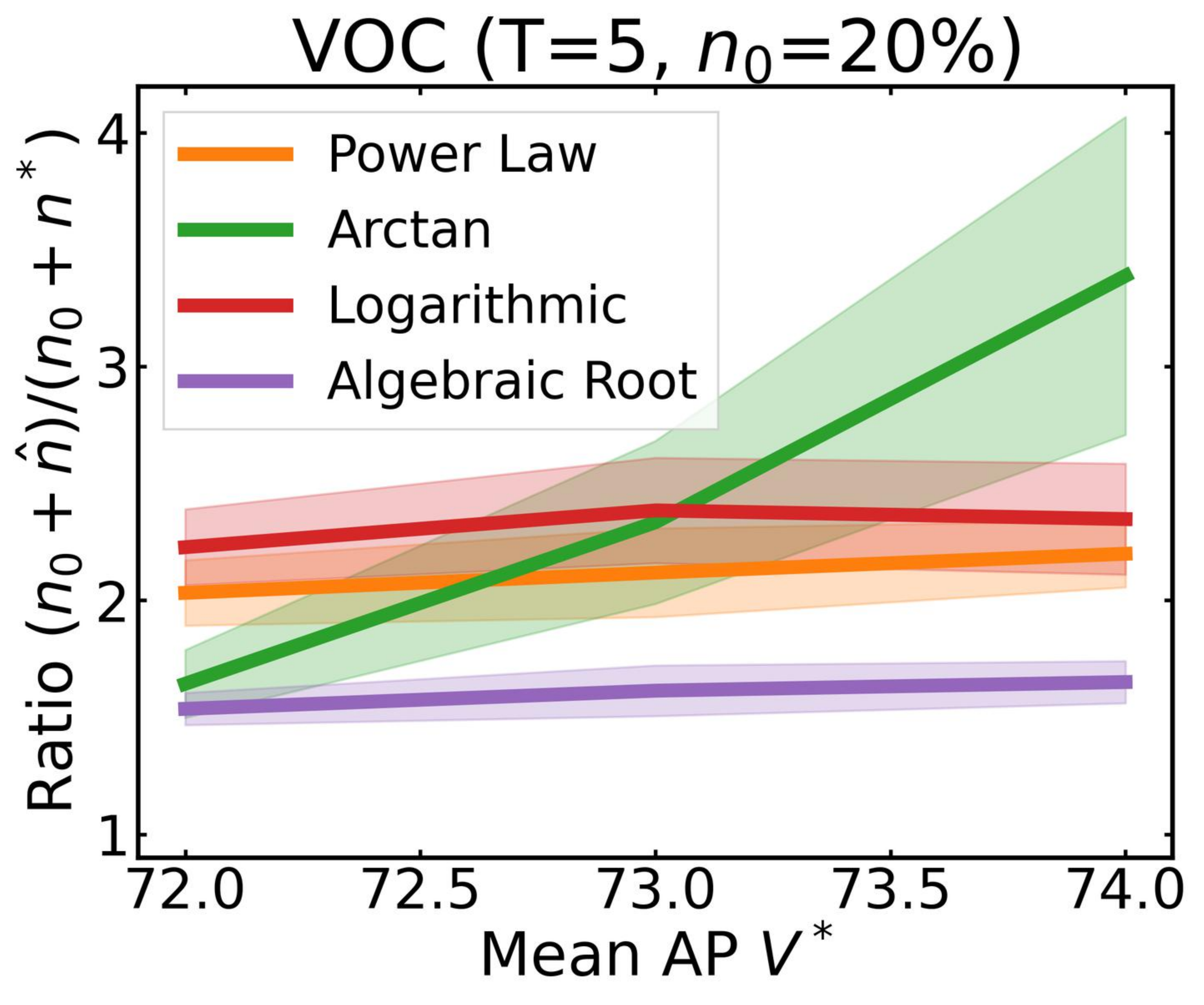} \end{minipage}
\begin{minipage}{0.16\linewidth}\includegraphics[width=1\textwidth]{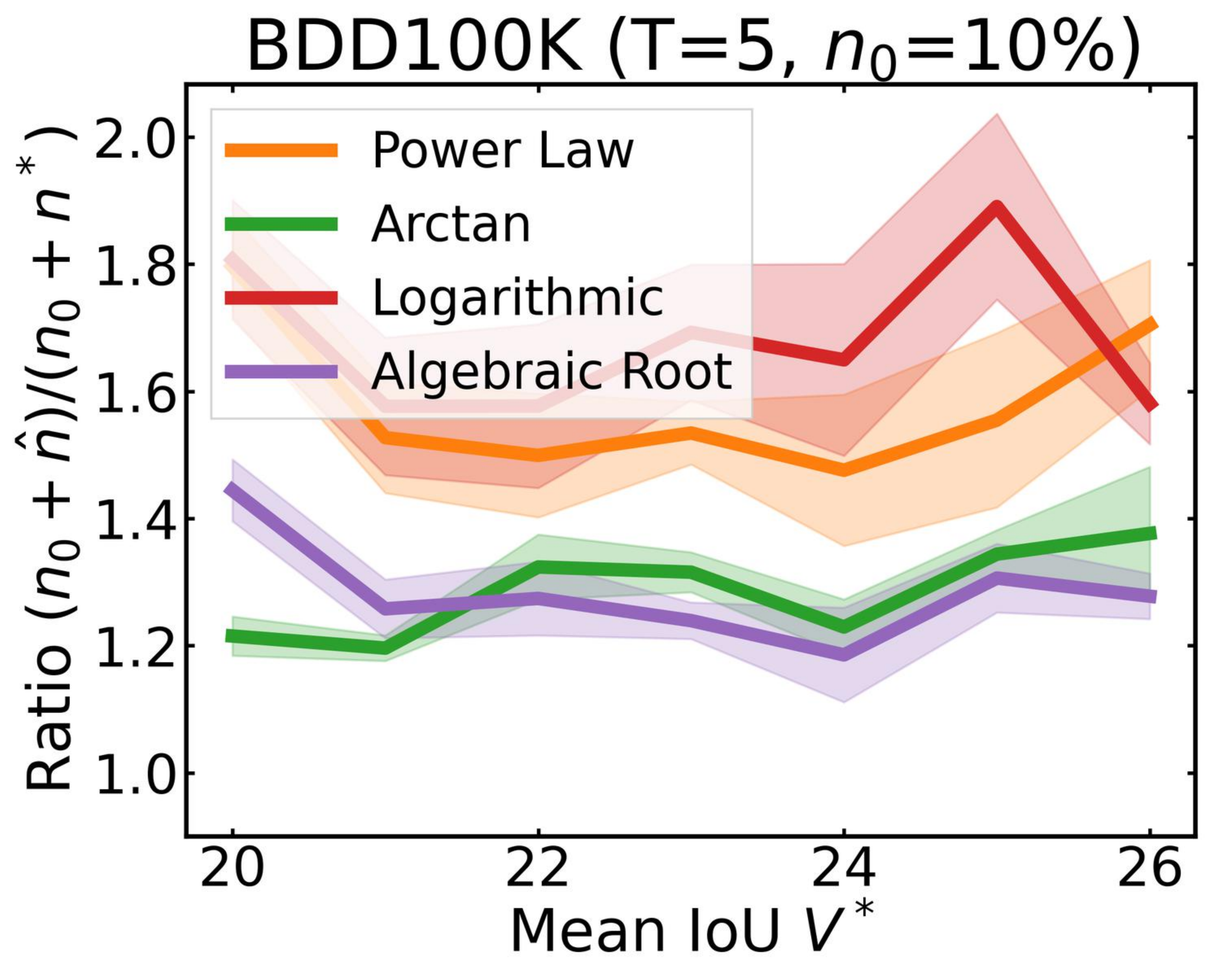} \end{minipage}
\begin{minipage}{0.16\linewidth}\includegraphics[width=1\textwidth]{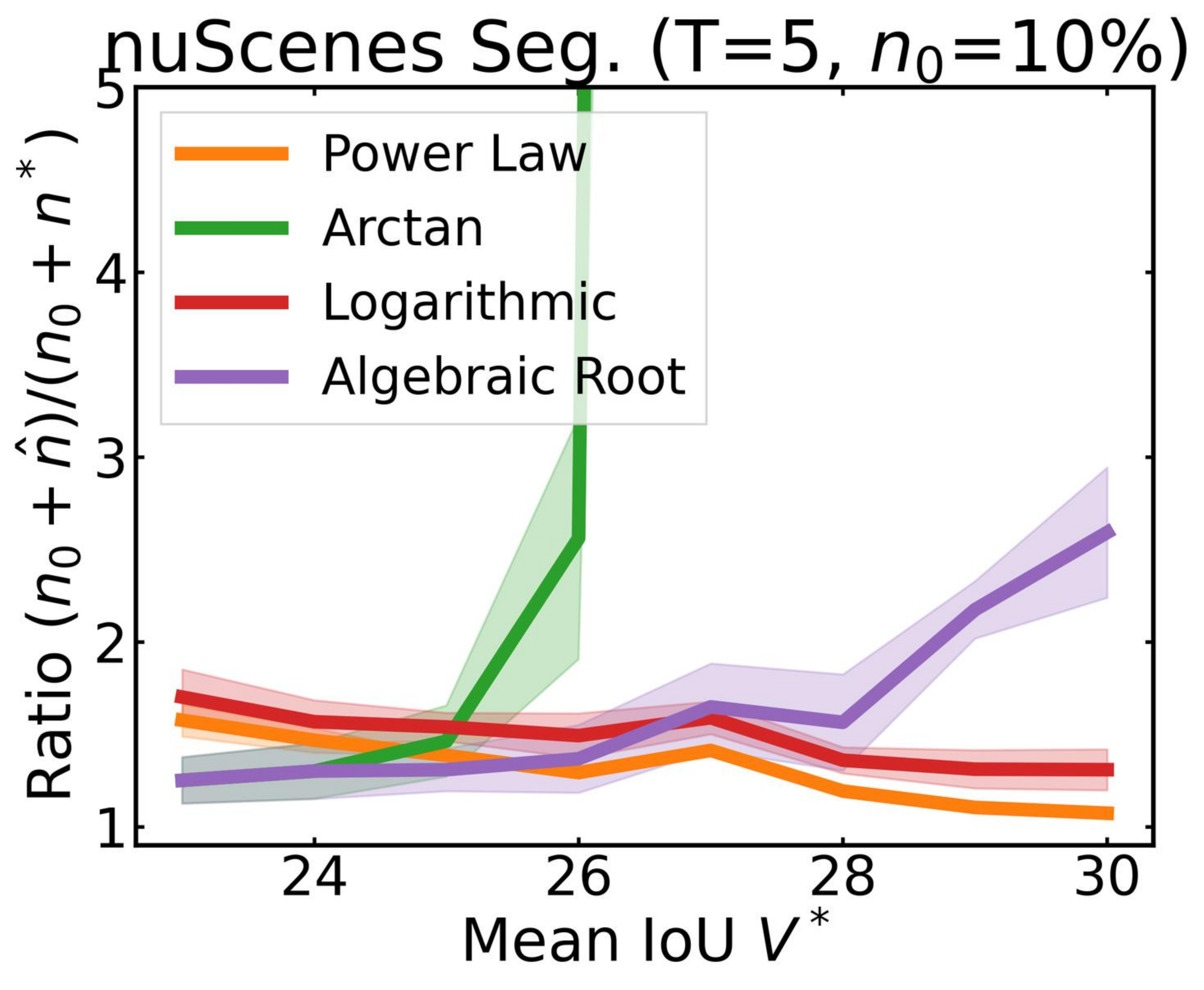} \end{minipage}
%
\vspace{-2mm}
\caption{\label{fig:simulation_after_tol}
For $T=5$, the ratio of the amount of data collected versus the minimum data needed to meet different target $V^*$ when using regression functions with correction factors fitted using CIFAR10.
}
\end{center}
\vspace{-9mm}
\end{figure*}

\subsection{A correction factor to help meet the target}

From Algorithm~\ref{alg:data_collection}, in each round of data collection, we minimize $\hn$ subject to $\hv(n_0 + \hn; \btheta^*) \geq V^*$. 
Ideally, we would want to minimize the true data requirement, \ie solving for $n^*$ satisfying $v(n_0 + n^*) = V^*$.
However, our simulations show that most of the regression functions are optimistic and under-estimate how much data is needed.
Intuitively, a simple way to correct for collecting less than the data needed to meet $V^*$ is to impose a correction factor $\tau \geq 0$ and instead estimate the data required to meet a ``corrected'' higher target $V^* + \tau$. 
As a result, we fix a constant $\tau$ and modify Algorithm~\ref{alg:data_collection} so that in each round, we now minimize $\hn$ subject to $\hv(n_0 + \hn; \btheta^*) \geq V^* + \tau$.


In order to determine how large this correction factor should be, we treat it as a hyper-parameter to fit. 
For instance, suppose that we have the full CIFAR10 data set and we want to construct a $T$-round collection policy for future data sets. 
We first simulate data collection with $\tau = 0$ for CIFAR10 with each regression function to obtain the plots in Figure~\ref{fig:simulation_all}.
We then increase $\tau$ until the entire ratio curve for that function is above $1$. In other words, we solve for the smallest $\tau$ such that the data collection policy will collect just enough data to meet all target values $V^*$ for CIFAR10 (for a given fixed $T$ and function). 
We then use this fitted $\tau$ as a correction factor for future data sets.

By combining the correction factor with multiple rounds of data collection, we can consistently collect just above the minimum data requirement. 
Table~\ref{tab:tolerance_values} compares the effect of using $\tau$ for each of the regression functions on the minimum ratio over all $V^*$ for each data set.
We use the CIFAR10 data set to fit $\tau$ for each setting of $T$ and regression function. 
Without correction, the Power Law, Logarithmic, and Algebraic Root functions achieve ratios less than $1$ for every data set except VOC. 
Using $\tau$, these functions almost always achieve ratios between $1$ to $2$. 
Furthermore for each data set, these three regression functions achieve their respective lowest ratios (above $1$) when $T=5$. 
Figure~\ref{fig:simulation_after_tol} further plots simulations using $\tau$ over all $V^*$ for each data set with $T=5$. 
Here, the Power Law, Logarithmic, and Algebraic Root functions achieve ratios between $1.03$ to $2.5$ for all $V^*$ with every data set.
Furthermore, there is no consistently best regression function for all data sets. For instance, the Algebraic Root function dominates over VOC, but the Power Law is particularly effective on nuScenes BEV segmentation when $V^*$ is large.
However, recall that Arctan naturally over-estimates the data requirement, so this function does not benefit from correction.
We conclude that correcting any of the three optimistic estimators, Power Law, Logarithmic, or Algebraic Root, and collecting data over five rounds is enough to approximately minimize the total data collected while still meeting the desired target.

\begin{figure*}[!t]
\begin{center}
\begin{minipage}{0.16\linewidth}\includegraphics[width=1\textwidth]{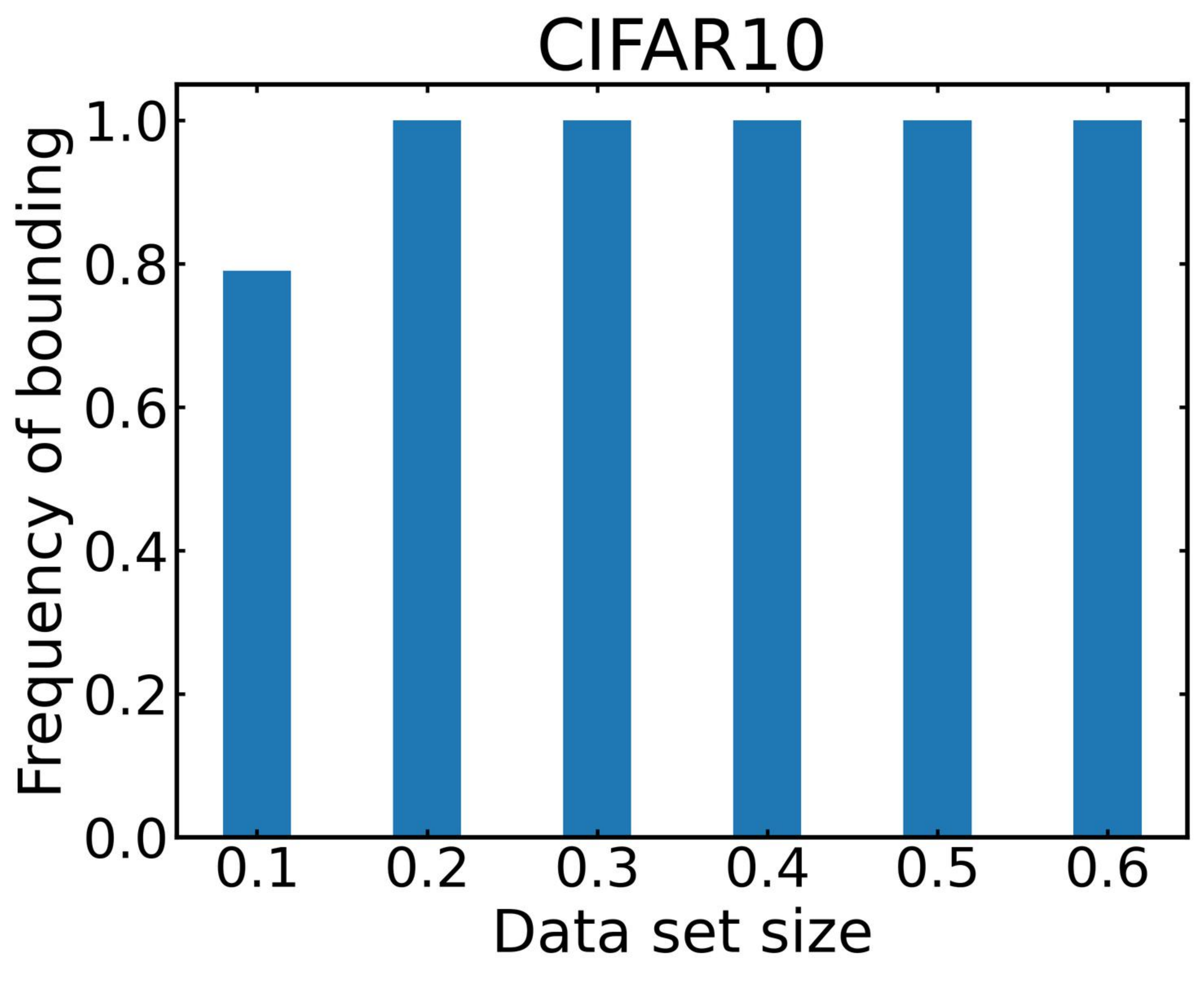} \end{minipage}
\begin{minipage}{0.16\linewidth}\includegraphics[width=1\textwidth]{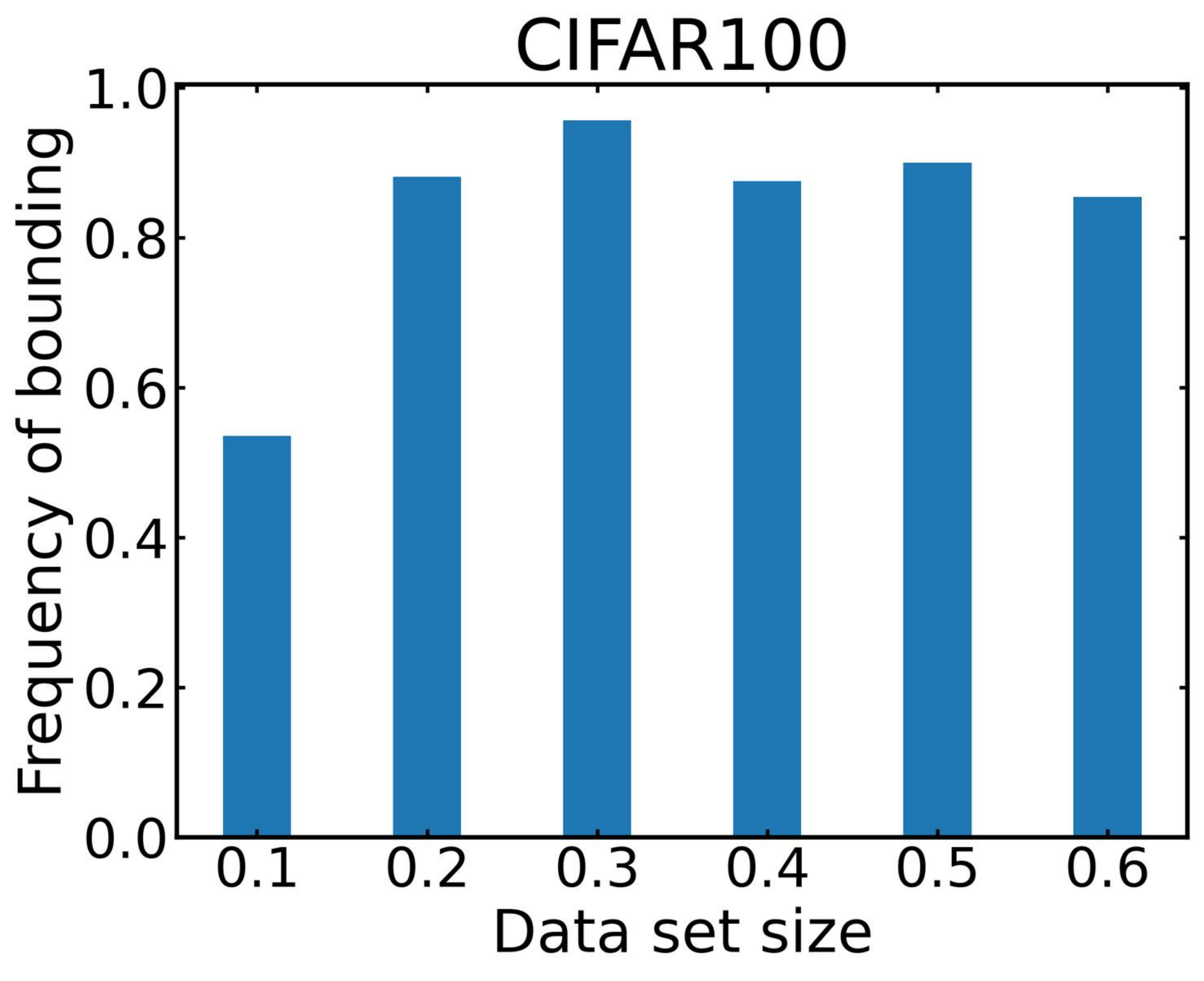} \end{minipage}
\begin{minipage}{0.16\linewidth}\includegraphics[width=1\textwidth]{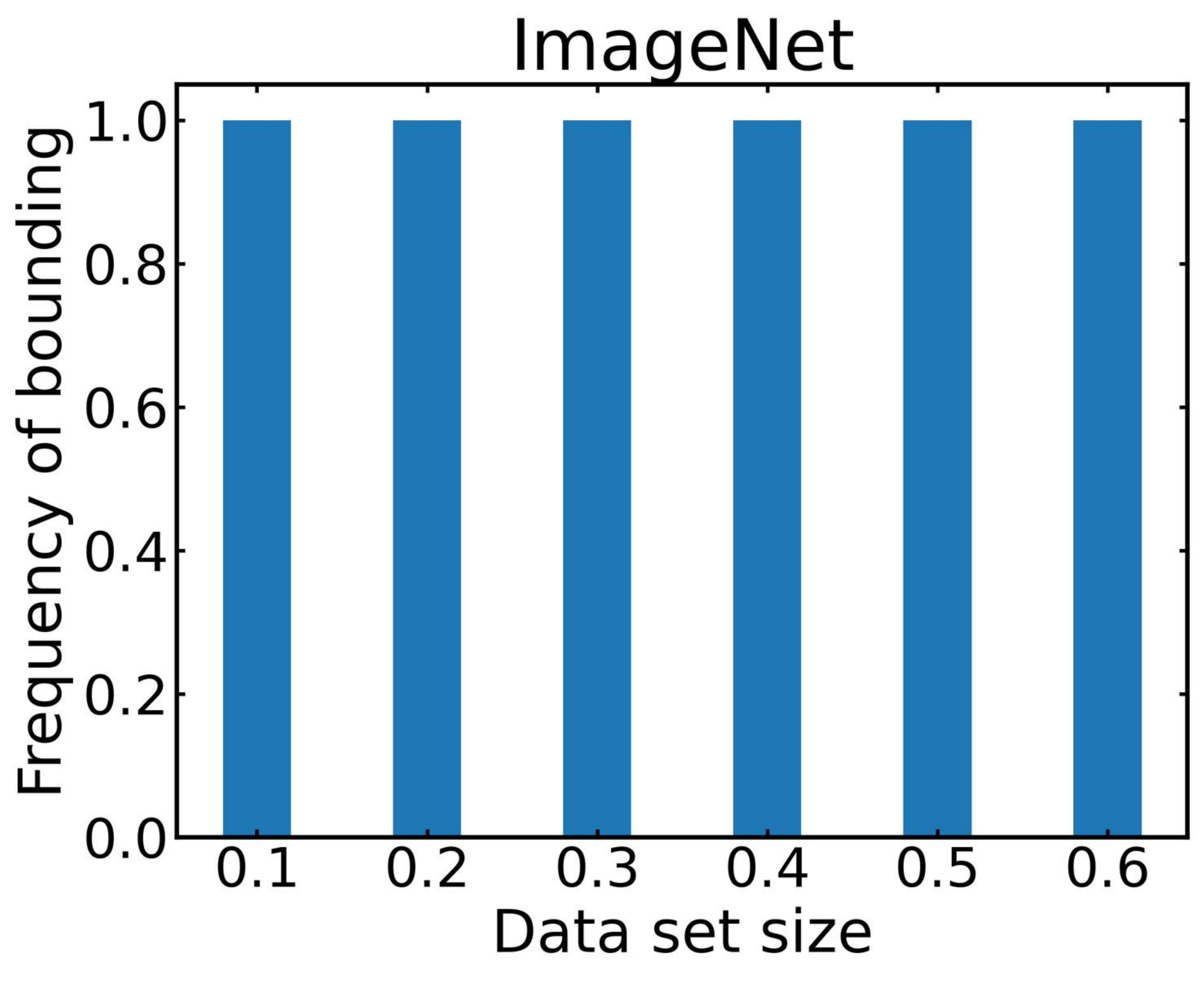} \end{minipage}
\begin{minipage}{0.16\linewidth}\includegraphics[width=1\textwidth]{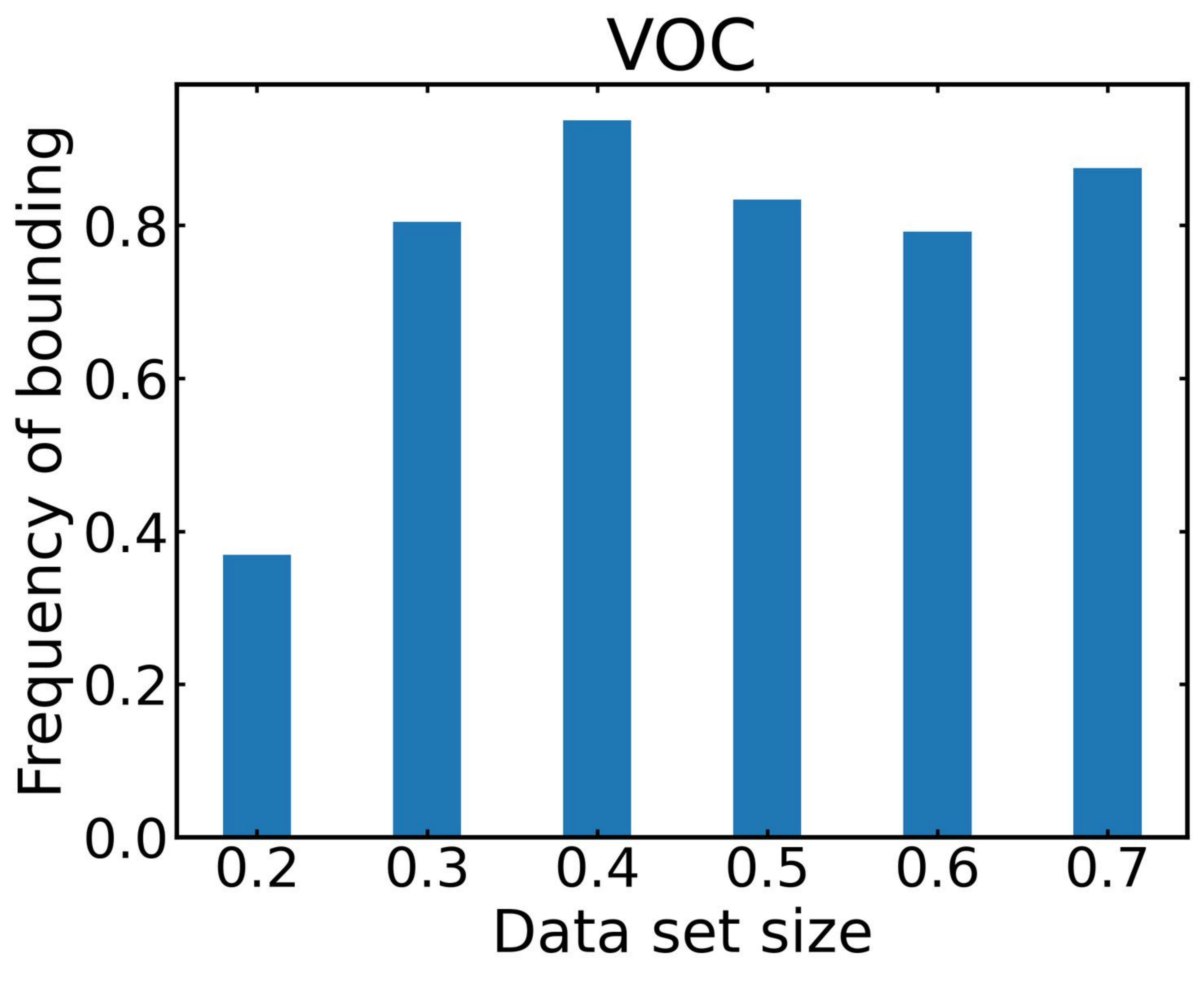} \end{minipage}
\begin{minipage}{0.16\linewidth}\includegraphics[width=1\textwidth]{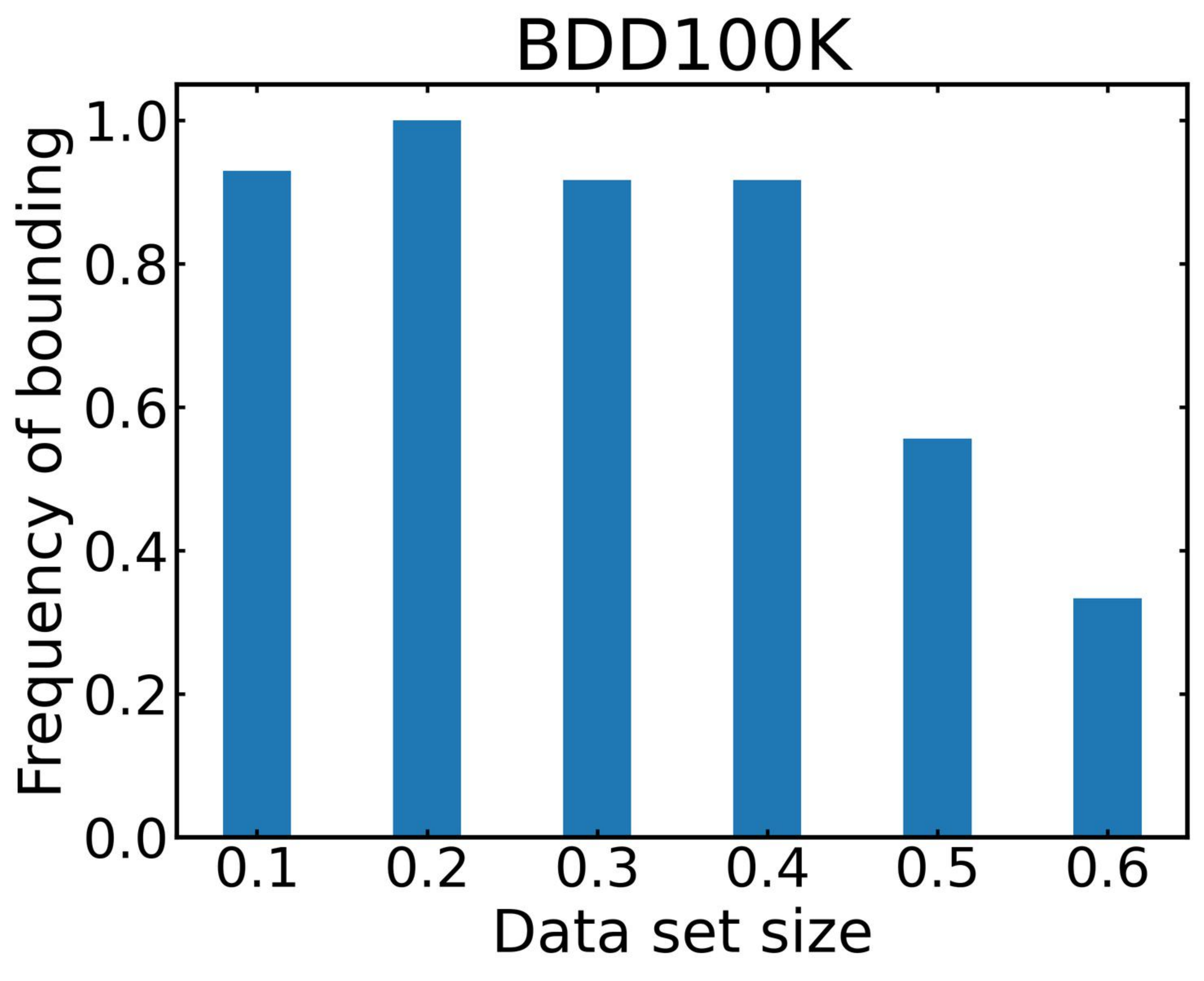} \end{minipage}
\begin{minipage}{0.16\linewidth}\includegraphics[width=1\textwidth]{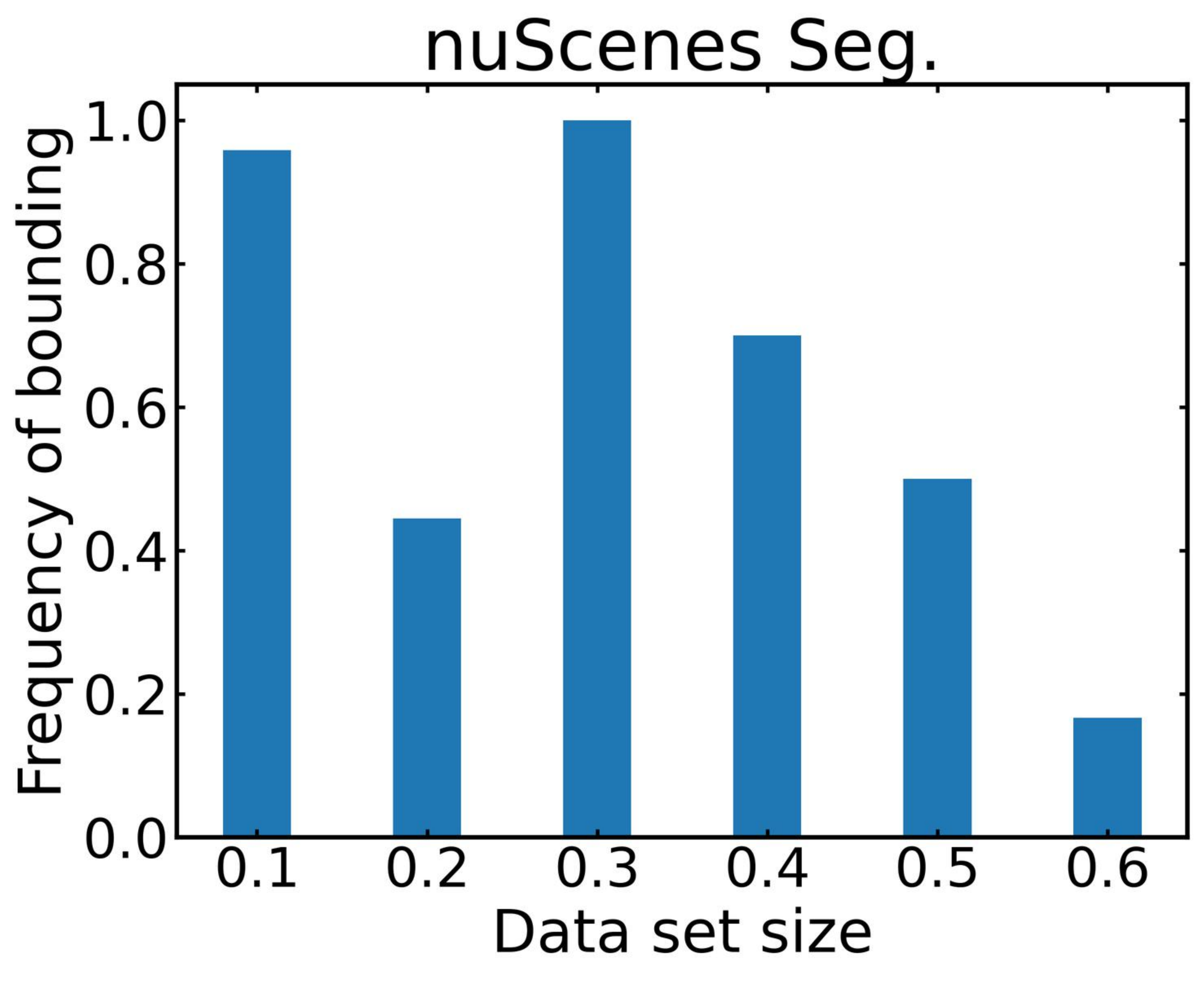} \end{minipage}
\begin{minipage}{0.16\linewidth}\includegraphics[width=1\textwidth]{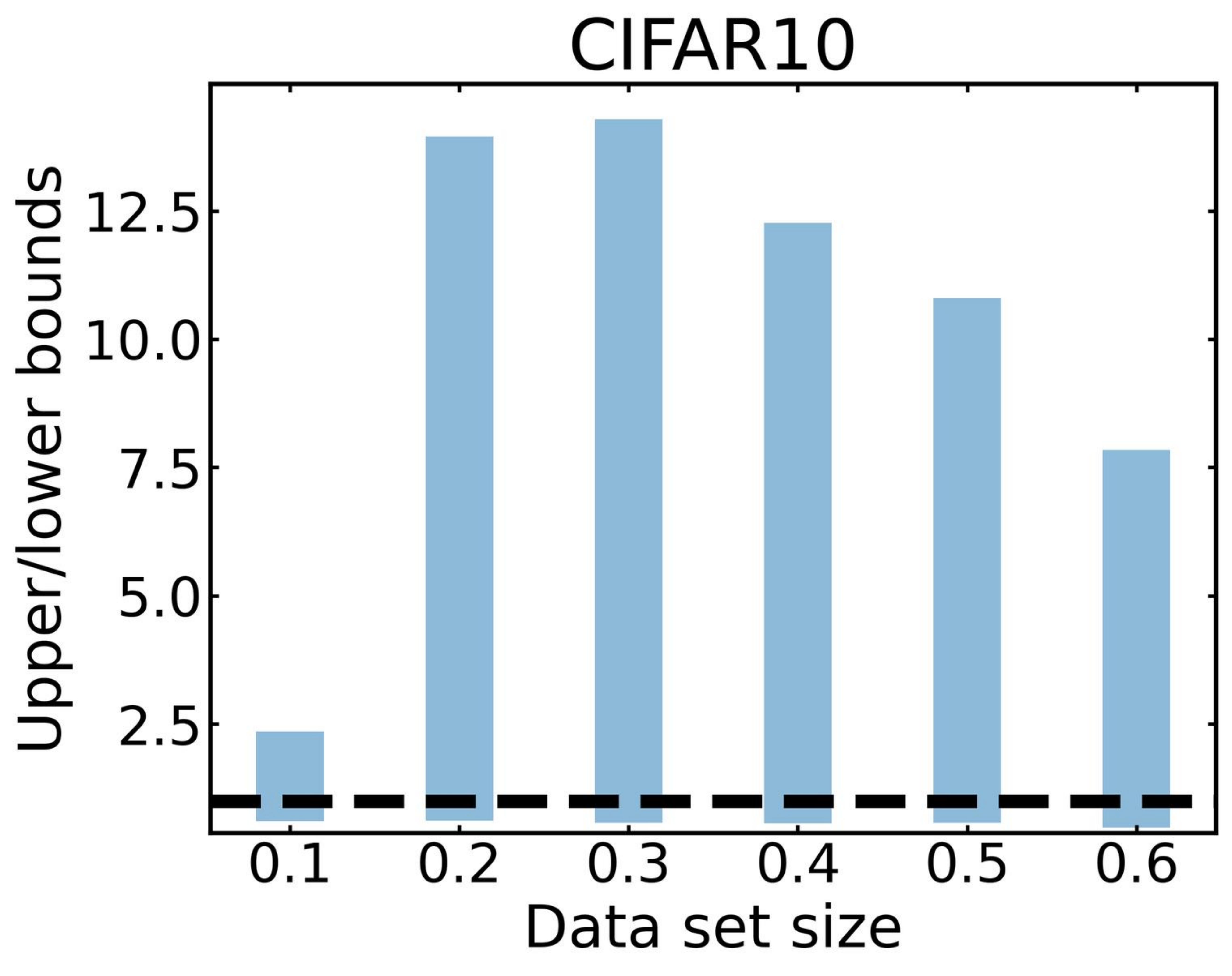} \end{minipage}
\begin{minipage}{0.16\linewidth}\includegraphics[width=1\textwidth]{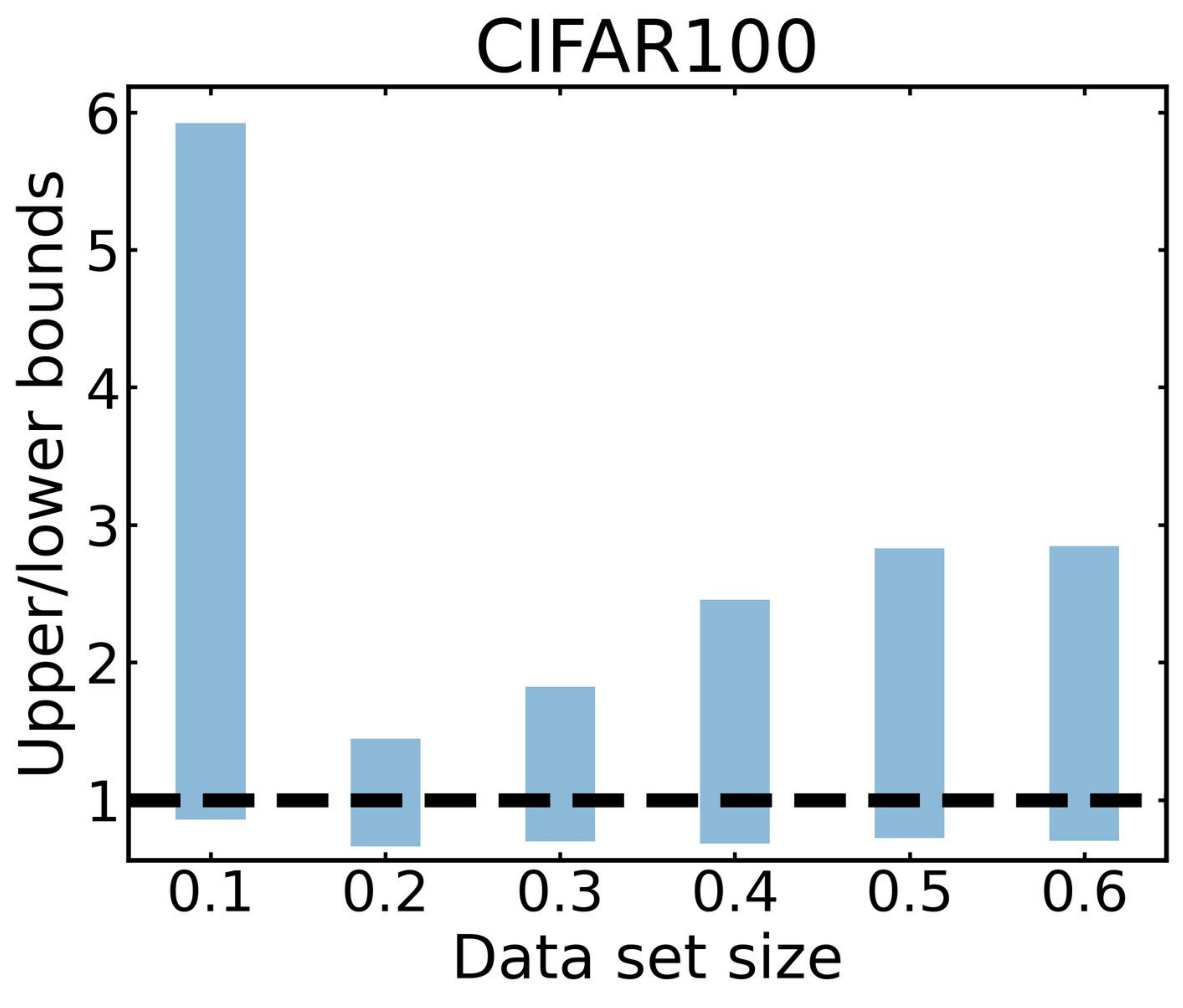} \end{minipage}
\begin{minipage}{0.16\linewidth}\includegraphics[width=1\textwidth]{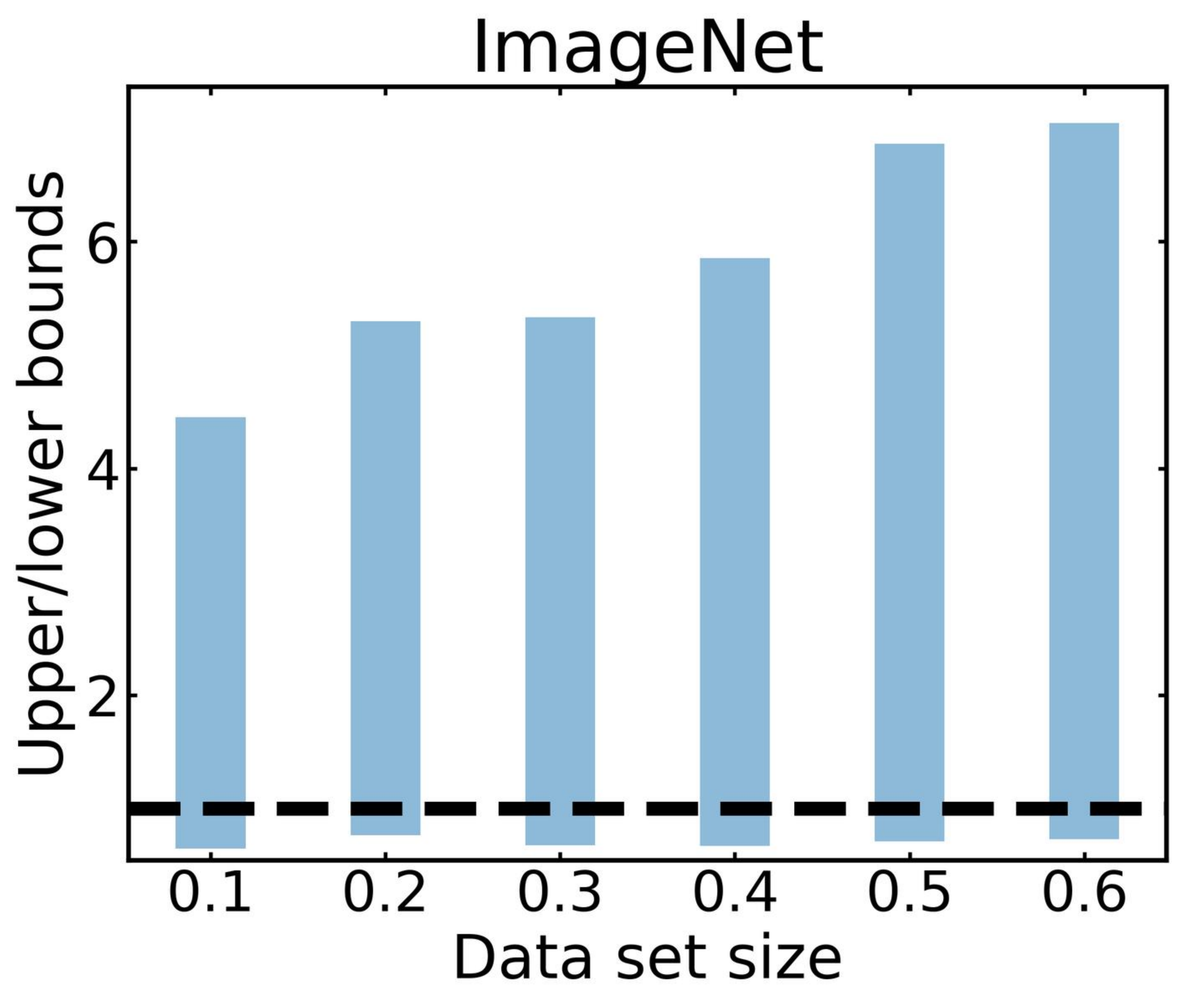} \end{minipage}
\begin{minipage}{0.16\linewidth}\includegraphics[width=1\textwidth]{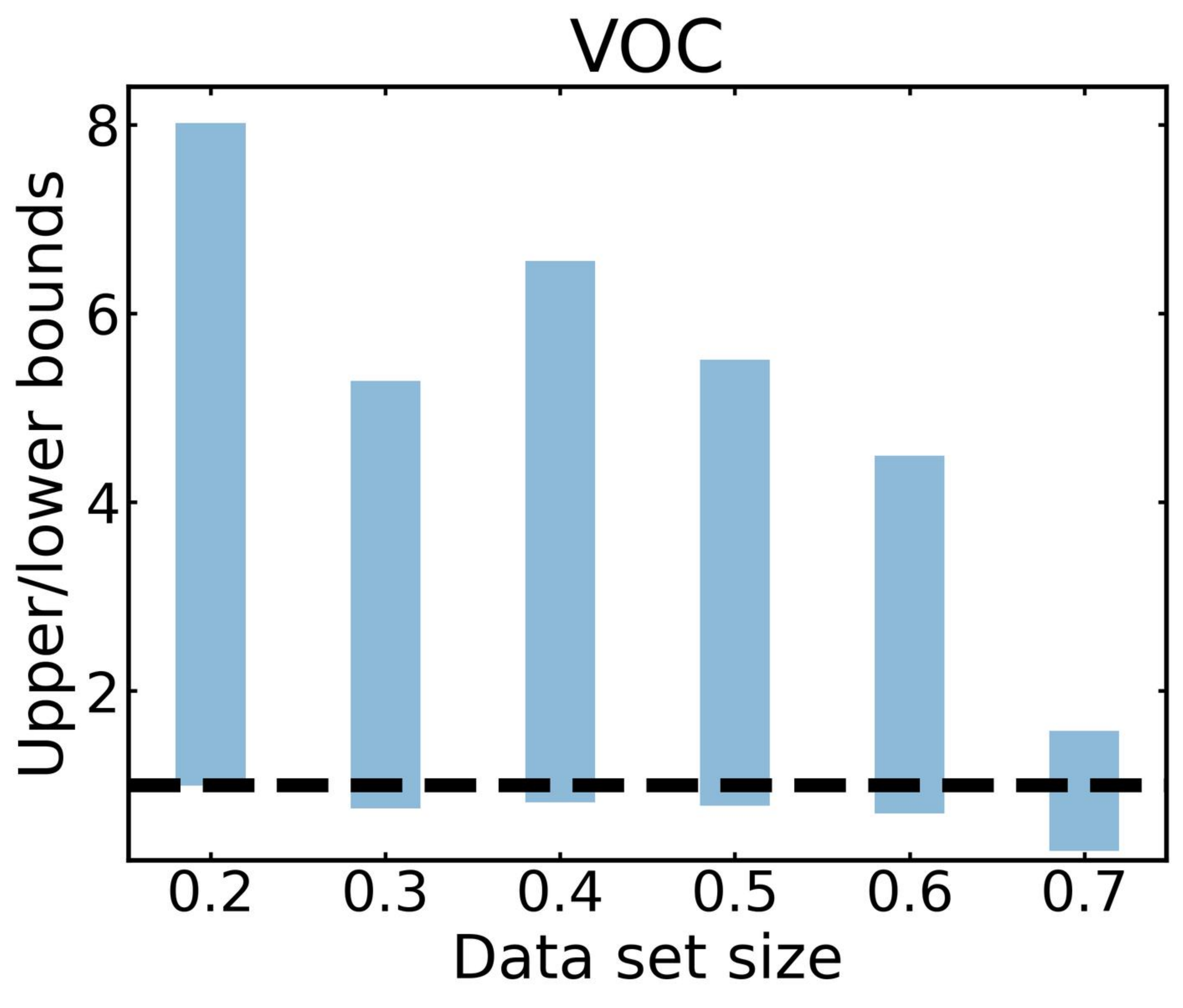} \end{minipage}
\begin{minipage}{0.16\linewidth}\includegraphics[width=1\textwidth]{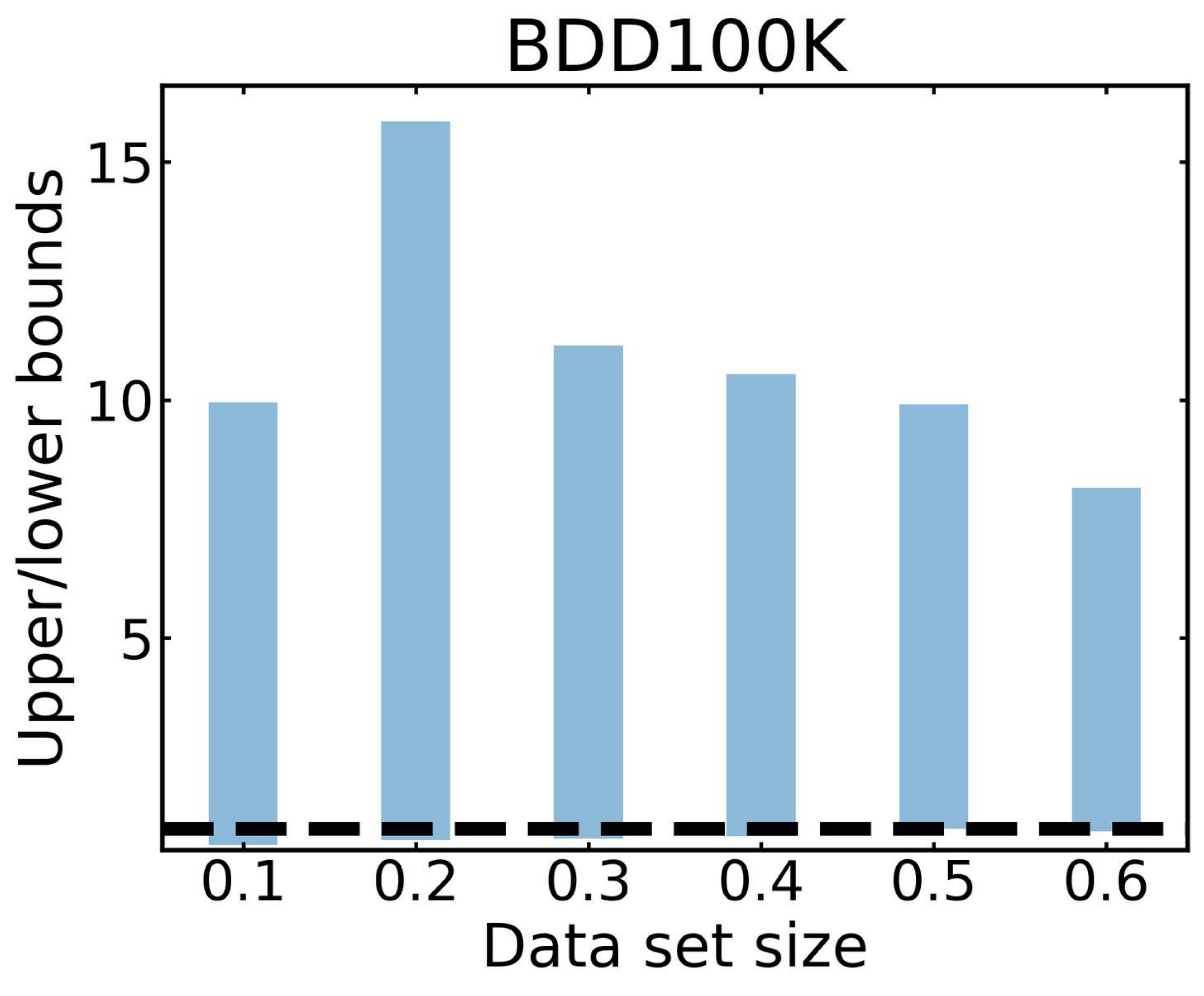} \end{minipage}
\begin{minipage}{0.16\linewidth}\includegraphics[width=1\textwidth]{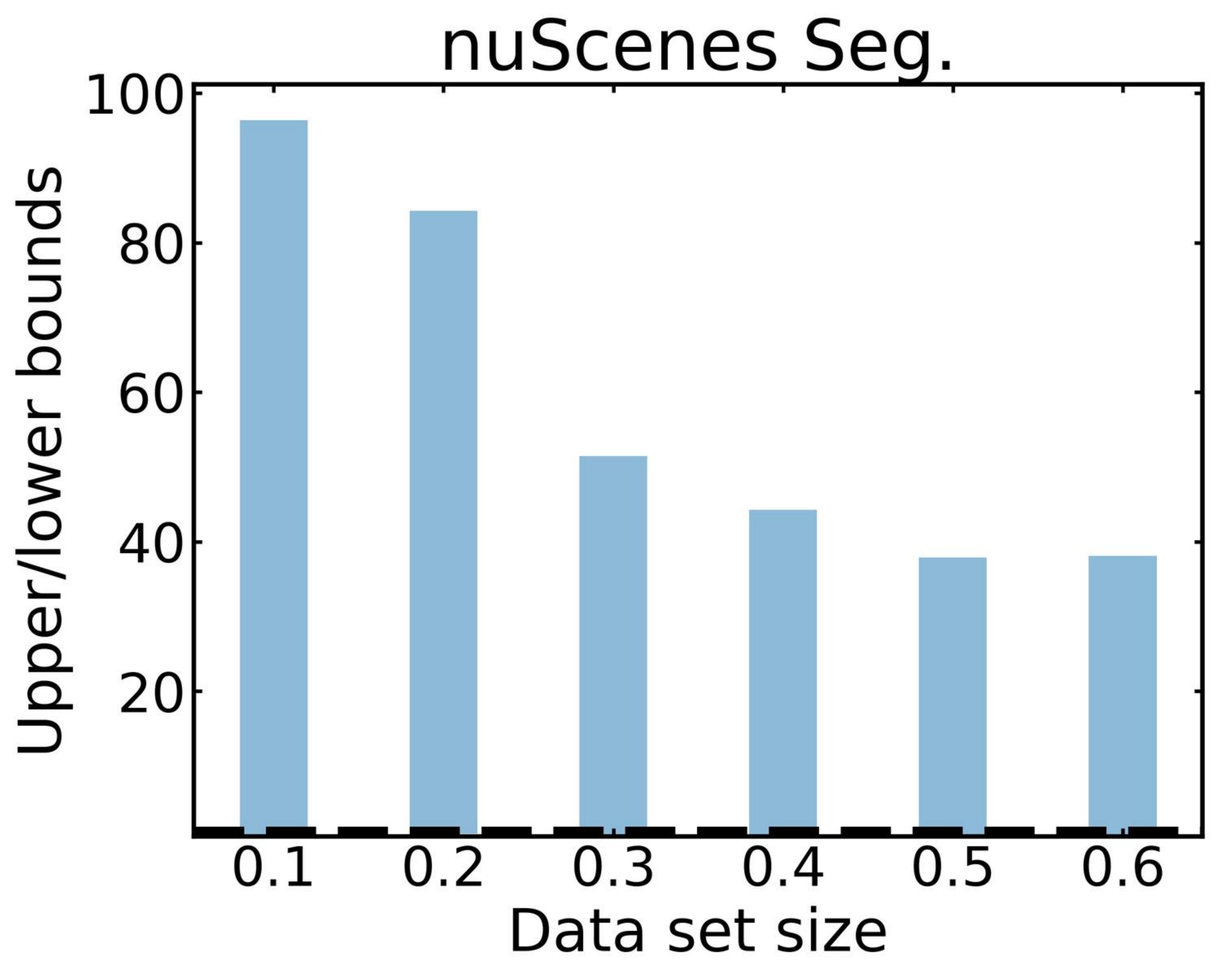} \end{minipage}
%
%
%
\vspace{-2mm}
\caption{\label{fig:bounds} 
(Top row) For $T=1$ and varying $n_0$, the frequency of instances where the largest and smallest $\hn$ estimated by the different functions upper and lower bound the true $n^*$. 
(Bottom row) The largest and smallest ratios $(n_0 + \hn) / (n_0 + n^*)$ estimated by the different functions. The dashed black line corresponds to ratio $1$. 
}
\end{center}
\vspace{-9mm}
\end{figure*}

\subsection{Empirical bounds on the data requirement}

If the correction factor is poorly fit or the number of rounds is constrained to be small, we may still under- or over-estimate the data requirement.
From Table~\ref{tab:tolerance_values}, on nuScenes segmentation with $T=1$, the Power Law without $\tau$ leads to estimating $58\%$ of the required data, whereas using $\tau$ leads to estimating $28$ times more data than needed.

In some applications, modelers may also desire rule-of-thumb estimates of the amount of data that they should immediately collect. 
We now consider the problem where we have a single $T=1$ round remaining with $n_0$ data points; in a single round or the final of multiple rounds, we must meet the data collection target.
As a result here, we seek to obtain worst and best-case estimates (\ie upper and lower bounds) on how much more data is needed.
All of the different regression functions yield an ensemble of predictions. Then, the largest prediction is the worst-case estimate and the smallest prediction is the best-case estimate.

For each data set, we set $T=1$ and sweep over $n_0$ and $V^*$ to estimate the data requirement with eight regression functions. 
Figure~\ref{fig:bounds} (top row) shows for each $n_0$, the frequency of instances of $V^*$ in which the most optimistic and pessimistic regression function bound the true data requirement. 
The bottom row further shows the average value of these upper and lower bounds. 
For image classification, our estimators bound the true requirement over $80\%$ of the time. 
This trend also holds for VOC wherein we bound the requirement over $80\%$ of the time for $n_0 \geq 30\%$ of the data set.
Since BDD100K and nuScenes BEV segmentation are more challenging data sets, our probability of bounding the data requirement can at times decrease. 
Because training the 3-D object detector on nuScenes is computationally far more expensive than the other tasks, we omit their plots and only report values for $n_0 = 10\%, 20\%, 50\%$.
Here, the range of estimators bound the true data requirement $88\%$, $91\%$, and $83\%$ of the time with ratios in the interval $[0.56, 31.1]$, $[0.76, 40.8]$, $[0.56, 26.9]$, respectively.
Nonetheless, the results show that if we are given a single round with a large initial data set, we will be able to accurately estimate upper and lower bounds on the data requirement. 
Moreover, even if we are given multiple rounds to collect data, on the final round, we should be able to obtain upper and lower intervals for the requirement. 
In practical applications, these bounds can guide modelers to make optimistic or pessimistic choices, for example if the real-world deadline for training and deploying a model is strict.

%% file: sections/discussion.tex
\section{Discussion}

In this work, we propose an effective solution to the problem of estimating how much data must be collected to meet a target performance. 
While the problem of predicting a model's performance has received growing research interest as a springboard for various design decisions, we find that estimating performance does not capture the downstream problem of estimating data requirements. 
Even small errors in predicting performance can yield large errors in data collection, meaning that the error permissible from a good data estimator is far smaller than intuition suggests.
Furthermore, errors are divided into under- or over-estimation, where each poses different challenges to data collection. 
To better analyze data collection strategies, we formulate an iterative data collection simulation. 
Our experiments draw several high-level insights:

$\bullet$ Different techniques estimate either far more data or far less data than needed. Using multiple rounds of data collection with techniques that under-estimate can lead to collecting up to $90\%$ of the true amount of data needed.

$\bullet$ By simulating on previous tasks, we can identify which approaches under-estimate data requirements and learn a correction factor to address this deficiency. Using a correction factor and collecting for up to five rounds allows us to collect at most one to two times the minimum amount of data needed for any desired performance.

$\bullet$ With only one round of data collection remaining, we can use all of the different regression functions to obtain an interval that often bounds the true data requirement. These bounds can guide modelers to collect data more or less aggressively with respect to practical requirements.

\noindent\textbf{Limitations.}
The data collection problem and the simulation proposed in this work approximate real collection practices. 
Our simulation relies on a pre-constructed ground truth $v(n)$ rather than sampling points, training a model, and evaluating $V_f(\dataset)$. 
The latter is computationally too expensive to perform for the range of settings explored in this paper. 
The quality of our simulation depends on the number of subsets used to construct $v(n)$. More subsets means $v(n)$ better approximates $V_f(\dataset)$ and from  inspection (see the supplementary content), all of our $v(n)$ appear to be visually smooth curves. 
Moreover in our data collection problem, we assume that the model $f$ and sampling strategy $p(z)$ are constant. In practice, designers may update $f$ in between rounds; this may be incorporated in a more complete model of the deep learning workflow.
In addition, secondary metrics can be used to optimize $p(z)$.
For example, if a classifier is particularly poor for a single class in a given round, modelers may seek to obtain more samples of that specific class in the next round. 
We leave these more sophisticated problem settings to future research.

%% file: sections/supplement.tex
\clearpage

\appendix

\onecolumn

\section*{Supplementary Content} 


\section{Experiment setup}
\label{sec:app_experiment_setup}


We first summarize the regression problem before detailing the data collection and training process for each data set and task.
All models were implemented using PyTorch and trained on machines with up to eight NVIDIA V100 GPU cards.

We fit each regression function by minimizing a least squares problem using the Levenberg-Marquardt algorithm as implemented by Scipy~\cite{more1978levenberg, 2020SciPy-NMeth}. The parameters for each function are initialized to either $1$ or $0$ depending on if they are product or bias terms. To further help fit the data, we use weighted least squares where each subsequent point is weighted twice as much as the previous point. This ensures that our regression model is tuned to better fit the curve for larger $n$.

\noindent\textbf{Image classification tasks.} For all experiments with CIFAR10 and CIFAR100, we use a ResNet18~\cite{he2016deep} following the same procedure as in~\cite{coleman2019selection}. For ImageNet, we use a ResNet34~\cite{he2016deep} using the procedure in~\cite{coleman2019selection}. 
All models are trained with cross entropy loss using SGD with momentum. We evaluate all models on Top-1 Accuracy.

For all experiments, we first create 10 subsets $\set{S}_0 \subset \set{S}_1 \subset \set{S}_2 \subset \cdots  \subset \set{S}_9 = \dataset_0$ containing $2\%, 4\%, 6\%, \dots 20\%$ of the training data set, respectively. For example on CIFAR10, $\set{S}_0$ contains 1000 images, $\set{S}_1$ contains 2000 images, and so on. This data is used to build our initial regression models. Thus, when we use $n_0 = 10\%$ of the training data, our initial regression data contains five points evaluating the score from training with $1000, 2000, \dots, 5000$ images.
For evaluation, we sample $\dataset_1 \subset \dataset_2 \subset \dataset_3 \subset \cdots  \subset \dataset_8$ containing $30\%, 40\%$, $50\%, \dots, 100\%$ of the training data set, respectively. In regression, we evaluate our estimators on predicting $V_f(\dataset_i)$ for each of these data sets.

\noindent\textbf{VOC.}
We use the Single-Shot Detector 300 (SSD300)~\cite{liu2016ssd} based on a VGG16 backbone~\cite{simonyan2014very}, following the same procedure as in~\cite{elezi2021towards}. All models are trained using SGD with momentum. We evaluate all models on mean AP.

For all experiments, we create 8 regression subsets $\set{S}_0 \subset \set{S}_1 \subset \set{S}_2 \subset \cdots  \subset \set{S}_7 = \dataset_0$ containing approximately $2.5\%, 5\%, 7.5\%, \dots 20\%$ of the training data set, meaning $n_0 = 10\%$ of the data corresponds to an initial regression data set of four points.
For evaluation, we sample $\dataset_1 \subset \dataset_2 \subset \dataset_3 \subset \cdots \subset \dataset_8$ containing approximately $30\%, 40\%, 50\%, \dots, 100\%$ of the full training data, respectively.

\noindent\textbf{nuSenes (Detection).}
We use the FCOS3D network architecture~\cite{wang2021fcos3d}, which received first place in the NeurIPS 2020 nuScenes 3-D detection challenge. We follow the same procedure from the original paper for training using SGD. We evaluate on mean AP.

Because training this 3-D detector is computationally expensive, we only use a small number of points for these experiments. We first create initial subsets $\set{S}_0 \subset \set{S}_1 \subset \set{S}_2 \subset \set{S}_3$ containing $5\%, 10\%, 15\%, 20\%$ of the training data set, respectively. For evaluation, we sample $\dataset_1 \subset \dataset_2 \subset \dataset_3 \subset \dataset_4$ containing $25\%, 50\%, 75\%, 100\%$ of the training data set, respectively.

\noindent\textbf{BDD100K.}
We use Deeplabv3 \cite{chen2017rethinking} with ResNet50 backbone. We use random initialization for the backbone. We use the original dataset split from \cite{yu2020bdd100k} with 7k train and 1k validation set. The evaluation metrics is mean Intersection over Union (IoU).

For all experiments, we first create 10 regression subsets $\set{S}_0 \subset \set{S}_1 \subset \set{S}_2 \subset \cdots \subset \set{S}_9 = \dataset_0$ containing $2\%, 4\%, 6\%, \dots, 20\%$ of the training data set, respectively. For evaluation, we sample $\dataset_1 \subset \dataset_2 \subset \dataset_3 \subset \cdots \subset \dataset_8$ containing $30\%, 40\%, 50\%, \dots, 100\%$ of the training data set, respectively.

\noindent\textbf{nuScenes (Segmentation).}
We use the ``Lift Splat'' architecture~\cite{liftsplat}, which is used for BEV segmentation from driving scenes, following the steps from the original paper to train this model. We evaluate on mean IoU. Our data collection procedure follows the same steps and percentages of the data set as used for BDD100K.

\begin{figure*}[!t]
\begin{center}
\begin{minipage}{0.16\linewidth}\includegraphics[width=1\textwidth]{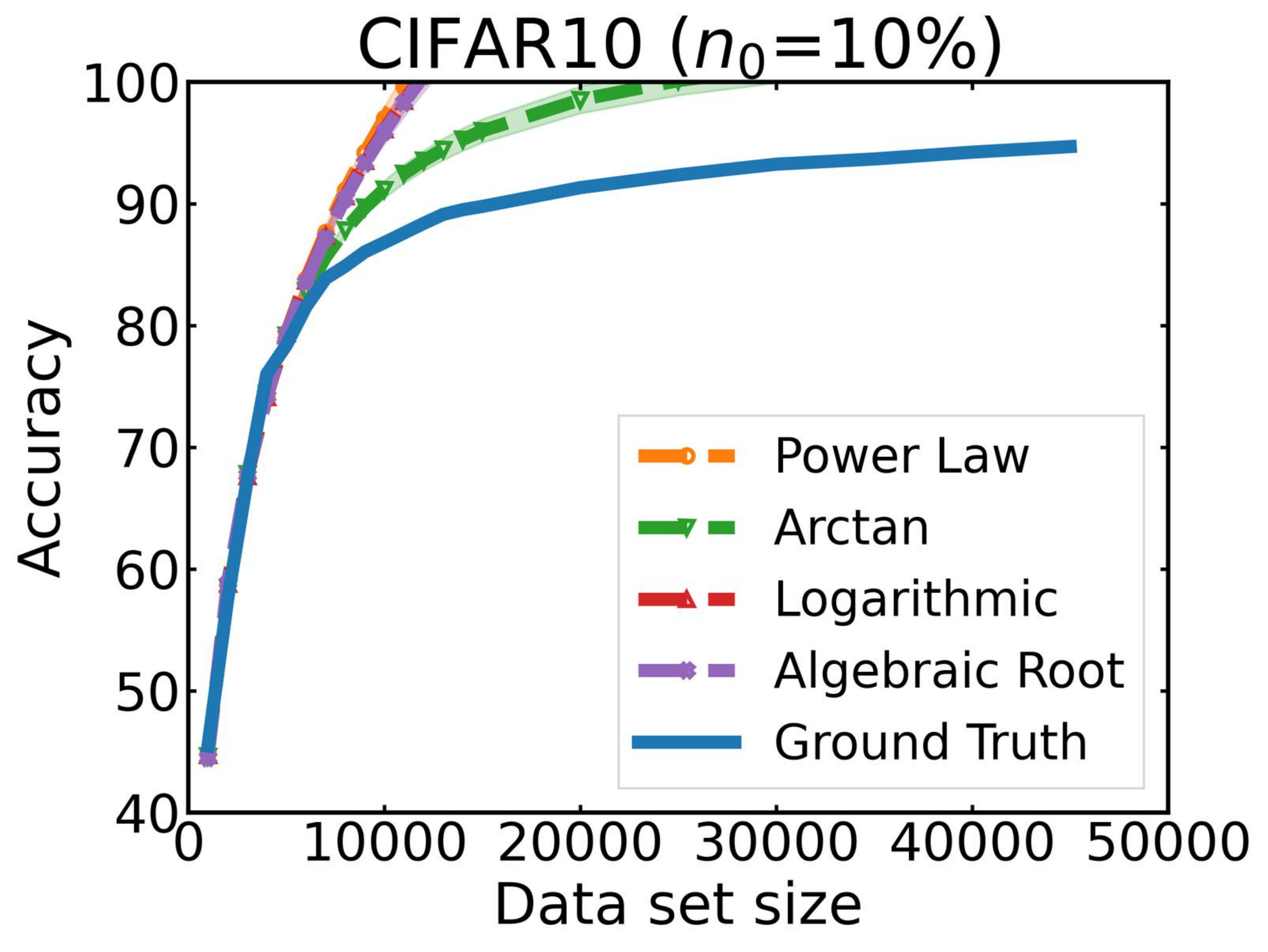} \end{minipage}
\begin{minipage}{0.16\linewidth}\includegraphics[width=1\textwidth]{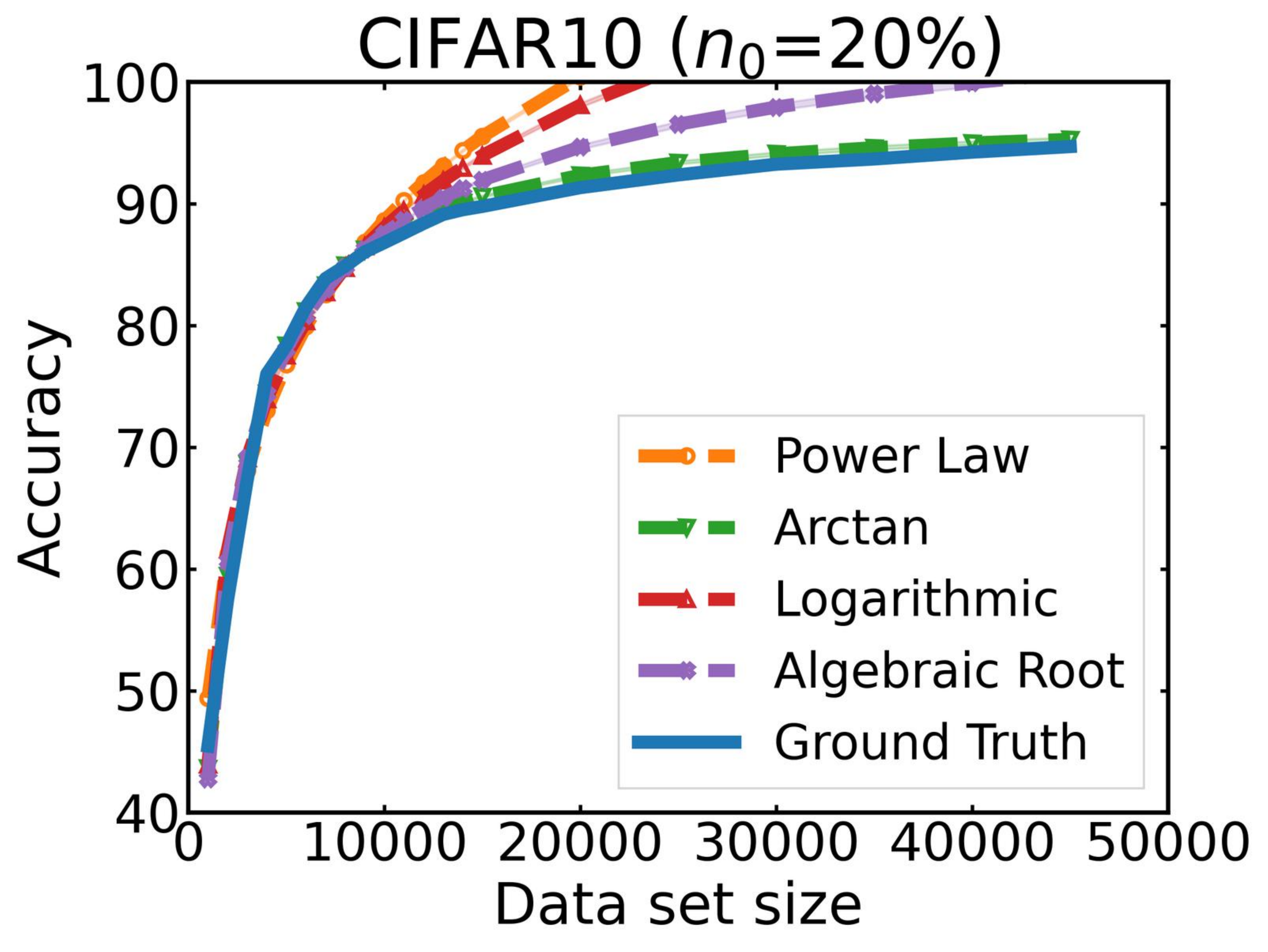} \end{minipage}
\begin{minipage}{0.16\linewidth}\includegraphics[width=1\textwidth]{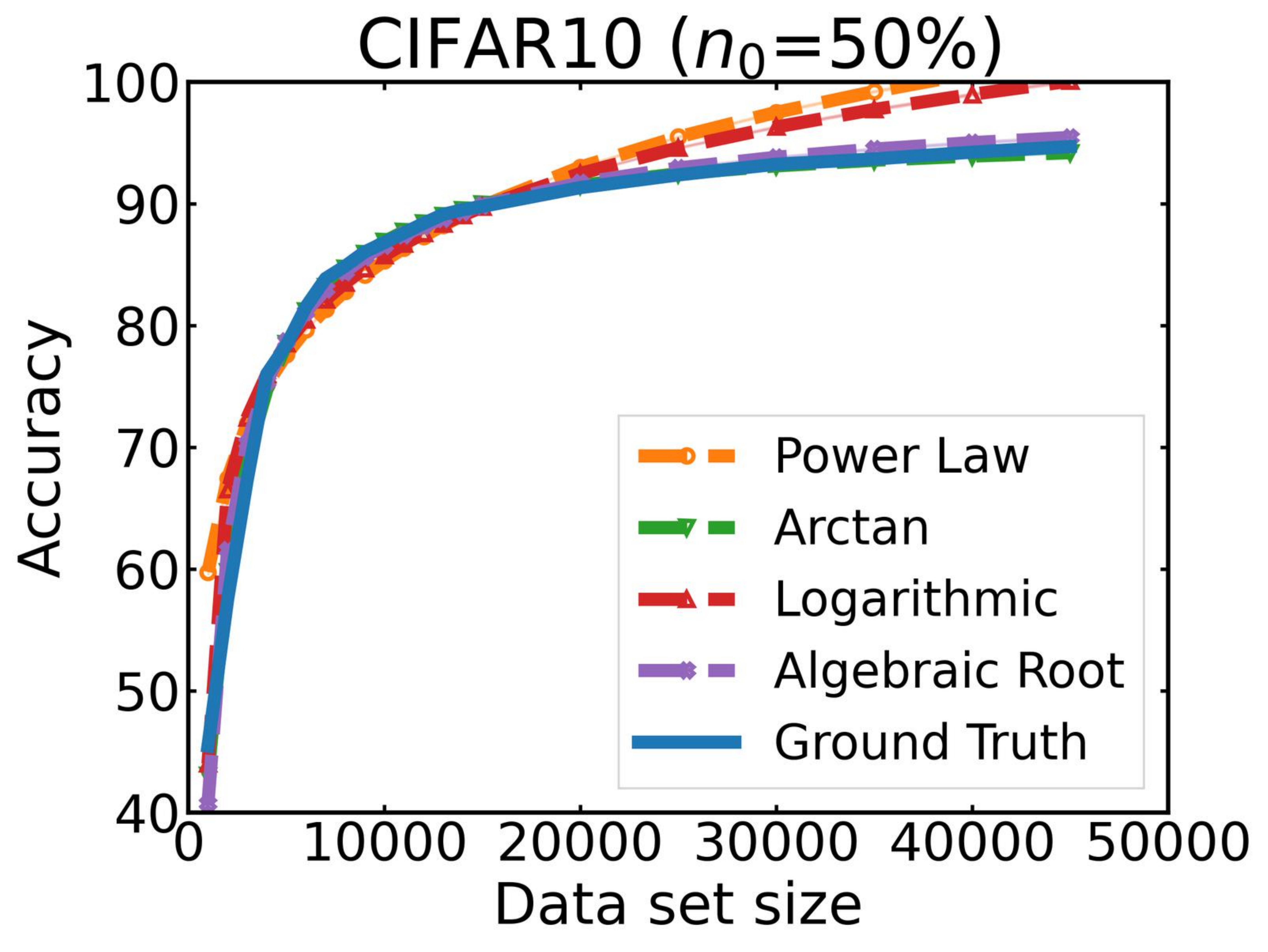} \end{minipage}
\begin{minipage}{0.16\linewidth}\includegraphics[width=1\textwidth]{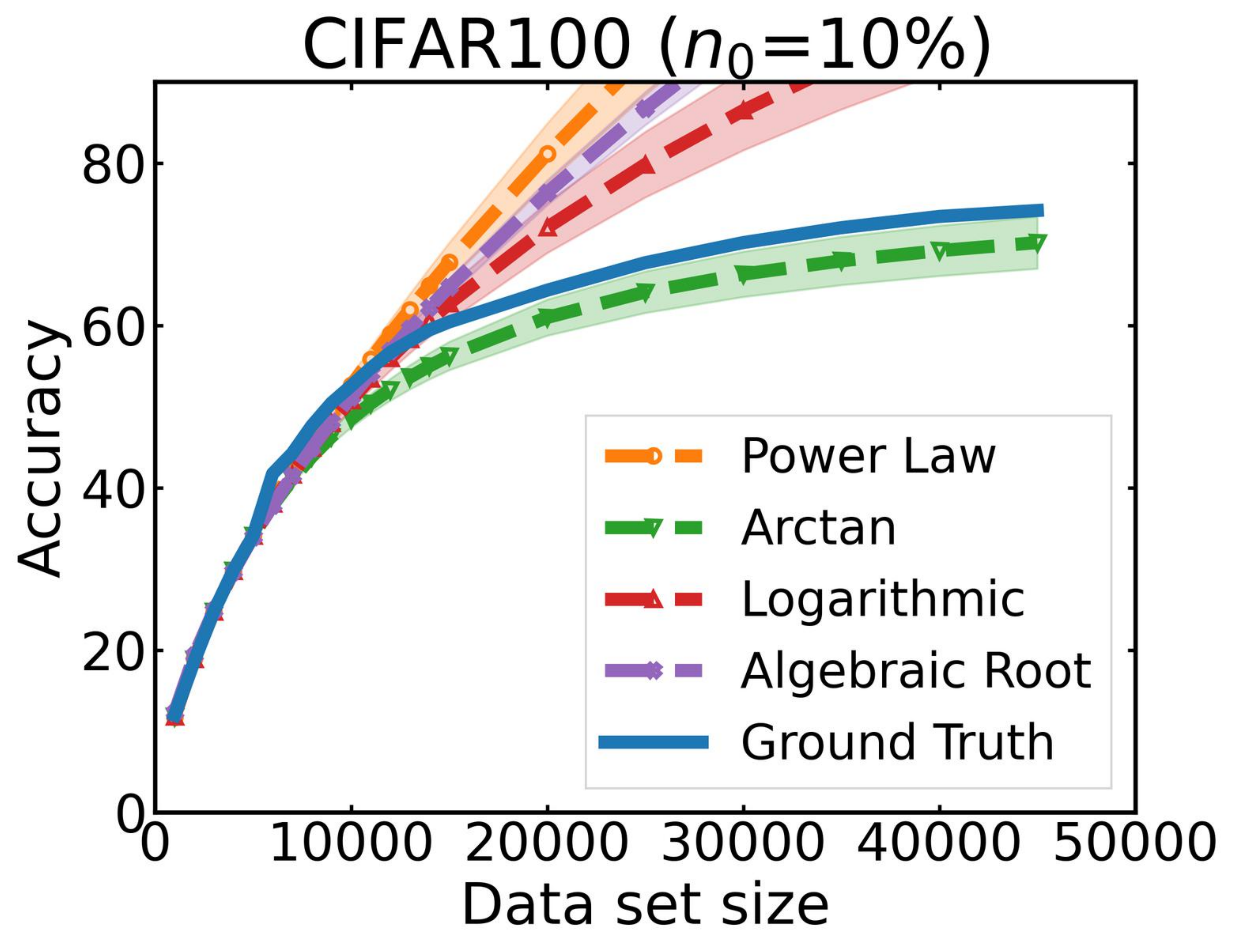} \end{minipage}
\begin{minipage}{0.16\linewidth}\includegraphics[width=1\textwidth]{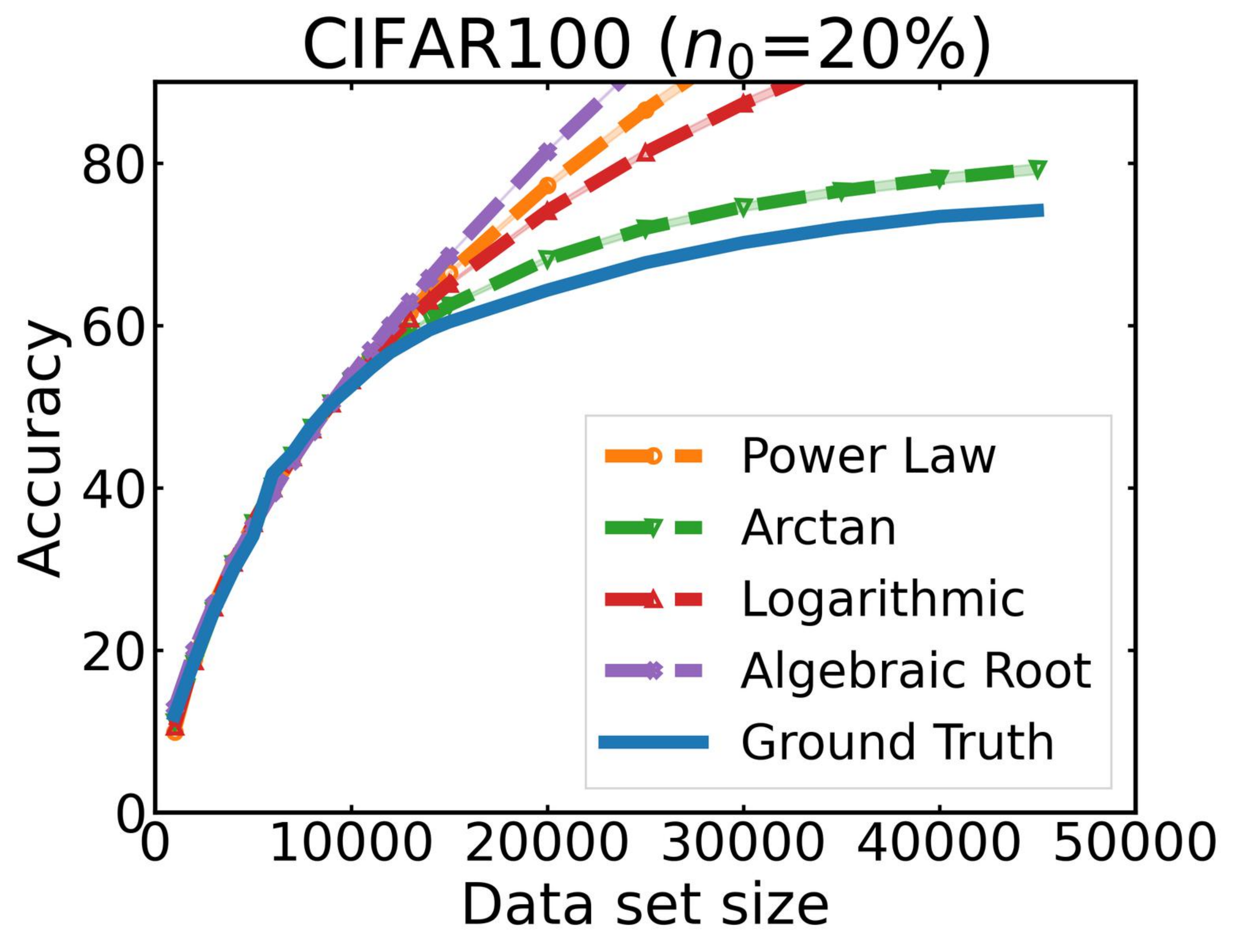} \end{minipage}
\begin{minipage}{0.16\linewidth}\includegraphics[width=1\textwidth]{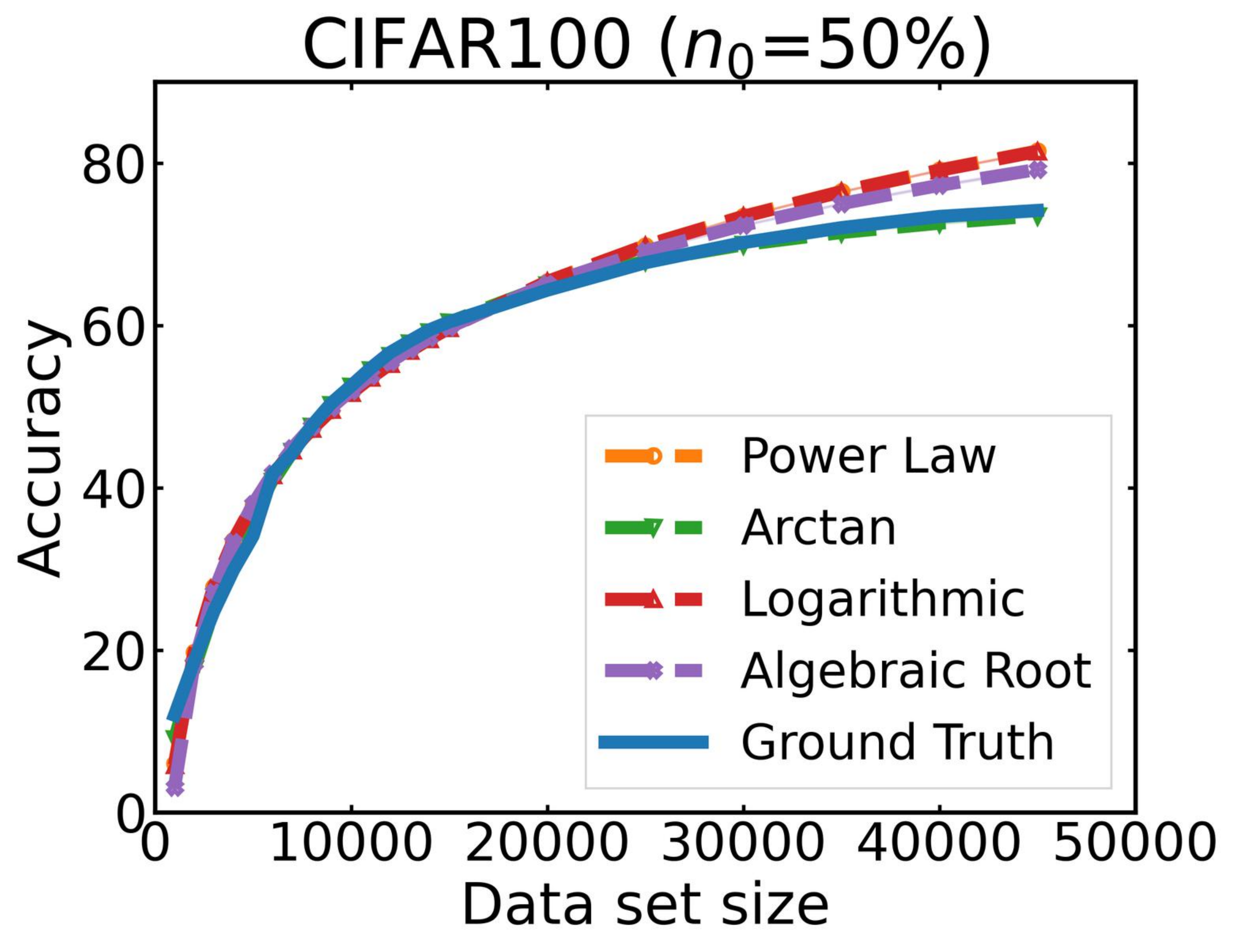} \end{minipage}
\begin{minipage}{0.16\linewidth}\includegraphics[width=1\textwidth]{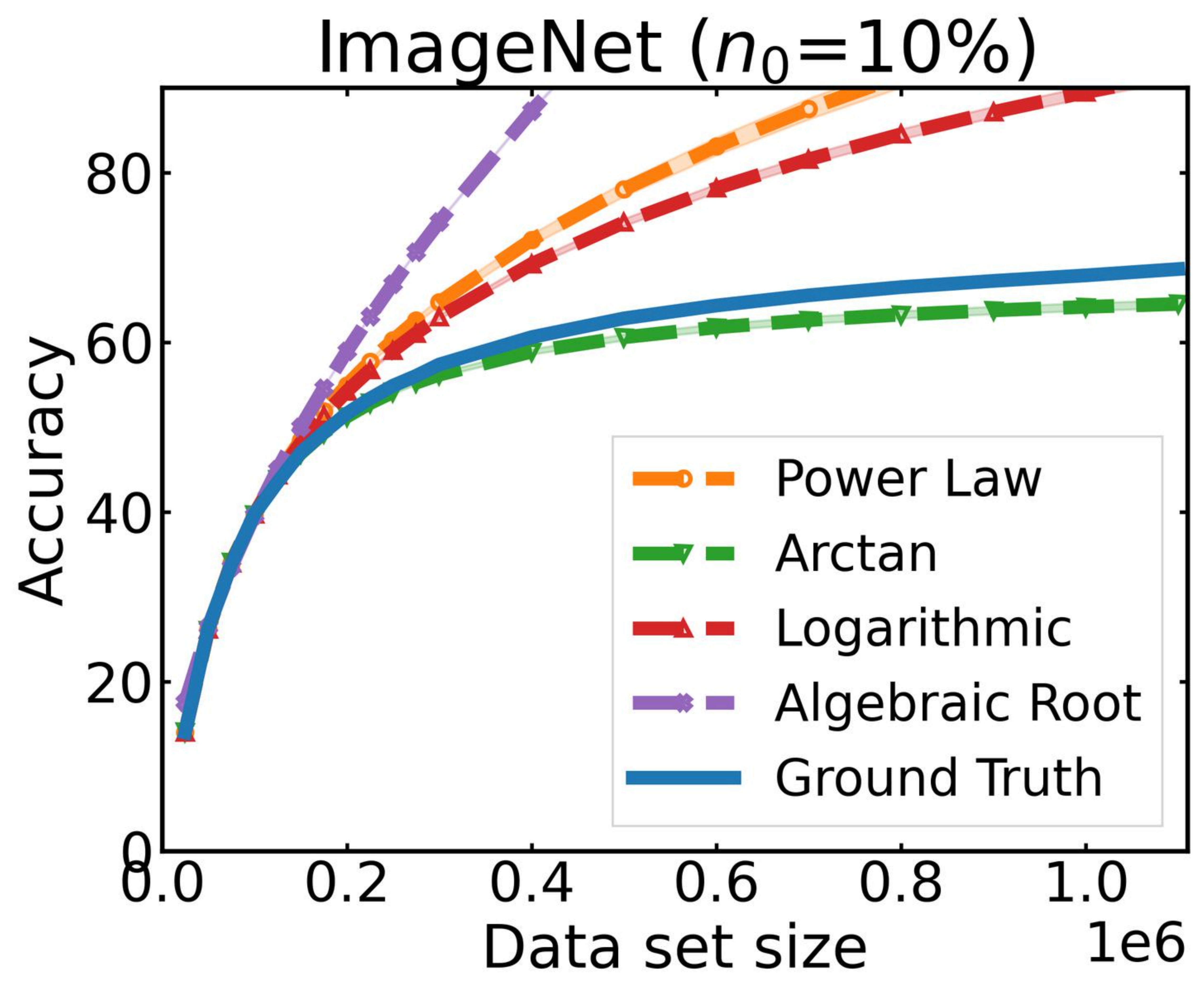} \end{minipage}
\begin{minipage}{0.16\linewidth}\includegraphics[width=1\textwidth]{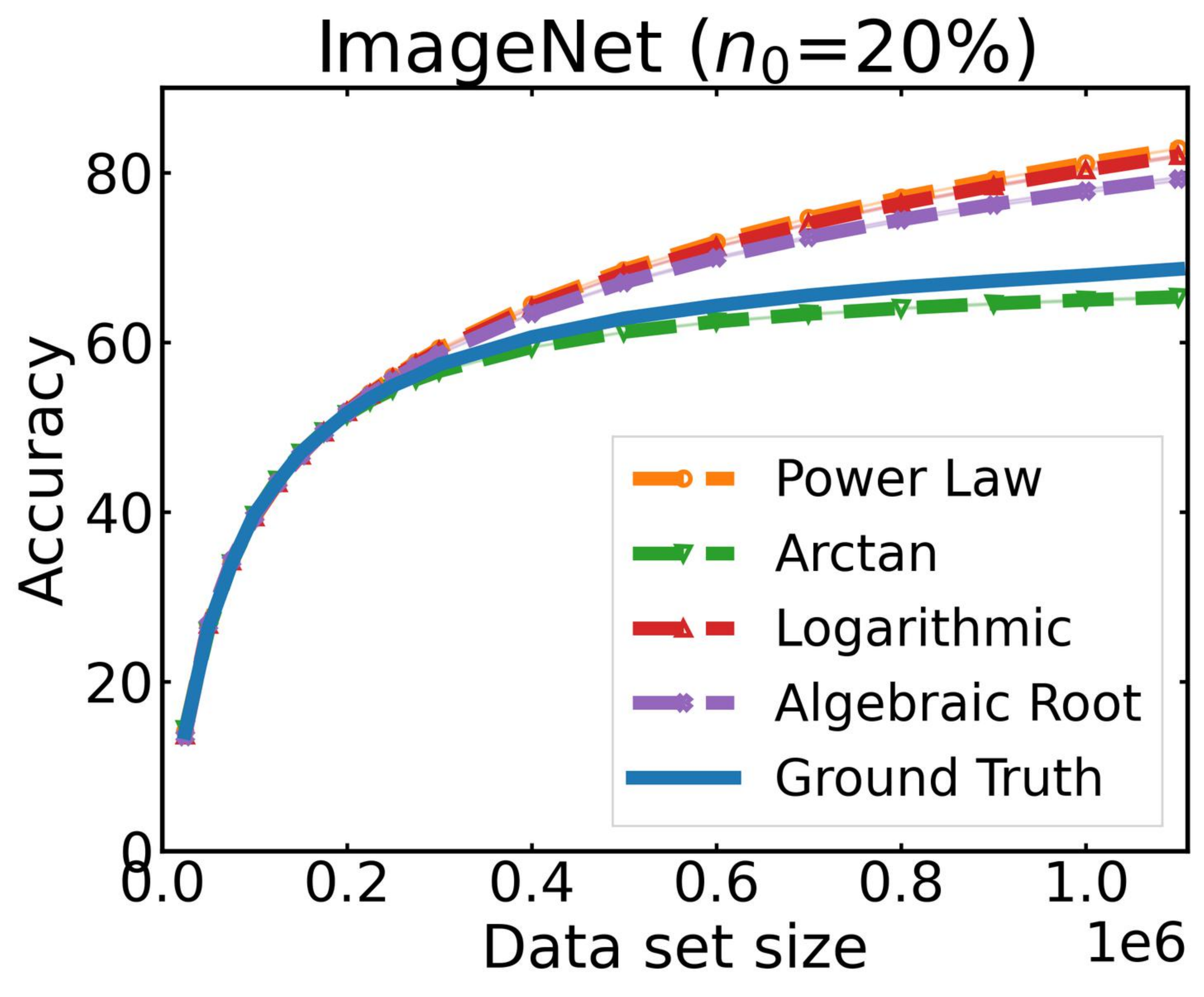} \end{minipage}
\begin{minipage}{0.16\linewidth}\includegraphics[width=1\textwidth]{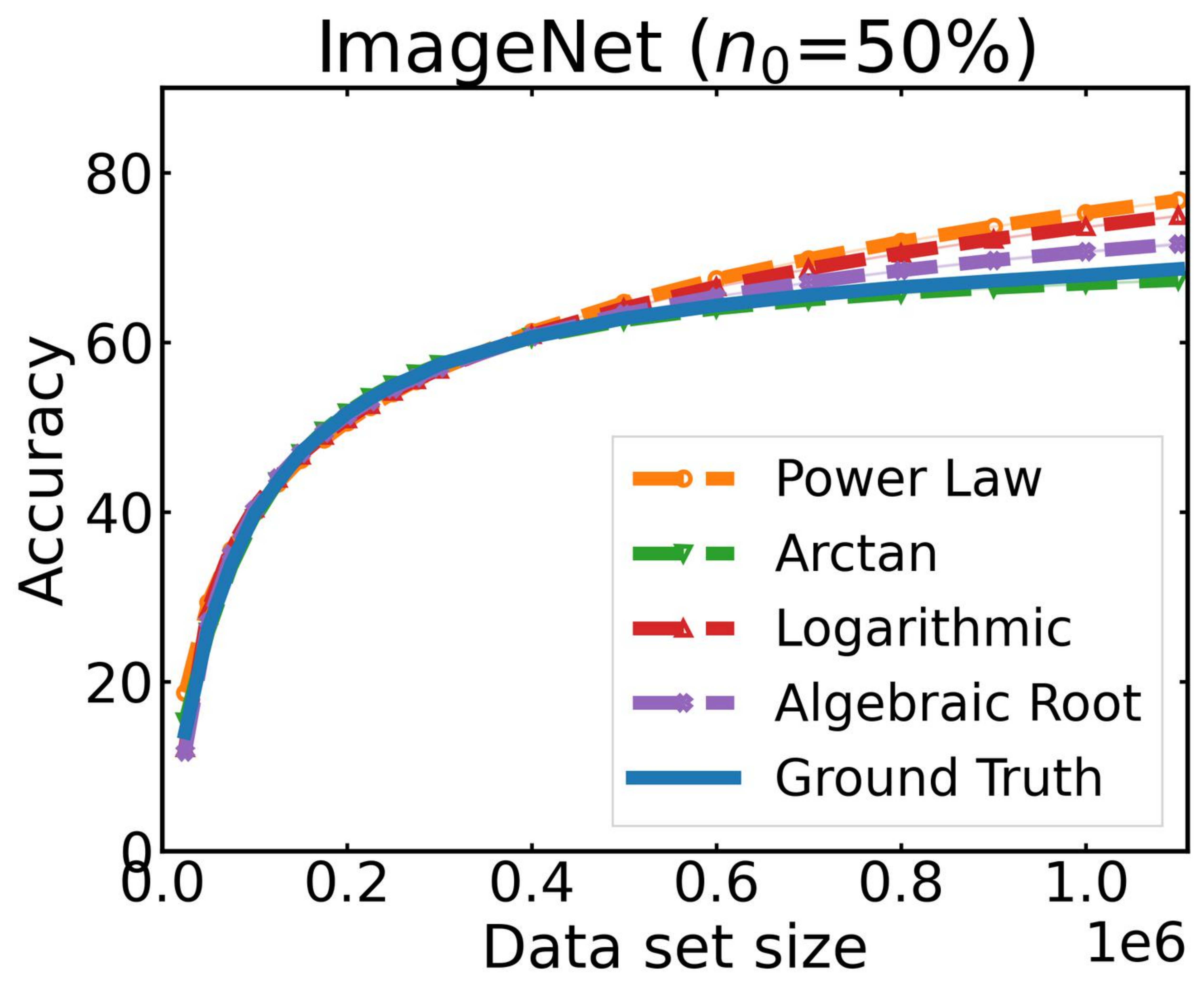} \end{minipage}
\begin{minipage}{0.16\linewidth}\includegraphics[width=1\textwidth]{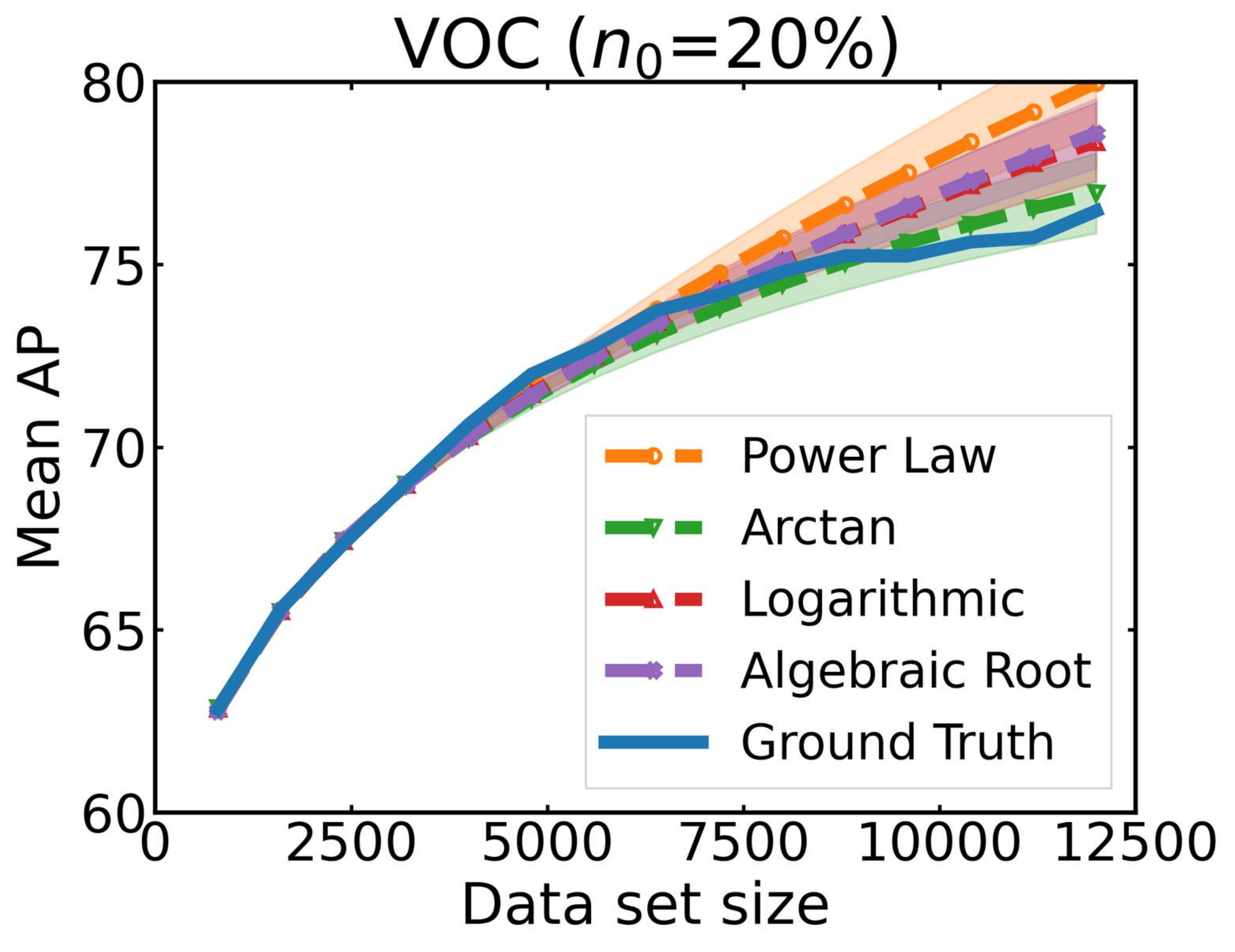} \end{minipage}
\begin{minipage}{0.16\linewidth}\includegraphics[width=1\textwidth]{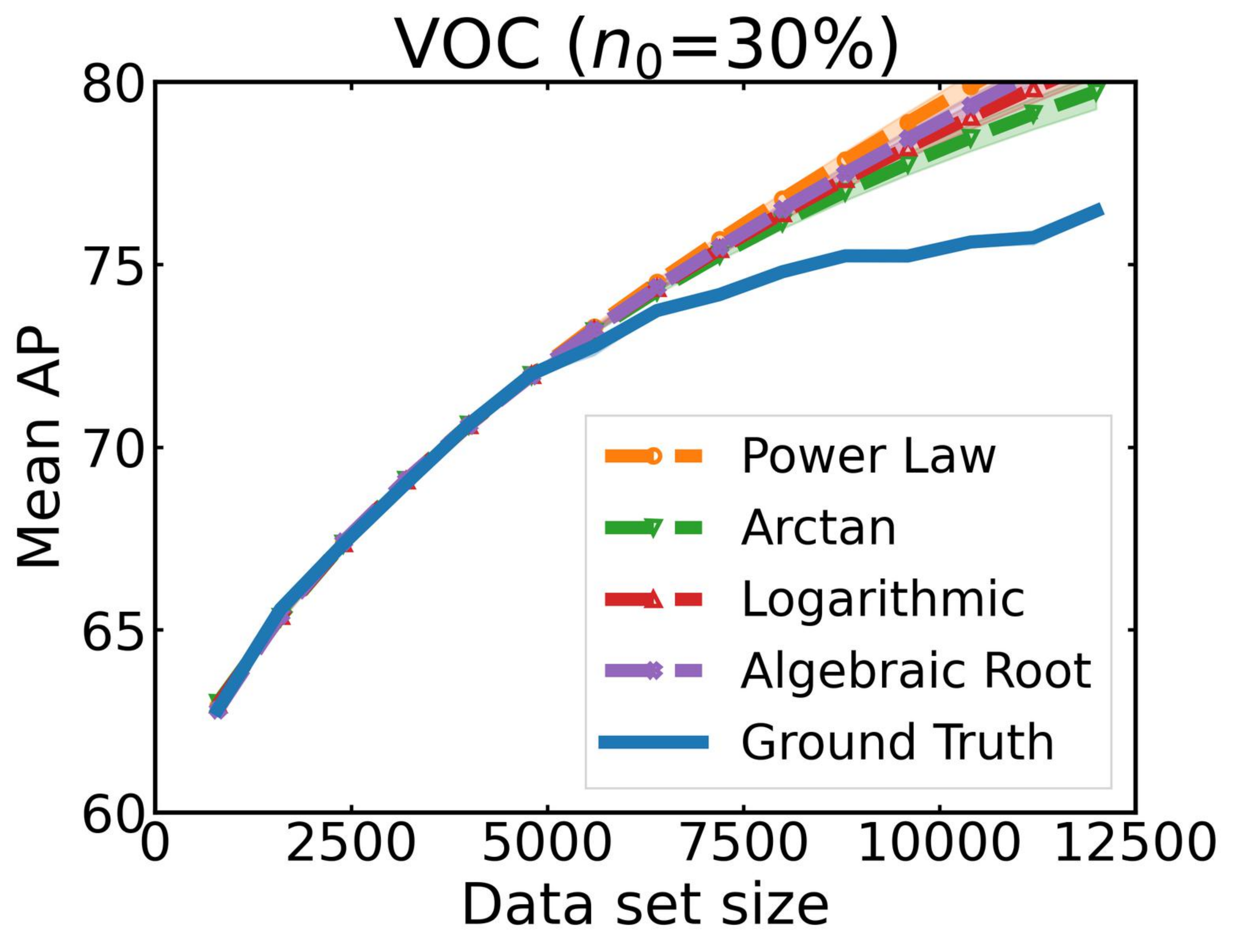} \end{minipage}
\begin{minipage}{0.16\linewidth}\includegraphics[width=1\textwidth]{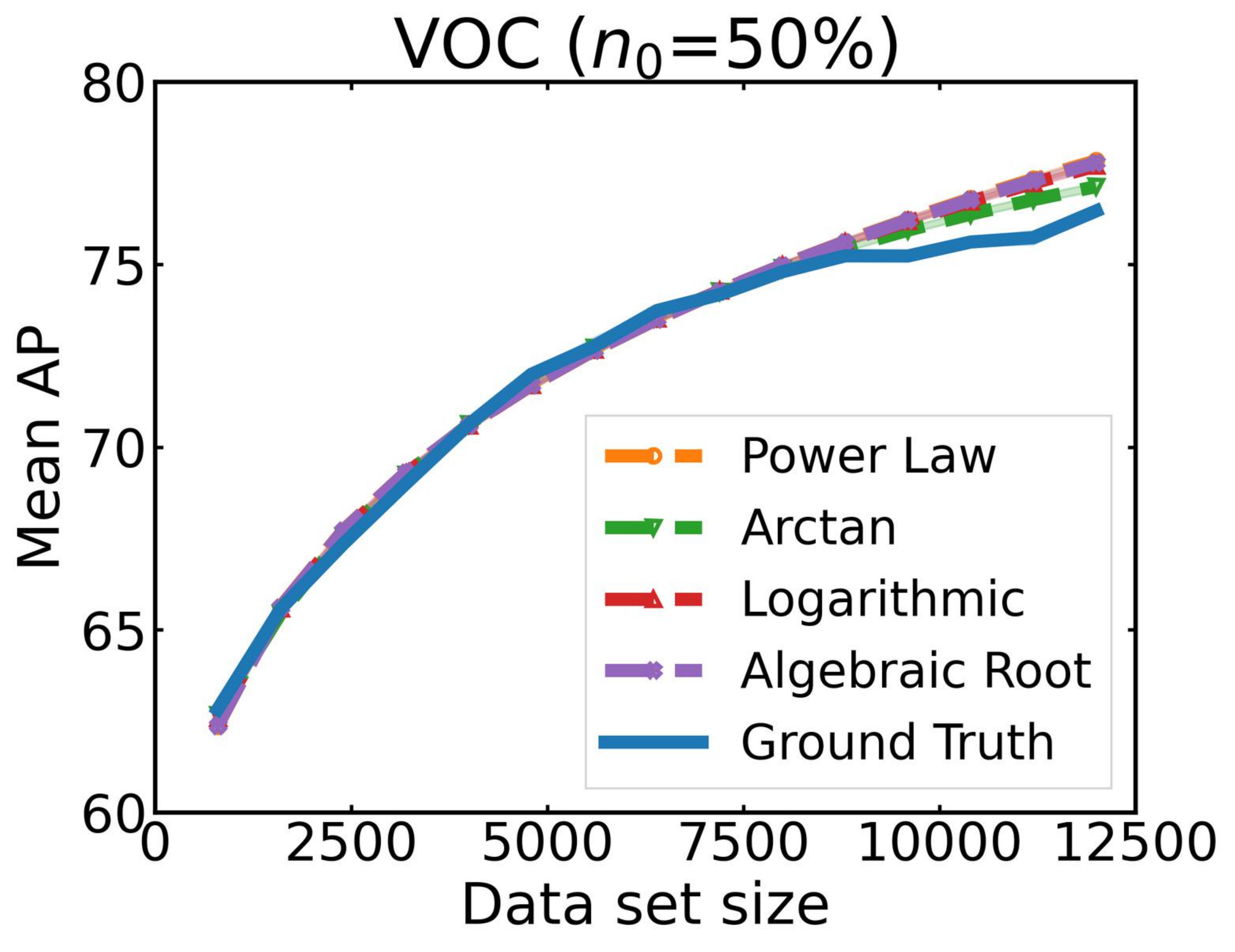} \end{minipage}
\begin{minipage}{0.16\linewidth}\includegraphics[width=1\textwidth]{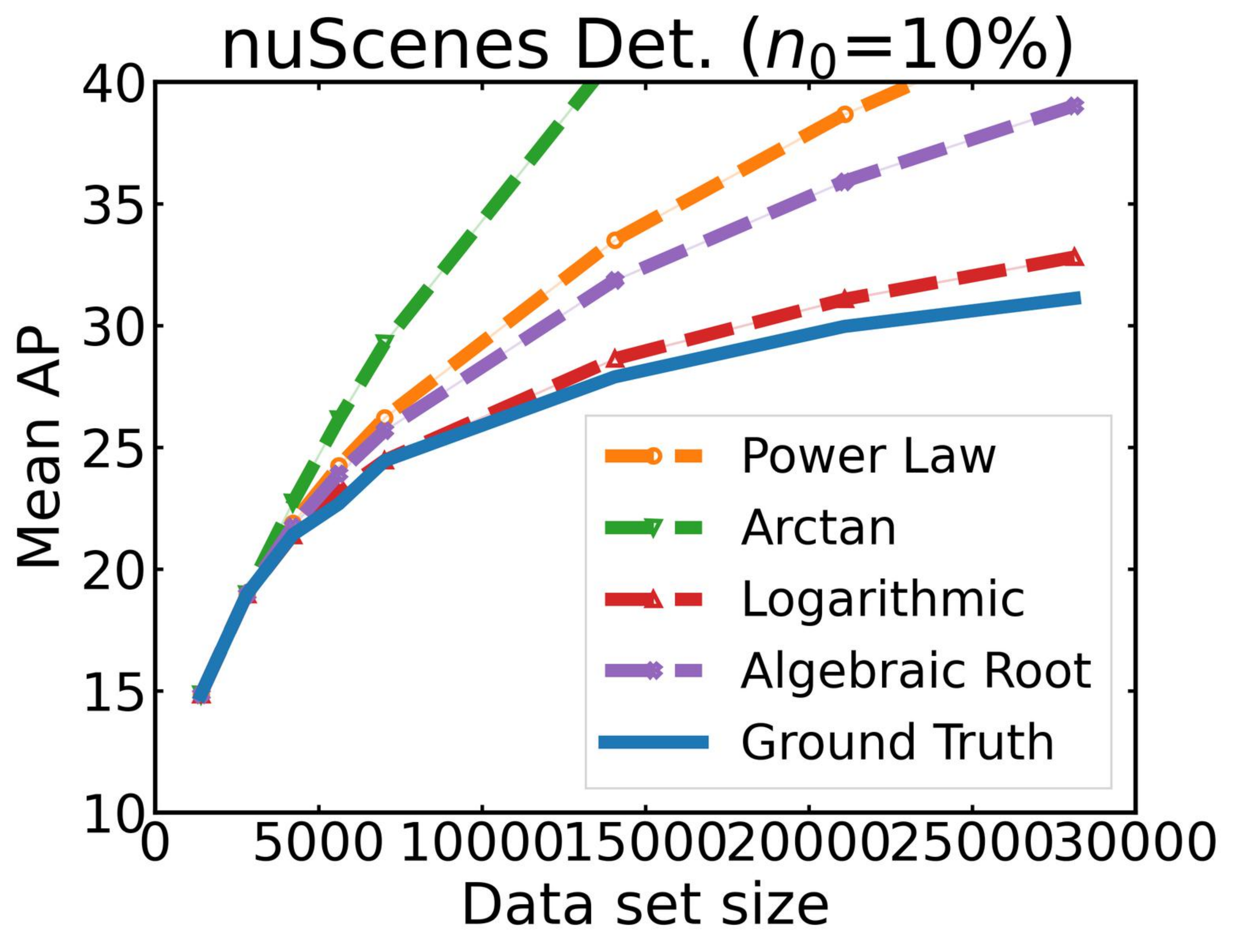} \end{minipage}
\begin{minipage}{0.16\linewidth}\includegraphics[width=1\textwidth]{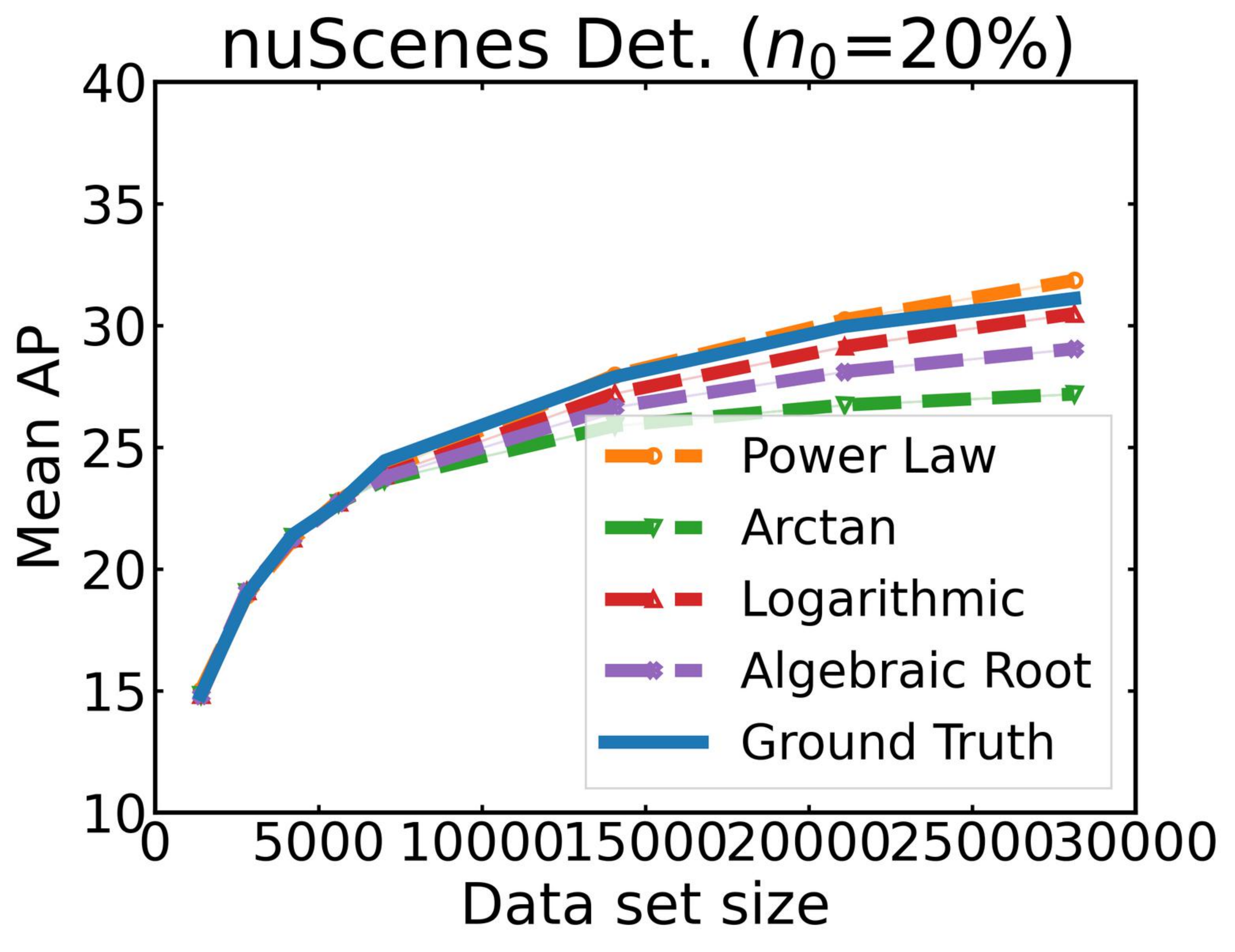} \end{minipage}
\begin{minipage}{0.16\linewidth}\includegraphics[width=1\textwidth]{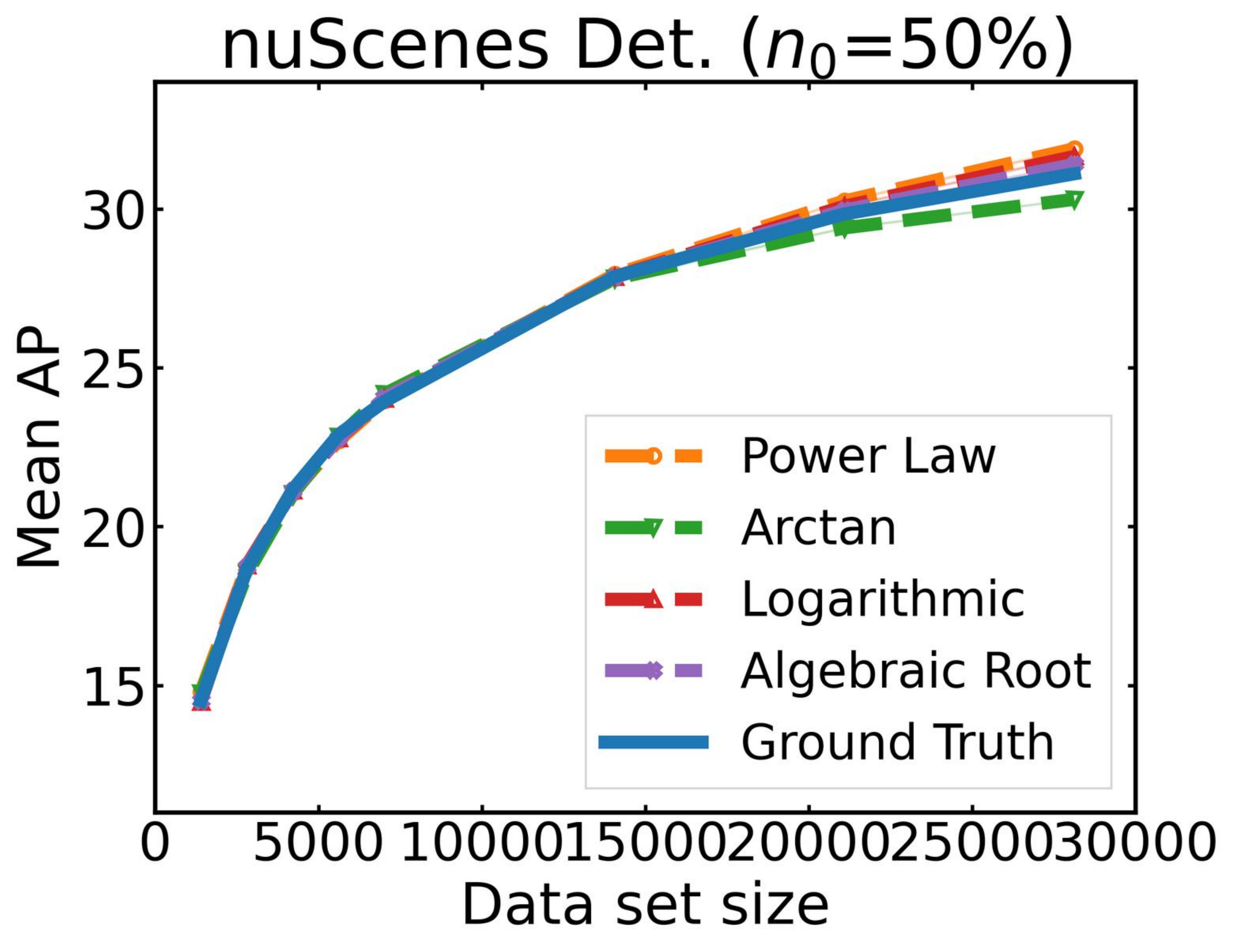} \end{minipage}
\begin{minipage}{0.16\linewidth}\includegraphics[width=1\textwidth]{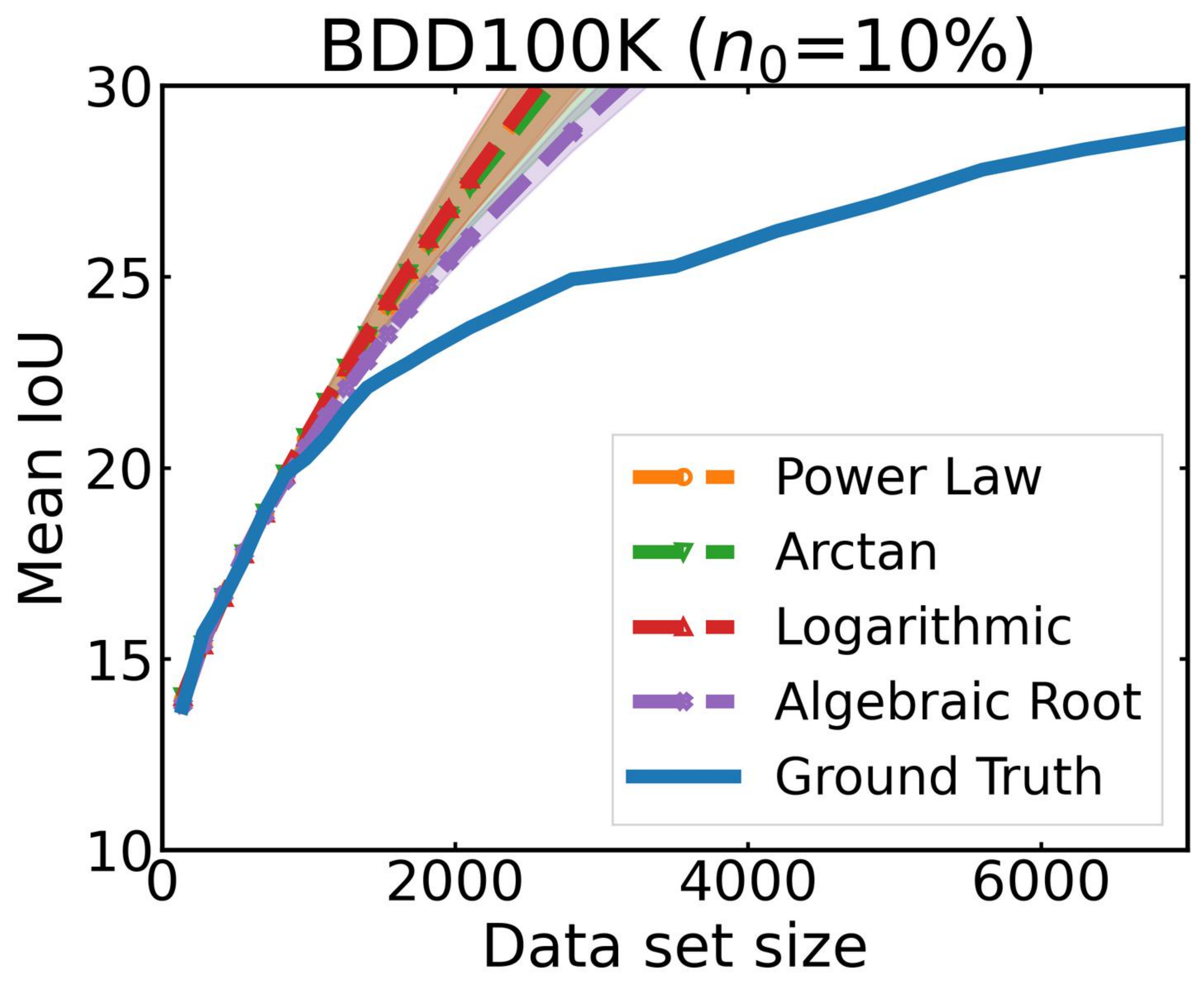} \end{minipage}
\begin{minipage}{0.16\linewidth}\includegraphics[width=1\textwidth]{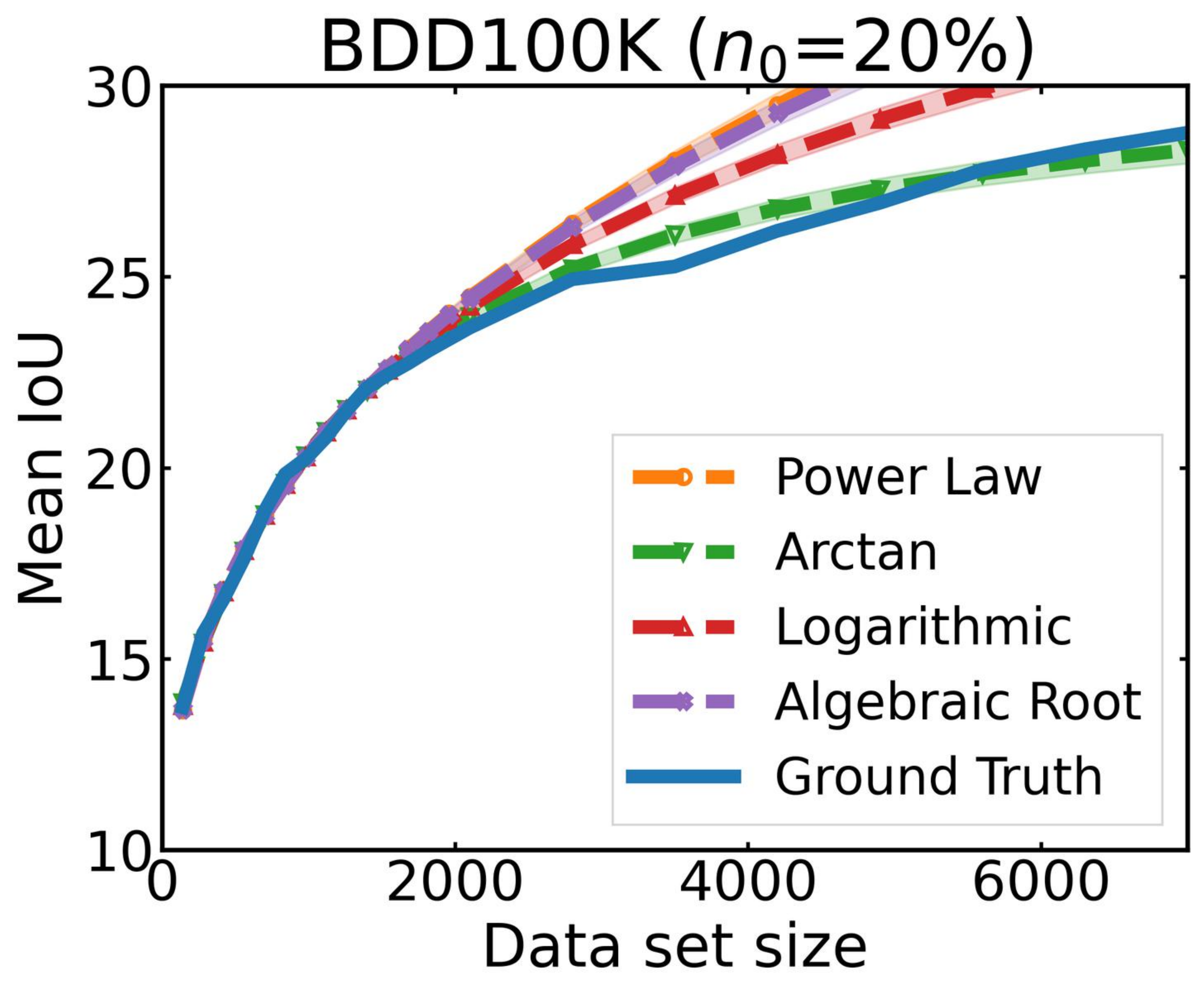} \end{minipage}
\begin{minipage}{0.16\linewidth}\includegraphics[width=1\textwidth]{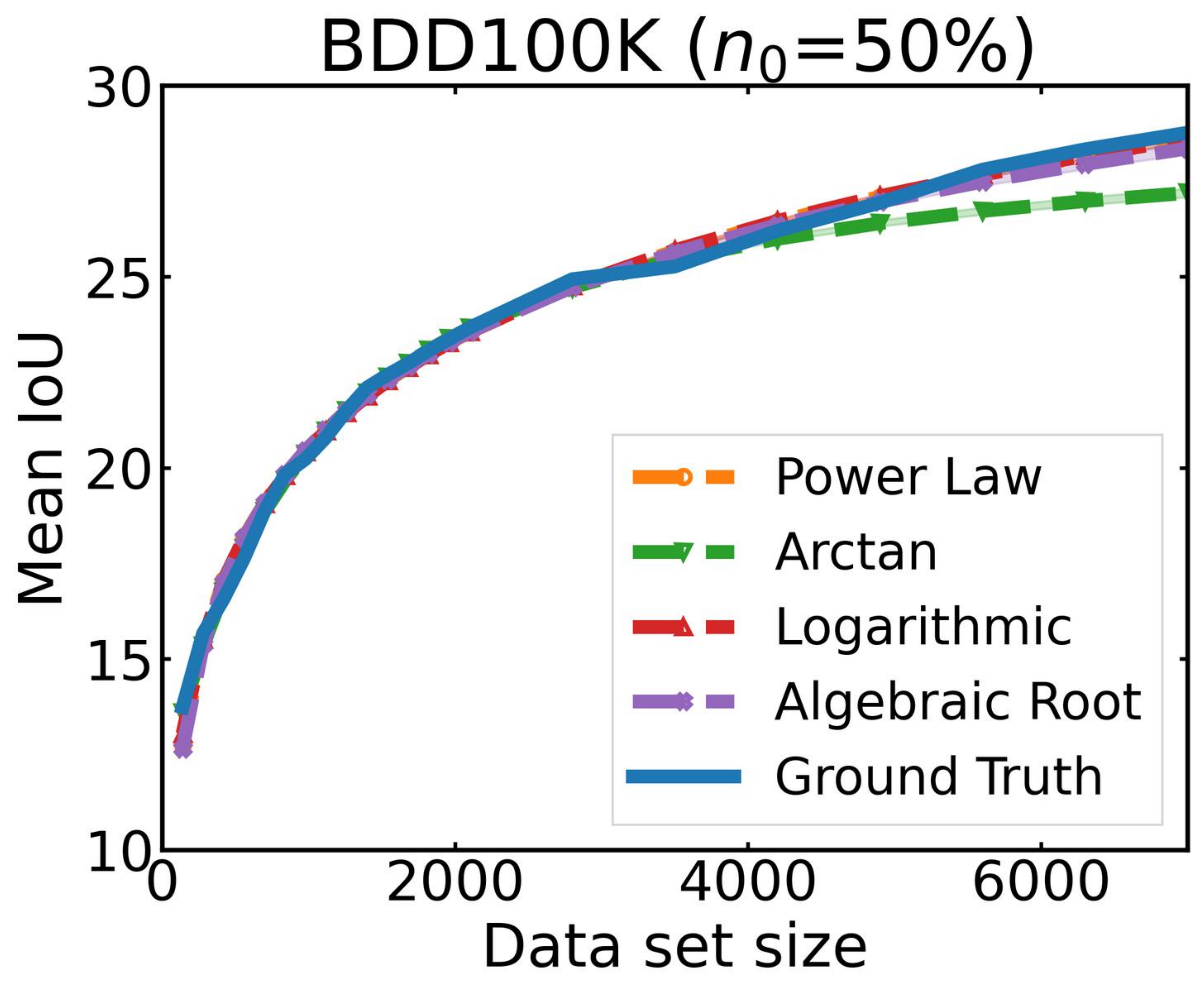} \end{minipage}
\begin{minipage}{0.16\linewidth}\includegraphics[width=1\textwidth]{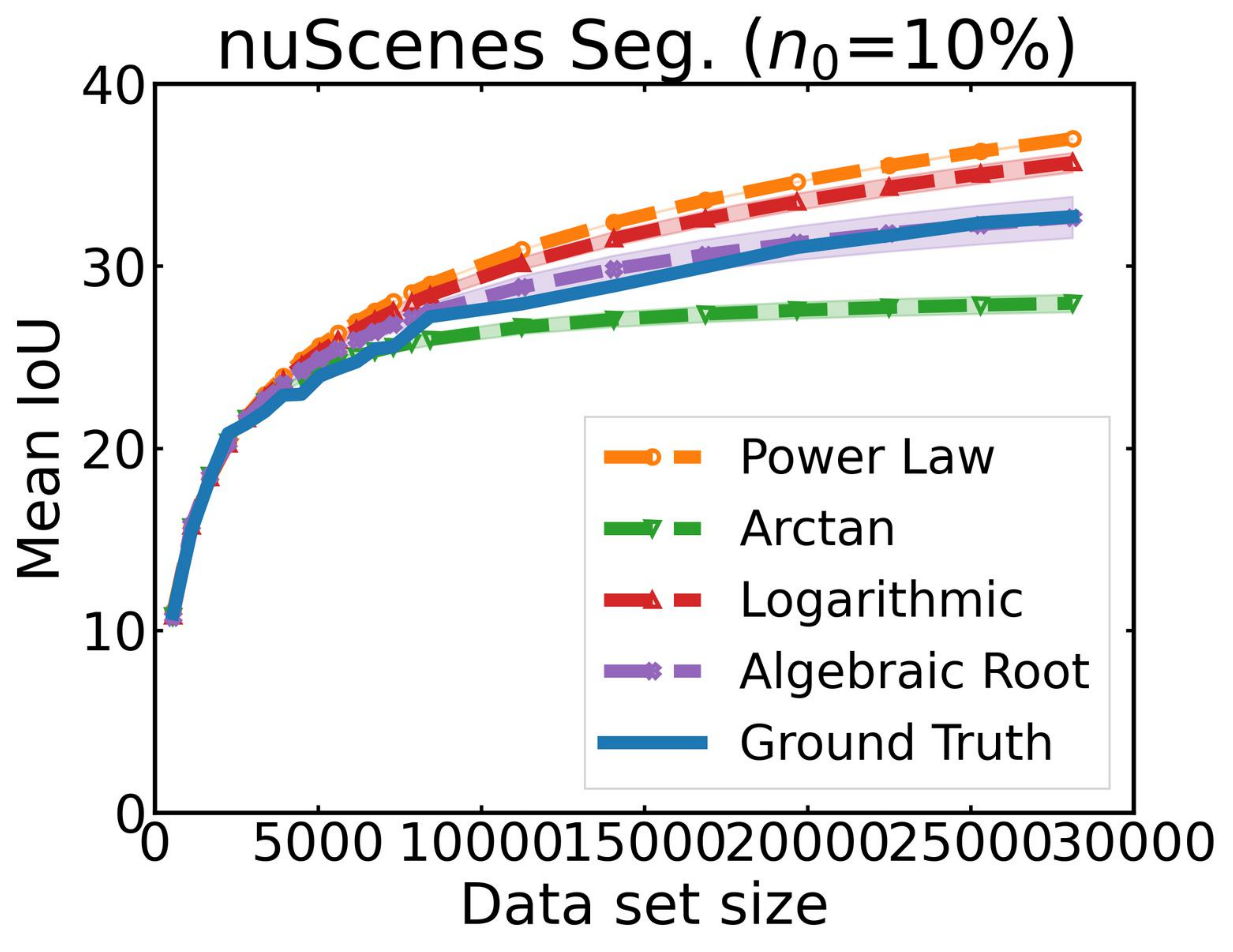} \end{minipage}
\begin{minipage}{0.16\linewidth}\includegraphics[width=1\textwidth]{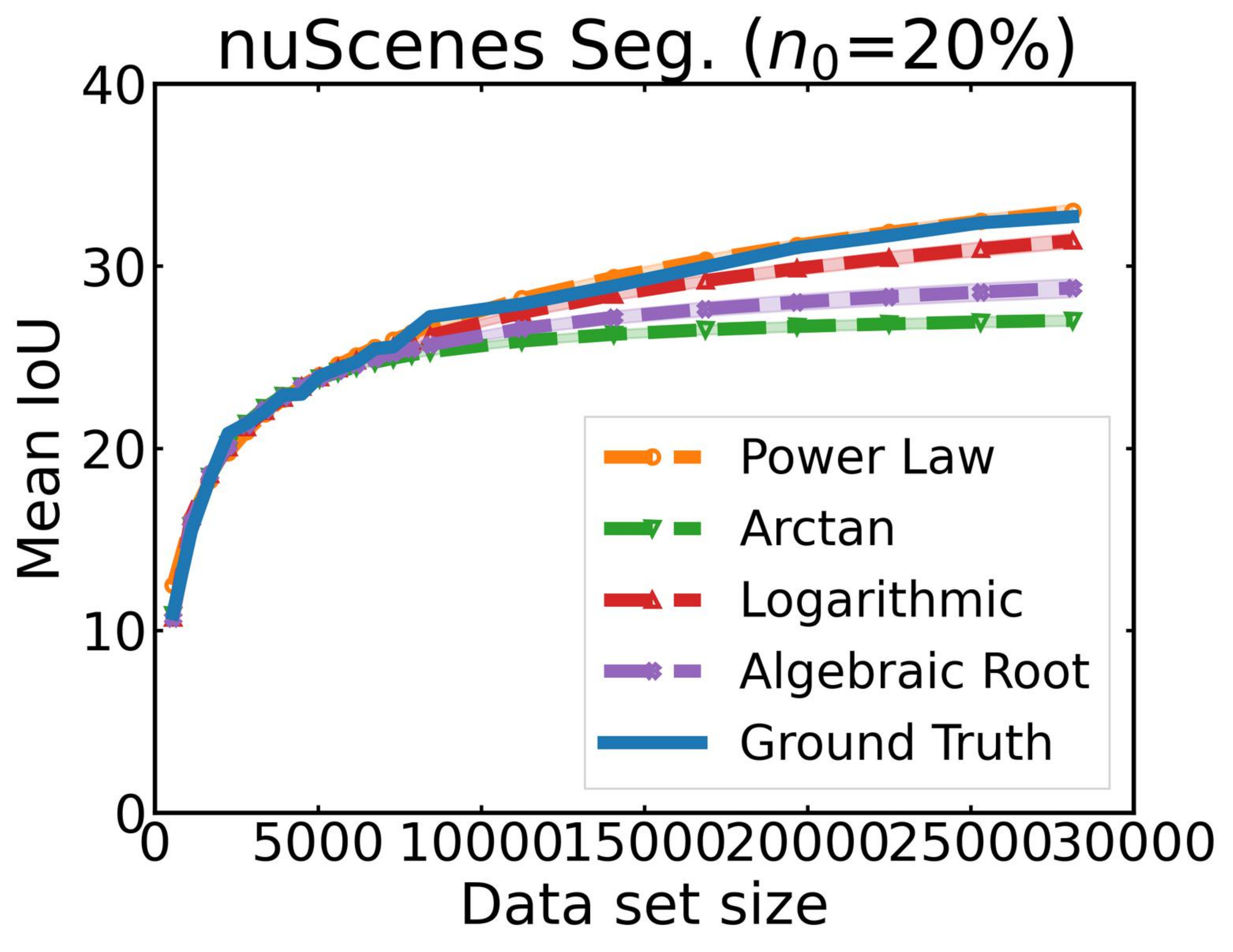} \end{minipage}
\begin{minipage}{0.16\linewidth}\includegraphics[width=1\textwidth]{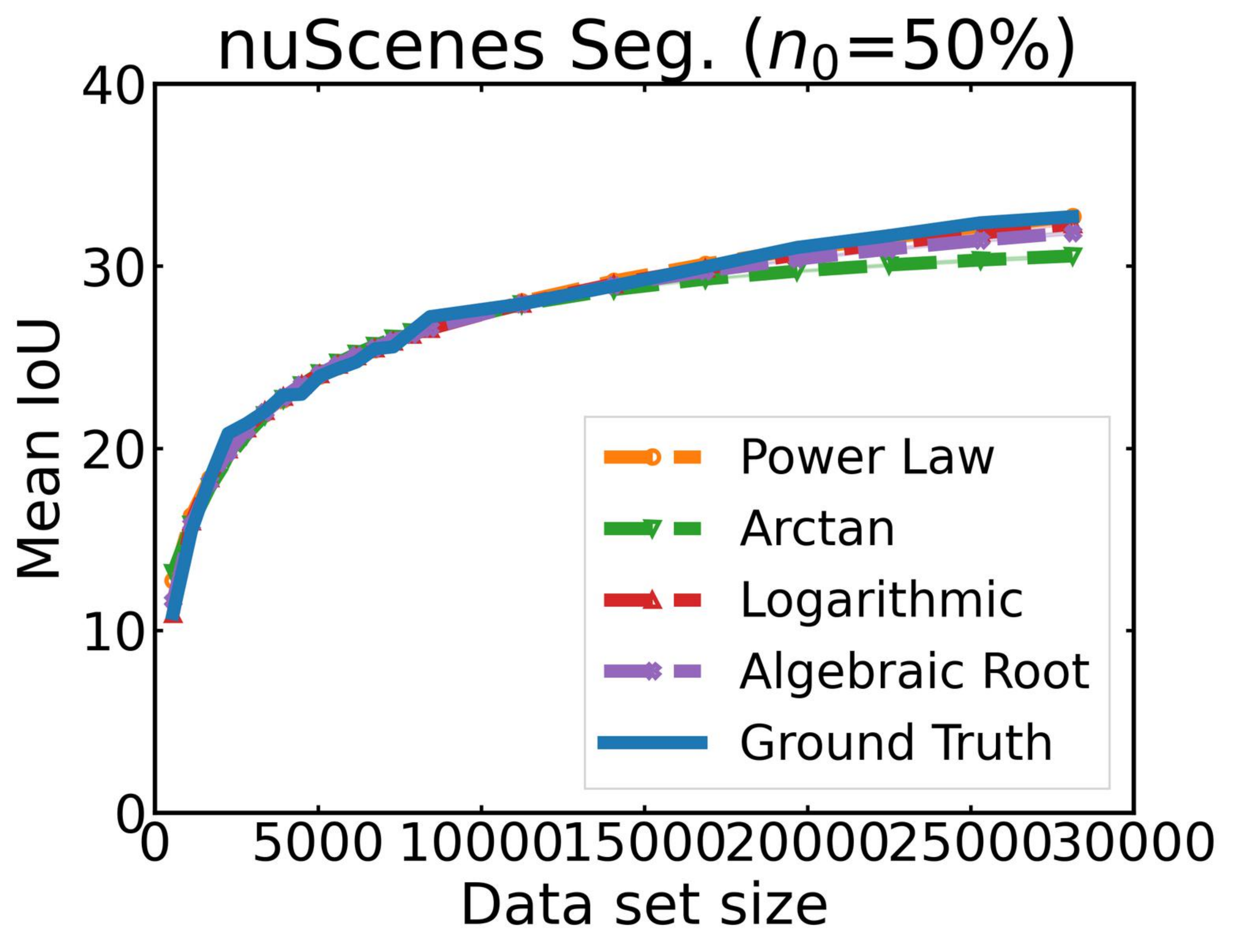} \end{minipage}
\begin{minipage}{0.5\linewidth}
\caption{\label{fig:regression_plots} 
Regression plots showing mean$\pm$standard deviation of multiple runs extrapolating performance in each task when trained on small subsets of the data. The solid blue line in each plot represents the ground truth performance.
}
\end{minipage}
\end{center}
\vspace{-7mm}
\end{figure*}

\begin{table*}[t!]
\begin{minipage}{0.73\linewidth}
    \centering
    \footnotesize
\begin{tabular}{clllcccc}
\toprule
         & Data set & $n_0$ & $r$ &       Power Law &           Arctan &      Logarithmic &   Algebraic Root \\
\midrule
\multirow{9}{*}{\rotatebox[origin=c]{90}{Classification}} &       
                    CIFAR10 &   $10\%$ &   5 &        $0.19 \pm 0.1$ &   $\fir{0.06 \pm 0.1}$ &         $0.17 \pm 0.1$ &   $0.17 \pm 0.1$  \\
               &    CIFAR10 &   $20\%$ &  10 &         $0.1 \pm 0.0$ &   $\fir{0.01 \pm 0.0}$ &         $0.08 \pm 0.0$ &   $0.04 \pm 0.0$  \\
               &    CIFAR10 &   $50\%$ &  17 &        $0.06 \pm 0.0$ &        $-0.01 \pm 0.0$ &         $0.04 \pm 0.0$ &   $\fir{0.01 \pm 0.0}$  \\ \cmidrule{2-8}
               &   CIFAR100 &   $10\%$ &   5 &         $0.1 \pm 0.2$ &        $-0.12 \pm 0.1$ &   $\fir{0.01 \pm 0.2}$ &   $0.06 \pm 0.1$  \\
               &   CIFAR100 &   $20\%$ &  10 &        $0.21 \pm 0.0$ &   $\fir{0.05 \pm 0.0}$ &         $0.15 \pm 0.0$ &   $0.27 \pm 0.0$  \\
               &   CIFAR100 &   $50\%$ &  17 &        $0.07 \pm 0.0$ &        $-0.01 \pm 0.0$ &         $0.07 \pm 0.0$ &   $\fir{0.05 \pm 0.0}$  \\ \cmidrule{2-8}
               &   ImageNet &   $10\%$ &   4 &        $0.18 \pm 0.1$ &        $-0.03 \pm 0.0$ &   $\fir{0.14 \pm 0.0}$ &   $0.38 \pm 0.0$  \\
               &   ImageNet &   $20\%$ &   8 &   $\fir{0.1 \pm 0.0}$ &        $-0.03 \pm 0.0$ &         $0.09 \pm 0.0$ &   $0.07 \pm 0.0$  \\
               &   ImageNet &   $50\%$ &  15 &        $0.07 \pm 0.0$ &        $-0.01 \pm 0.0$ &         $0.05 \pm 0.0$ &   $\fir{0.02 \pm 0.0}$  \\ \midrule
\multirow{6}{*}{\rotatebox[origin=c]{90}{Detection}} &       
                        VOC &   $20\%$ &   4 &        $0.02 \pm 0.0$ &         $-0.0 \pm 0.0$ &   $\fir{0.01 \pm 0.0}$ &   $\fir{0.01 \pm 0.0}$  \\
               &        VOC &   $30\%$ &   6 &        $0.04 \pm 0.0$ &   $\fir{0.02 \pm 0.0}$ &         $0.03 \pm 0.0$ &   $0.03 \pm 0.0$  \\
               &        VOC &   $50\%$ &  10 &  $\fir{0.01 \pm 0.0}$ &   $\fir{0.01 \pm 0.0}$ &   $\fir{0.01 \pm 0.0}$ &   $\fir{0.01 \pm 0.0}$  \\ \cmidrule{2-8}
               &   nuScenes &   $10\%$ &   2 &        $0.15 \pm 0.0$ &          $0.3 \pm 0.0$ &   $\fir{0.02 \pm 0.0}$ &   $0.11 \pm 0.0$  \\
               &   nuScenes &   $20\%$ &   4 &        $0.06 \pm 0.1$ &        $-0.03 \pm 0.1$ &         $0.04 \pm 0.1$ &   $\fir{0.03 \pm 0.1}$  \\
               &   nuScenes &   $50\%$ &   6 &        $0.02 \pm 0.0$ &        $-0.02 \pm 0.0$ &   $\fir{0.01 \pm 0.0}$ &   $\fir{0.01 \pm 0.0}$  \\ \midrule
\multirow{6}{*}{\rotatebox[origin=c]{90}{Segmentation}} &       
                   BDD100K  &   $10\%$ &   5 &         $0.2 \pm 0.1$ &   $\fir{0.17 \pm 0.2}$ &         $0.19 \pm 0.2$ &   $0.14 \pm 0.1$  \\
               &   BDD100K  &   $20\%$ &  10 &        $0.08 \pm 0.0$ &   $\fir{0.01 \pm 0.0}$ &         $0.05 \pm 0.0$ &   $0.08 \pm 0.0$  \\
               &   BDD100K  &   $50\%$ &  15 &        $0.05 \pm 0.0$ &        $-0.03 \pm 0.0$ &   $\fir{0.03 \pm 0.0}$ &   $0.04 \pm 0.0$  \\ \cmidrule{2-8}
               &  nuScenes  &   $10\%$ &   5 &        $0.09 \pm 0.0$ &        $-0.04 \pm 0.0$ &         $0.07 \pm 0.0$ &   $\fir{0.03 \pm 0.1}$  \\
               &  nuScenes  &   $20\%$ &  10 &  $\fir{0.01 \pm 0.0}$ &         $-0.1 \pm 0.0$ &        $-0.02 \pm 0.0$ &  $-0.07 \pm 0.0$  \\
               &  nuScenes  &   $50\%$ &  15 &  $\fir{0.01 \pm 0.0}$ &        $-0.08 \pm 0.0$ &        $-0.01 \pm 0.0$ &   $0.03 \pm 0.1$  \\
\bottomrule
\end{tabular}
\end{minipage} 
\begin{minipage}{0.27\linewidth}
    \caption{\label{tab:regression_logratio_tasks}
    Mean$\pm$standard deviation of multiple runs evaluating the mean log relative ratio for extrapolating performance in each task when trained on small subsets of the data. We report $n_0$ in terms of the percentage of the true data set. The lowest error (\ie smallest positive value) for each setting is bolded. 
    Given $50\%$ of the data, there is always at least one regression function achieving a log ratio less than $0.03$, whereas with $n_0 = 10\%$ of the data, we may achieve ratios as low as $-0.12$ or as high as $0.2$ (\ie an order of magnitude higher error compared to performance with $n_0 = 50\%$).
    }
    
\end{minipage}
\vspace{-5mm}
\end{table*}

\section{Further regression analysis}
\label{sec:app_further_regression}

In this section, we provide regression plots visualizing each of the functions from Table~\ref{tab:submodular_funcs} as well as error measurements using the log of the prediction ratio, which is an alternative metric to RMSE. This analysis supports the three challenges observed in Section~\ref{sec:setup_regression} and reinforces the necessity of our simulation analysis.

\noindent\textbf{Visualizing the regression models. }
Figure~\ref{fig:regression_plots} plots the regression functions versus the ground truth for each of the regression experiments in Section~\ref{sec:empirical_analysis}. These results support the RMSE errors given in Table~\ref{tab:regression_RMSE_archs}. That is, when $n_0 = 50\%$ of the full data set, all of the functions are close to the ground truth curve, but when $n_0 = 10\%$, the regression functions can diverge significantly. Moreover, the Arctan function is often the closest function to the ground truth, as reflected by RMSE.

We observe that the regression functions are almost always either optimistic (\ie their curve is above the ground truth) or pessimistic (\ie their curve is below) over the entire range of the data set size. 
In particular, the Arctan function is one of the two most pessimistic estimators for $19/21$ plots, Algebraic Root for $13/21$ plots, Logarithm for $7/21$ plots, and Power Law for $0/21$. 
This leads us to conclude that Arctan is generally a pessimistic estimator whereas Logarithmic and Power Law are generally optimistic. Algebraic Root lies in the middle.
We hypothesize that this is because when $\theta_3$ is held constant, both Arctan and Algebraic Root converge to a finite value as $n\rightarrow +\infty$ (\ie they flatten), but Power Law and Logarithm are unbounded.

Finally, we remark on the shape of the ground truth curves. Recall from Section~\ref{sec:setup_regression} that we observe the ground truth score function $v(n)$ to be piecewise linear, concave, and monotonically increasing. Figure~\ref{fig:regression_plots} shows that this observation generally holds for CIFAR10, CIFAR100, ImageNet, and the two nuScenes tasks. However for VOC and BDD100K, we observe that the curves are not always concave and monotonically increasing. 
Nonetheless, our observations about using regression to evaluate data requirement estimation persists. Moreover, our proposed techniques for estimating data requirements remain effective, emphasizing that the initial observation of $v(n)$ from Section~\ref{sec:setup_regression} is not a theoretical requirement but a motivating trend.

\noindent\textbf{Alternative metrics to RMSE. }
When evaluating regression functions on their ability to estimate data requirements, we must be able to differentiate between estimators that over- versus under-estimate model performance (and therefore under- or over-estimate data requirements). Since RMSE is not a signed metric, it does not provide this information, and we consequently explore alternative signed metrics. 
Table~\ref{tab:regression_logratio_tasks} evaluates each regression function on the log of the relative ratio of estimation error $\log\hv(n;\btheta^*) - \log v^*(n)$ when extrapolating to larger data sets. The log relative ratio is a signed metric where negative values means that we are pessimistic (\ie $\hv(n;\btheta^*) < v^*(n)$ on average) and positive values means that we are optimistic. Ideally, we want the smallest positive log relative ratio.

Unlike Table~\ref{tab:regression_RMSE_tasks}, the Arctan function does not consistently dominate on any individual task when evaluating on the log relative ratio. However, Table~\ref{tab:regression_logratio_tasks} also does not reveal any clear best performing regression function. Specifically, Power Law, Arctan, Logarithmic, and Algebraic Root each rank the best 4, 7, 7, and 8 times, respectively.
This result supports our belief that evaluating regression error alone does not permit us to identify a good data collection policy.

\begin{table*}[]
\begin{minipage}{0.76\linewidth}
    \centering
    \footnotesize
\begin{tabular}{lllcccc}
\toprule
         Architecture & $n_0$ & $r$ & Power Law &          Arctan &       Logarithmic &    Algebraic Root \\
\midrule
ResNet18 (baseline) &  10\% &  5 &  $34.38 \pm 35.1$ &   $\fir{13.3 \pm 5.3}$ &  $17.25 \pm 21.8$ &  $26.29 \pm 16.8$  \\
ResNet18 (baseline) &  20\% & 10 &   $29.52 \pm 3.9$ &   $\fir{4.71 \pm 2.0}$ &   $19.87 \pm 2.5$ &   $40.33 \pm 1.5$  \\
ResNet18 (baseline) &  50\% & 17 &    $5.49 \pm 0.2$ &   $\fir{0.69 \pm 0.2}$ &    $5.42 \pm 0.2$ &    $3.65 \pm 0.3$  \\ \midrule
ResNet50            &  10\% &  5 &$275.98 \pm 521.1$ & $44.94 \pm 41.7$ &  $49.43 \pm 72.1$ &  $\fir{24.01 \pm 7.1}$  \\
ResNet50            &  20\% & 10 &  $31.04 \pm 13.1$ &   $\fir{4.14 \pm 4.4}$ &   $21.48 \pm 8.3$ &  $37.73 \pm 3.4$  \\
ResNet50            &  50\% & 17 &   $6.57 \pm 1.0$  &   $\fir{1.07 \pm 0.8}$ &   $6.71 \pm 1.4$ &  $4.8 \pm 1.7$  \\ \midrule
ResNet101           &  10\% &  5 & $88.47 \pm 115.3$ & $\fir{25.53 \pm 17.1}$ &  $46.73 \pm 60.9$ &  $26.96 \pm 4.5$  \\
ResNet101           &  20\% & 10 &  $47.34 \pm 12.1$ &  $\fir{10.23 \pm 4.7}$ &   $32.35 \pm 8.5$ &  $40.89 \pm 1.5$  \\
ResNet101           &  50\% & 17 &    $8.07 \pm 0.3$ &   $\fir{0.78 \pm 0.3}$ &   $7.95 \pm 0.3$ &  $5.57 \pm 0.5$ \\ \midrule
WideResNet-16-4     &  10\% &  5 &  $34.38 \pm 35.1$ &   $\fir{13.3 \pm 5.3}$ &  $17.25 \pm 21.8$ &  $26.29 \pm 16.8$  \\
WideResNet-16-4     &  20\% & 10 &   $13.78 \pm 1.8$ &   $\fir{0.99 \pm 0.1}$ &  $12.86 \pm 0.6$ &  $12.55 \pm 1.8$ \\
WideResNet-16-4     &  50\% & 17 &    $5.35 \pm 0.3$ &   $\fir{1.61 \pm 0.4}$ &   $4.24 \pm 0.3$ &  $1.55 \pm 0.5$  \\ \midrule
WideResNet-16-8     &  10\% &  5 &   $57.7 \pm 19.3$ &    $\fir{5.0 \pm 3.8}$ & $33.34 \pm 10.2$ &  $65.31 \pm 2.2$ \\ 
WideResNet-16-8     &  20\% & 10 &   $14.01 \pm 1.8$ &   $\fir{1.04 \pm 0.3}$ &  $13.62 \pm 1.2$ &  $12.74 \pm 1.7$ \\
WideResNet-16-8     &  50\% & 17 &    $5.57 \pm 0.2$ &   $1.87 \pm 0.1$ &   $4.32 \pm 0.2$ &  $\fir{1.4 \pm 0.2}$  \\ 
\bottomrule
\end{tabular}
\end{minipage}
\begin{minipage}{0.19\linewidth}
    \caption{For different architectures with CIFAR100, mean$\pm$standard deviation of multiple runs evaluating the RMSE on extrapolating performance when trained on small subsets of the data. The lowest error for each architecture is bolded. These results reinforce the initial results for CIFAR100 in Table~\ref{tab:regression_RMSE_tasks}, as the Arctan function consistently dominates in nearly every setting.
    }
    \label{tab:regression_RMSE_archs}
\end{minipage}
\vspace{-4mm}
\end{table*}

\begin{figure*}[!t]
\begin{center}
\begin{minipage}{0.16\linewidth}\includegraphics[width=1\textwidth]{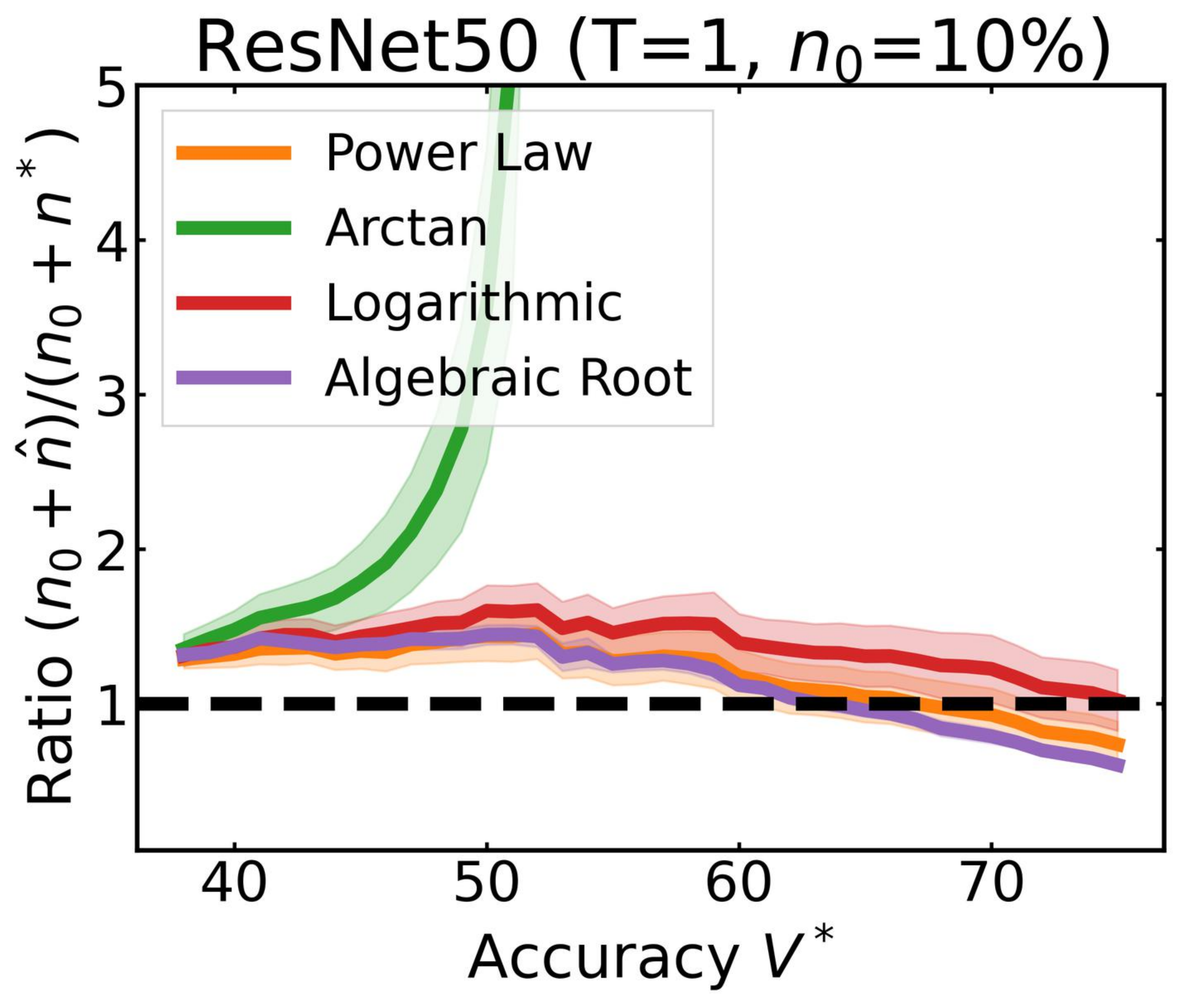} \end{minipage}
\begin{minipage}{0.16\linewidth}\includegraphics[width=1\textwidth]{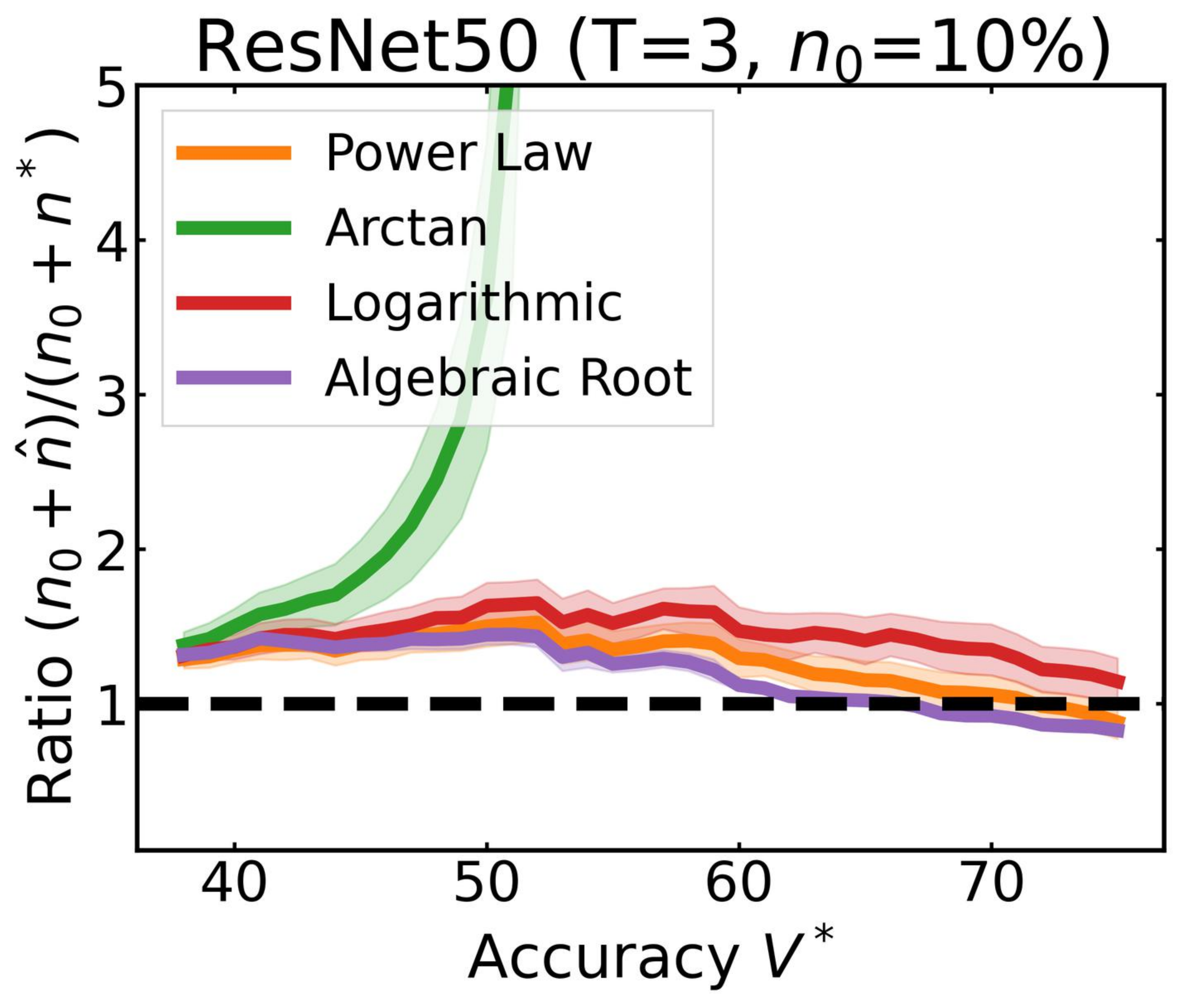} \end{minipage}
\begin{minipage}{0.16\linewidth}\includegraphics[width=1\textwidth]{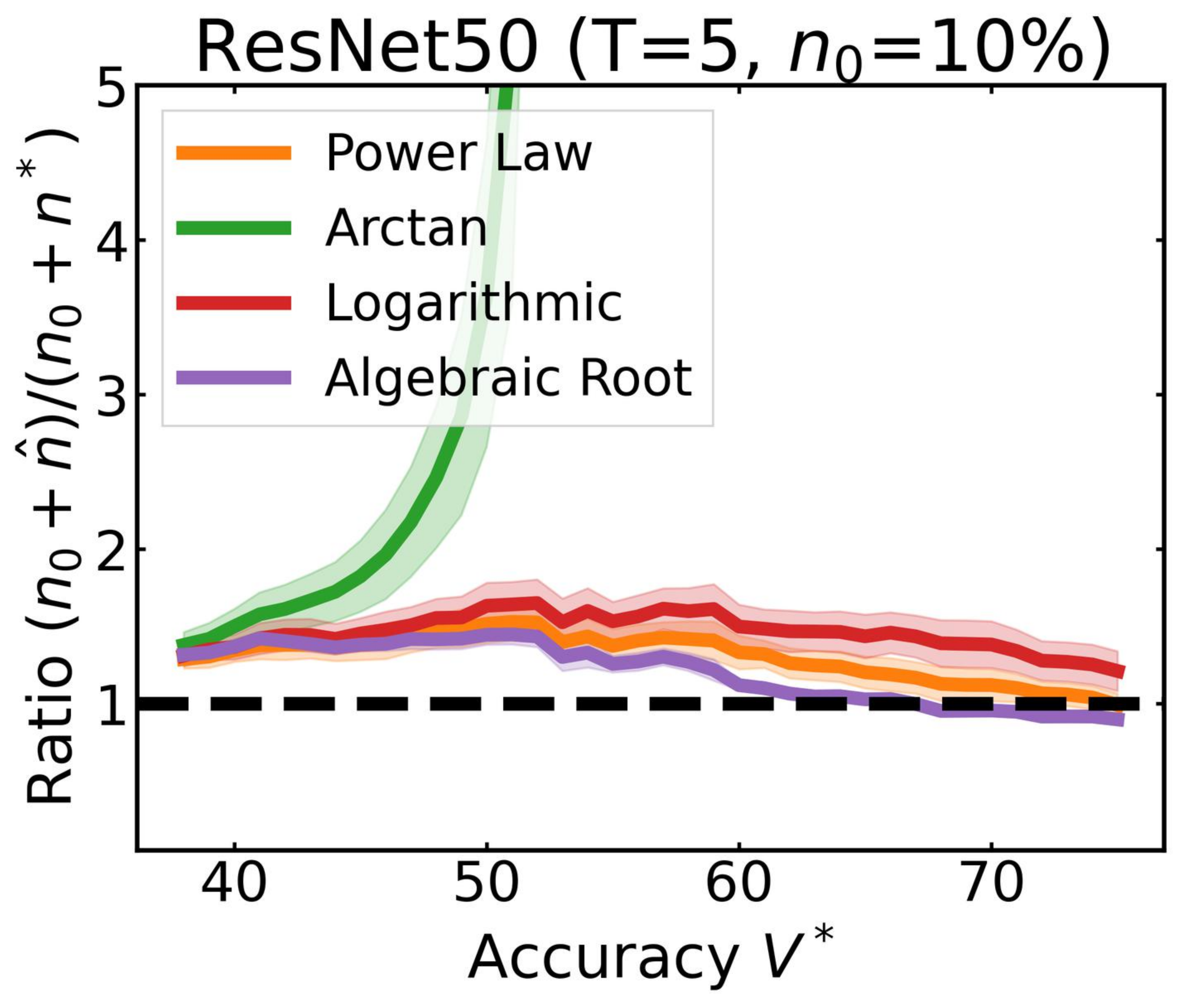} \end{minipage}
\begin{minipage}{0.16\linewidth}\includegraphics[width=1\textwidth]{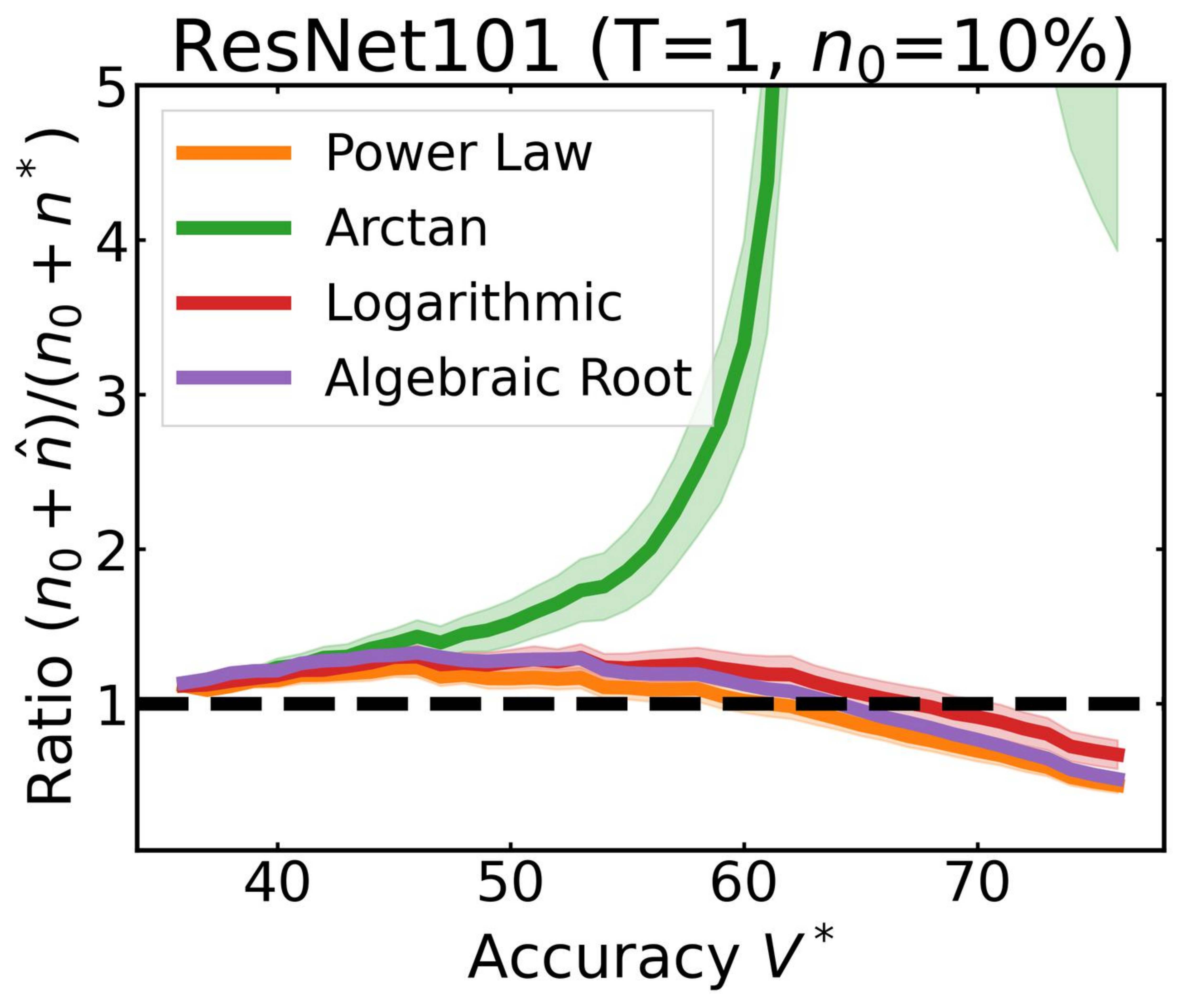} \end{minipage}
\begin{minipage}{0.16\linewidth}\includegraphics[width=1\textwidth]{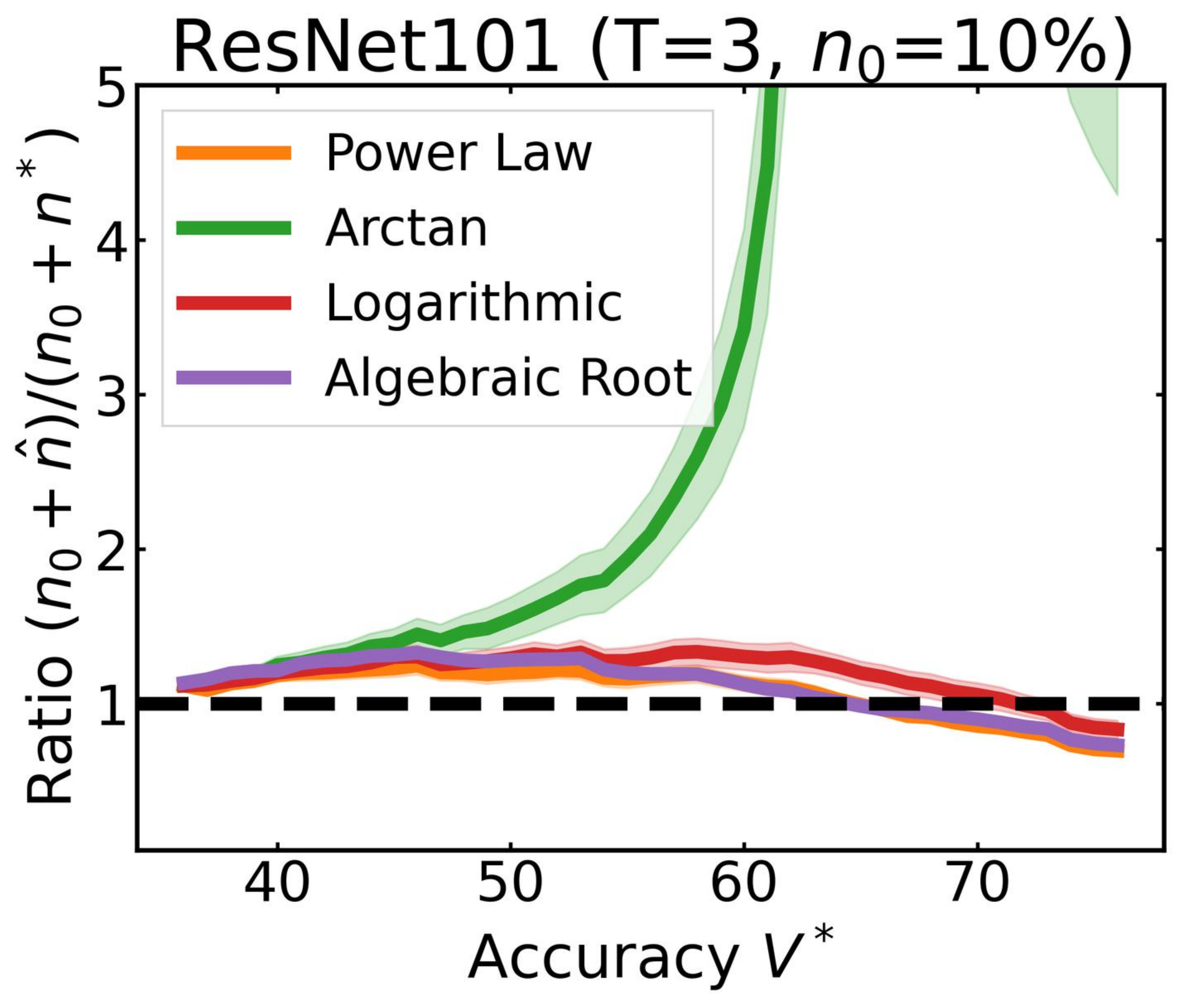} \end{minipage}
\begin{minipage}{0.16\linewidth}\includegraphics[width=1\textwidth]{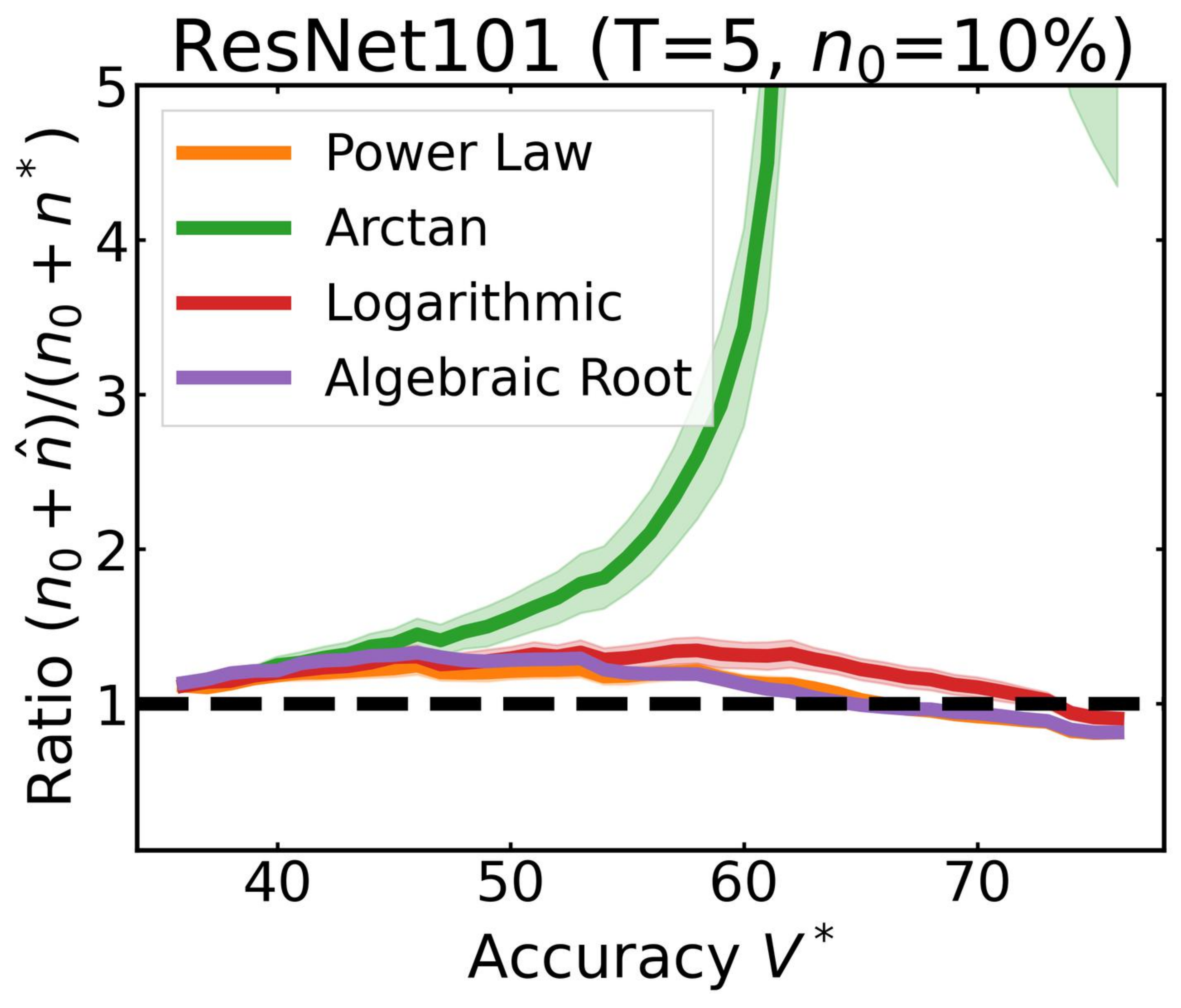} \end{minipage}
\begin{minipage}{0.16\linewidth}\includegraphics[width=1\textwidth]{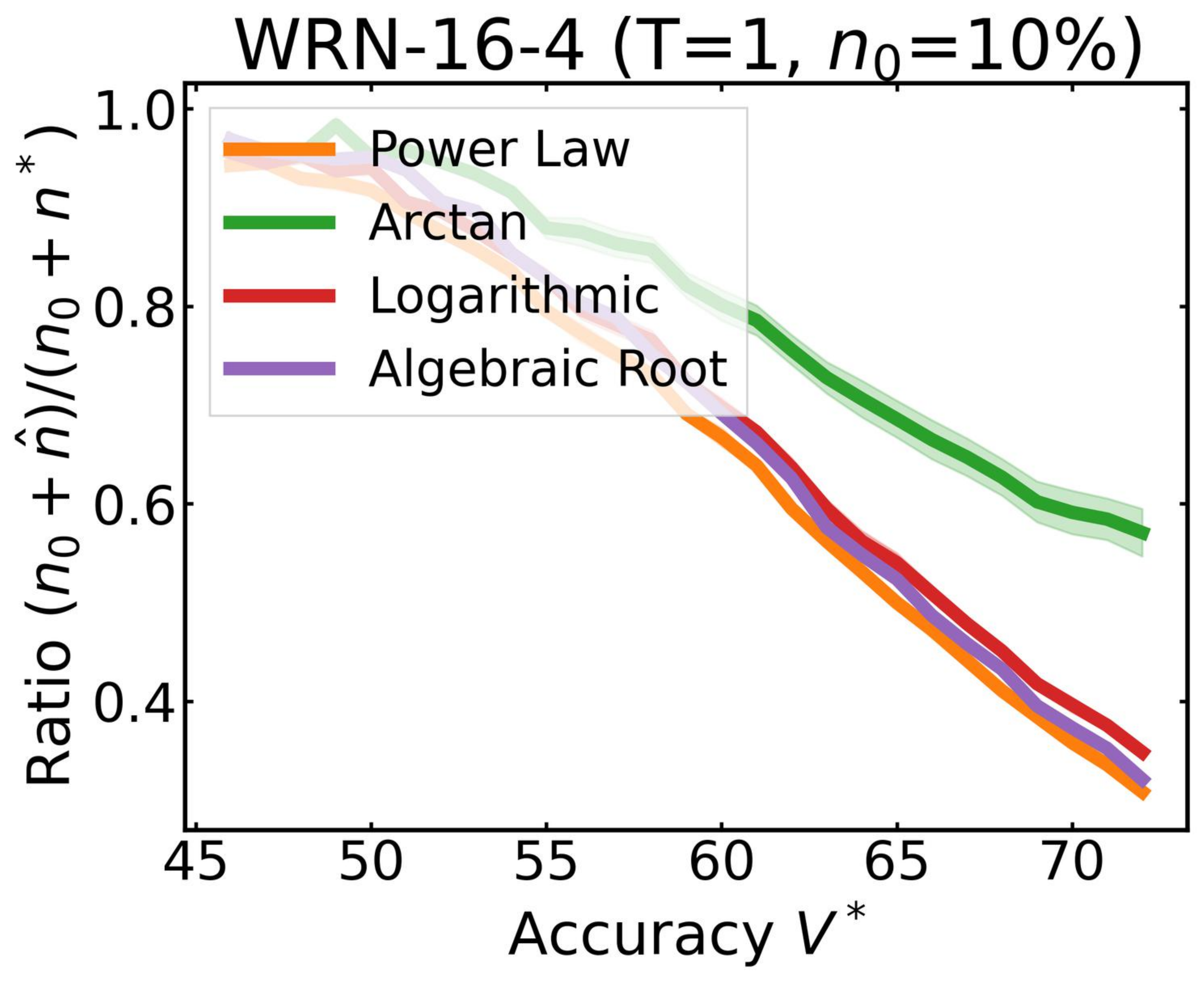} \end{minipage}
\begin{minipage}{0.16\linewidth}\includegraphics[width=1\textwidth]{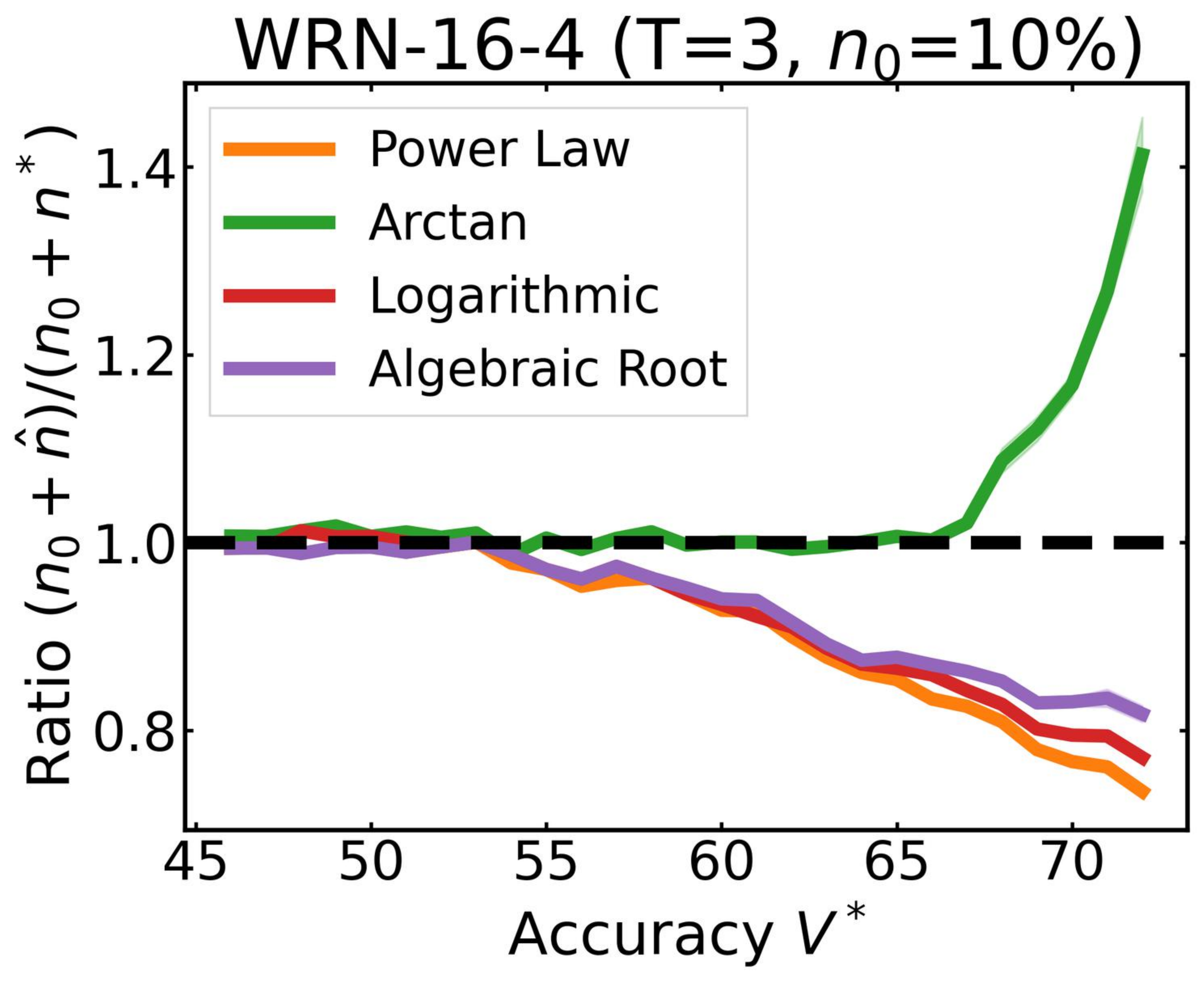} \end{minipage}
\begin{minipage}{0.16\linewidth}\includegraphics[width=1\textwidth]{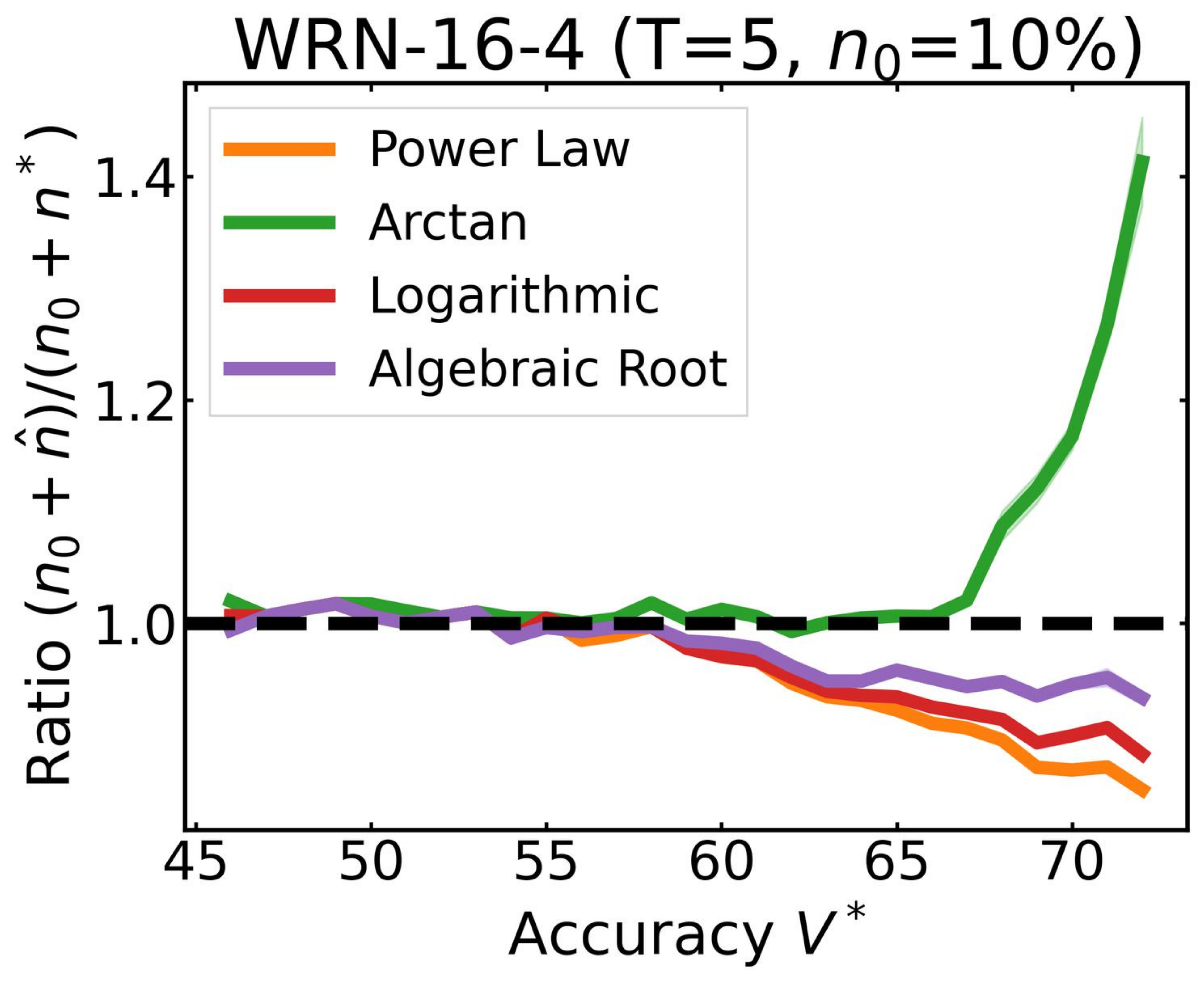} \end{minipage}
\begin{minipage}{0.16\linewidth}\includegraphics[width=1\textwidth]{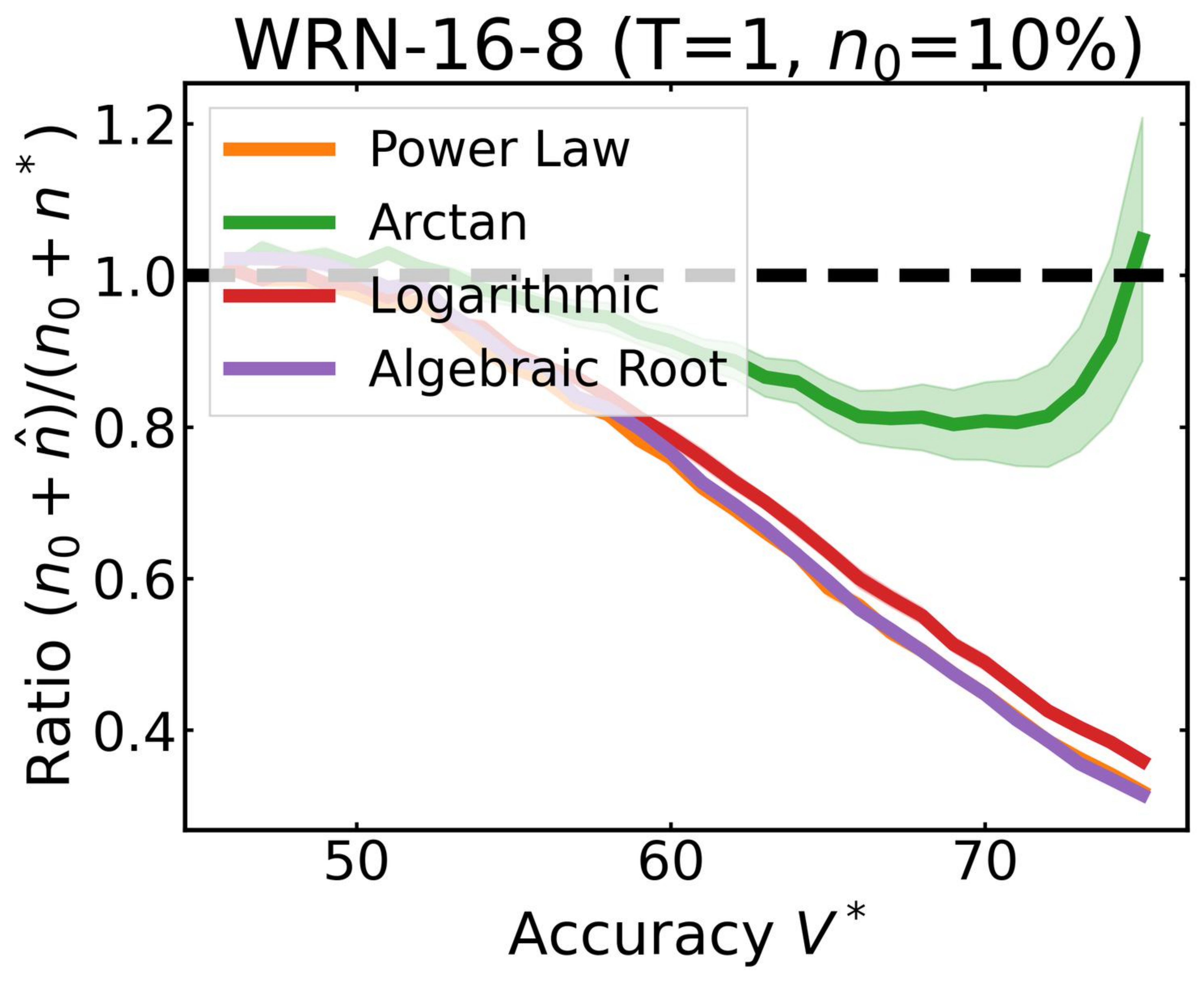} \end{minipage}
\begin{minipage}{0.16\linewidth}\includegraphics[width=1\textwidth]{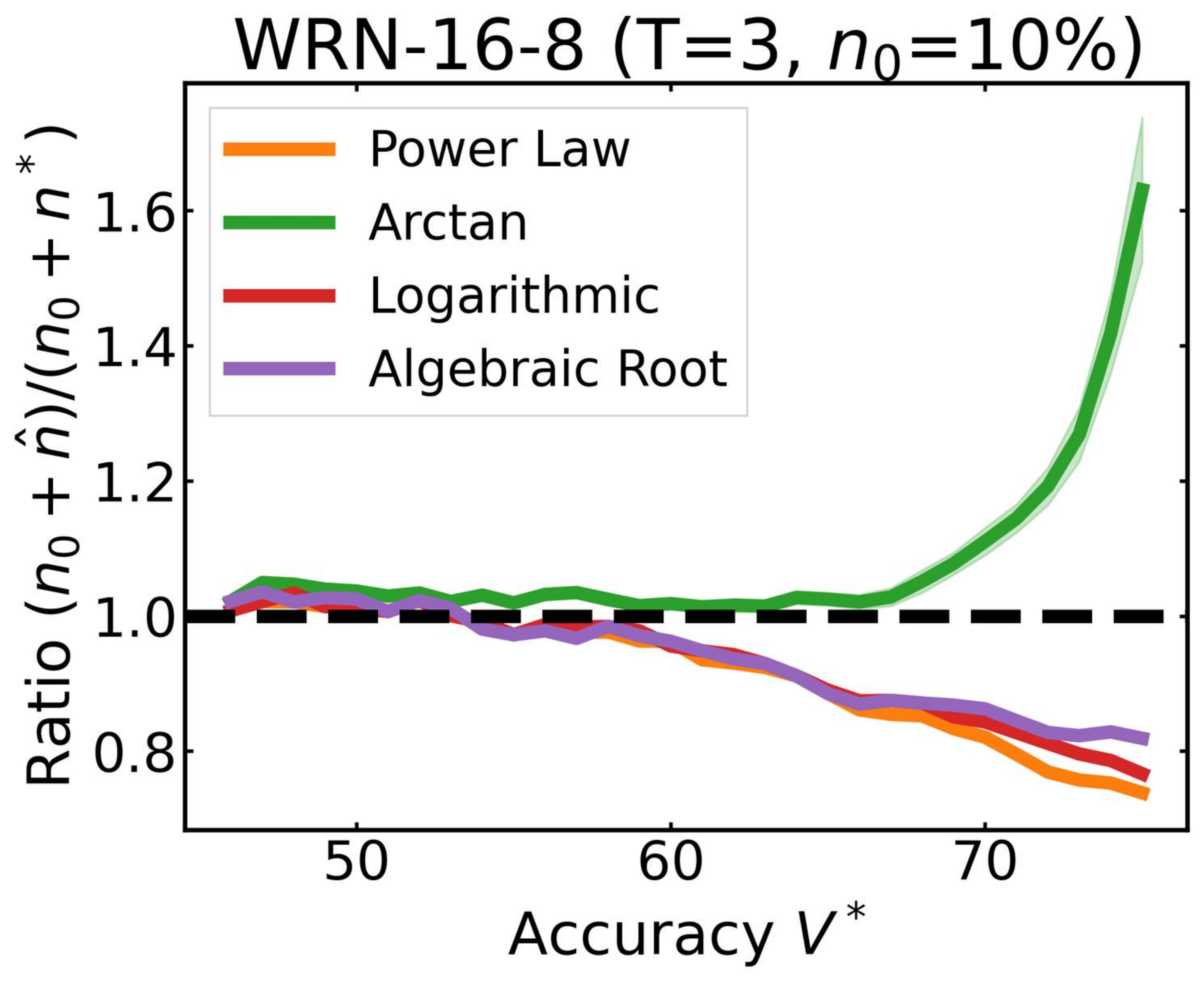} \end{minipage}
\begin{minipage}{0.16\linewidth}\includegraphics[width=1\textwidth]{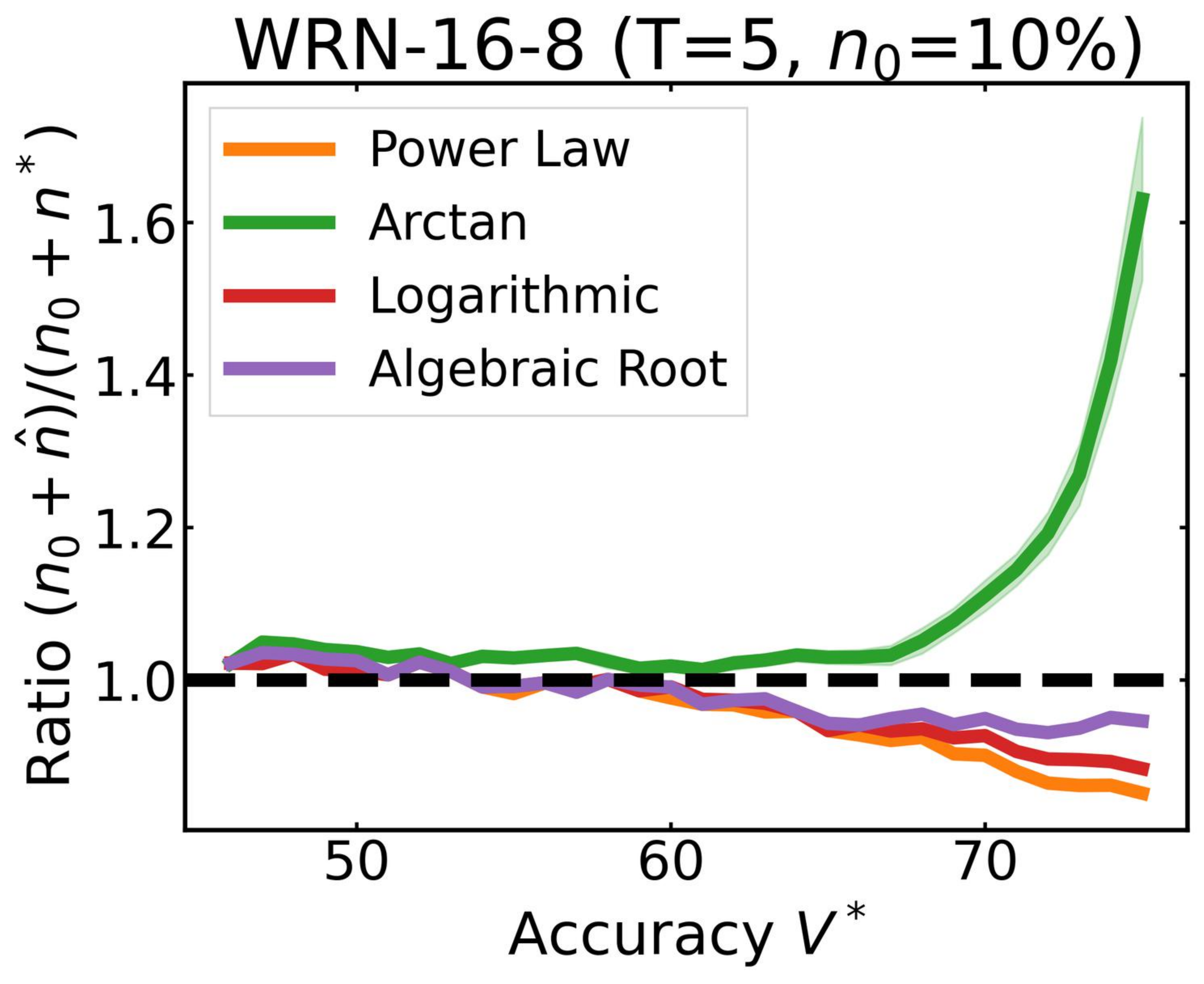} \end{minipage}
\caption{\label{fig:simulation_cifar100_architectures} The ratio of the amount of data collected versus the minimum data needed (y-axis) for different target $V^*$ (x-axis) in simulations initializing with $n_0 = 10\%$ of the data set. For each data set, we show simulations for $T=1, 3, 5$ maximum rounds. The dashed black line corresponds to collecting the least amount of data needed to reach $V^*$.
}
\end{center}
\vspace{-7mm}
\end{figure*}

\section{Ablations on different architectures}
\label{sec:app_ablations_architectures}

In this section, we further explore CIFAR100 and repeat the previous experiments but with different architectures, to show that our results are consistent even with larger models. 
Specifically, we observe that with small data sets, estimating data requirements is more difficult and that moreover, regression error does not give a complete picture in terms of determining a good data collection policy. 
Comparing against the baseline ResNet18, we consider ResNet50 and ResNet101 as well as WideResNet-16-4 and WideResNet-16-8~\cite{zagoruyko2016wide}. All models are trained in the exact same setup.

Table~\ref{tab:regression_RMSE_archs} reports RMSE from fitting each regression function with $n_0 = 10\%, 20\%, 50\%$ of the data. 
Similar to the baseline, the Arctan function is almost always the best performing regression function. 
Figure~\ref{fig:simulation_cifar100_architectures} plots the ratios of the amount of data collected versus the minimum data required for each of the alternative architectures. ResNet50 and ResNet101 follows the same trends as the baseline ResNet18 (see Figure~\ref{fig:simulation_all}) in that Arctan significantly over-estimates the data requirement, even though it presents the lowest RMSE among the regression functions. The WideResNets show a slightly different picture. Here, although Arctan does not always over-estimate the requirement for all $V^*$, it is nonetheless, the most pessimistic estimator and recommends collecting more data than the other functions.

\begin{figure*}[!t]
\begin{center}
\begin{minipage}{0.16\linewidth}\includegraphics[width=1\textwidth]{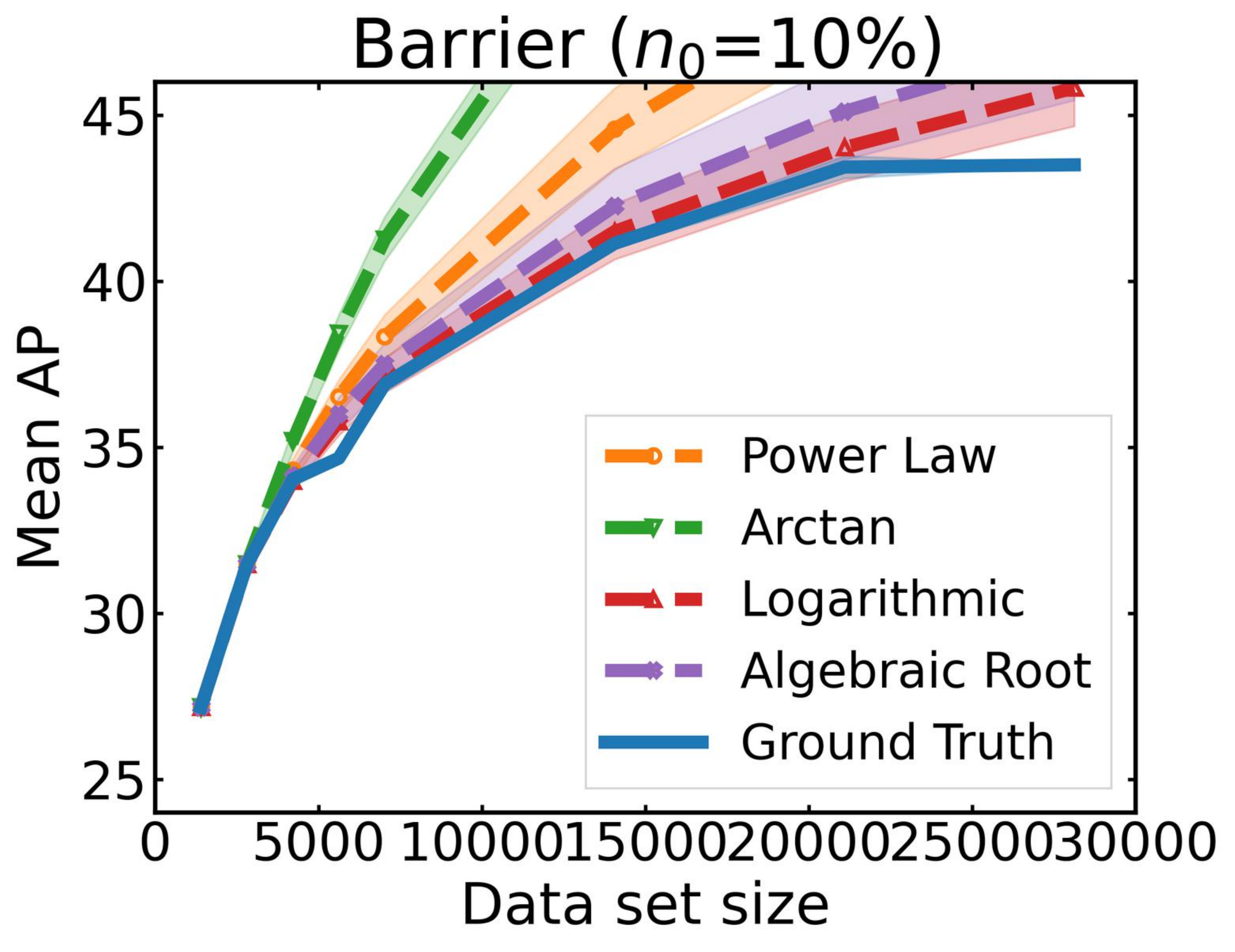} \end{minipage}
\begin{minipage}{0.16\linewidth}\includegraphics[width=1\textwidth]{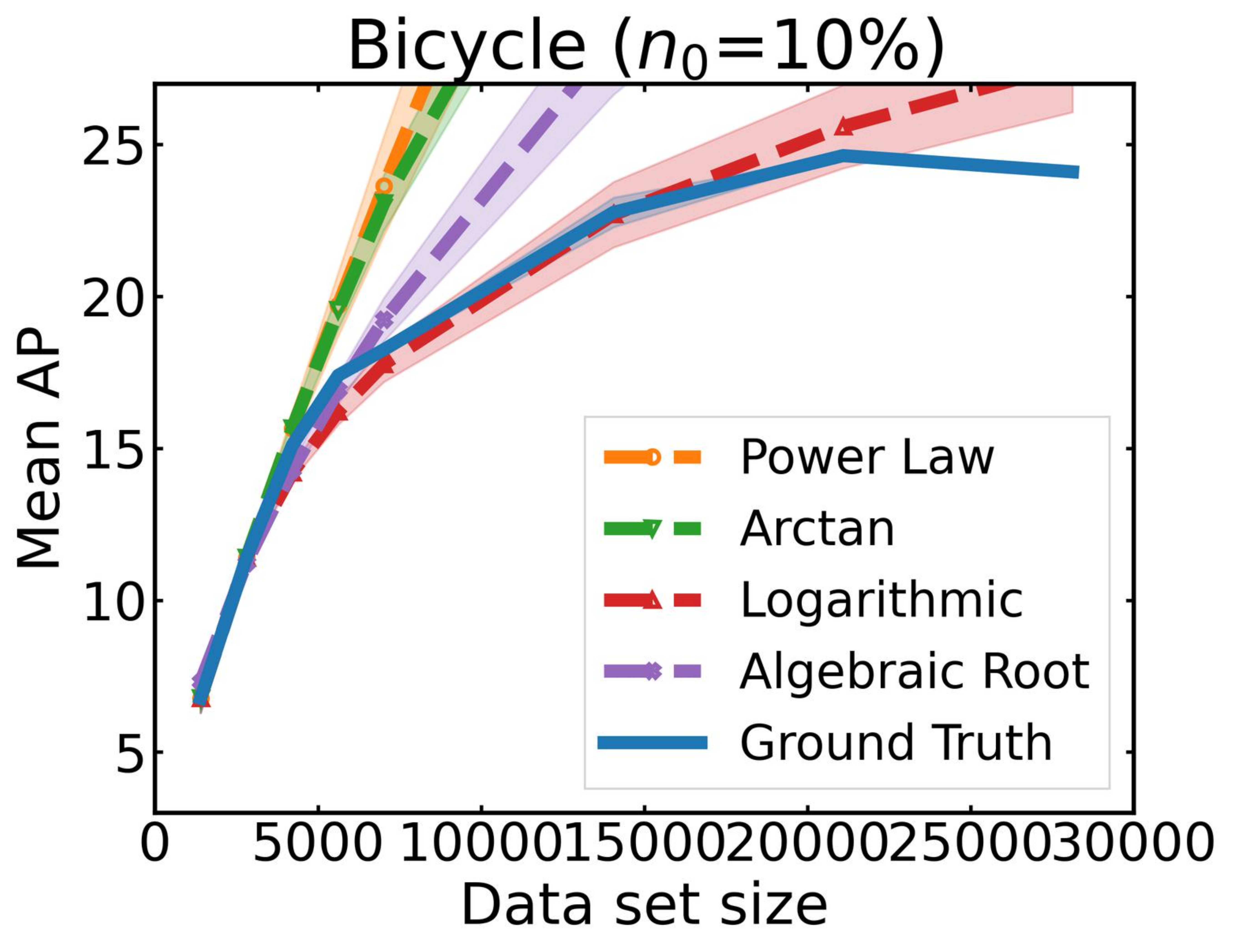} \end{minipage}
\begin{minipage}{0.16\linewidth}\includegraphics[width=1\textwidth]{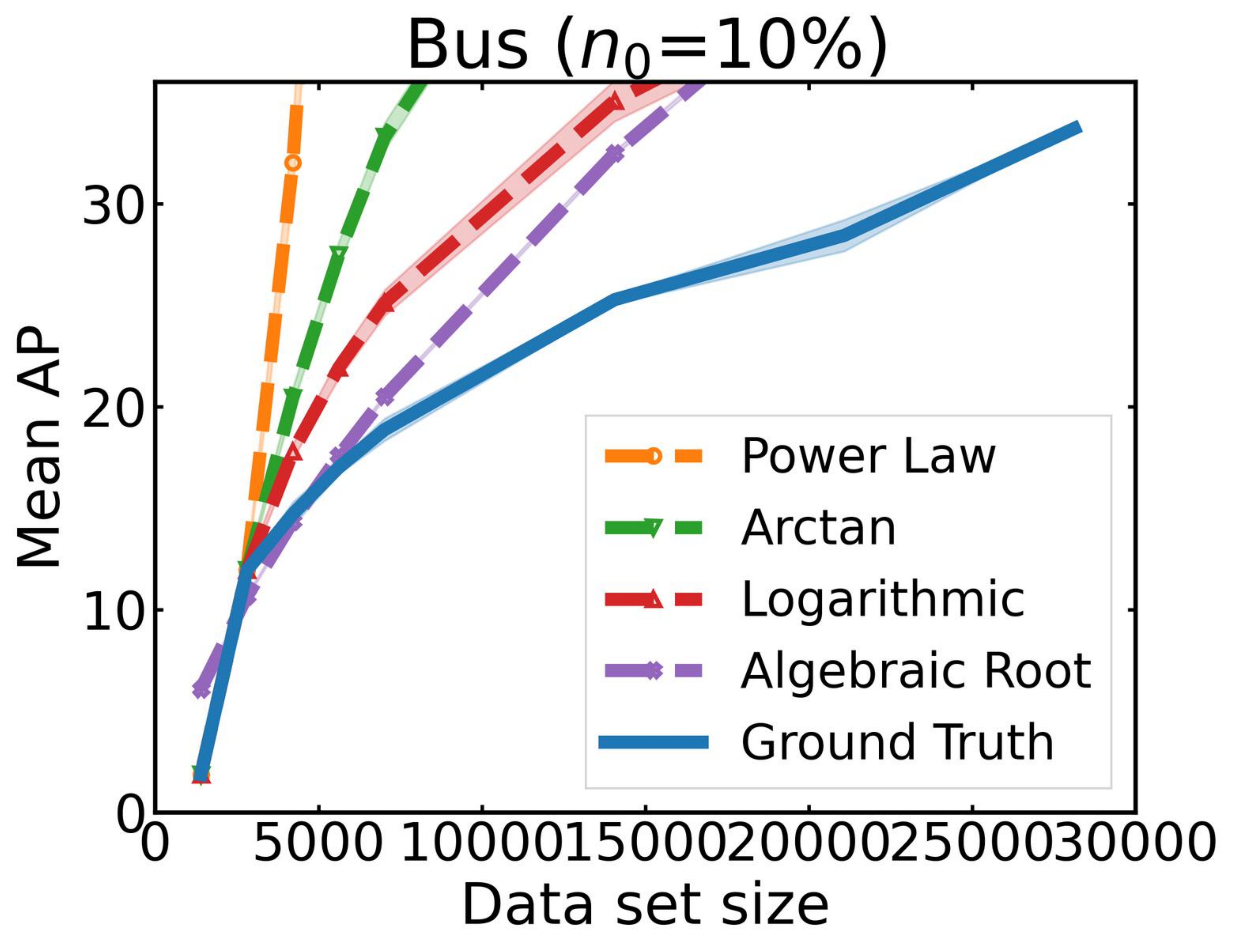} \end{minipage}
\begin{minipage}{0.16\linewidth}\includegraphics[width=1\textwidth]{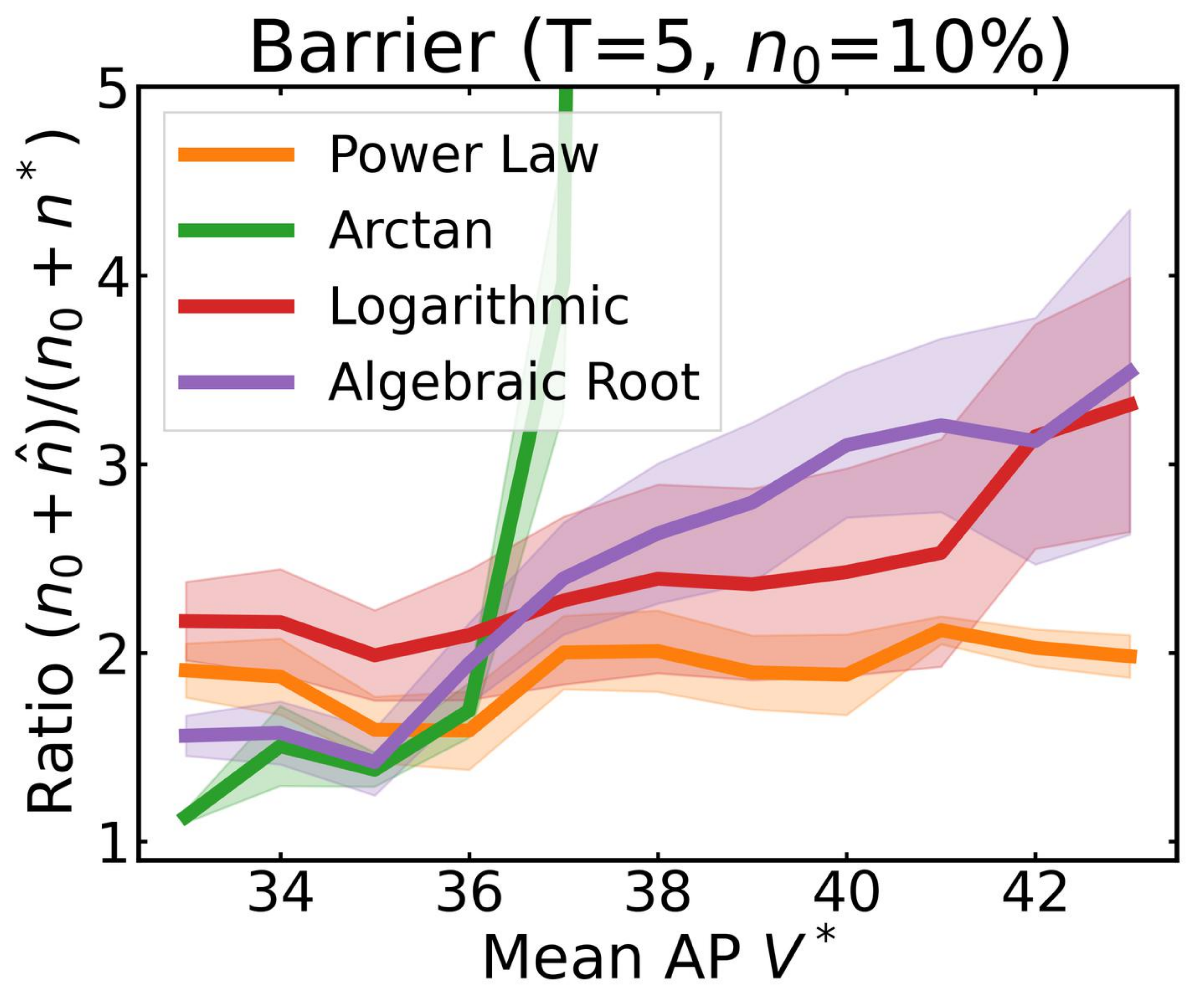} \end{minipage}
\begin{minipage}{0.16\linewidth}\includegraphics[width=1\textwidth]{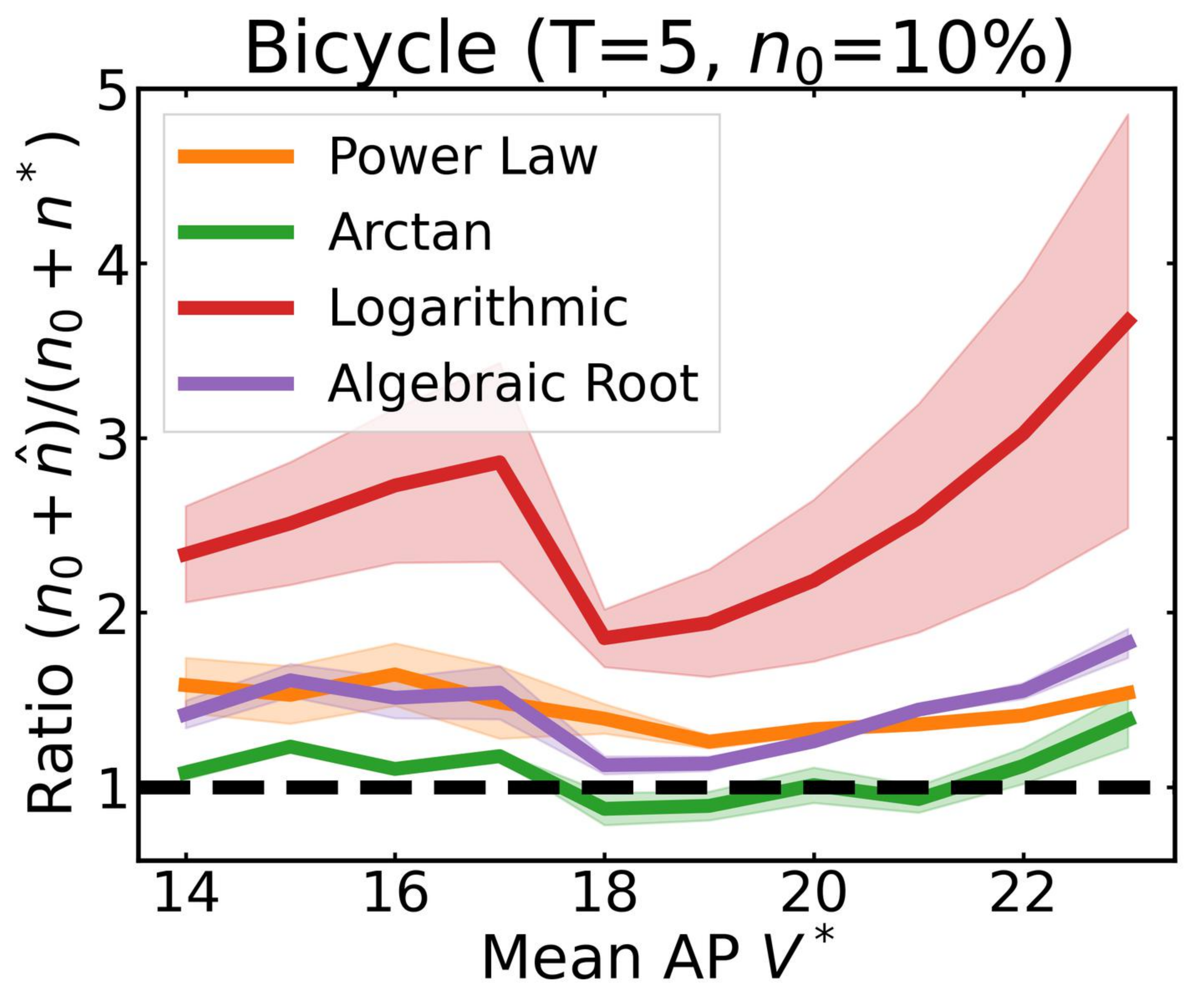} \end{minipage}
\begin{minipage}{0.16\linewidth}\includegraphics[width=1\textwidth]{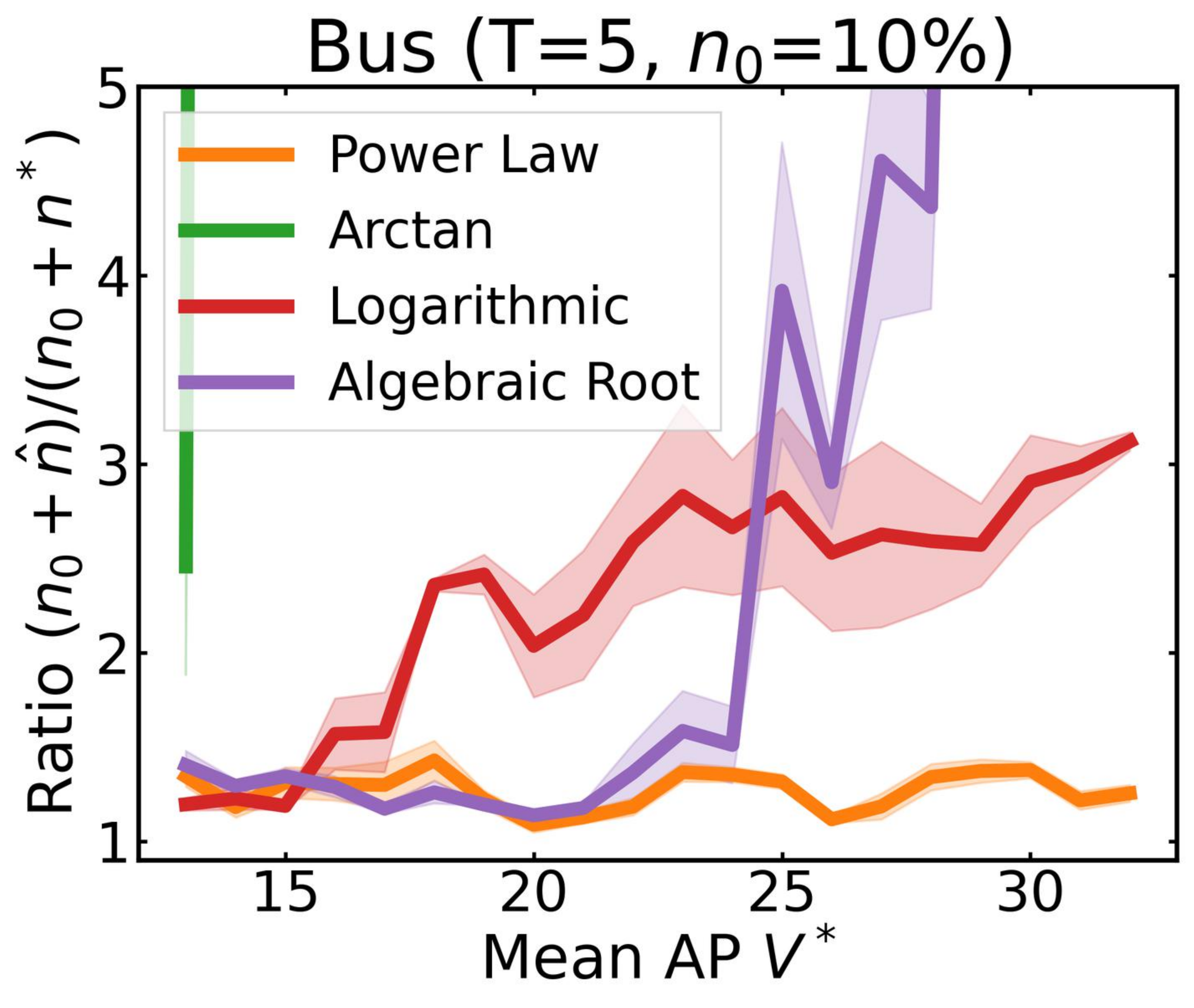} \end{minipage}
\begin{minipage}{0.16\linewidth}\includegraphics[width=1\textwidth]{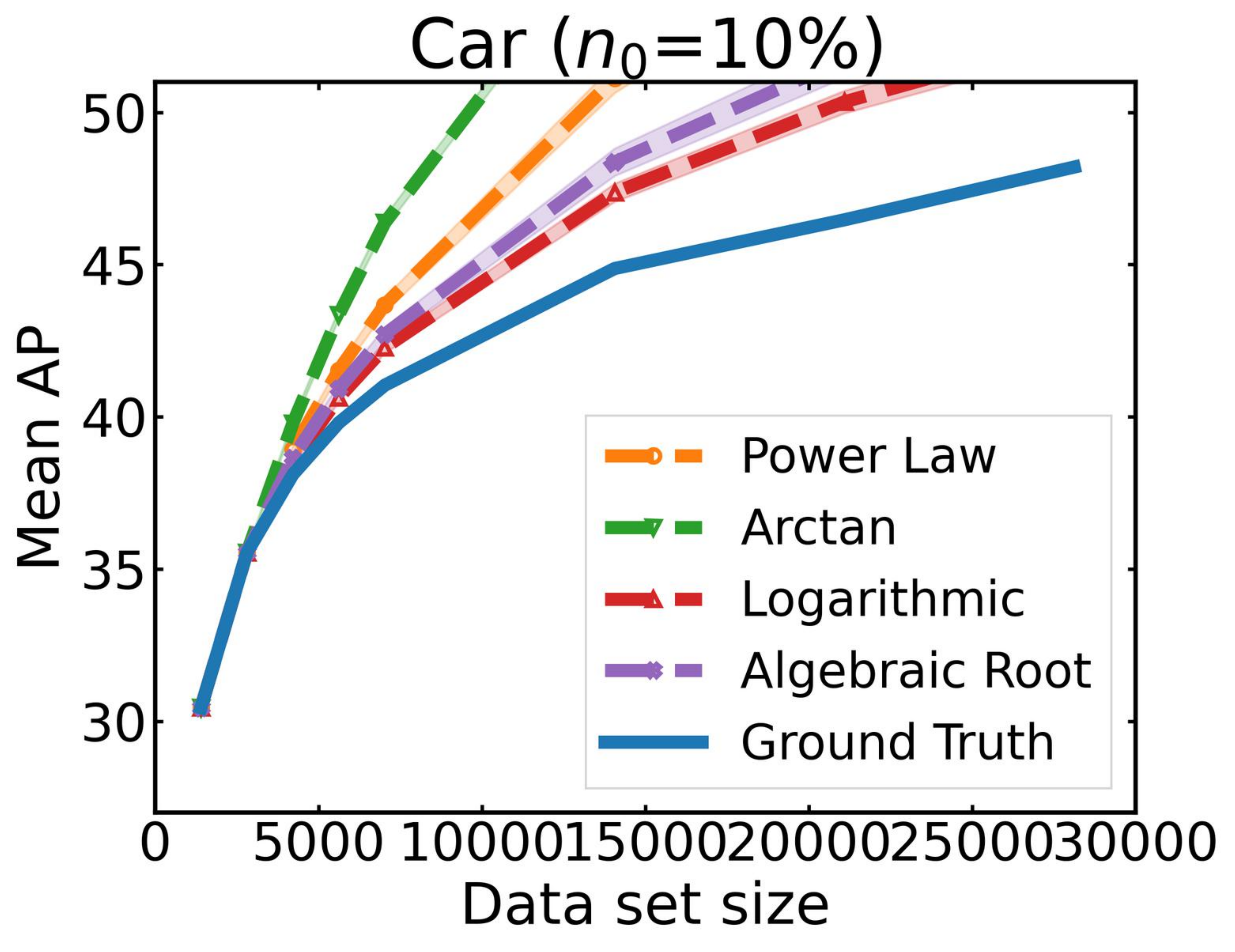} \end{minipage}
\begin{minipage}{0.16\linewidth}\includegraphics[width=1\textwidth]{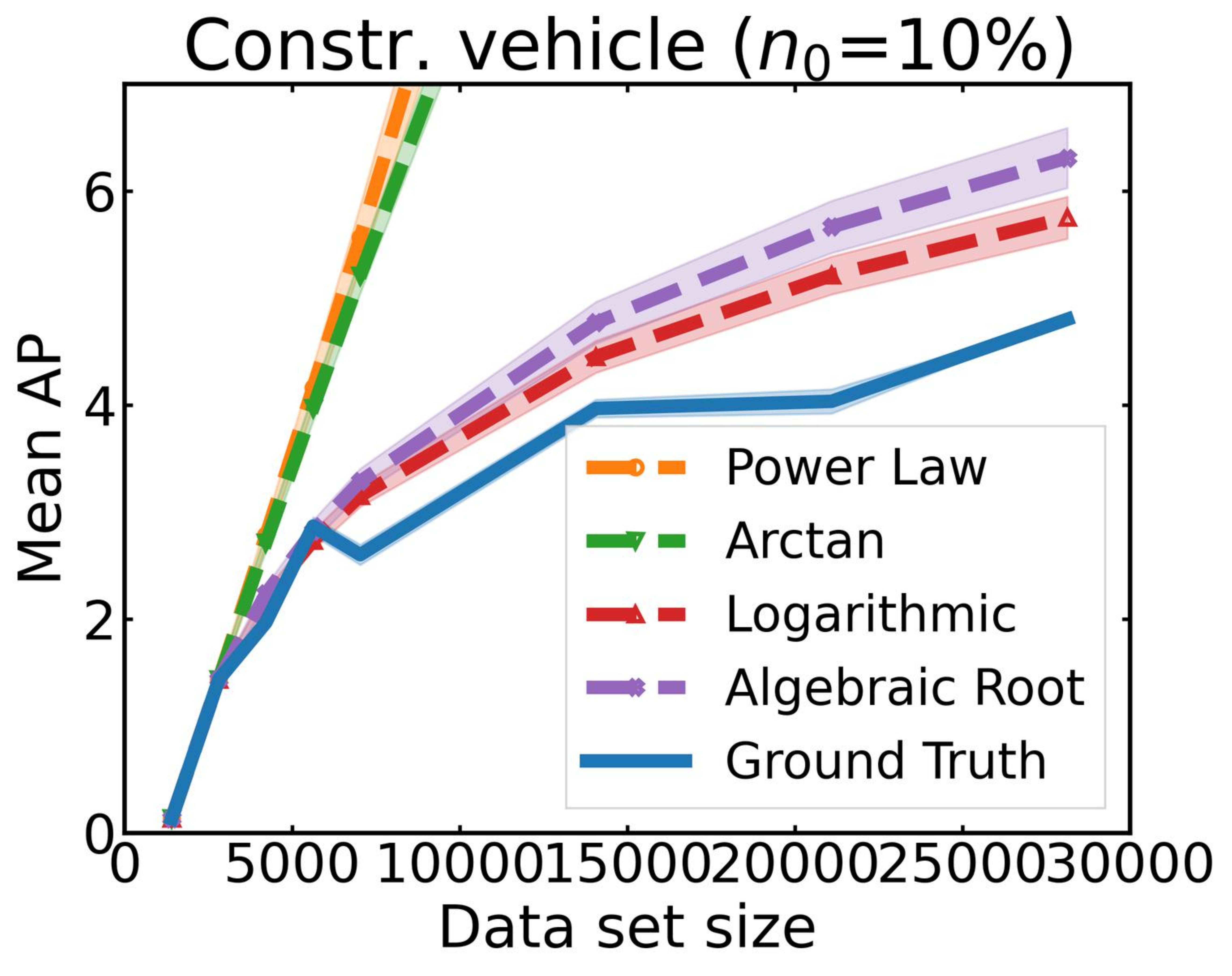} \end{minipage}
\begin{minipage}{0.16\linewidth}\includegraphics[width=1\textwidth]{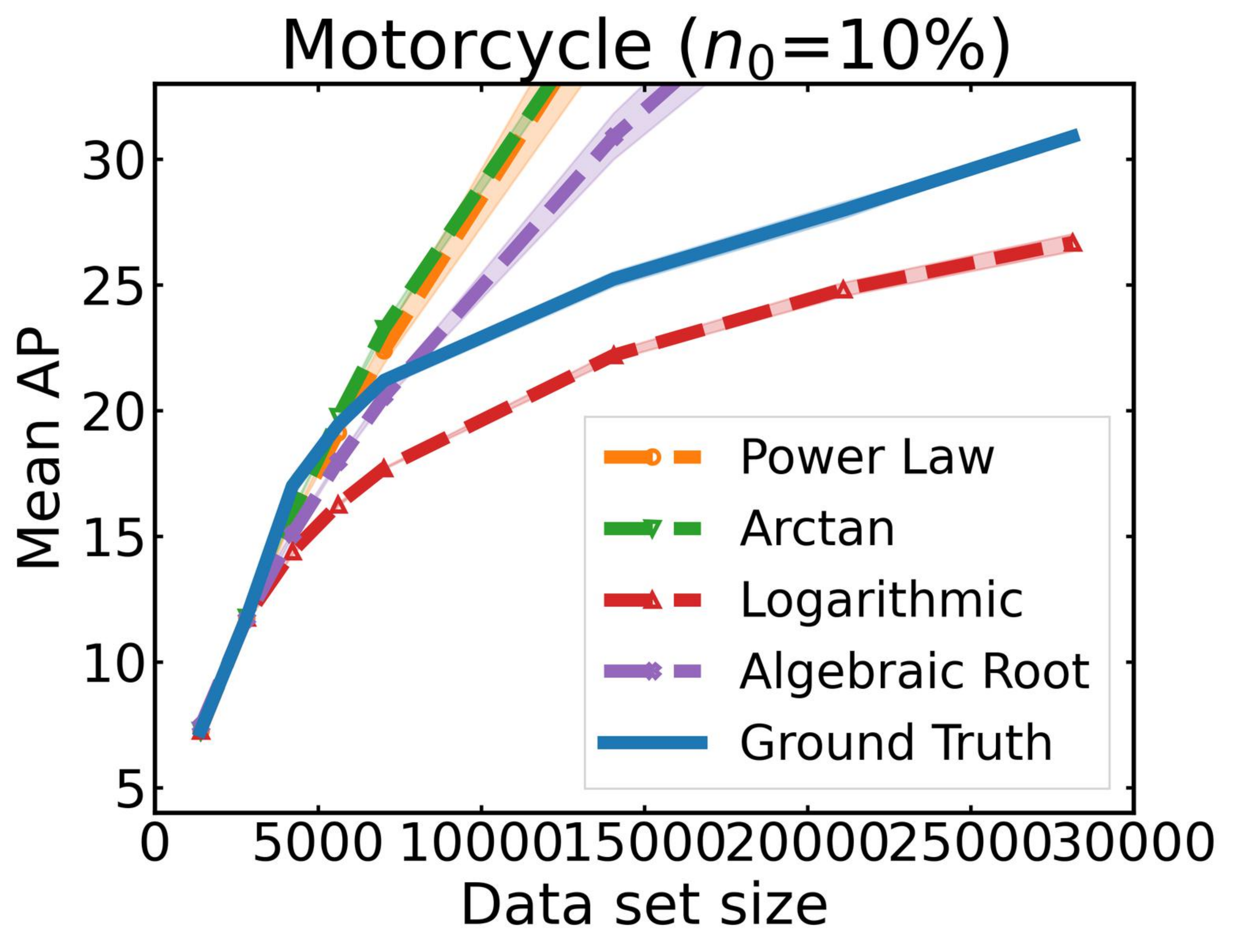} \end{minipage}
\begin{minipage}{0.16\linewidth}\includegraphics[width=1\textwidth]{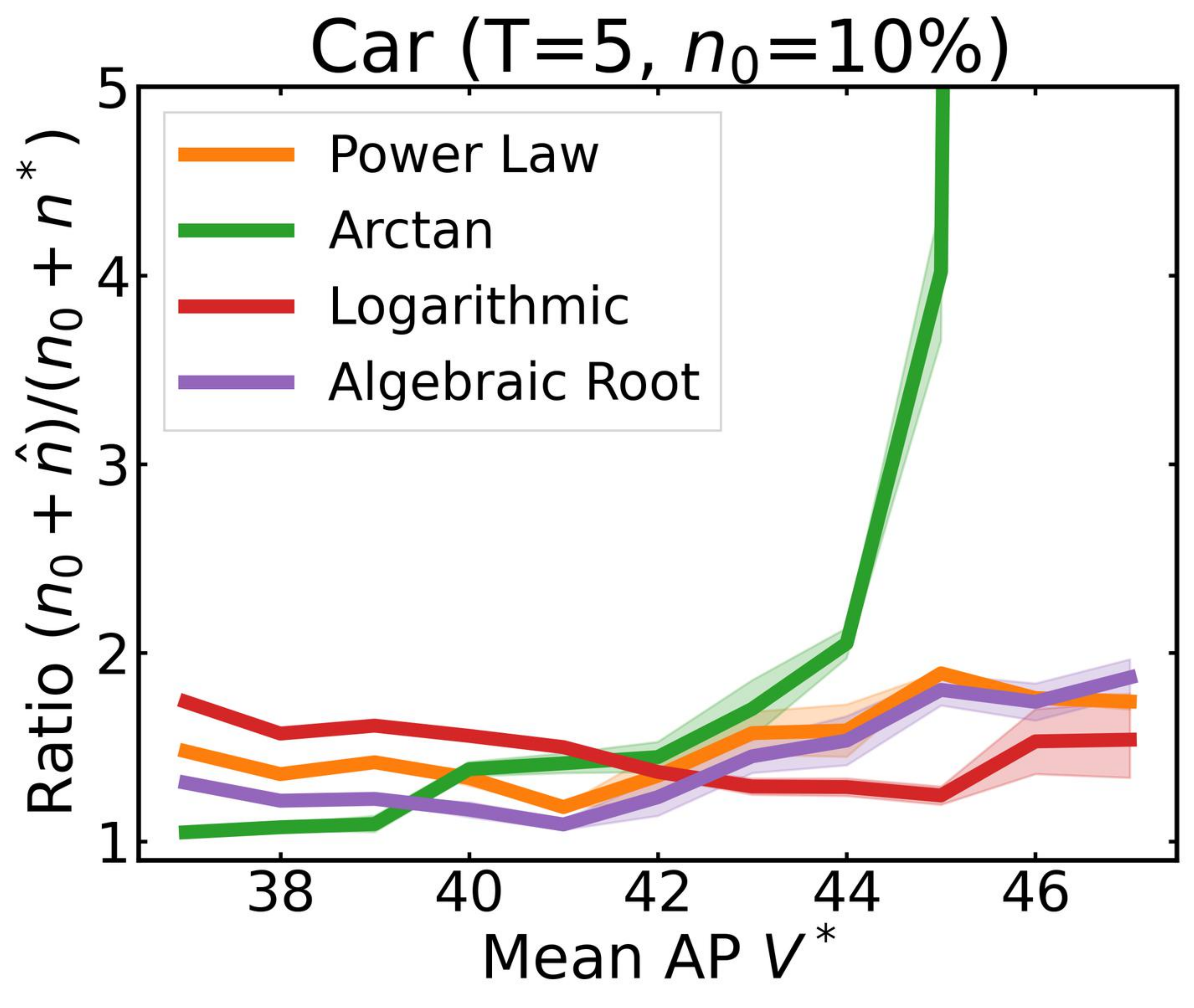} \end{minipage}
\begin{minipage}{0.16\linewidth}\includegraphics[width=1\textwidth]{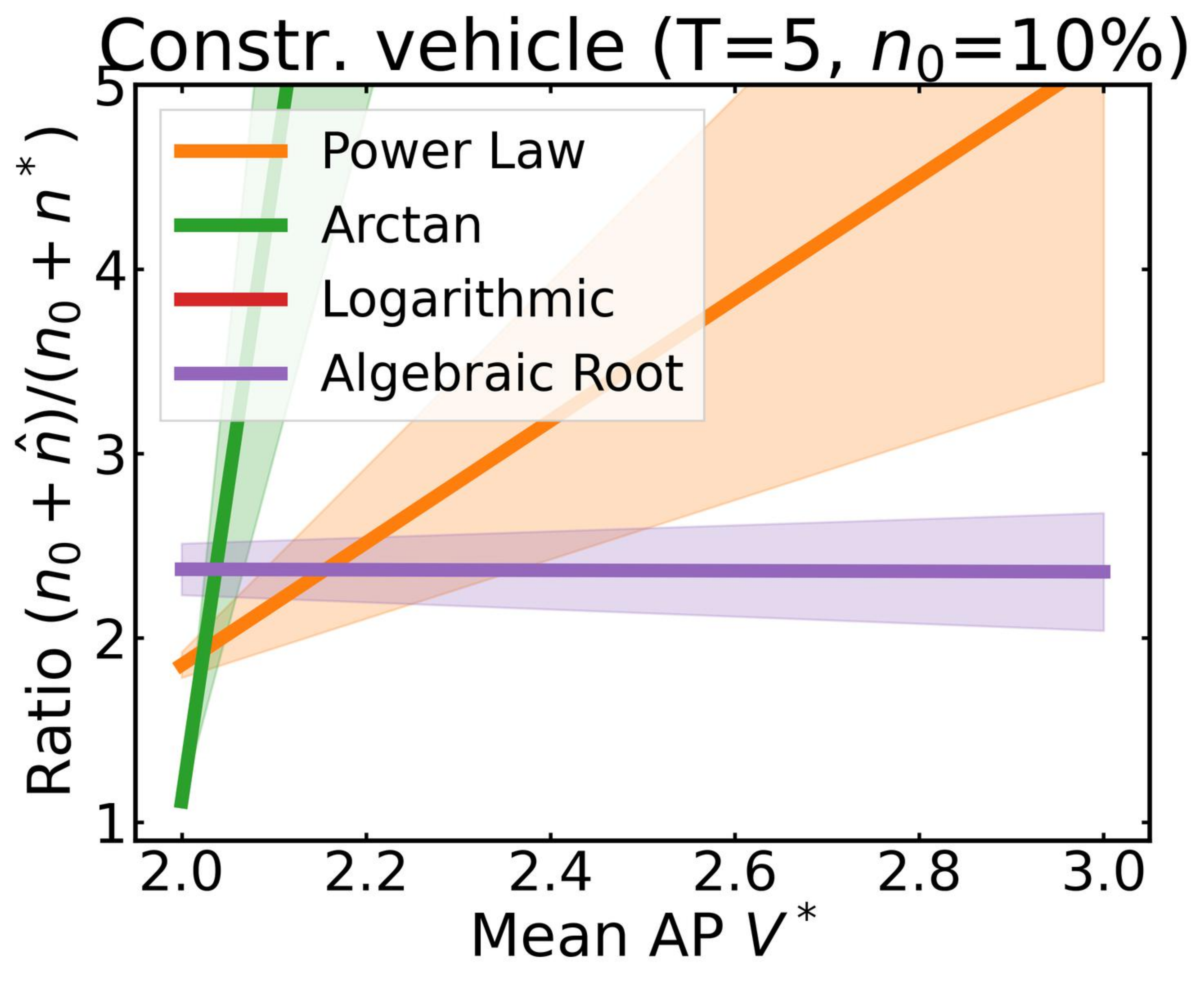} \end{minipage}
\begin{minipage}{0.16\linewidth}\includegraphics[width=1\textwidth]{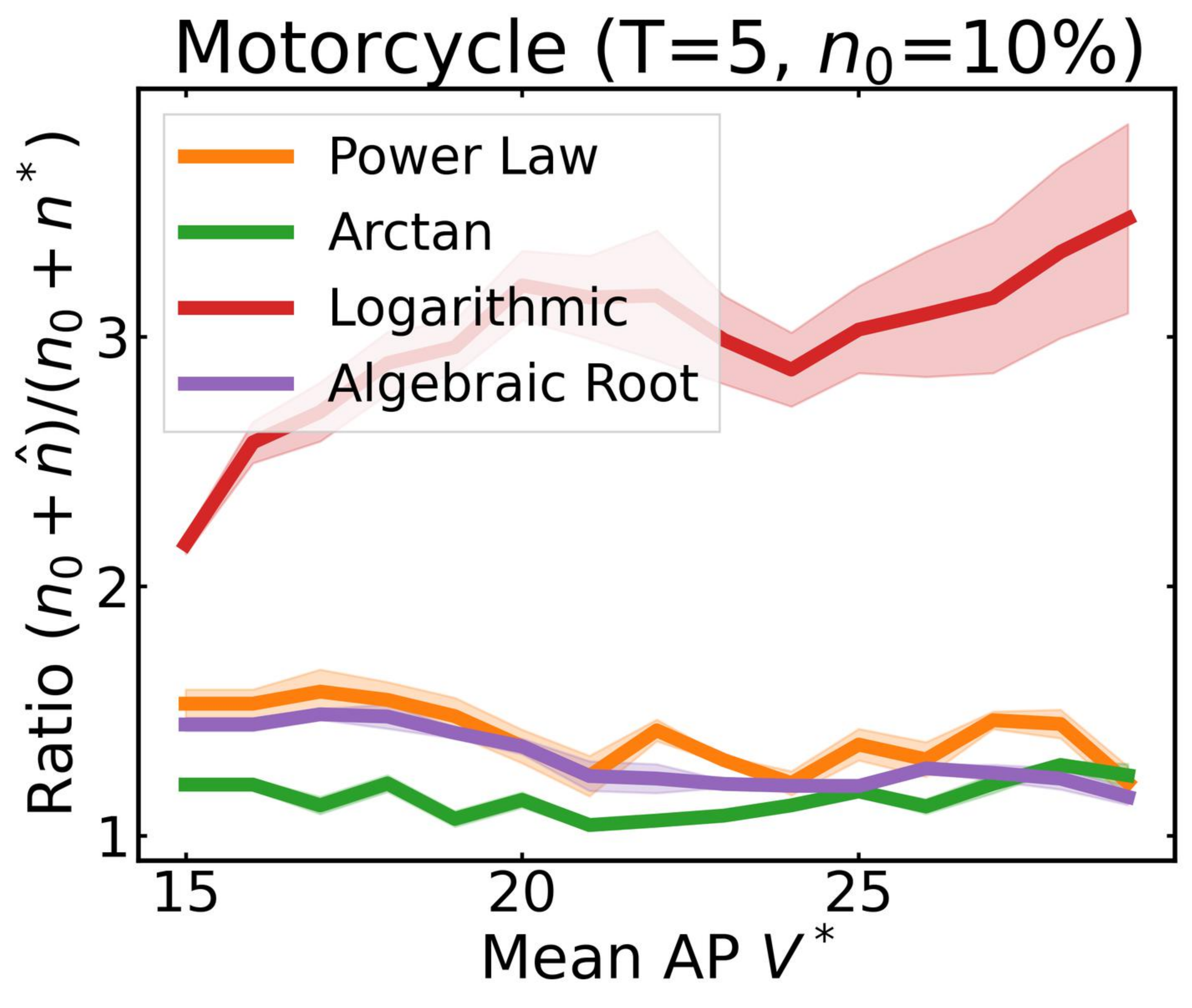} \end{minipage}
\begin{minipage}{0.16\linewidth}\includegraphics[width=1\textwidth]{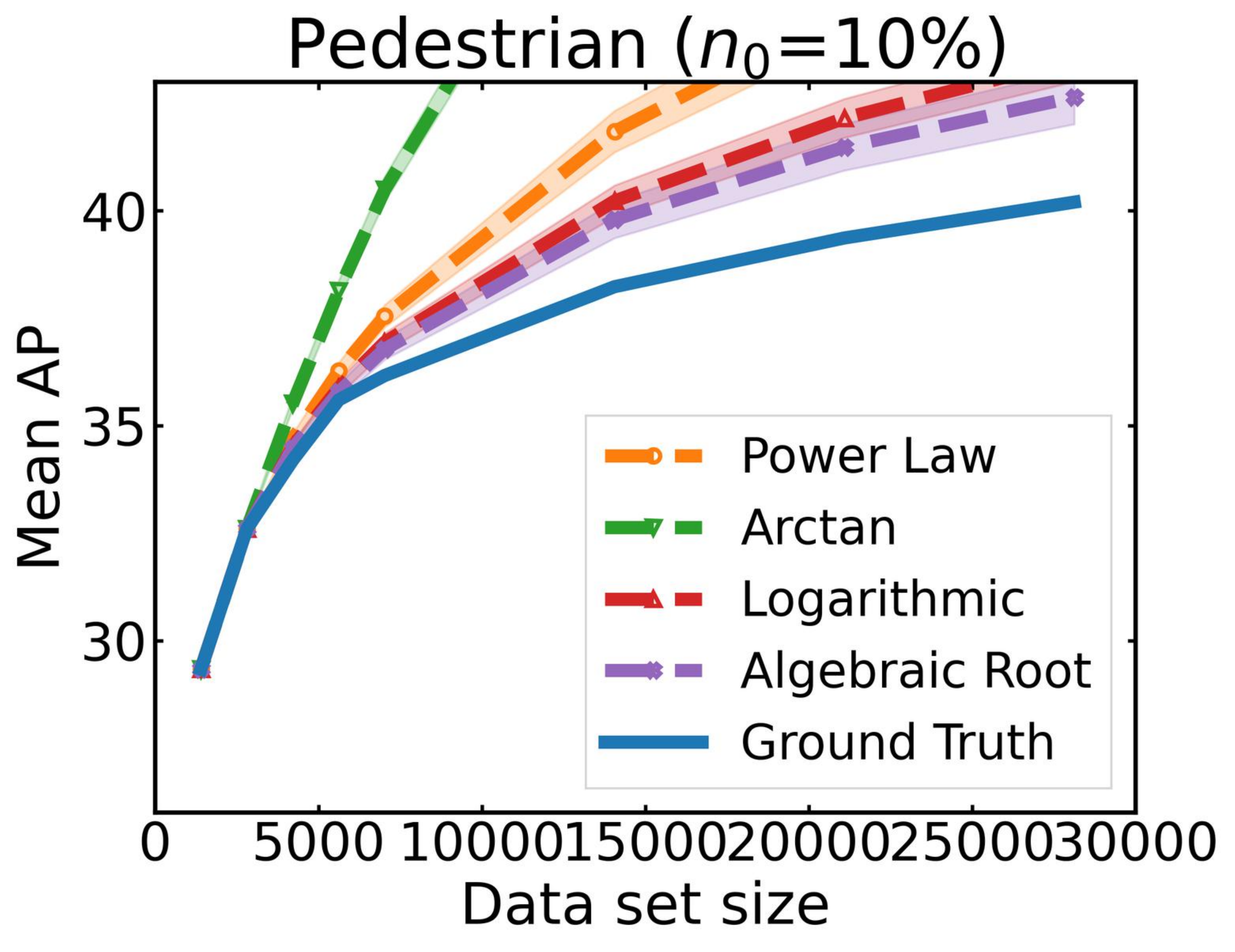} \end{minipage}
\begin{minipage}{0.16\linewidth}\includegraphics[width=1\textwidth]{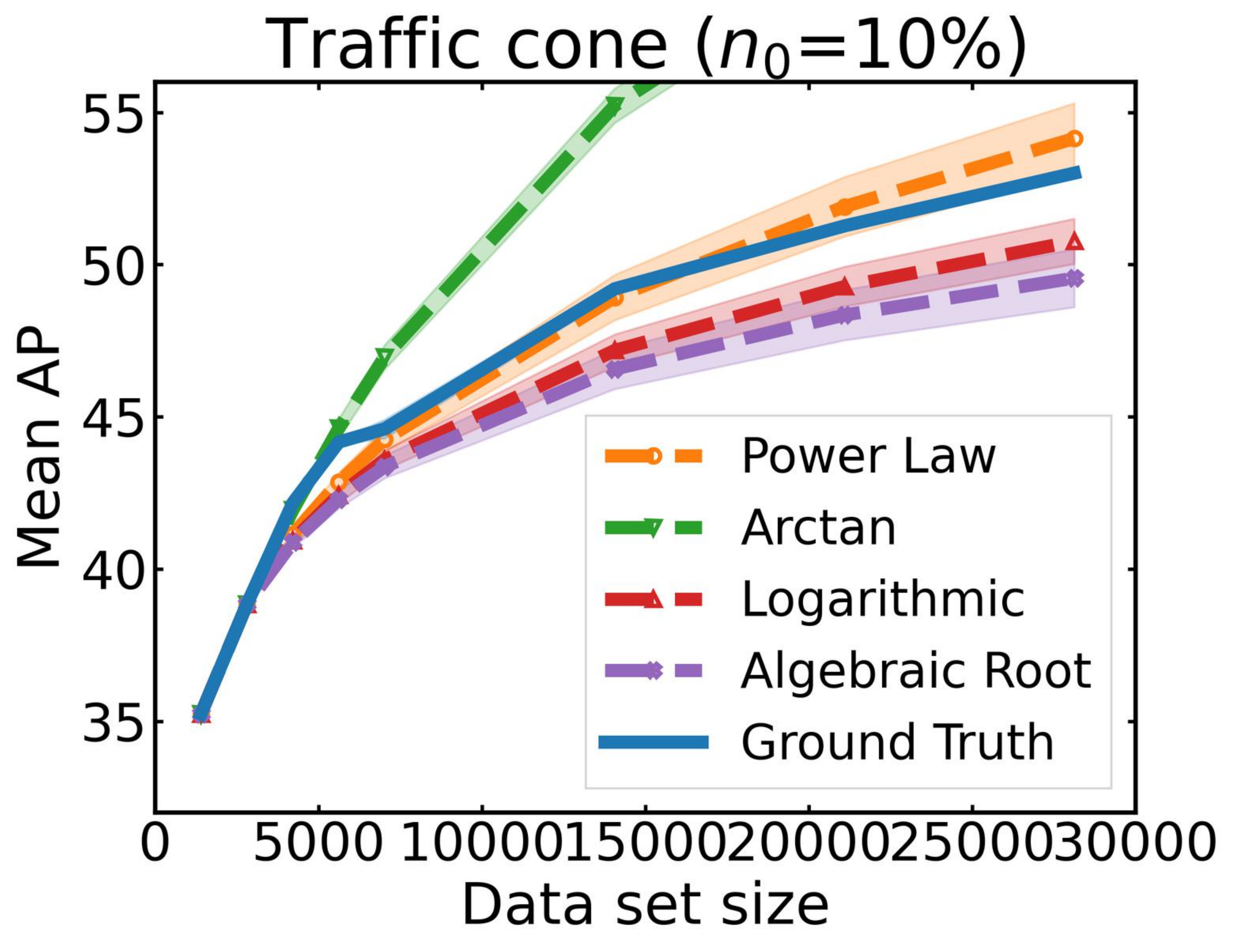} \end{minipage}
\begin{minipage}{0.16\linewidth}\includegraphics[width=1\textwidth]{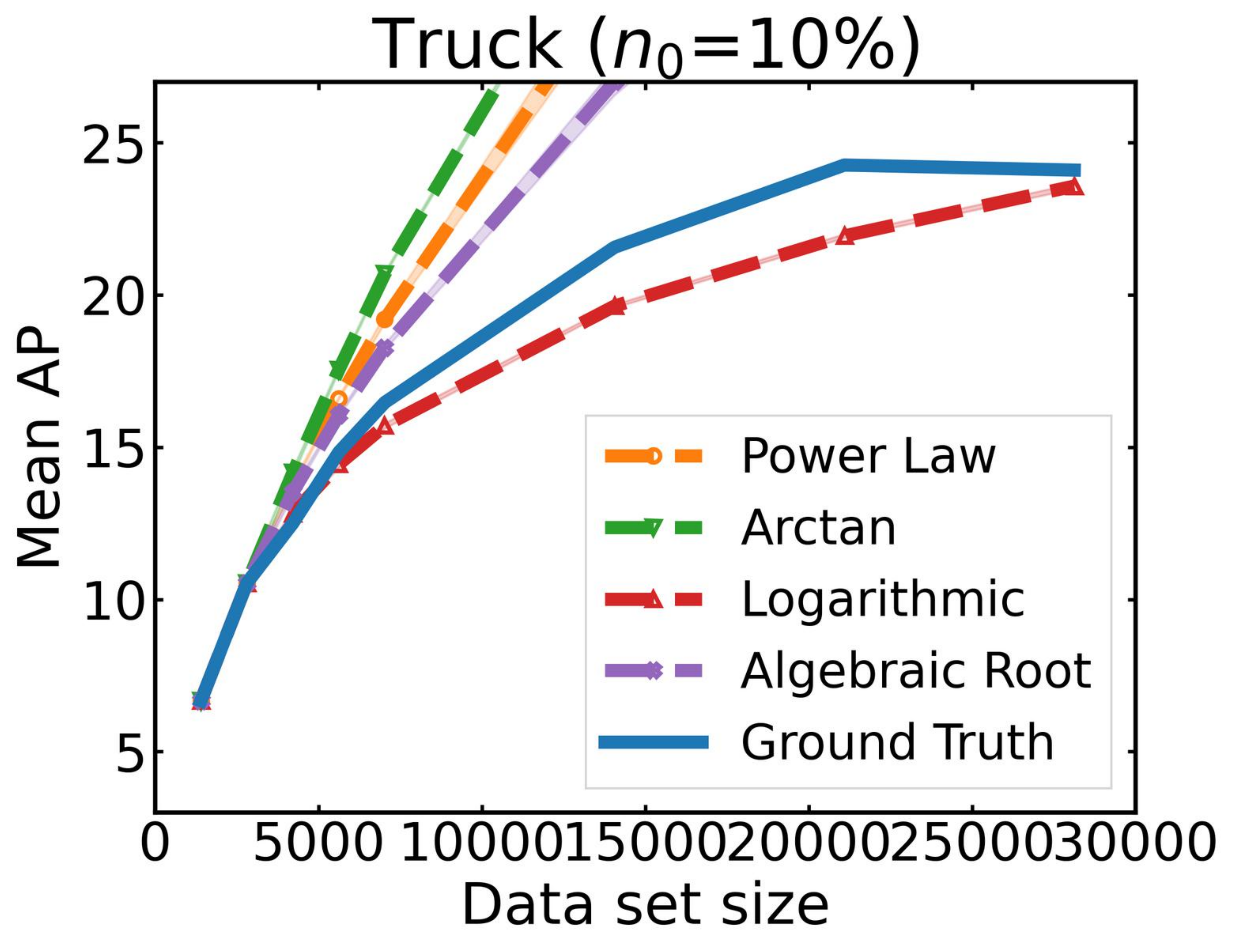} \end{minipage}
\begin{minipage}{0.16\linewidth}\includegraphics[width=1\textwidth]{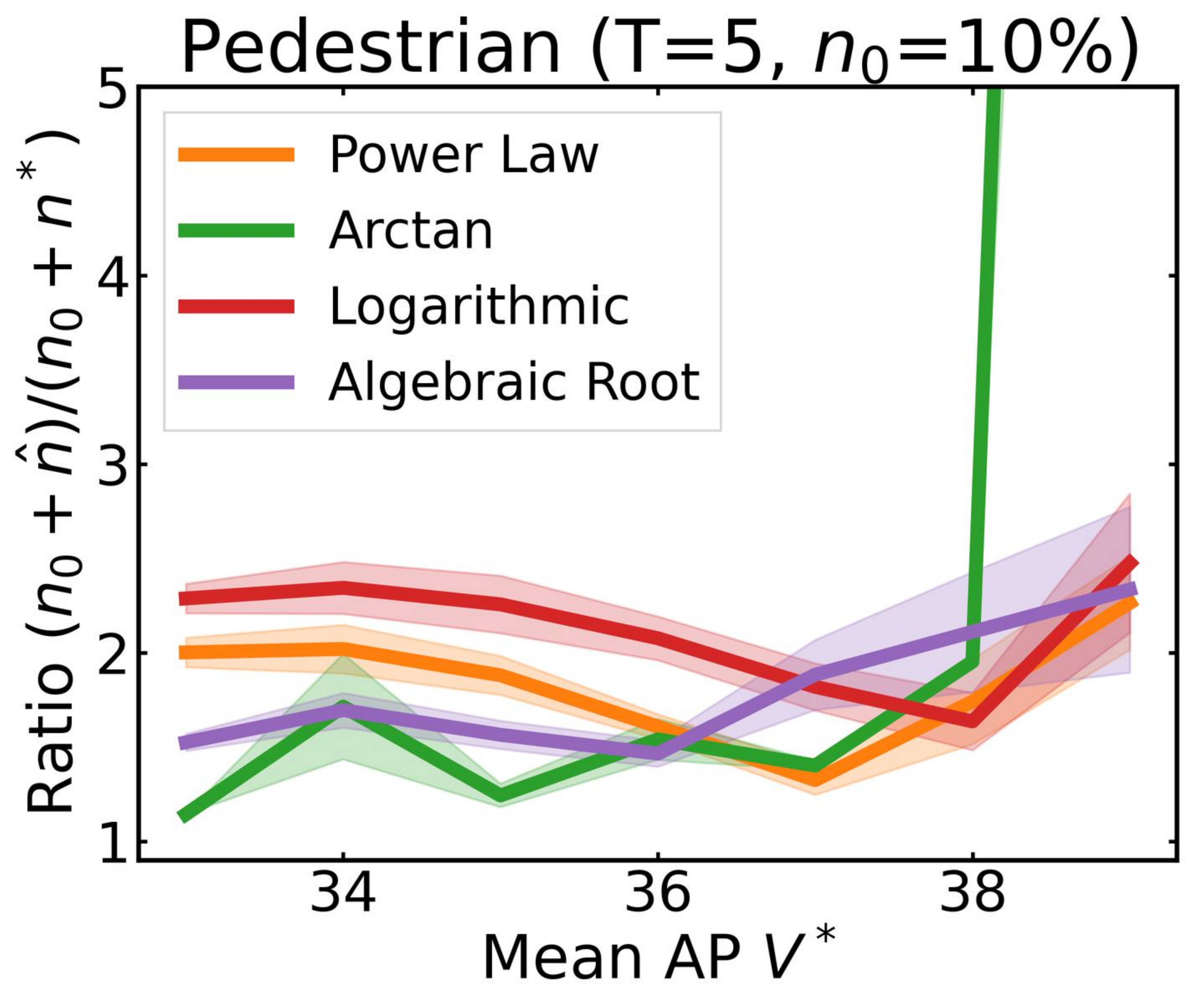} \end{minipage}
\begin{minipage}{0.16\linewidth}\includegraphics[width=1\textwidth]{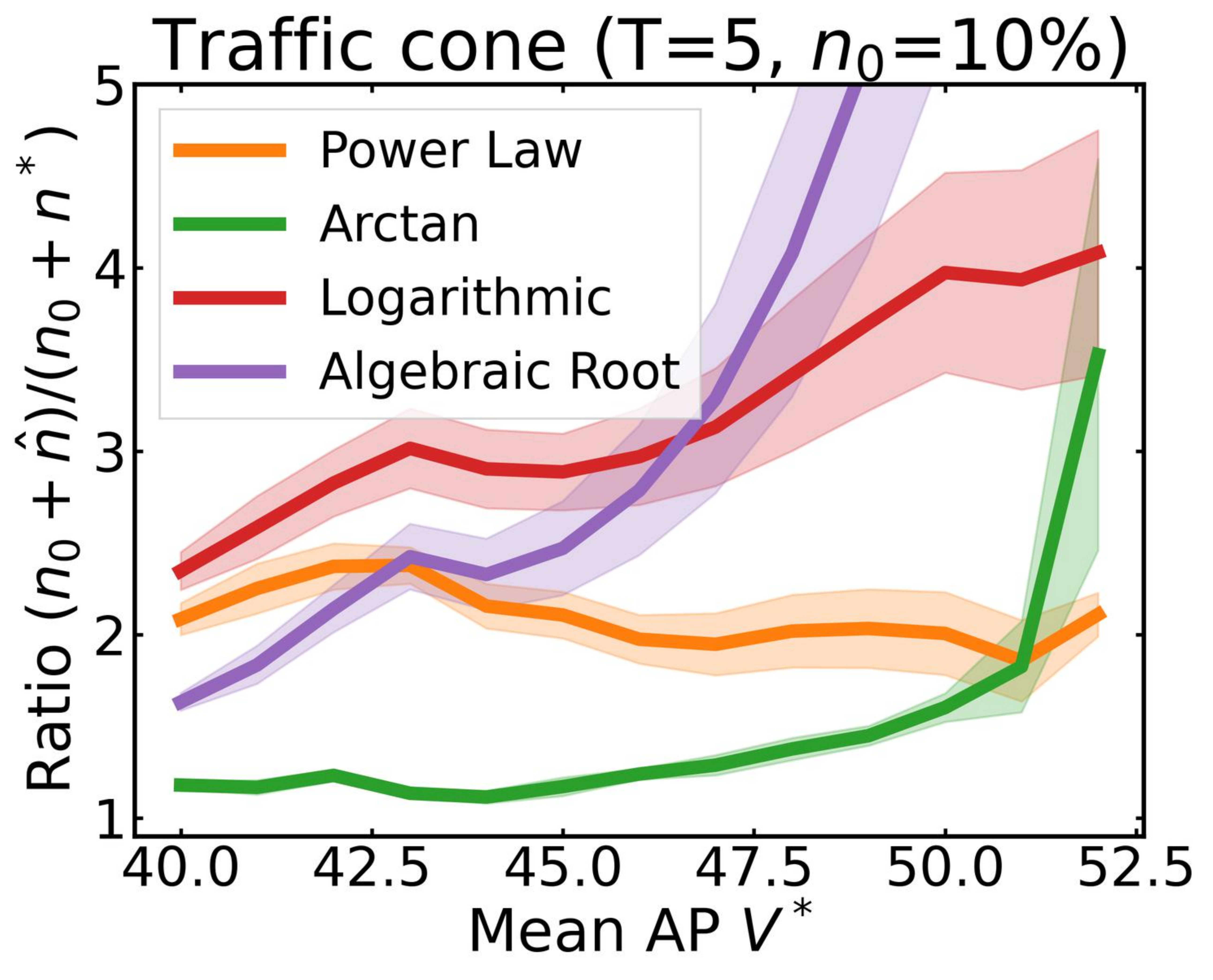} \end{minipage}
\begin{minipage}{0.16\linewidth}\includegraphics[width=1\textwidth]{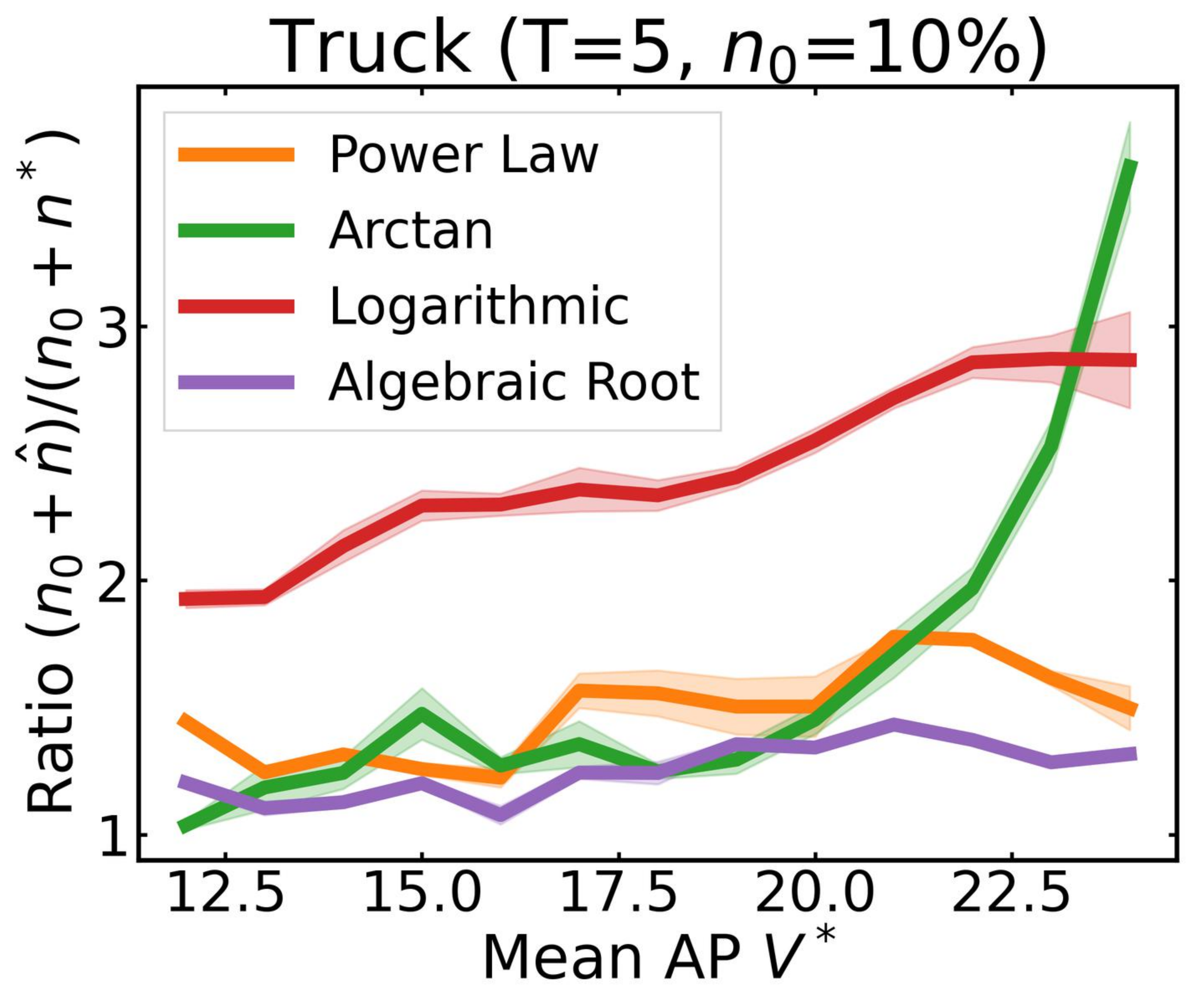} \end{minipage}
\caption{\label{fig:simulation_nuscenes_classes}
Experiments where the score function $V_f(\dataset)$ corresponds to the AP of each individual class from the nuScenes data set. All plots show mean$\pm$standard deviation.
(Left columns) Regression plots extrapolating performance of each individual class.
(Right columns) The ratio of the amount of data collected versus the minimum data needed for different target $V^*$ in simulations initializing with $n_0 = 10\%$ of the data set. We fix $T=5$ and apply the correction factor fit using CIFAR10. The dashed black line corresponds to collecting the least amount of data needed to reach $V^*$.
}
\end{center}
\vspace{-9mm}
\end{figure*}

\section{The data collection problem for class specific metrics}
\label{sec:app_class_specific_metrics}

In many applications, we may be motivated to collect data in order to improve on a specific class.
In this section, we explore scenarios where the score function $V_f(\dataset)$ is a class-specific metric.
We consider 3-D object detection over the nuScenes data set. For every class, we perform regression and simulation over the entire data set after setting $V_f(\dataset)$ to be the AP for that specific class. That is, we use the same setup and data collection as described in Appendix~\ref{sec:app_experiment_setup} but now fit our regression models to the class-specific metric.

Figure~\ref{fig:simulation_nuscenes_classes} (left) plots regression analysis for each class when $n_0=10\%$ of the data. We first observe that the ground truth $v(n)$ are not always concave, monotonically increasing functions. In particular, the bicycle, construction vehicle, and truck classes contain situations where performance slightly decreases after increasing data. Moreover, these three classes are also the classes with the lowest AP even after training with the full data set. 
Finally, all of the regression functions, including the Arctan function, tend to be optimistic for most of the classes. In particular, only the traffic cone class features multiple pessimistic regression functions. These results suggest that in general, it can be more difficult to fit regression curves for individual class-specific metrics as opposed to fitting for mean AP.

Figure~\ref{fig:simulation_nuscenes_classes} (right) plots simulation results after employing the $\tau$ fit from CIFAR10 data for $T=5$ rounds. Here, we demonstrate the effectiveness of our general recipe (\ie to use $T=5$ rounds and incorporate a correction factor $\tau$). That is, for most of the classes, the Power Law, Logarithmic, and Algebraic Root functions consistently achieve ratios between $1$ to $3$. These methods achieve their poorest performance for the bicycle and construction vehicle classes, as these classes are naturally the most noisy and thus more challenging for estimating the data requirement.

It is worth noting that Arctan continues to over-estimate the data requirement by large margins for six out of nine classes, even though the regression plots themselves show that the Arctan function is optimistic. This behavior is due to the fact that we have $T=5$ rounds to meet the data requirement. For instance in the first round when our initial data set contains $2,813$ images, we will significantly under-estimate the data requirement. After the first round however, we will have added a new data point into our regression set and re-fit the Arctan function. It is likely that after this first round, the next Arctan function will be pessimistic. As a result in the second round, we would over-estimate the data requirement, leading to the corresponding curve.
Since the simulations show Arctan to over-estimate, we conclude the pessimistic nature of the Arctan function is generally consistent across data sets and tasks, likely being a factors of the function and the model fitting process.

\begin{figure}[!t]
\begin{center}
\begin{minipage}{0.16\linewidth}\includegraphics[width=1\textwidth]{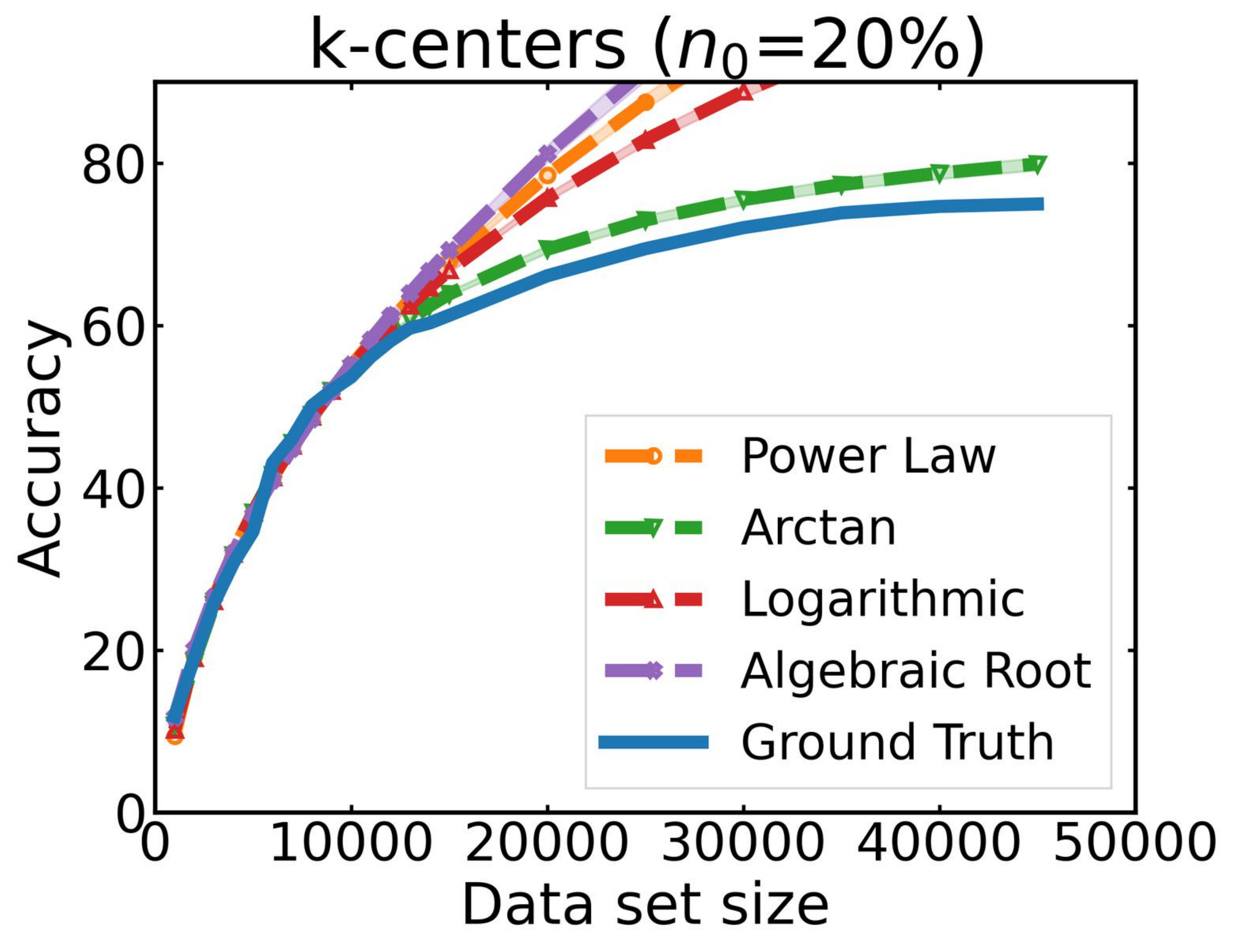} \end{minipage}
\begin{minipage}{0.16\linewidth}\includegraphics[width=1\textwidth]{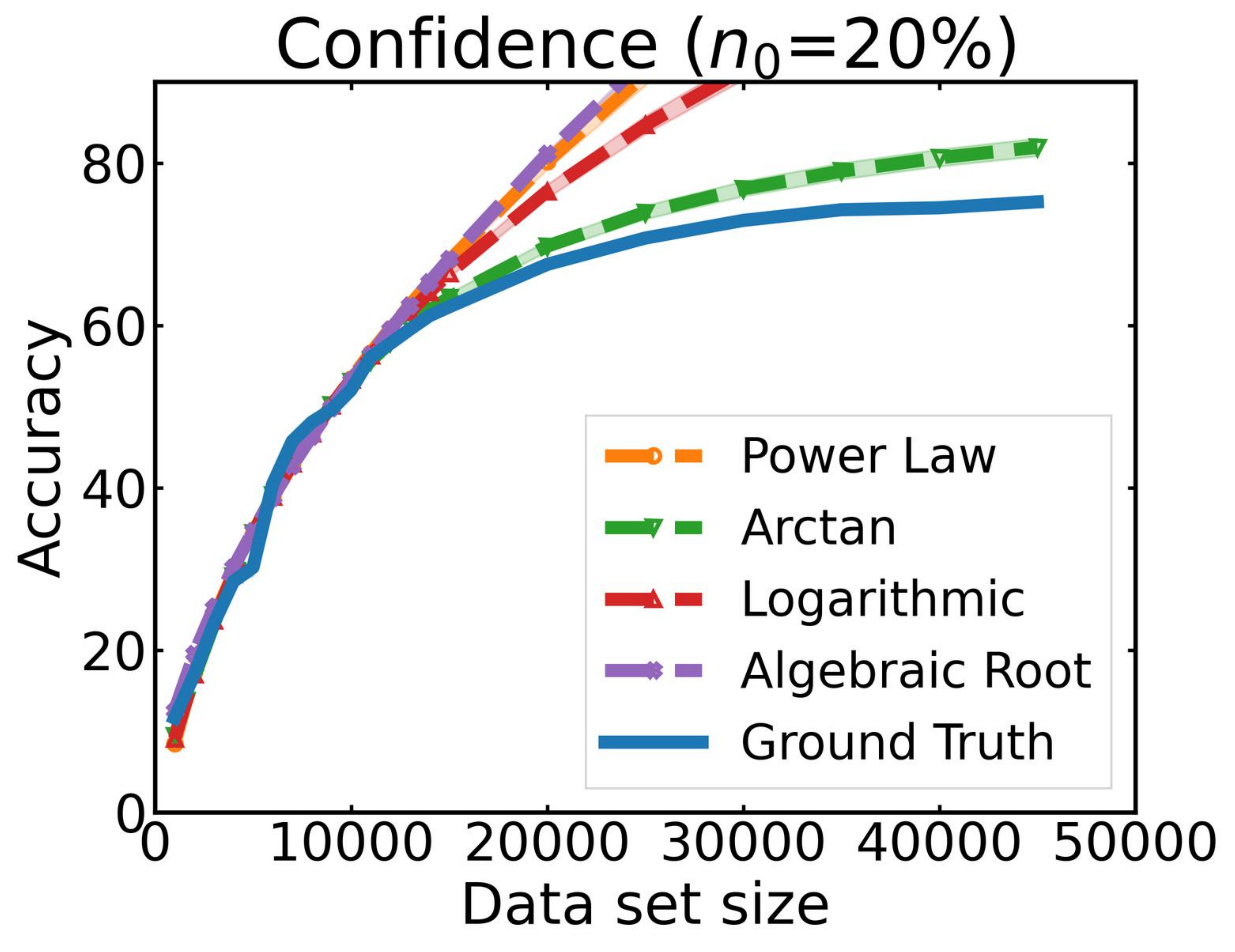} \end{minipage}
\begin{minipage}{0.16\linewidth}\includegraphics[width=1\textwidth]{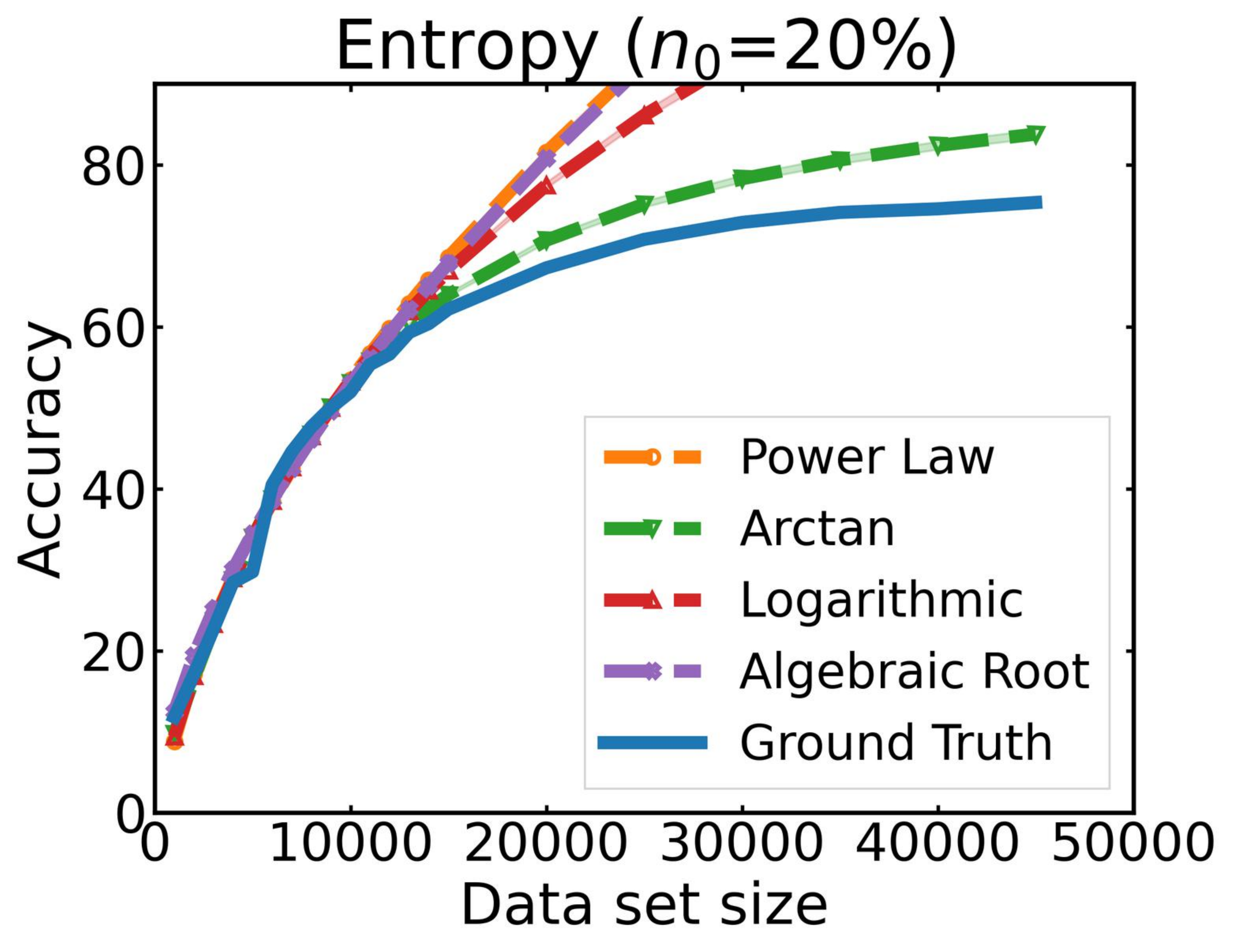} \end{minipage}
\begin{minipage}{0.16\linewidth}\includegraphics[width=1\textwidth]{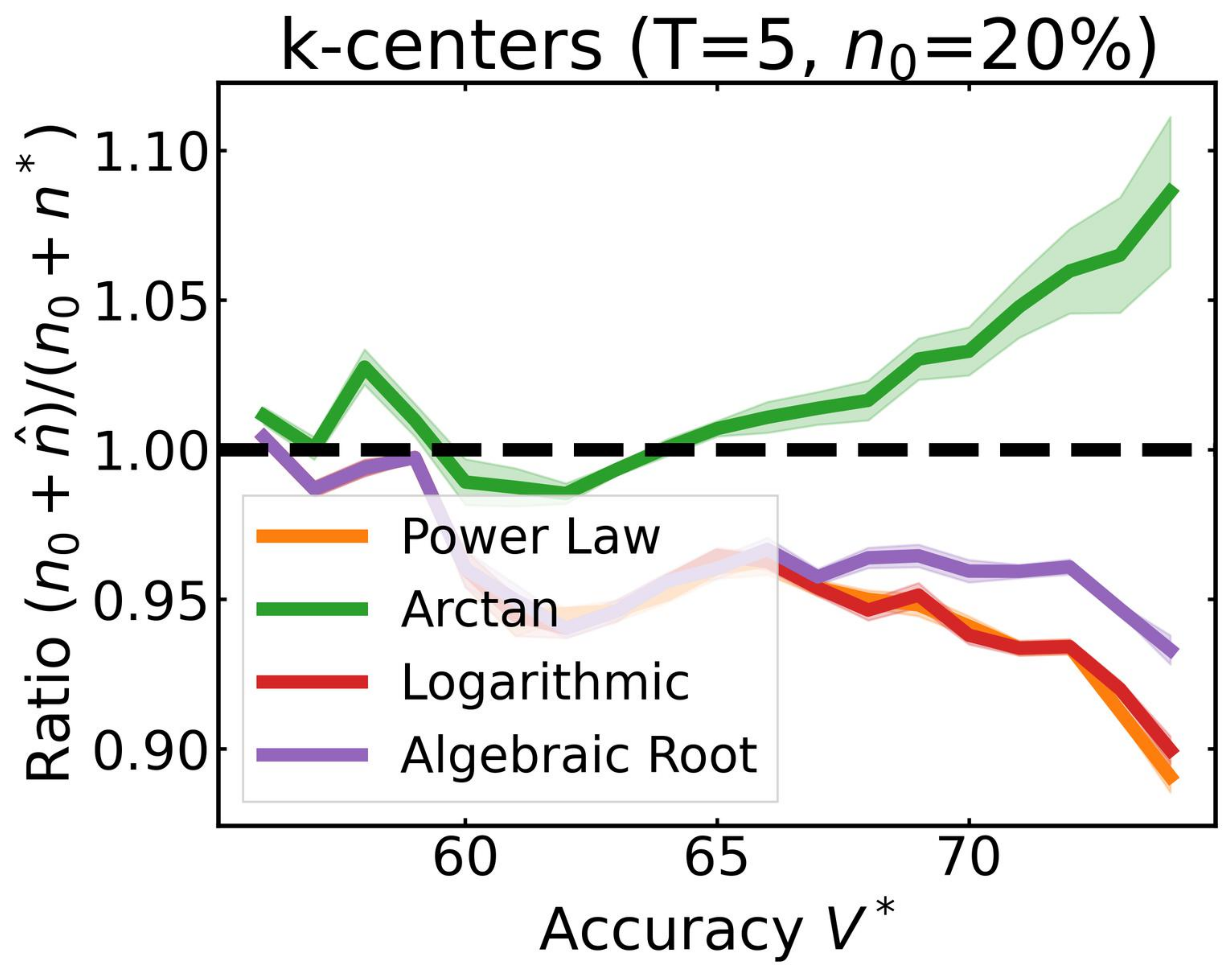} \end{minipage}
\begin{minipage}{0.16\linewidth}\includegraphics[width=1\textwidth]{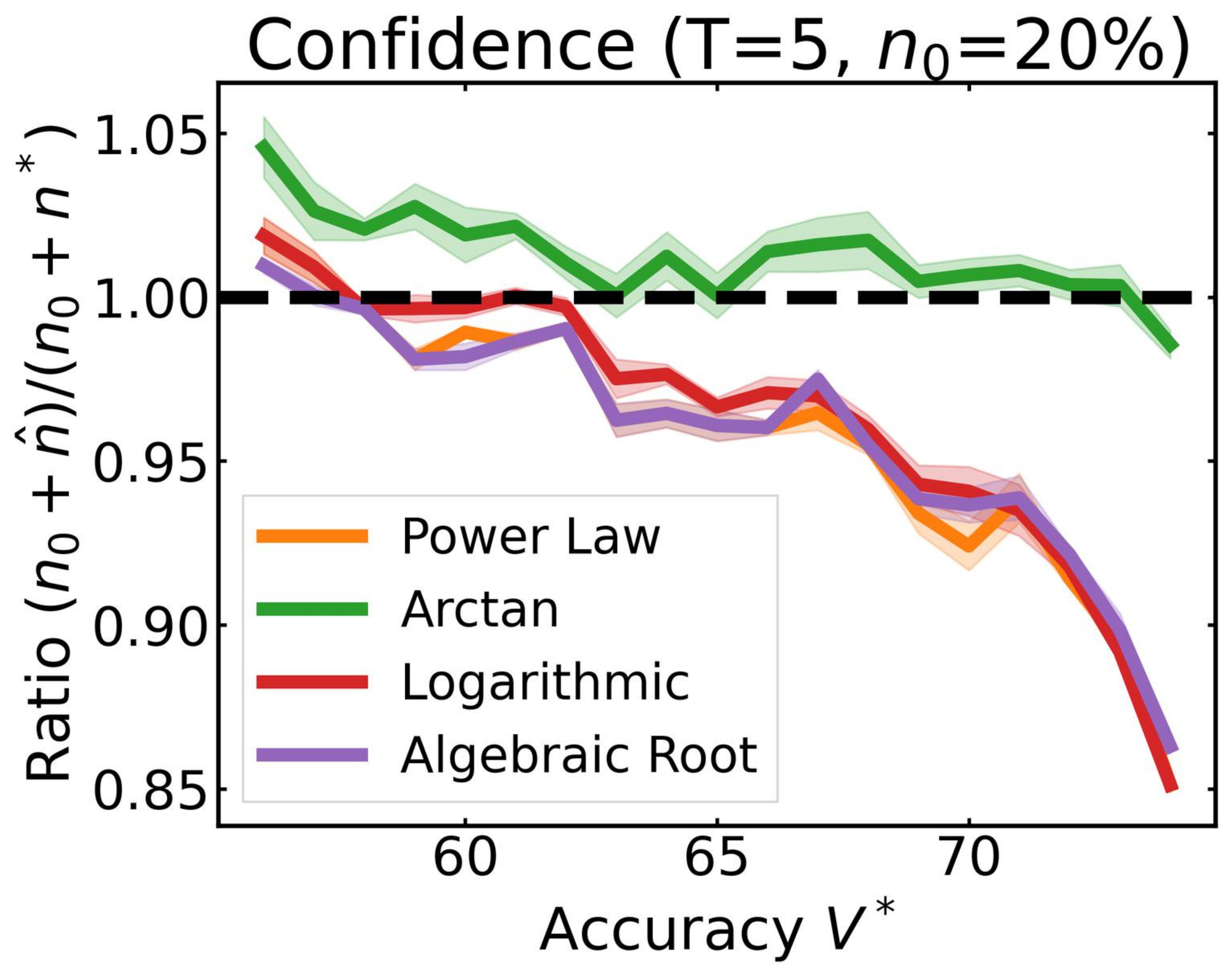} \end{minipage}
\begin{minipage}{0.16\linewidth}\includegraphics[width=1\textwidth]{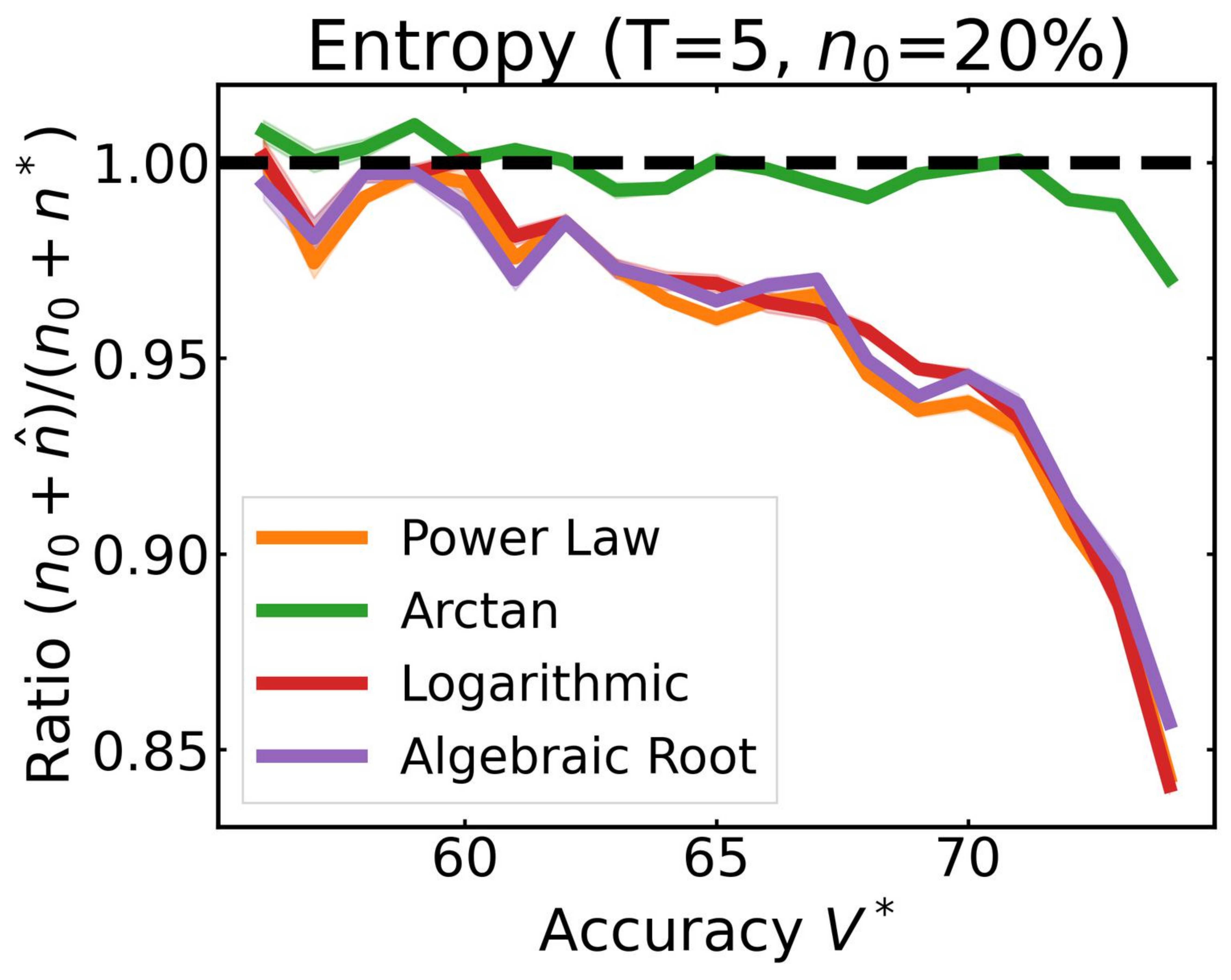} \end{minipage}
%
%
%
\vspace{-8mm}
\begin{minipage}{\linewidth}
\caption{
Experiments evaluating three different active learning strategies on CIFAR100 with $n_0 = 20\%$ of the data set and $T=5$ rounds.
(Left) regression plots showing mean$\pm$standard deviation extrapolating performance. (Right) The ratio of data collected versus the minimum data needed (y-axis) for different target $V^*$ (x-axis) in simulations. 
}
\label{fig:active_learning} 
\end{minipage}
\end{center}
\vspace{-4mm}
\end{figure}

\section{The data collection problem with active learning}
\label{sec:app_active_learning}

Although all of our experiments so far have considered collecting data by randomly sampling data points from a fixed training data set, we can also consider structured techniques such as active learning, where data is collected and labeled using sample efficient strategies. In conventional active learning, we would collect a pre-determined budget of $B$ data points each round over $T$ rounds to attain the best possible final model. Instead, if we are given a performance target $V^*$, we can use our data collection framework to dynamically choose a budget $B_t$ to collect in each round for up to $T$ rounds.

Figure~\ref{fig:active_learning} plots experiments of data collection for CIFAR100 when using three active learning strategies, $k$-centers~\cite{sener2017active}, Least Confidence~\cite{settles2009active}, and Max Entropy~\cite{settles2009active}, rather than random sampling. Figure~\ref{fig:active_learning} (left) demonstrate regression analysis for each strategy when initializing with $n_0 = 20\%$ of the data set. We use a larger initial set than in the previous experiments because the ground truth learning curves are not always concave monotonically increasing, especially in the lower data regimes. By giving a sufficiently large initial data set, the regime of extrapolation is stable and does not contain erratic trends.

Figure~\ref{fig:active_learning} (right) show our simulation analysis for $T=5$ rounds of data collection using each active learning strategy. 
We set $\tau = 0$ since the previous correction factor values were designed from random sampling and with an initial $n_0 = 10\%$ of the data set.
Nonetheless, this simulation approximates the scenario where we would dynamically choose the budget before each active learning round for up to $T$ rounds. These plots validate that our main empirical findings all hold regardless of the specific techniques used in collecting data. For each active learning strategy, all four of the regression functions achieve ratios between $0.8$ and $1.1$. Furthermore, Arctan typically over-estimates the data requirement whereas the others are more likely to under-estimate the requirement.